\documentclass[11pt]{article}
\usepackage{mathrsfs,amsmath,amsfonts,amssymb,bm,bbm,dsfont,pifont,amscd,
amsthm,stmaryrd,euscript,color,xcolor}
\usepackage{algorithmic}
\usepackage{algorithm}
\usepackage{url}
\usepackage{mathtools}

\newcommand\setItemnumber[1]{\setcounter{enumi}{\numexpr#1-1\relax}}

\definecolor{myred}{rgb}{0.8,0,0}   
\definecolor{mygreen}{rgb}{0.0,0.7,0.0}  
\definecolor{myblue}{rgb}{0.0,0.1,0.7}  

\newcommand{\sdsize}[1]{\tiny{#1}}
\newcommand{\mypm}{ $\pm$ }

\newcommand{\betterentry}[1]{{\textcolor{orange}{#1}}}
\newcommand{\significantbetterentry}[1]{{\textcolor{myred}{#1}}}

\newcommand{\worseentry}[1]{{\textcolor{mygreen}{#1}}}
\newcommand{\significantworseentry}[1]{{\textcolor{myblue}{#1}}}

\newcommand{\tablefontsize}{\tiny}

\newlength{\myleftmargin}
\usepackage{amsmath}
\usepackage{amssymb}
\DeclareSymbolFontAlphabet{\Bbb}{AMSb}

\newtheorem{theorem}{Theorem}[section]
\newtheorem{lemma}[theorem]{Lemma}
\newtheorem{proposition}[theorem]{Proposition}
\newtheorem{corollary}[theorem]{Corollary}

\newtheorem{definition}[theorem]{Definition}

\theoremstyle{definition}

\newtheoremstyle{myremark}{}{}{}{}{\bfseries}{.}{ }{\thmname{#1}\thmnumber{ #2}\thmnote{ (#3)}}
\theoremstyle{myremark}
\newtheorem{remark}[theorem]{Remark}
\newtheorem{example}[theorem]{Example}

\newcommand{\atob}[2]{\emph{#1)} $\Rightarrow$ \emph{#2)}.} 
\newcommand{\aquib}[2]{\emph{#1)} $\Leftrightarrow$ \emph{#2)}.} 
\newcommand{\ada}[1]{\emph{#1).}}

\newenvironment{proofof}[1]{\noindent{\bf Proof of #1:}}{\qed\medskip}

\newcommand{\remarkend}{\hfill$\blacktriangleleft$}

\newlength{\fixboxwidth}
\newcommand{\ca}[1]{{\cal #1}}

\newcommand{\mycdot}{\,\cdot\,}
\newcommand{\descriptpar}[1]{\vspace*{1ex}\textbf{#1.}\hspace*{1ex}}
\newcommand{\scalingname}[1]{\textbf{#1}}
\newcommand{\softwarename}[1]{{\tt #1}}
\newcommand{\datasetname}[1]{{\sc#1}}

\newcommand{\N}{\mathbb{N}}

\newcommand{\R}{\mathbb{R}}
\newcommand{\Rd}{{\mathbb{R}^d}}

\newcommand{\Sd}{{\mathbb{S}^{d-1}}}

\newcommand{\E}{\mathbb{E}}
\DeclareMathOperator{\mode}{mode}
\DeclareMathOperator{\var}{Var}

\renewcommand{\a}{\alpha}
\renewcommand{\b}{\beta}
\newcommand{\g}{\gamma}
\newcommand{\G}{\Gamma}
\renewcommand{\d}{\delta}

\newcommand{\e}{\varepsilon}

\newcommand{\lb}{\lambda}

\renewcommand{\r}{\rho}

\newcommand{\s}{\sigma}

\renewcommand{\t}{\tau}
\newcommand{\p}{\varphi}

\DeclareMathOperator{\rank}{rank}

\DeclareMathOperator{\spann}{span}
\DeclareMathOperator{\vol}{vol}

\DeclareMathOperator{\co}{co}

\newcommand{\RD}[2]{{{\cal R}_{#1,D}(#2)}}
\newcommand{\RDp}[2]{{{\cal R}_{#1,D'}(#2)}}

\newcommand{\fpb}{{f_{L,P}^*}}

\newcommand{\eins}{\boldsymbol{1}}

\newcommand{\snorm}[1]{\Vert #1 \Vert}
\newcommand{\nsnorm}[1]{|\!|\!| #1 |\!|\!|}

\newcommand{\inorm}[1]{\Vert #1 \Vert_\infty}

\newcommand{\Lx}[2]{{L_{#1}(#2)}}

\newcommand{\relu}[1]{|#1|_+}
\newcommand{\brelu}[1]{\bigl|#1\bigr|_+}

\newcommand{\eul}{\mathrm{e}}

\newcommand{\xmin}{x_{\mathrm{min}}}
\newcommand{\xmax}{x_{\mathrm{max}}}

\newcommand{\archx}[1]{\ca A_{#1}}
\newcommand{\arch}[2]{\archx{#1, #2, 1}}

\newcommand{\heetal}{\emph{He-et-al.}}

\DeclareMathOperator{\ate}{ATE}
\DeclareMathOperator{\te}{TE}
\DeclareMathOperator{\rate}{RATE}

\DeclareMathOperator{\att}{ATT}
\DeclareMathOperator{\ati}{ATI}

\DeclareMathOperator{\ico}{ico}
\DeclareMathOperator{\coni}{coni}
\DeclareMathOperator{\ray}{ray}
\newcommand{\Aip}{A_i^+}
\newcommand{\Aim}{A_i^-}

\newcommand{\sd}{\s^{d-1}}
\newcommand{\lbd}{\lb^{d}}

\newcommand{\sgnorm}[1]{\snorm{#1}_{\Psi_2}}

\makeatletter
\def\moverlay{\mathpalette\mov@rlay}
\def\mov@rlay#1#2{\leavevmode\vtop{%
   \baselineskip\z@skip \lineskiplimit-\maxdimen
   \ialign{\hfil$\m@th#1##$\hfil\cr#2\crcr}}}
\newcommand{\charfusion}[3][\mathord]{
    #1{\ifx#1\mathop\vphantom{#2}\fi
        \mathpalette\mov@rlay{#2\cr#3}
      }
    \ifx#1\mathop\expandafter\displaylimits\fi}
\makeatother

\usepackage[font={footnotesize}]{caption}

\begin{document}
\title{A Sober Look at Neural Network Initializations}
\author{Ingo Steinwart\\
Institute for Stochastics and Applications\\
Faculty 8: Mathematics and Physics\\
University of Stuttgart\\
D-70569 Stuttgart Germany \\
\texttt{\small ingo.steinwart@mathematik.uni-stuttgart.de}
}

\maketitle

\begin{abstract}
 Initializing the weights and the biases is a key 
part of the training process of a neural network.
Unlike the subsequent optimization phase, however, 
the initialization phase has gained only limited 
attention in the literature. 
In this paper we discuss some
consequences of commonly used initialization strategies
for vanilla DNNs with ReLU activations. Based on 
these insights we   then develop an alternative 
initialization strategy.  Finally, we present
some large scale experiments assessing the quality 
of the new initialization strategy.
\end{abstract}

\section{Introduction}

Improving and understanding the training phase of deep neural networks 
has attracted a lot of attention in the last couple of years. This training phase
mostly consists of minimizing an empirical risk term, and due to the structure 
of deep neural networks, the corresponding optimization landscape is 
convoluted and highly non-convex. To avoid getting stuck in local minima
several variants of stochastic gradient descent have been proposed and 
successfully applied. These success stories suggest that the initialization
of neural networks, that is, choosing the starting point of the optimization,
has become less important. In fact, the two commonly used heuristics proposed 
in \cite{GlBe10a, HeZhReSu15a} both 
focus on normalizing the variance 
of the weights of the neural network to ensure that the gradients of deep
networks do not exponentially explode or implode. So far, however,  
positive or negative 
side-effects of these 
initialization strategies have not been investigated in depth. 
This is the first goal of our paper, and the second goal is to use 
these insights to develop a new initialization strategy.

To be a bit more specific let  $\relu\cdot :\R\to [0,\infty)$ be the ReLU function, that is 
$\relu t := \max\{0, t\}$. For $d\in \N$,  a single neuron is then given by 
% 
% 
% 
% Now, the considered hidden layer consists of $m$ neurons of the form
\begin{align*} 
 h:\R^d & \to [0,\infty)\\  
 x&\mapsto  \relu{\langle a, x\rangle + b} \, ,
\end{align*}
where $a \in \R^d$ and $b \in \R$ are the weight vector and the bias of the neuron. A layer of width $m$ is 
a function $H:\R^d \to [0,\infty)^m$, whose coordinate functions are  neurons. Finally, a deep neural network 
is the composition of layers followed by an affine linear function, that is, a function $g:\R^d \to \R$ of the form
\begin{align}\label{general-dnn}
 g = v\circ H_L \circ H_{L-1}\circ \dots\circ H_1\, ,
\end{align}
where $H_l:\R^{m_{l-1}}\to [0,\infty)^{m_l}$ are layers with $m_0 := d$ and the output neuron $v:\R^{m_L}\to \R$
is a function given by $v(x) = \langle w, x\rangle + c$, where $w$ and $c$ are the weight vector and the bias of the output 
neuron.
Clearly, $g$ is always a continuous and piecewise linear function, which is fully described by all its weight vectors and biases.
Moreover, the architecture of a deep neural network is described by the number $L$ of hidden layers, the input dimension $d$,
and the widths $m_1, \dots, m_L$. In the following, we write 
\begin{displaymath}
 \archx{d, m_1, \dots, m_L, 1} := \bigl\{ g:\R^d\to \R\, \bigl|\,  g \mbox{ is of the form \eqref{general-dnn} with layers $H_l:\R^{m_{l-1}}\to [0,\infty)^{m_l}$ and  $m_0 := d$ }  \bigr\}\, .
\end{displaymath}
To train a neural network of fixed architecture, we need a labeled data set 
 $D:=((x_1,y_1),\dots,(x_n,y_n)) \in (X\times \R)^n$, where  $X\subset \R^d$ is called the input space,
 as well as a loss function $L:\R\times \R\to [0,\infty)$. For a function $f:X\to \R$, we then define 
the empirical $L$-risk 
by 
\begin{displaymath}
   \RD Lf := \frac 1 n \sum_{j=1}^n L(y_j,f(x_j))\, .
\end{displaymath}
Now, training a network  seeks an (approximate) empirical risk minimizer within the given architecture, that 
is a network $g_D\in \archx{d, m_1, \dots, m_L, 1}$ such that 
\begin{align}\label{learn-dnn}
 \RD L {g_D} \approx \inf  \bigl\{ \RD L g : g\in \archx{d, m_1, \dots, m_L, 1}\bigr\}\, .
\end{align}
Usually, the considered loss function is differentiable in its second argument and the networks $g\in \archx{d, m_1, \dots, m_L, 1}$
are parameterized by their weights and biases. The optimization problem is then executed on these parameters
with the help of some variant of stochastic gradient descent (SGD). Consequently, the training
produces a sequence $g_0,g_1,\dots,g_T \in \archx{d, m_1, \dots, m_L, 1}$ from which a $g_D$ is chosen, e.g.~$g_D := g_T$.
Unfortunately, however, the optimization problem \eqref{learn-dnn} is, in general, highly non-convex, and
therefore, the final $g_D$ may depend on the initial $g_0$. Initializing the network, i.e., choosing 
an initial $g_0$, is therefore a potentially crucial part of the entire training.

It is well-known, that initializing  all weights and biases to the same value, e.g.~to zero, hinders 
training by SGD since all neurons in the same layer will be updated in the same way. For this reason, the 
weights (and biases) are typically initialized randomly. More precisely, the most common initialization
strategies proposed in \cite{GlBe10a} and \cite{HeZhReSu15a} both 
fix some random variable $A$ with distribution $\mu$, that is $A\sim \mu$, and then initialize 
the weights of the layer $H_l$ by realizations of independent copies of $\s_{m_{l-1}, m_l}A$, where
$\s_{m_{l-1}, m_l}$ is a suitable scaling factor. In fact, $\mu$ is usually either the standard normal 
distribution or the uniform distribution on e.g.~$[-1,1]$. Moreover, both 
papers propose to initialize the biases to zero, but some other heuristics also recommend a small
positive value such as $0.1$ or $0.01$, or a some small random value, 
instead. We refer to \cite[Ch.~8.4]{GoBeCo16} for a more detailed discussion on these and other 
initialization strategies.

In any case, the resulting initial function $g_0$ is a random function, and one may ask how suitable this 
starting point $g_0$ is. So far, this question has not been answered in a satisfying manner, in fact,
most papers dealing with this question only apply some heuristic arguments, mostly centered around
effects on SGD updates on the weights, and report some empirical findings, mostly on a few data sets
related to images. 

The goal of this paper is to go beyond this by investigating 
how different initialization strategies influence the shape of the function $g_0$.
To this end, we first investigate the most simple case of one-dimension input data and one 
hidden layer, that is 
$d=L=1$ in Section \ref{sec:one-d-sit}. Here it turns out that we can explicitly compute several key quantities 
such as the probability of initializing a neuron into an inactive state. 
As a consequence, we can also compare the effects of different initialization strategies,
for example, we will see why it is better to choose a small positive value for the bias 
instead of a small negative value. Finally,  based on these insights, we will develop
a first alternative initialization strategy.
In Section \ref{sec:general} we will 
then investigate the significantly more complicated general situation.
Here we will compute, for example, the influence of $\s_{m_{l-1}, m_l}A$ on the size and the direction
of the weight vector, as well as on the size of the output of $H_l$. In addition, we will 
investigate the effect of different initialization strategies for the bias term.
Based on these insights we will then develop a new initialization strategy that spreads
the active and inactive regions of each neuron more widely across the space spanned by the input data
of the layer.
Finally, in Section \ref{sec:experiments} we present some experiments that compare to 
the new initialization strategy to the one of \cite{HeZhReSu15a}.

\section{The simplest case: One-dimensional data and one hidden layer}\label{sec:one-d-sit}

In this section we explore the effects of different initialization strategies 
in the simplest case of one-dimensional input data and neural networks with one hidden
layer consisting of ReLU-neurons.
To be more precise, we assume that our input space $X$ is 
a subset of 
$\R$
 and that our hidden layer has $m$ neurons $h_1,\dots,h_m:\R\to \R$
of the form 
\begin{displaymath}
   h_i(x) = \brelu{a_i x + b_i}\, ,\qquad \qquad x\in \R,
\end{displaymath}
where $a_i,b_i\in \R$ are the weights and the biases of these neurons.
Consequently, our network can represent exactly those functions 
  $g:\R\to \R$ that are of the form
\begin{equation}\label{first-layer}
   g(x) = \sum_{i=1}^m w_i \brelu{a_i x + b_i} +c\, , \qquad \qquad x\in \R, 
\end{equation}
where $w_1,\dots,w_m\in \R$ are the weights and $c\in \R$ is the bias of the output neuron.
The goal of the training process is then to find suitable values for 
 $a_1,b_1, \dots, a_m, b_m\in \R$, 
 $w_1,\dots,w_m\in \R$, and $c\in \R$.
Let us denote the set of all functions that can be represented by our network by 
$\arch 1m$, that is 
\begin{displaymath}
 \arch 1m 
:= \bigl\{ g:\R\to \R\, \bigl|\, \mbox{ $g$ has a representation 
\eqref{first-layer} for suitable $a_i,b_i,w_i\in\R$ and $c\in \R$ }\bigr\}\, .
\end{displaymath}
It is not hard to see that given a $g\in \arch 1m$, the representing 
parameters in \eqref{first-layer}
are anything than unique. 
% For example, if $g$ has a representation of the form
% \eqref{first-layer} and $\pi:\{1,\dots,m\}\to \{1,\dots,m\}$ is a permutation, then $g$ 
% is also given by 
% \begin{equation}\label{first-layer-permut}
%    g(x) = \sum_{i=1}^m w_{\pi(i)} \brelu{a_{\pi(i)} x + b_{\pi(i)}} +c\, , \qquad \qquad x\in \R.
% \end{equation}
% In other words, we can freely rearrange the order of the neurons
% without changing the represented function
%  as long as we rearrange
% the correspnding weights $w_i$ of the subsequent layer in the same way.  

Now notice that for $a_i=0$ the neuron $h_i$ is a constant function, namely $h_i \equiv \relu{b_i}$.
Moreover, if $a_i\neq  0$, then $h_i$ is a continuous, piecewise linear function
with exactly one kink, and this kink is located at $x_i^* := -b_i/a_i$.
Inspired by spline interpolation
we call $x_i^*$ a \emph{knot} throughout this section.
A simple calculation shows that in the case $a_i<0$, the function
$h_i$ is given by 
		\begin{align}\label{left-neuron}
		   h_i(x) = 
		\begin{cases}
		   a_ix+b_i & \mbox{ if } x\in (-\infty, x_i^*]\\
			0 &\mbox{ if } x\in [x_i^*, \infty)\, ,
		\end{cases}
		\end{align}
while for $a_i> 0$, it  is given by 
		\begin{align}\label{right-neuron}
		   h_i(x) = 
		\begin{cases}
		   0 & \mbox{ if } x\in (-\infty, x_i^*]\\
			a_ix+b_i &\mbox{ if } x\in [x^*_i, \infty)\, .
		\end{cases}
\end{align}
To describe the corresponding behavior of the function $g$ with representation \eqref{first-layer}
we now write $I= \{1,\dots,m\}$, $I_* :=\{i\in I: a_i\neq 0\}$, and 
\begin{align*} 
I_- & := \{ i\in I_*: a_i < 0\}\\  
I_+ & := \{ i\in I_*: a_i > 0\}\, .
\end{align*}
Moreover, throughout the rest of this section we write $x_i^*:= -b_i/a_i$  for $i\in I_*$.

Now, we immediately obtain the following 
result, which provides a different representation of $g\in \arch 1 m$.

\begin{proposition}\label{result:one-layer-repres}
For $m\geq 1$ we   fix a $g\in \arch 1m$ with the representation
\eqref{first-layer}.
Then for all $x\in \R$ we have 
\begin{equation*}%\label{result:one-layer-repres-h1}
   g(x) = 
\sum_{i\in I_-: x\leq x_i^*} \,w_i(a_ix + b_i) + \sum_{i\in I_+: x\geq x_i^*} \,w_i(a_ix + b_i)
 + \sum_{i\in I\setminus I_*} w_i \relu{b_i} + c
\end{equation*}
\end{proposition}

Our next goal is to derive explicit formulas for the partial derivatives
considered during training of our neural network. To this end, we say that a loss function
$L:\R\times \R\to [0,\infty)$ is \emph{differentiable}, if for all $y\in \R$ the function
\begin{displaymath}
   t\mapsto L(y,t)
\end{displaymath}
is differentiable. In this case we write 
\begin{displaymath}
L'(y,t) := \frac{\partial L}{\partial t} (y,t)  \, .  
\end{displaymath}
Since the function $t\mapsto \relu t$ is not differentiable at $0$, we formally need to exclude 
all occasions, at which we would need to use its derivative at $0$. 
However, from a practical point of view this is not feasible, since there are actually realistic situations
in which the ``derivative'' of $t\mapsto \relu t$ at $t=0$ is needed,
see e.g.~Example \ref{example:He-init-1d} below.
For this reason, we pick a $\partial_0\in [0,1]$, which will serve as a surrogate for the 
missing derivative.\footnote{In ``native'' PyTorch, for example, we find $\partial_0 := 0$, see 
\url{https://github.com/pytorch/pytorch/issues/11662#issuecomment-423138052}, and the same 
choice is taken in Tensorflow, see \url{https://github.com/tensorflow/tensorflow/blob/e39d8feebb9666a331345cd8d960f5ade4652bba/tensorflow/core/kernels/relu_op_functor.h#L54}.}
To be more precise, in all formulas involving derivatives of $t\mapsto \relu t$ we will use $\partial_0$, whenever
we would actually need the derivative $t\mapsto \relu t$ at $t=0$. 
In addition, to allow for compact formulas, we define $\partial _t = 0$ for $t<0$ and $\partial_t = 1$ for $t>0$.
Then,
our approach gives
\begin{align}\label{relu-chain-rule-a}
 \frac{\partial \relu{a x +b}  }{\partial a}(a_0)
 = 
 \partial_{a_0 x + b } \cdot x
 =
 \begin{cases}
  0 & \mbox{ if } a_0 x + b < 0\\
  \partial_0 \cdot x &\mbox{ if } a_0 x + b = 0\\
  x &\mbox{ if } a_0 x+b>0\, ,
 \end{cases}
\end{align}
where the first and third case is covered by the usual chain rule and in the second case
we used $\partial_0$ as a formal surrogate. Similarly, we get 
\begin{align}\label{relu-chain-rule-b}
 \frac{\partial \relu{a x +b}  }{\partial b}(b_0)
 = 
 \partial_{a x + b_0 }
 =
 \begin{cases}
  0 & \mbox{ if } a x + b_0 < 0\\
  \partial_0  &\mbox{ if } a x + b_0 = 0\\
  1 &\mbox{ if } a x+b_0>0\, .
 \end{cases}
\end{align}
Moreover, if $f:\R\to \R$ is a differentiable function, then we formally apply the chain rule 
in the following sense
\begin{align}\label{chain-rule-extension-a}
  \frac{\partial f\bigl(\relu{a x +b} \bigr)  }{\partial a}(a_0)
 &= 
 f'\bigl( \relu{a_0 x +b}  \bigr)\cdot 
  \frac{\partial \relu{a x +b}  }{\partial a}(a_0)
 = f'\bigl( \relu{a_0 x +b}  \bigr)\cdot 
 \partial_{a_0 x + b } \cdot x   \\ \label{chain-rule-extension-b}
 \frac{\partial f\bigl(\relu{a x +b} \bigr)  }{\partial b}(b_0)
 &= 
 f'\bigl( \relu{a x +b_0}  \bigr)\cdot 
  \frac{\partial \relu{a x +b}  }{\partial b}(b_0)
 = f'\bigl( \relu{a x +b_0}  \bigr)\cdot 
 \partial_{a x + b_0}   \, .
\end{align}
In particular,
 given a $g\in \arch 1m$, these extended chain rules are used when computing 
partial derivatives 
of $\RD L g$ with respect to the parameters in  \eqref{first-layer}. The next proposition
executes these computations.

% 
% 
% The next  lemma computes the partial derivatives for
% $\RD Lg$ for $g\in \arch 1m$
% with respect to the 
% parameters in .

\begin{proposition}\label{result:one-layer-deriv}
Let $L:\R\times \R\to [0,\infty)$ be a differentiable loss function.
   For $m\geq 1$ we  further fix a $g\in \arch 1m$ with the representation
\eqref{first-layer}. 
Then for   $i\in I_-$ we have
\begin{align*}
   \frac{\partial \RD L{g}}{\partial a_i} (w,c,a,b)
&=  \frac {w_i}n \sum_{j:x_j < x_i^*} L'\bigl(y_j, g(x_j, w,c,a,b)\bigr)\cdot  x_j +  \frac {\partial_0\cdot w_i\cdot  x_{i}^*}n \sum_{j:x_j = x_i^*} L'\bigl(y_j, g(x_{i}^*, w,c,a,b)\bigr)  \\
   \frac{\partial \RD L{g}}{\partial b_i} (w,c,a,b)
&=  \frac {w_i}n \sum_{j:x_j < x_i^*} L'\bigl(y_j, g(x_j, w,c,a,b)\bigr)   +   \frac {\partial_0\cdot w_i}n \sum_{j:x_j = x_i^*} L'\bigl(y_j, g(x_{i}^*, w,c,a,b)\bigr)   \\
   \frac{\partial \RD L{g}}{\partial w_i} (w,c,a,b)
&=  \frac 1n \sum_{j:x_j < x_i^*} L'\bigl(y_j, g(x_j, w,c,a,b)\bigr) \cdot (a_i \cdot x_j + b_i) \, .
\end{align*}
Moreover, for   $i\in I_+$ we have
\begin{align*}
   \frac{\partial \RD L{g}}{\partial a_i} (w,c,a,b)
&=   \frac {w_i}n \sum_{j:x_j > x_i^*} L'\bigl(y_j, g(x_j, w,c,a,b)\bigr)  \cdot  x_j  +  \frac {\partial_0\cdot w_i\cdot  x_{i}^*}n \sum_{j:x_j = x_i^*} L'\bigl(y_j, g(x_{i}^*, w,c,a,b)\bigr) \\
   \frac{\partial \RD L{g}}{\partial b_i} (w,c,a,b)
&=   \frac {w_i}n \sum_{j:x_j > x_i^*} L'\bigl(y_j, g(x_j, w,c,a,b)\bigr)   +   \frac {\partial_0\cdot w_i}n \sum_{j:x_j = x_i^*} L'\bigl(y_j, g(x_{i}^*, w,c,a,b)\bigr) \\
   \frac{\partial \RD L{g}}{\partial w_i} (w,c,a,b)
&=  \frac 1n \sum_{j:x_j > x_i^*} L'\bigl(y_j, g(x_j, w,c,a,b)\bigr) \cdot (a_i \cdot x_j + b_i) \, .
\end{align*}
In addition, for   $i\in I\setminus I_*$ we have  
\begin{align*}
   \frac{\partial \RD L{g}}{\partial a_i} (w,c,a,b)
&=  \frac {\partial_{b_i} \cdot w_i }n \sum_{j=1}^n L'\bigl(y_j, g(x_j, w,c,a,b)\bigr)\cdot x_j  \\
   \frac{\partial \RD L{g}}{\partial b_i} (w,c,a,b)
&=  \frac {\partial_{b_i} \cdot w_i }n \sum_{j=1}^n L'\bigl(y_j, g(x_j, w,c,a,b)\bigr)   \\
   \frac{\partial \RD L{g}}{\partial w_i} (w,c,a,b)
&=  \frac {\relu{b_i}}n \sum_{j=1}^n L'\bigl(y_j, g(x_j, w,c,a,b)\bigr)  \, .
\end{align*}
Finally, we have 
\begin{align*}
      \frac{\partial \RD L{g}}{\partial c} (w,c,a,b)
&=  \frac 1n \sum_{j=1}^n L'\bigl(y_j, g(x_j, w,c,a,b)\bigr)   \, .
\end{align*}
\end{proposition}

Inspired by  Propositions \ref{result:one-layer-repres}
and Proposition \ref{result:one-layer-deriv} we now introduce the following 
classification for the state of a neuron $h_i$ in \eqref{first-layer}.

\begin{definition}\label{def:neuron-states}
   Let $D=( (x_1,y_1),\dots,(x_n,y_n)) \in (\R\times \R)^n$ be a data set, and
   \begin{displaymath}
    \xmin := \min_{1\leq j\leq n} x_i 
    \qquad \qquad 
    \mbox{ and }
    \qquad\qquad
    \xmax := \max_{1\leq j\leq n} x_i \, ,
   \end{displaymath}
   and
$g\in \arch 1m$ be a function with representation \eqref{first-layer}.
For $i\in I_*$ we then say that the neuron $h_i$ is: 
\begin{enumerate}
   \item \emph{Fully active}, if  
   $\xmin<x_i^* < \xmax$.
	\item \emph{Semi-active}, if $i\in I_-$ and $ x_i^* \geq \xmax> \xmin$, or if 
$i\in I_+$ and $x_i^* \leq \xmin< \xmax$.
\item \emph{Inactive}, if $i\in I_-$ and $x_i^* \leq \xmin$, or
if 
$i\in I_+$ and $x_i^* \geq \xmax$. 
\end{enumerate}
Moreover, if  $L:\R\times \R\to [0,\infty)$ is  a differentiable loss function  and $i\in I_*$, then we say that the neuron $h_i$ 
is \emph{dead}, if $h_i$ is inactive
% $h_i(x_j) = 0$ for all $j=1\dots,n$, 
and for all sub-samples $D'$ of $D$ we have
	\begin{align*}%\label{def:neuron-states-dead}
	    \frac{\partial \RDp L{g}}{\partial a_i} (w,c,a,b) =  \frac{\partial \RDp L{g}}{\partial b_i} (w,c,a,b)  = 0\, .
	\end{align*}

\end{definition}

The following corollary shows that the state of a neuron $h_i$ determines how 
$h_i$ influences the entire function $g$.

% shows that the state of a neuron influences both its 
% contribution to $g$ and the partial derivative of $\RD Lg$ with respect to its parameters. 

\begin{corollary}\label{result:one-layer-states}
  Let   $D=((x_1,y_1),\dots,(x_n,y_n)) \in (\R\times \R)^n$ be a data set
and
$g\in \arch 1m$ be a function with representation \eqref{first-layer}. Then for 
all $i\in I_*$ the following statements are true:
\begin{enumerate}
   \item If $h_i$ is fully active and 
   there exists an $x_{j_0}$ with $\xmin < x_{j_0} < \xmax$,
%    
%    $x_k\neq x_l$ for all $k\neq l$,
 then $h_i$ does not behave linearly on the data set, that 
		is, for all  $\tilde a, \tilde b\in \R$ there exists a $j\in \{1\dots,n\}$ such that 
	\begin{displaymath}
	   h(x_j) \neq \tilde a x_j + \tilde b\, .
	\end{displaymath}
	\item If $h_i$ is semi-active, then $h_i$    behaves linearly on the data set, namely 
	for all $j=1,\dots,n$ we have 
	\begin{displaymath}
	   h_i(x_j) = a_i x_j + b_i\, .
	\end{displaymath}
	\item The neuron $h_i$ is inactive, if and only if  for all $j=1,\dots,n$ we have 
	\begin{displaymath}
	   h_i(x_j) = 0 \, .
	\end{displaymath}
	Moreover, if $h_i$ is inactive and $L:\R\times \R\to [0,\infty)$ is a differentiable loss function, 
	then
	for all sub-samples $D'=((x_{j_1},y_{j_1}),\dots,(x_{j_k},y_{j_k}))$ of $D$ we have
	\begin{align*}
	  \frac{\partial \RDp L{g}}{\partial a_i} (w,c,a,b) 
	  &= \frac {\partial_0\cdot w_i\cdot  x_i^*}n \sum_{l:x_{j_l} = x_i^*} L'\bigl(y_{j_l}, g(x_i^*, w,c,a,b)\bigr)  \\
	  	  \frac{\partial \RDp L{g}}{\partial b_i} (w,c,a,b) 
	  &= \frac {\partial_0\cdot w_i}n \sum_{l:x_{j_l} = x_i^*} L'\bigl(y_{j_l}, g(x_i^*, w,c,a,b)\bigr)\\
			\frac{\partial \RDp L{g}}{\partial w_i} (w,c,a,b) 
			&= 0 	\, .
	\end{align*}
	Consequently, $h_i$ is dead independently of the specific choice of $L$, 
	if $\partial_0=0$ or if $x_j \neq x_i^*$ for all $j=1,\dots,n$.
% 	\item If $h_i$ is dead, then $h_i$ is inactive.
\end{enumerate}    
\end{corollary}

Corollary \ref{result:one-layer-states} shows that, depending on its state, a neuron has 
a rather different impact on the entire network. Indeed, fully active neurons contribute 
in a truly non-linear manner, while semi-active neurons all contribute in a linear fashion.
Once training is completed, all semi-active neurons could therefore be replaced by a 
single  semi-active neuron weighted with new weight $w=1$ and given by
\begin{displaymath}
   h(x) := \Bigl(\sum_{j\in I_{\mathrm{SA}}} w_j a_j\Bigr) \cdot x + \sum_{j\in I_{\mathrm{SA}}} w_j b_j\, ,
\end{displaymath}
where $I_{\mathrm{SA}}$ denotes the set of all indices of semi-active neurons, and where we assume that future 
inputs $x$ satisfy $x\in [\xmin,\xmax]$.
In addition, all inactive neurons do not contribute  to the network, and can therefore 
be removed after training. 
Finally, all dead neurons do not contribute to the network, either, and since the partial 
derivatives of their parameters
vanish, 
any training algorithm that uses 
these derivatives in a gradient-descent-type step will never change the parameters of these 
neurons. 
Consequently, these neurons can be removed during training without changing the final 
decision function $g\in \arch 1m$. 
Finally, note that if $\partial_0=0$, then all inactive neurons are actually dead.
These observations 
raise the following  question:
\begin{description}
   \item[Q1.] How many neurons are semi-active, inactive, or dead due to their initialization?
% 		\item[Q2.] How many neurons become inactive or dead during training? 
% 	\item[Q3.] How many neurons become semi-active during training?
\end{description}

To answer this question, 
% Let us first consider \textbf{Q1}. 
% To this end, 
we write $\lb$ for the Lebesgue measure on $\R$ and
$F_\nu$ for the cumulative distribution function of a given probability measure $\nu$ on $\R$. 
Moreover, 
if $\nu$ is $\lb$-absolutely continuous, then $f_\nu$ denotes a density of $\nu$.

Now, we consider the following 
generic initialization strategy for our simple neural networks $g\in \arch 1m$.

\begin{definition}
   Let $P_a$ and $P_w$ be  probability measures on $\R$ with $P_a(\{0\}) = P_w(\{0\}) = 0$ 
	and $P_b,   P_c$ be  probability measures on $\R$. Then we say that a $g\in \arch 1m$
	with representation \eqref{first-layer} is initialized by $(P_w,P_c, P_a,P_b)$, if 
	the parameter 
	vector $(w,c,a,b)$ is a realization of a random variable with distribution $P:=P_w^m\otimes P_c\otimes P_a^m\otimes P_b^m$.
\end{definition}

Essentially all commonly used initialization methods are of the above type for suitably 
chosen $(P_w,P_c, P_a,P_b)$. We will discuss a few examples after we have investigated 
the generic initialization method. 

Now recall that the state of a  neuron $h_i$ is defined by
  the position of its knot $x_i^* = -b_i/a_i$ relative to the data set $D$.
  This motivates the following definitions.

\begin{definition}
   Let $P$ and $Q$ be two probability measures on $\R$ with $Q(\{0\}) = 0$ and
  $X,Y$ be two independent random variables with $X\sim P$ and $Y\sim Q$.
  Then the ratio distribution $P/Q$ is the probability measure $\mu$ on $\R$ that is given by 
  \begin{displaymath}
      \frac XY \sim \mu\, .
  \end{displaymath}
  Moreover, we define the functions $F_{P,Q}^-:\R\to [0,1]$ and $F_{P,Q}^+:\R\to [0,1]$
  by
  \begin{align*}
   F_{P,Q}^-(z) &:=  P\otimes Q \bigl(\{ (x,y)\in \R^2: x\geq  zy \mbox{ and } y<0   \} \bigr)\\
   F_{P,Q}^+(z) &:=  P\otimes Q \bigl(\{ (x,y)\in \R^2: x\leq zy \mbox{ and } y>0   \} \bigr)\, .
  \end{align*}
\end{definition}

To motivate the functions $F_{P,Q}^\pm$ we consider 
the product measure $P\otimes Q$ on $\R^2$ and the two projections $\pi_X, \pi_Y:\R^2\to \R$
defined by $\pi_X(x,y) := x$ and $\pi_Y(x,y) := y$. Then $\pi_X$ and $\pi_Y$ are
independent random variables  and their distributions are $P$ and $Q$.
Using $Q(\{0\}) = 0$ this leads to 
\begin{align}\nonumber
   F_{P/Q}(z) 
&= P\otimes Q \Bigl( \frac{\pi_X}{\pi_Y} \leq z\Bigr) \\ \nonumber
&= P\otimes Q \bigl( \{ (x,y) \in \R^2 : x\leq zy \mbox{ and } y>0 \} \bigr)
  + P\otimes Q \bigl( \{ (x,y) \in \R^2 : x\geq zy \mbox{ and } y<0 \} \bigr) \\ \label{basic-ratio-equation}
&= F_{P,Q}^+(z) + F_{P,Q}^-(z)
\end{align}
for all $z\in \R$. Moreover, 
 the functions 
$F_{P/Q}$, $F_{P,Q}^-$, and $F_{P,Q}^+$ can be used to 
describe the probability 
 for
a neuron to be initialized into a fully active, semi-active, or inactive state, respectively.
This is done in the following lemma.

\begin{lemma}\label{result:one-layer-distribution-knots}
   Let $g\in \arch 1m$ be initialized by $(P_w,P_c, P_a,P_b)$. 
	Then 
	$P$-almost surely we have $I= I_*$. Moreover,  $-x_i^*$ is, for all $i\in I_*$, 
	  a realization of a random variable with distribution $P_b/P_a$.
	  In particular, if we have a data set $D$, 
	  then for all $i\in I_*$
	  we have  
	 \begin{displaymath} 
	  P\bigl(\{\mbox{neuron }  h_i \mbox{ is fully active}    \} \bigr)
	  = P_b/P_a \bigl( (-\xmax, -\xmin)  \bigr) \, .
	\end{displaymath}  
	Moreover, if $F_{P_b/P_a}$ is  continuous, then the following equations hold:
	\begin{align} \label{result:one-layer-distribution-knots-fa}
	 P\bigl(\{\mbox{neuron }  h_i \mbox{ is fully active}    \} \bigr) &= F_{P_b/P_a}(-\xmin) - F_{P_b/P_a}(-\xmax)\, , \\ \label{result:one-layer-distribution-knots-sa} 
	 P\bigl(\{\mbox{neuron }  h_i \mbox{ is semi-active}    \} \bigr) &= P_a([0,\infty))  + F_{P_b, P_a}^-(- \xmax)  - F_{P_b,P_a}^+(- \xmin)\, ,\\ \label{result:one-layer-distribution-knots-ia}
	 P\bigl(\{\mbox{neuron }  h_i \mbox{ is inactive}    \} \bigr) &= P_a((-\infty, 0]) + F_{P_b,P_a}^+(-\xmax) - F_{P_b,P_a}^-(-\xmin) \, ,
	\end{align}
	and,  in addition, the following equivalence holds $P_b/P_a$-almost surely:
	\begin{displaymath}
	 \mbox{neuron }  h_i \mbox{ is inactive } \qquad \qquad \Longleftrightarrow \qquad \qquad \mbox{neuron }  h_i \mbox{ is dead.}    
	\end{displaymath}
\end{lemma}

Lemma \ref{result:one-layer-distribution-knots} shows that answering  Question \textbf{Q1}
reduces to computing the functions $F_{P/Q}$, $F_{P,Q}^-$, and $F_{P,Q}^+$. Fortunately, ratio distributions
have a rather long history in probability and their 
first systematic treatment can be found in 
\cite{Curtiss41a}. Consequently, computing the probability for neurons being fully active after initialization
can be directly computing using those results. Distinguishing between semi-active and inactive neurons
neurons, however, also requires knowledge about $F_{P,Q}^-$ and $F_{P,Q}^+$. 
For this reason, Proposition \ref{result:ration-distributions} collects 
several useful  results on $F_{P/Q}$ as well as some 
results on $F_{P,Q}^-$ and $F_{P,Q}^+$. 
In particular, it is shown there  $Q$ is Lebesgue absolutely continuous and $P$ is either
also Lebesgue absolutely continuous or a Dirac distribution, then $F_{P/Q}$ is 
continuous, and hence \eqref{result:one-layer-distribution-knots-fa}, \eqref{result:one-layer-distribution-knots-sa}, and \eqref{result:one-layer-distribution-knots-ia} hold.
Moreover, in both cases, simplified formulas for computing 
$F_{P/Q}$, $F_{P,Q}^-$, and $F_{P,Q}^+$ are presented.
Finally, if $Q$ is symmetric, that is $Q(A) = Q(-A)$ for all measurable $A\subset \R$,
then $P/Q$ is symmetric, too.

The next theorem, which relies on both Lemma \ref{result:one-layer-distribution-knots}
and Proposition  \ref{result:ration-distributions}, characterizes distributions $P_b$
that prevent either  inactive neurons or semi-active neurons during initialization.

% 
% 
% With the help of Proposition \ref{result:ration-distributions}, we can now establish the following theorem,
% which explains, why, for the networks we consider in this section,  it is better to initialize 
% the biases of the hidden layer with non-negative (random) values.

\begin{theorem}\label{result:pos-bias-better}
    Let $g\in \arch 1m$ be initialized by $(P_w,P_c, P_a,P_b)$ and assume that 
    $F_{P_b/P_a}$ is  continuous and that $P_a((-\e,\e))>0$ holds for all $\e>0$. 
    Moreover, let $D$ be a data set with $\xmin \leq 0\leq \xmax$.
    Then the following statements are equivalent:
    \begin{enumerate}
     \item $P_b$ only assigns positive values, that is  $P_b((0,\infty)) = 1$.
     \item For all $i\in I$ we have $P (\{\mbox{neuron }  h_i \mbox{ is inactive}    \} ) = 0$. 
    \end{enumerate}
    In addition, we also have the equivalence of the following two statements:
    \begin{enumerate}
    \setItemnumber{3}
     \item $P_b$ only assigns negative values, that is  $P_b((-\infty, 0)) = 1$.
     \item For all $i\in I$ we have $P (\{\mbox{neuron }  h_i \mbox{ is semi-active}    \} ) = 0$. 
    \end{enumerate}
\end{theorem}

Note that without the 
continuity of $F_{P_b/P_a}$ Theorem \ref{result:pos-bias-better} does not hold in general.
In particular, if $P_b$ is the Dirac measure at zero, that is 
$P_b = \d_{\{0\}}$, then $P_b / P_a = \d_{\{0\}}$, and hence 
we have $x_i^* = 0$ almost surely. For data sets with  $\xmin < 0< \xmax$, 
all neurons are therefore fully active after initialization.

For other commonly used distributions, such as $P_b = \d_{\{0.01\}}$, 
$P_{b} = \ca U[\a,\b]$, or $P_{b} = \ca N(\mu, \s_b^2)$, and 
$P_{a} = \ca U[-\g,\g]$ or $P_{a} = \ca N(0, \s_a^2)$, however, 
the assumptions of Theorem \ref{result:pos-bias-better}  are satisfied. 
In this case, 
Theorem \ref{result:pos-bias-better} shows that the only way to prevent 
inactive neurons during initialization is to enforce 
strictly positive biases by $P_b$. For such $P_b$, however,
Theorem \ref{result:pos-bias-better} further shows  that the initialization
necessarily produces  some semi-active neurons. By
combining both equivalences of 
Theorem \ref{result:pos-bias-better} we thus find
\begin{align}\label{not-all-fully-active}
   P \bigl(\{\mbox{neuron }  h_i \mbox{ is fully active}    \} \bigr) < 1\, .
\end{align}
However, this result requires, as already mentioned, the continuity of 
$F_{P_b/P_a}$. The next theorem in particular shows that 
for data sets with $\xmin < 0< \xmax$, Inequality 
\eqref{not-all-fully-active} actually
holds for all $P_b \neq \d_{\{0\}}$ and all commonly used $P_a$.
% 
% 
% The next theorem complements these considerations by showing that
%  the Dirac measure 
% $\d_{\{0\}}$ is the only distribution for $P_b$ that guarantees all knots $x_i^*$ to 
% be contained in $[\xmin, \xmax]$, provided that  $\xmin \leq 0\leq \xmax$.

\begin{theorem}\label{result:nonzero-bias}
       Let $g\in \arch 1m$ be initialized by $(P_w,P_c, P_a,P_b)$ and assume that 
     $P_a((-\e,0))>0$ and $P_a((0,\e))>0$ hold for all $\e>0$. 
    Moreover, let $D$ be a data set with $\xmin \leq 0\leq \xmax$.
    Then for all $i\in I$ the following statements are equivalent:
    \begin{enumerate}
       \item $P_b(\{0\}) < 1$.
				\item $P(\{x_i^* > \xmax  \}) > 0$.
				\item $P(\{x_i^* < \xmin  \}) > 0$.
    \end{enumerate}
    Moreover, if the data set $D$ satisfies $\xmin < 0<  \xmax$, then these conditions are also equivalent to:
    \begin{enumerate}
    \setItemnumber{4}
    \item $P(\{x_i^* \geq \xmax  \}) > 0$.
  \item $P(\{x_i^* \leq \xmin  \}) > 0$.
    \end{enumerate}
\end{theorem}

\begin{figure}[t]
\begin{center}
\includegraphics[width=0.32\textwidth]{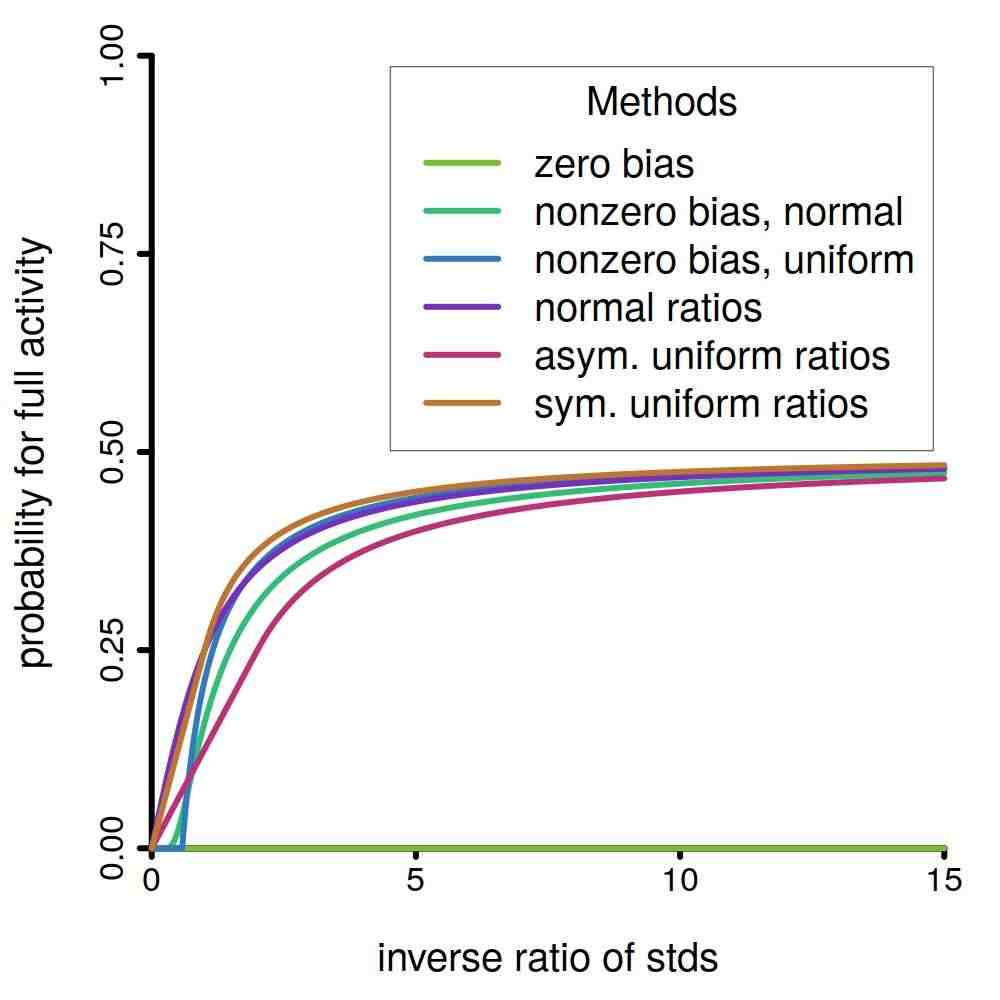}
\hspace*{-0.01\textwidth}
\includegraphics[width=0.32\textwidth]{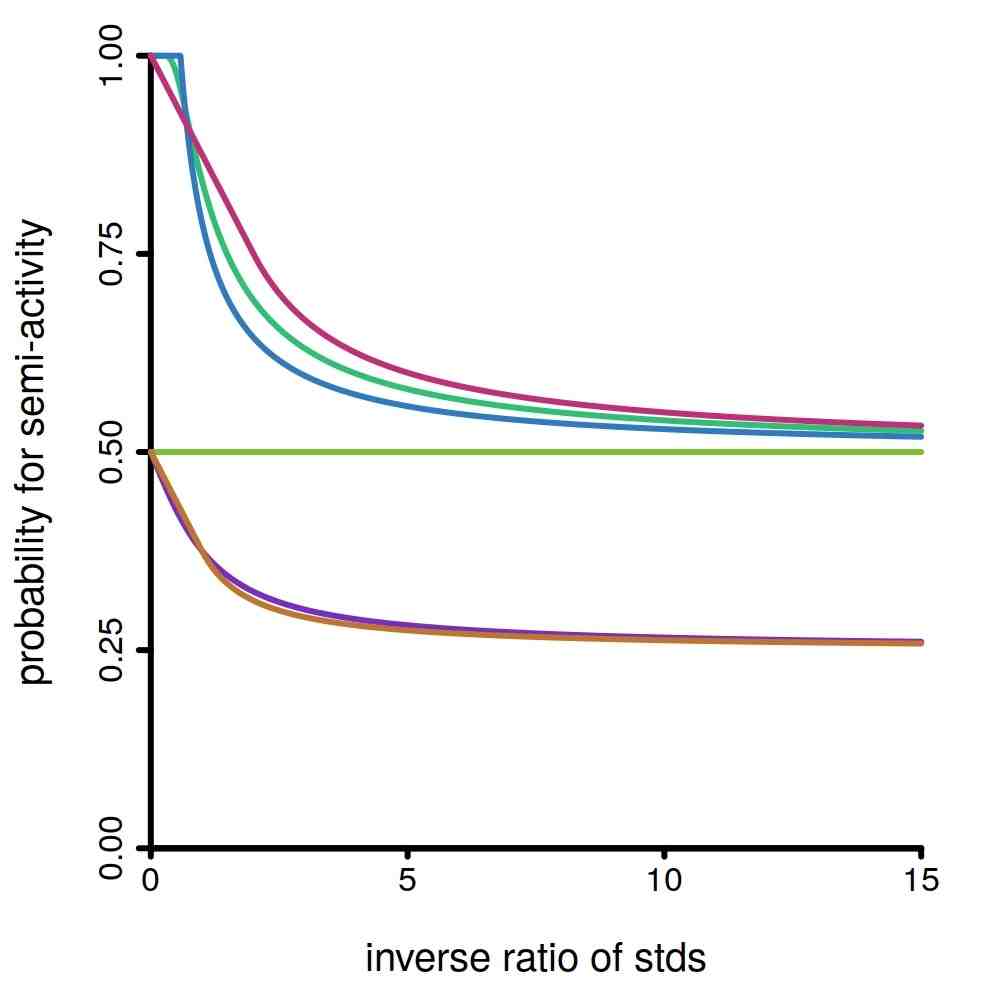}
\hspace*{-0.01\textwidth}
\includegraphics[width=0.32\textwidth]{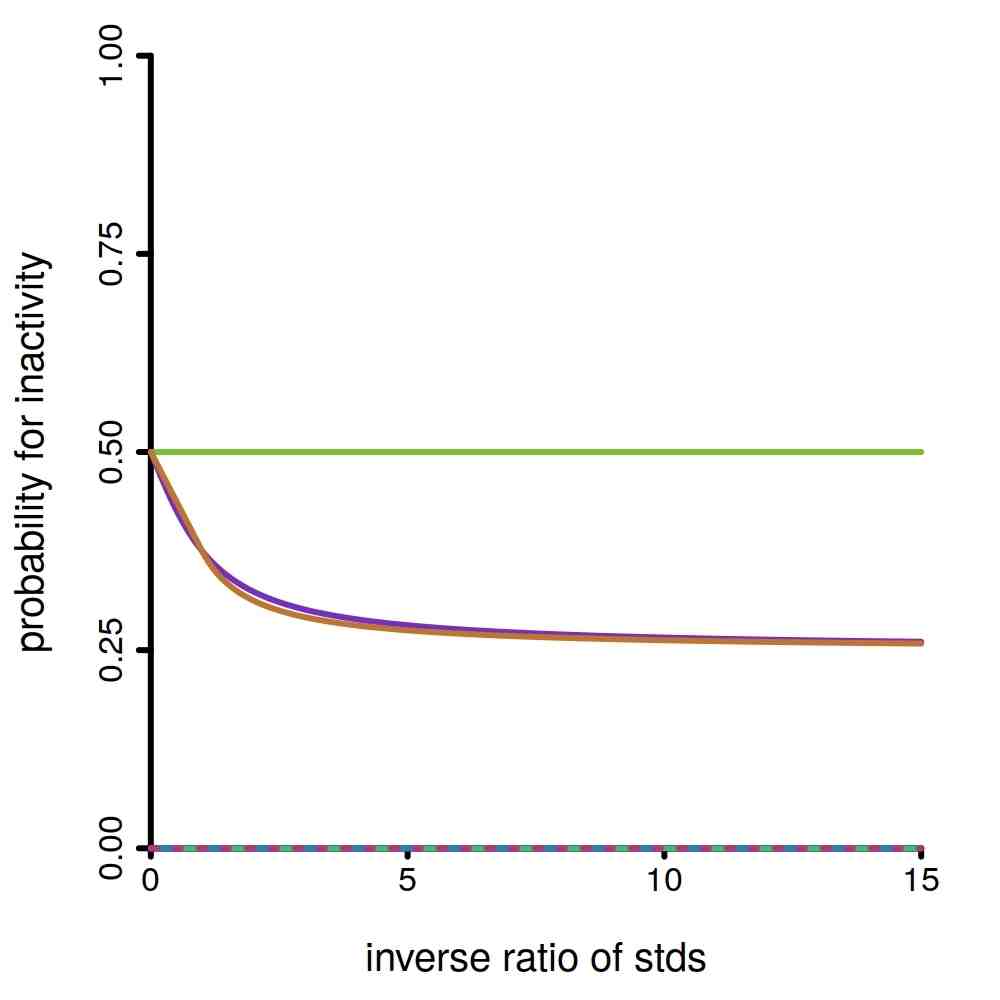}
\vspace*{-6ex}
\end{center}
\caption{Probability for a knot being  fully active  (left), semi-active (middle), and  inactive (right) for 
``inverse ratio of standard deviations'' 
$\r\in (0,15]$ and data sets 
with $[\xmin, \xmax] = [0,1]$.
Six different initialization methods, which are discussed in Example \ref{example:He-init-1d} (``zero bias''), 
Example  \ref{example:init-non-zero} (``nonzero bias \dots''), and 
Example \ref{example:random-init} (``\dots ratios''), are displayed. For each method, the probability of initializing a fully 
active neuron is bounded from above by $0.5$, and for typical choices of $\rho$ this upper bound
is actually almost attained. Moreover,  the probability of 
initializing an inactive neuron is either approximately $0.25$ or equal to $0$. As shown in 
Theorem \ref{result:pos-bias-better} the latter case occurs  exactly for those distributions $P_b$, which
only produce positive values for the bias.}\label{figure:prob-active-knots-1d}
\end{figure}

\begin{figure}[t]
\begin{center}
\includegraphics[width=0.32\textwidth]{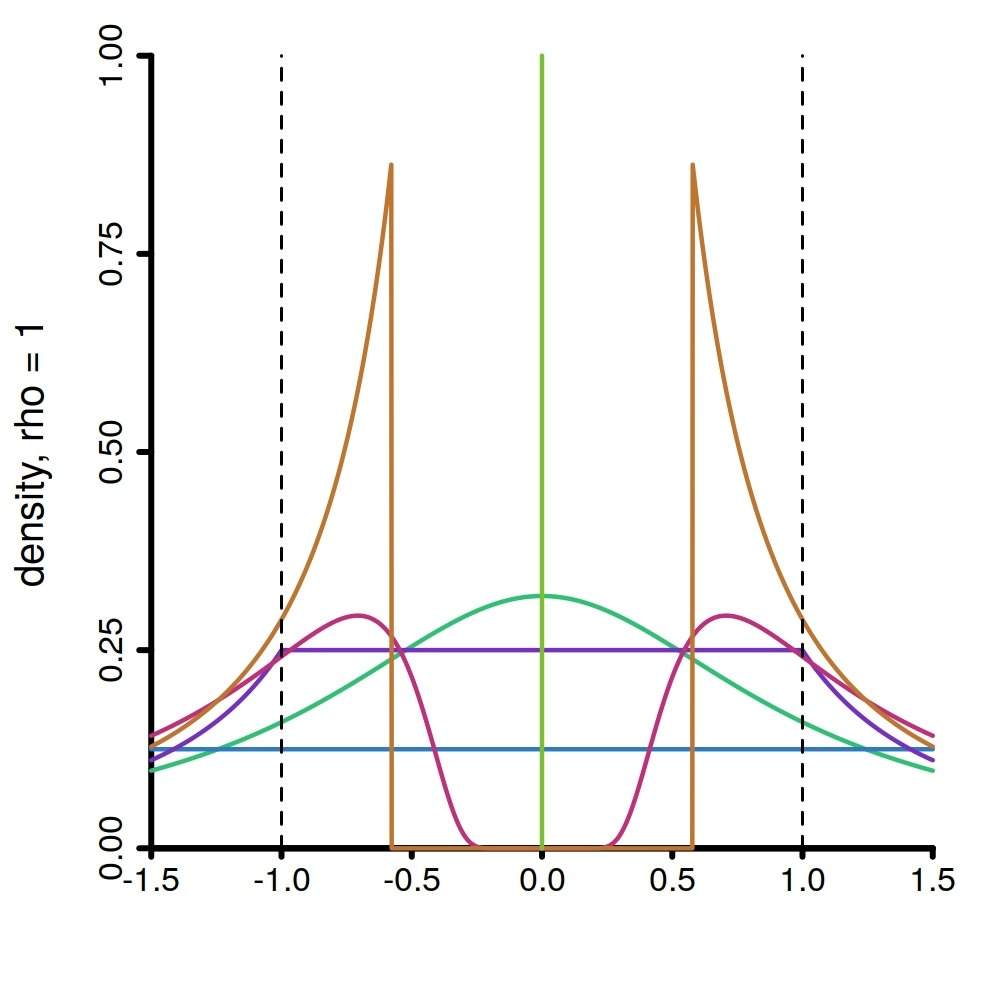}
\hspace*{-0.01\textwidth}
\includegraphics[width=0.32\textwidth]{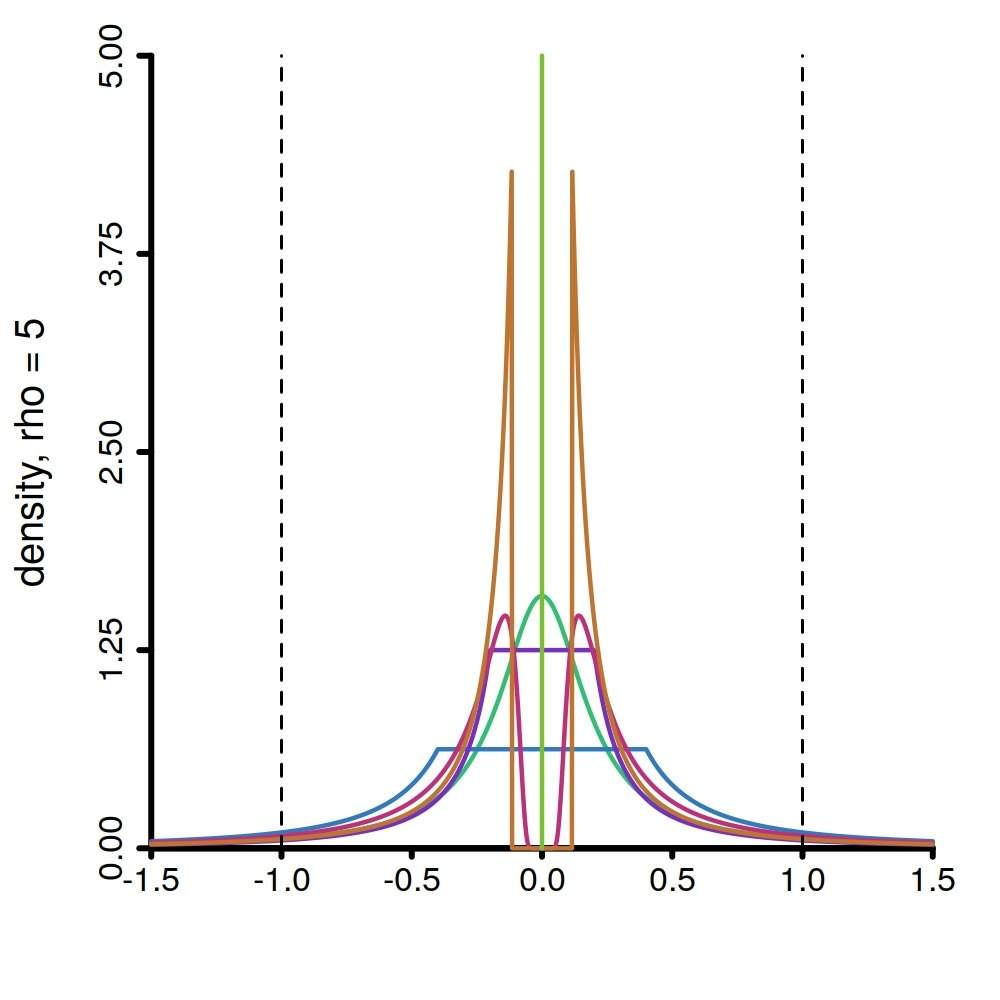}
\hspace*{-0.01\textwidth}
\includegraphics[width=0.32\textwidth]{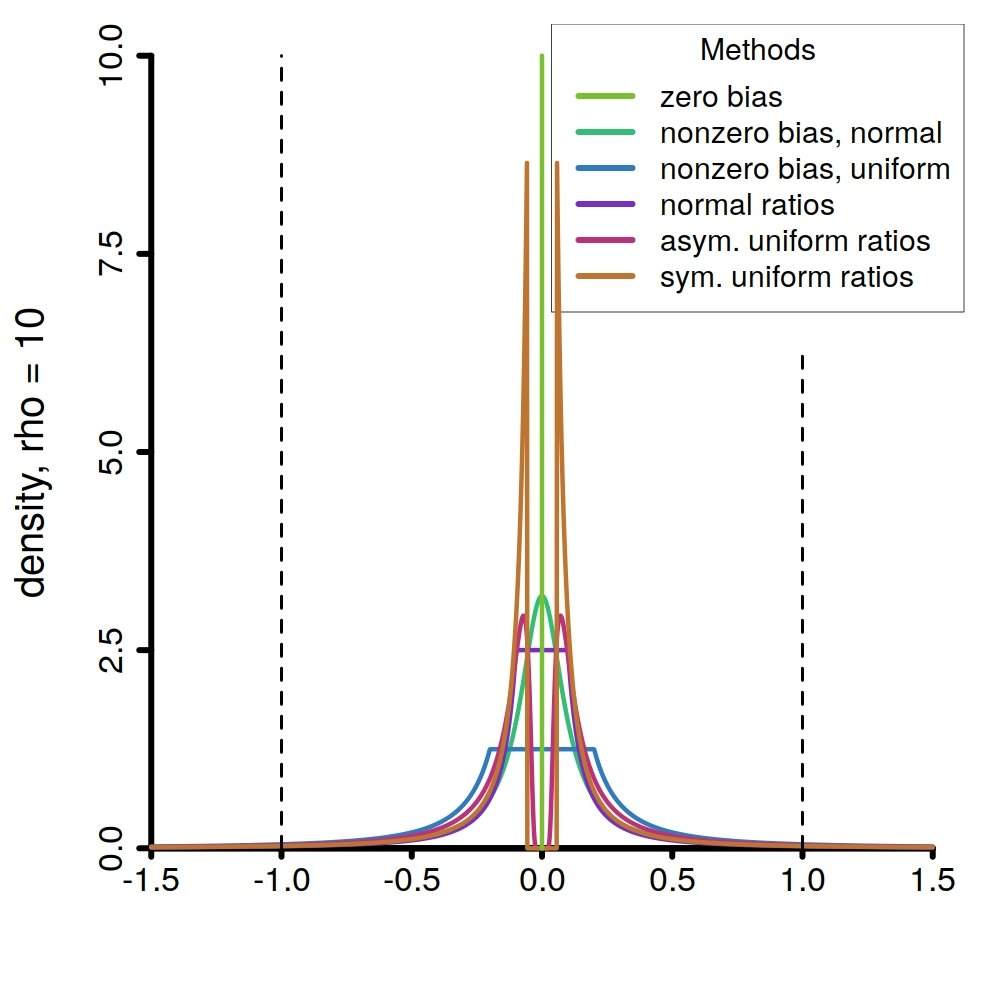}
\vspace*{-6ex}
\end{center}
\caption{Densities $f_{P_b/P_a}$  of the knot distributions for 5 different 
initialization strategies considered in Examples  
\ref{example:init-non-zero} and \ref{example:random-init} and the 
initialization with $P_b = \delta_{\{0\}}$ 
of Example \ref{example:He-init-1d}, which is 
indicated by a vertical line at $x=0$.
% 
% is not plotted
% since the resulting distribution $P_b/P_a$ is, in any case, the Dirac measure 
% $\delta_{\{0\}}$.
Left to Right: 3 inverse ratios of standard deviations $\rho = 1$, $5$, and $10$ as defined in the examples.
For fixed $P_b$, distributions $P_a$
with larger variance lead to larger $\rho$ and, as the graphics show,
to 
a higher concentration of $P_a/P_b$ around zero. All densities are 
symmetric, and hence at least have of the initialized knots fall outside  $[\xmin, \xmax] = [0,1]$. In addition,
all densities have a fat tail and are far from being uniform on $[\xmin, \xmax] = [0,1]$.
}\label{figure:densities-1d}
\end{figure}

For the most commonly used distributions for $P_a$ and $P_b$, the ratio distribution
as well as the functions $f_{P/Q}$, $F_{P/Q}$, $F_{P,Q}^-$, and $F_{P,Q}^+$ can be explicitly 
derived, see Examples, \ref{example:normal-ratio}, \ref{example:normal-denominator},
\ref{example:asym-uniform-ratio}, \ref{example:sym-uniform-ratio}, and \ref{example:uniform-denominator}.
Consequently, the probabilities for initializing  fully active, semi-active, and inactive 
neurons can be explicitly with the help of Lemma \ref{result:one-layer-distribution-knots}.
This is the goal of the next couple of examples, see also 
Figure \ref{figure:prob-active-knots-1d} for the probabilities of not fully active and inactive knots
and
Figure \ref{figure:densities-1d} for the densities $f_{P_b/P_a}$ of the knot distributions.
In these examples, we restrict our considerations
to data sets with $\xmin = 0$ and $\xmax = 1$, since \emph{a)} this describes one of the two most 
commonly used data pre-scalings, and \emph{b)} the non-negativity of $\xmin$ will also play a key
role when considering hidden layers in the middle of deeper neural networks in Section \ref{sec:general}.
In addition, considering 
 the other commonly used data pre-scaling 
$\xmin = -1$ and $\xmax = 1$ in the examples below is merely more than a straight forward exercise.

The first two examples consider initialization strategies that assign constant values to the bias.
These strategies are probably the most commonly used ones.\footnote{For example, \cite[p.~302]{GoBeCo16},
writes \emph{``Typically, we set the biases for each unit to heuristically chosen constants, and
initialize only the weights randomly.''}}

% 
% 
% 
% 
% The most commonly used distributions for initializing neural networks include uniform, normal, and Dirac distributions.
% The following examples provide, with the help of Proposition \ref{result:ration-distributions}, 
% the crucial quantities for corresponding ratio distributions.

\begin{example}[Zero bias initialization]\label{example:He-init-1d}
  In recent years, the importance of proper initialization of neural networks 
  and in particular of their weight parameters has been observed in e.g.~\cite{GlBe10a}
  and \cite{HeZhReSu15a}. To be more precise, in \cite{GlBe10a} it was proposed to initialize 
  the weights of the $l$-th layer 
  with the help
  of the following distributions $P_w :=  \ca U[-\a,\a ]$ with $\a = \sqrt{6/(m_l + m_{l-1})}$,
  where $m_{l-1}$ denotes the number of neurons in layer   $l-1$. Moreover, all 
  bias entries are initialized using 
  $\d_{\{0\}}$. This initialization method is known as \emph{Xavier} and is strictly speaking not for 
  ReLUs. For this reason,  \cite{HeZhReSu15a} adapted the insights of \cite{GlBe10a} to ReLU-Functions.
  To be more precise,   \cite{HeZhReSu15a}  proposes to initialize 
  the  weight entries of the $l$-th layer 
  using a symmetric distribution  whose variance is $2/m_{l-1}$. An
  explicitly mentioned example of such a distribution is 
   $\ca N(0, \s_l^2)$ with $\s_l = \sqrt{2/m_{l-1}}$.
  Moreover, \cite{HeZhReSu15a} again proposed to use $\d_{\{0\}}$  for  all 
  bias entries. This initialization method is known as \heetal.
%   and is probably the most popular 
%   initialization strategy for neural networks using ReLUs.

  Let us now analyze the effect of this and similar initialization methods. To this end, we 
  we consider a $g\in \arch 1m$, and assume that $P_b = P_c = \d_{\{0\}}$ and that $P_a$ and $P_w$ are some 
  Lebesgue-absolutely continuous, symmetric distributions.
  Then we have  $P_b/P_a = \d_{\{0\}}$, and therefore 
  the initialization almost surely yields
    $x_i^* = 0$  for all $i=1,\dots,m$.
  Moreover, we have  $P_w(\{0\}) = 0$ and 
  consequently, independent of the number of neurons $m$, our initialized $g$ has almost surely exactly one knot,
  which is located at $0$. Our next goal is to investigate the states of the neurons after initialization.

%   Let us first assume that $[\xmin, \xmax]=[-1,1]$. Then it is easy
%   to see that all neurons are fully active.

  To this end we assume that our data set $D$ is normalized such that it satisfies $[\xmin, \xmax]=[0,1]$.
  Since $x_i^* = 0$, we then see that 
    each neuron is either semi-active or inactive, and therefore Corollary \ref{result:one-layer-states}
  shows that 
%   that is
  for all $i=1,\dots,m$ we either have 
    $h_i(x_j) = 0$ for all $j=1,\dots,n$ or 
  $h_i(x_j) = a_i x_j + b_i$ for all $j=1,\dots,n$. Moreover, a neuron $h_i$ is semi-active if and only if $i\in I_+$,
  and it is inactive if and only if $i\in I_-$. 
  By the symmetry of $P_a$ we then find 
  \begin{displaymath}
   P\bigl(\{\mbox{neuron }  h_i \mbox{ is semi-active}    \} \bigr) = P\bigl(\{\mbox{neuron }  h_i \mbox{ is inactive}    \} \bigr) = 0.5\, .
  \end{displaymath}
% 
%   
%   
%   In addition, since $P_a$ is symmetric, we have 
%   \begin{displaymath}
%       P_a\bigl(\{a_i \in \R: a_i < 0\}\bigr) =  P_a\bigl(\{a_i \in \R: a_i > 0\}\bigr) = 0.5\, .  
%   \end{displaymath}
  Let us now consider an inactive neuron $h_i$, that is $i\in I_-$.
  For the most commonly used choice
  $\partial_0=0$,  part \emph{iii}) of Corollary \ref{result:one-layer-states} then shows that $h_i$ is dead.
  Therefore, the probability of $h_i$ being initialized into a dead state is $0.5$ and the 
  total number $|I_{\mathrm{dead}}|$ of neurons that are initialized as dead is a 
  random variable with
  \begin{align*}%\label{dead-neuron-distribution}
      |I_{\mathrm{dead}}| \sim B(m, 0.5)\, .
  \end{align*}
  Let us now consider the case 
   $\partial_0 >0$. To this end, we first observe that for a sub-sample  $D'=((x_{j_1},y_{j_1}),\dots,(x_{j_k},y_{j_k}))$
   of $D$,   Corollary \ref{result:one-layer-states} and Proposition \ref{result:one-layer-deriv} show
	  \begin{align}\nonumber
  \frac{\partial \RDp L{g}}{\partial a_i} (w,c,a,0) 
  = \frac{\partial \RDp L{g}}{\partial w_i} (w,c,a,0) 
	  & =0 \\ \label{He-h1}
	  \frac{\partial \RDp L{g}}{\partial b_i} (w,c,a,0) 
  & = \frac {\partial_0\cdot w_i}n \sum_{l:x_{j_l} = 0} L'\bigl(y_{j_l}, c\bigr) \\ \nonumber
  \frac{\partial \RDp L{g}}{\partial c} (w,c,a,0)
&=  \frac 1n \sum_{l=1}^k L'\bigl(y_{j_l}, g(x_{j_l}, w,c,a,0)\bigr) \, ,
  \end{align}
%   and 
% 	  \begin{align*}
%     \frac{\partial \RDp L{g}}{\partial b_i} (w,c,a,b) 
%   = \frac {\partial_0\cdot w_i}n \sum_{x_{j_l} = 0} L'\bigl(y_{j_l}, 0\bigr)\, ,
%   \end{align*}
  where  we used $b_i = x_i^* = 0$ and $g(x_i^*, w,c,a,0) = g(0, w,c,a,0) = c$.
  Note that our initialization actually ensures $c=0$ but for the arguments below, we actually need
  general $c\in \R$.
  Let us now consider a gradient-descent type algorithm that uses a sub-sample $D'$ of $D$.
  In the case
  \begin{align}\label{He-init-subsample-gradient}
  \sum_{l:x_{j_l} = 0} L'(y_{j_l}, c) = 0\, ,
  \end{align}
  this algorithm does not change the values of $a_i$, $b_i$ and $w_i$, and hence the knots $x_i^*$
  are not changed, either. Note that \eqref{He-init-subsample-gradient} in particular holds,
  whenever 
  the sub-sample $D'$ does not contain a sample $x_{j_l}=0$.
%   , then \eqref{He-init-subsample-gradient}
%   is satisfied and therefore $a_i$, $b_i$ and $w_i$ are not changed.
%   Moreover, if \eqref{He-init-subsample-gradient} actually holds for all sub-samples $D'$, then 
%   all inactive neurons are dead and we  again get \eqref{dead-neuron-distribution} right after 
%   initialization. 
  Therefore let us now consider the first iteration of the training algorithm 
  that uses 
  sub-sample $D'$ for which \eqref{He-init-subsample-gradient} does not hold. 
  Clearly, such a $D'$ needs to contain a sample $x_{j_l} = 0$.
  Our previous considerations then show that $a_i$, $b_i$ and $w_i$ have not been changed 
  since their initialization. 
  Without loss of generality we may thus assume that 
  we are in the first iteration of the algorithm with $c$ having some arbitrary value.
  Then \eqref{He-h1}  together with the symmetry of the distribution $P_w$ and $P = P_w^m\otimes P_c\otimes P_a^m\otimes P_b^m$
  shows that 
  \begin{displaymath}
    P\biggl(\Bigl\{  (w,c,a,b):    \frac{\partial \RDp L{g}}{\partial b_i} (w,c,a,0)  < 0      \Bigr\}\biggr) = 0.5\, .
  \end{displaymath}
    Since $P_a$ is also symmetric we conclude that 
%   
%   
%   since $a_i$ and $w_i$ are independently sampled from  symmetric distributions $P_a$ and $P_w$,
% %   and $a_i$ is sampled independently of $w_i$,
%    we then see by \eqref{He-h1}  that 
  \begin{align*}
  0.25&=
  P\biggl(\Bigl\{  (w,c,a,b): a_i<0 \mbox{ and }  \frac{\partial \RDp L{g}}{\partial b_i} (w,c,a,0)  < 0      \Bigr\}\biggr) \\
  &=
  P\biggl(\Bigl\{  (w,c,a,b): a_i<0 \mbox{ and }  \frac{\partial \RDp L{g}}{\partial b_i} (w,c,a,0)  > 0      \Bigr\}\biggr)\, .
  \end{align*}
  In the case $a_i < 0$ and $\frac{\partial \RDp L{g}}{\partial b_i} (w,c,a,0)  > 0$, our gradient-descent-type
  algorithm will keep the values of $a_i$ and $w_i$ 
  by Proposition \ref{result:one-layer-deriv} since we still have $x_i^* = 0$.
  Moreover, it 
    will
  update $b_i$ to some negative value $b_i^{\mathrm{new}}$. Therefore we find 
  $x_i^* = - b_i^{\mathrm{new}}/a_i < 0$ after this update. Since all samples satisfy $x_j\geq 0$, we conclude by 
  part \emph{iii)} of
  Corollary \ref{result:one-layer-states} that 
  $h_i$ is dead after the update. 
  Similarly, 
  $a_i < 0$ and $\frac{\partial \RDp L{g}}{\partial b_i} (w,c,a,0)  < 0$, then the update yields $x_i^*>0$ and therefore
  the neuron is either semi-active or fully active. 
  The latter case occurs if the learning rate has been taken sufficiently small, and in the following considerations
  we only treat this ``optimistic'' case. 
  Furthermore, the two analogous sub-cases of $a_i>0$ can be treated similarly, showing that we obtain a 
  semi-active neuron if $\frac{\partial \RDp L{g}}{\partial b_i} (w,c,a,0)  < 0$, and, following our optimistic view,
  a fully active neuron in the remaining case.

%   Finally, for a neuron that has been initialized with $a_i>0$, no simple prediction on its state after the first iteration can be made.
  
  Summing up, if $\partial_0 >0$ and the learning rate is sufficiently small, 
  for each neuron the probabilities of being dead  or semi-active after the first 
  iteration, in which $x_i^*$ is changed, are  $0.25$ each, while the probability of having a fully active neuron is $0.5$.
\end{example}

\begin{figure}[t]
\begin{center}
\includegraphics[width=0.32\textwidth]{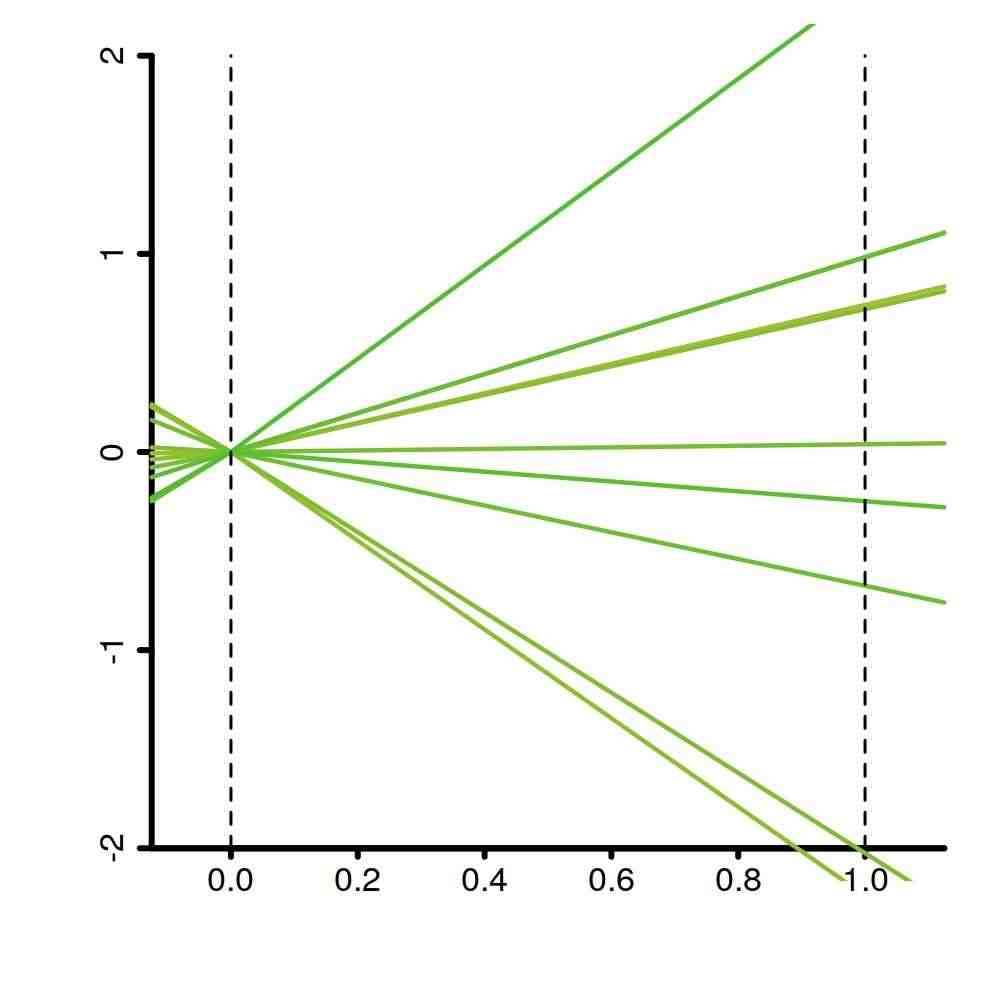}
\hspace*{-0.01\textwidth}
\includegraphics[width=0.32\textwidth]{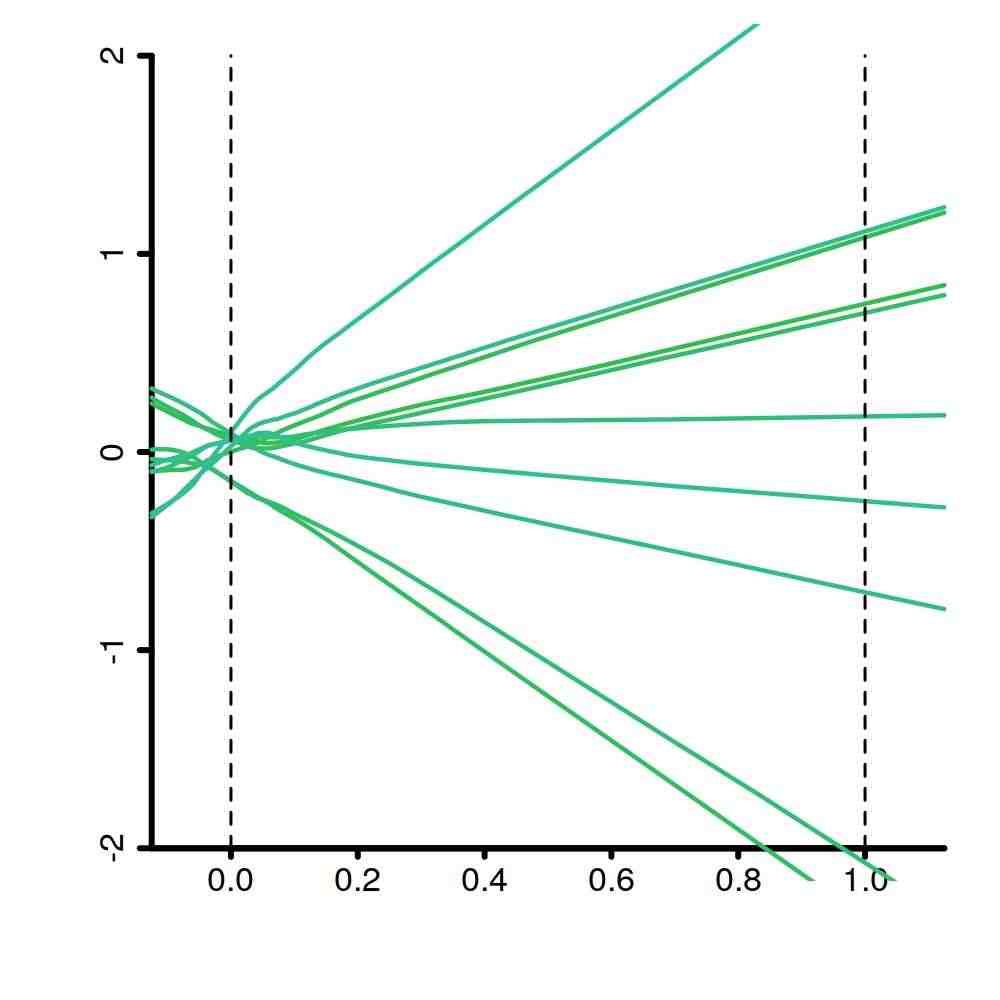}
\hspace*{-0.01\textwidth}
\includegraphics[width=0.32\textwidth]{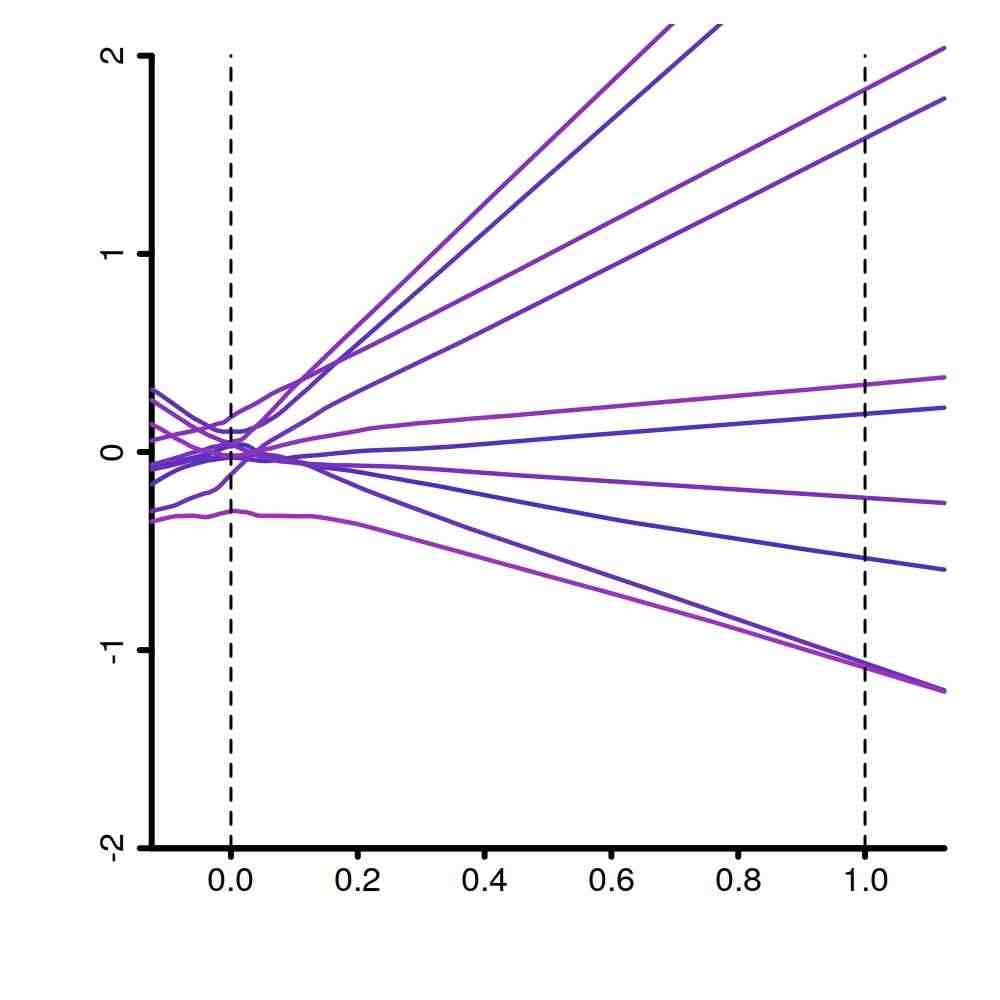}
% \hspace*{0.1\textwidth}
% \includegraphics[width=0.4\textwidth]{./images/init_funct_uniform_denom.jpg}
\vspace*{-6ex}
\end{center}
\caption{Ten randomly initialized predictors of a neural network with 128 hidden neurons
with weights initialized by a zero-mean normal distribution with variance
according to \cite{HeZhReSu15a}. In each case, we set $c=0$.
Left: zero bias initialization, i.e.~$b= 0$.
Middle: nonzero bias initialization with $b= 0.1$, which   leads to  $\rho \approx 14.1$.
Right: $b$ is initialized by $\ca N(0, \s_b^2)$ with $\s_b = 0.1$, which again results in $\rho \approx 14.1$.
As discussed in Example \ref{example:He-init-1d}, the zero bias initialization lead to a linear behavior 
on $[0,1]$, while the other two initialization methods only lead to an ``almost'' linear behavior on the right-hand side
of the interval.
}\label{figure:random-functions-1d}
\end{figure}

\begin{figure}[t]
\begin{center}
\includegraphics[width=0.32\textwidth]{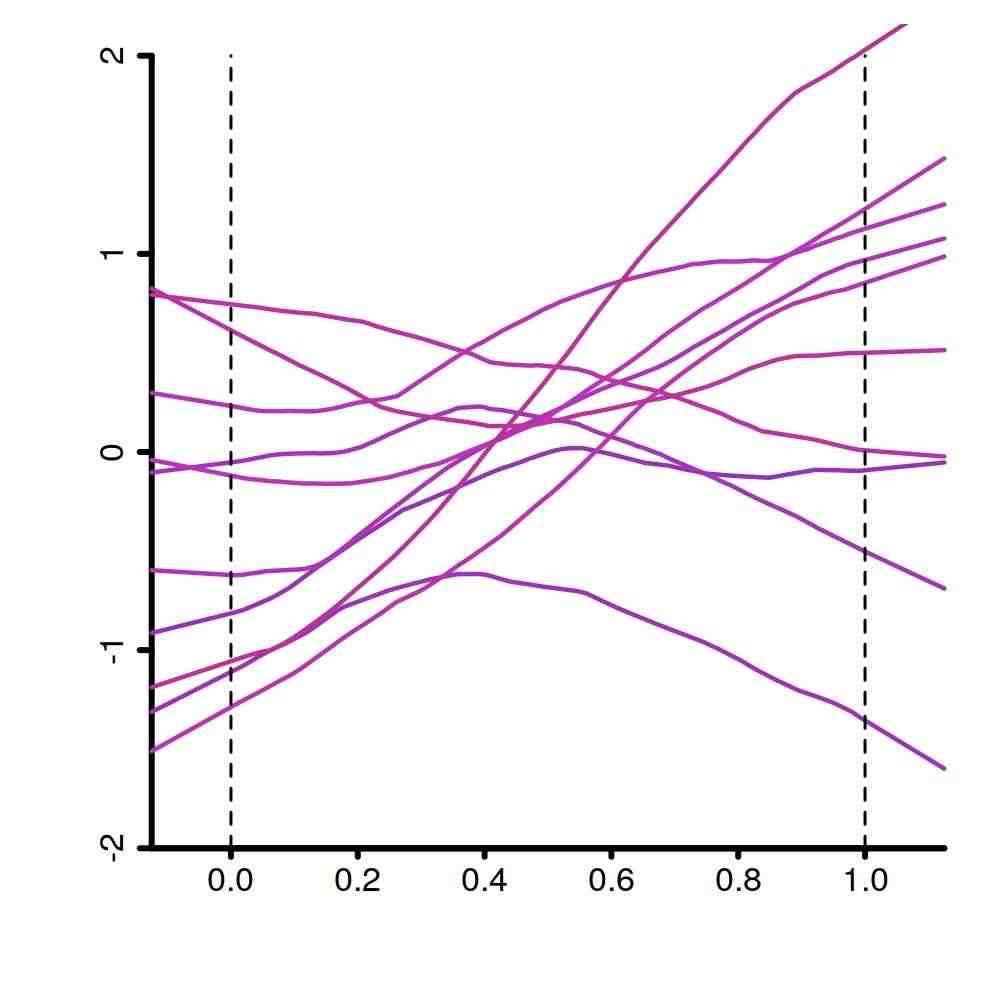}
\hspace*{-0.01\textwidth}
\includegraphics[width=0.32\textwidth]{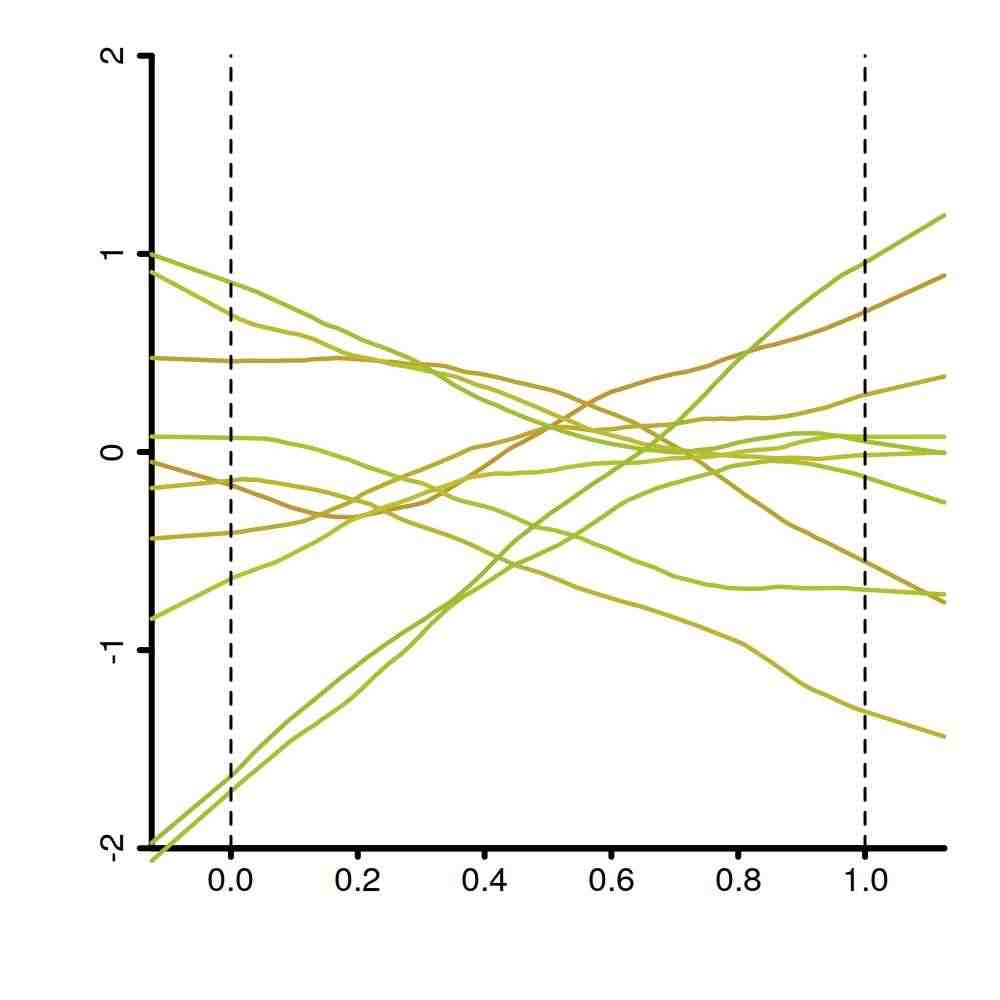}
\hspace*{-0.01\textwidth}
\includegraphics[width=0.32\textwidth]{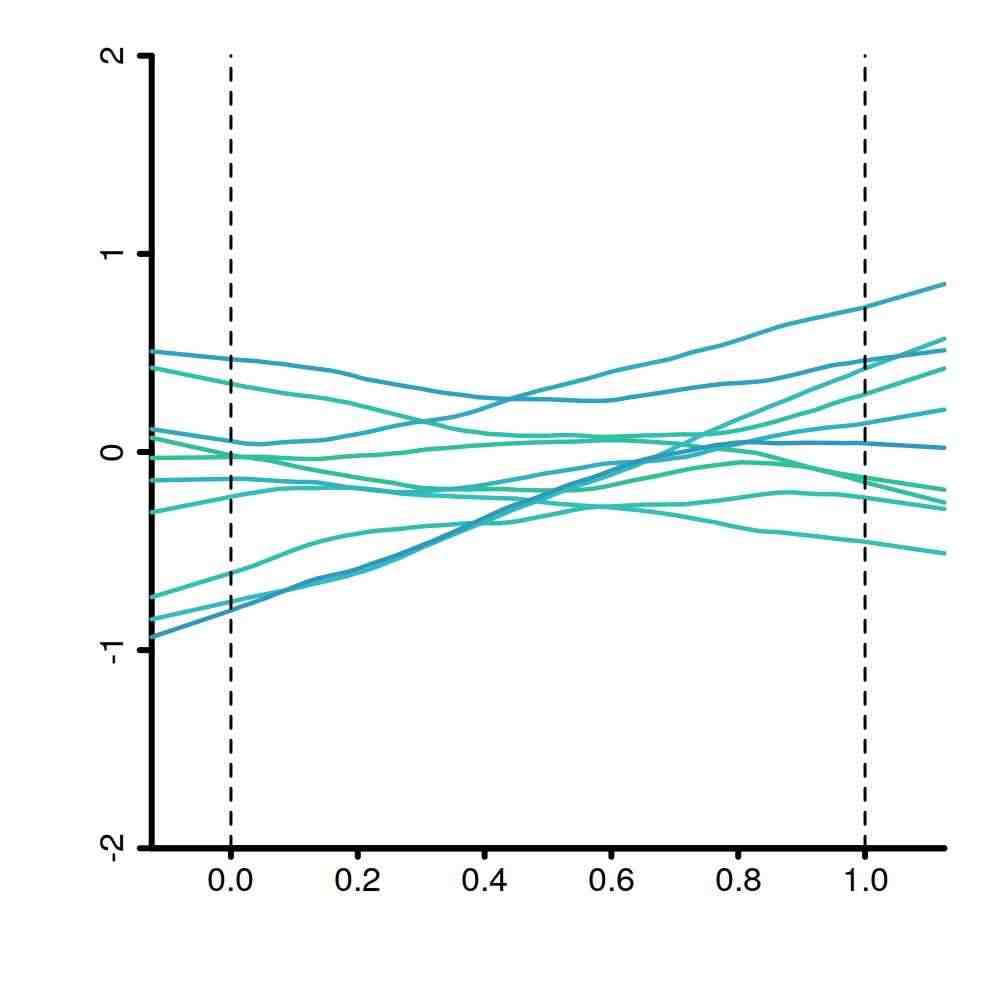}
\vspace*{-6ex}
\end{center}
\caption{Ten randomly initialized predictors of a neural network with 128 hidden neurons. In all three cases,
the knots $x_i^*\in [0,1]$ are  sampled from $\ca U[0,1]$ and the weights $a_i$ and $b_i$ are sampled from   symmetric
distributions. The biases are then set to $b_i = -a_i x_i^*$ and $c=0$.
Left: Uniform weight distributions with variance according \heetal.  
Middle: Normal weight distributions with variance according to \heetal.
Right: Weight distributions, which ensure $|a_i| = \snorm w_2 = 1$.}\label{figure:random-functions-own}
\end{figure}

\begin{example}[Non-zero bias]\label{example:init-non-zero}
      Initializing the weights according to  \cite{HeZhReSu15a}  seems to be one of the most common strategies.
 	Sometimes, however, the bias is initialized differently by $P_b :=\d_{\{b\}}$ for some small $b>0$.
 	For example, \cite{KrSuHi12a} uses $b=0.01$, and 
 	and 
 	\cite[p.~192]{GoBeCo16} discusses $b=0.1$.
	 Let us now investigate the consequences 
	of this initialization method. To this end, we assume that we have fixed an arbitrary  $b>0$ and $P_b :=\d_{\{b\}}$.
	
	Let us first consider the case $P_a := \ca N(0,\s_a^2)$, where $\s_a$ can, e.g.~be
	initialized according to \cite{HeZhReSu15a}. 
	Moreover, we write $\varrho := \s_a / b$ for the ``inverse ratio of standard deviations'', where for 
	$P_b$ we used the standard deviation of its symmetrized version $\frac12 (\d_{\{b\}} + \d_{\{-b\}})$. 
	Note that for the method proposed by \cite{HeZhReSu15a}, we have $\s_a = \sqrt{2}$ and hence 
	$b=0.1$ leads to $\rho \approx 14.1$ and $b=0.01$ leads to $\rho \approx 141$. 
	By Example \ref{example:normal-denominator} the distribution of each knot $x_i^*$ has the Lebesgue density
	\begin{displaymath}
	  f_{P_b/P_a}(z)  
	  =
	  \frac{1}{\sqrt{2\pi}\,\varrho  z^2} \exp\Bigl(-\frac{1}{2\varrho^2 z^2}\Bigr) \, , \qquad \qquad z\in \R.
	\end{displaymath}
	and Figure \ref{figure:densities-1d} indicates  that for  $b=0.1$  and $b=0.01$ 
	the corresponding distributions are highly concentrated around 0.
	Furthermore, 
	 Example \ref{example:normal-denominator} also provides  the 
	functions $F_{P_b/P_a}$, and $F_{P_b/P_a}^+$. 
	For a   data set with $[\xmin, \xmax] = [0,1]$,  Lemma \ref{result:one-layer-distribution-knots} 
	and Theorem \ref{result:pos-bias-better}
	then
	give
% % % 	\begin{align*}
% % % 	  P\bigl(\{\mbox{neuron }  h_i \mbox{ is fully active}    \} \bigr) &=  2\Phi\Bigl( - \frac 1 {\varrho} \Bigr)  \\
% % % 	    P\bigl(\{\mbox{neuron }  h_i \mbox{ is semi-active}    \} \bigr) &=  1 - 2\Phi\Bigl( - \frac 1 {\varrho} \Bigr)  \\
% % % 	 P\bigl(\{\mbox{neuron }  h_i \mbox{ is inactive}    \} \bigr) &= 0 \, ,
% % % 	\end{align*}
% % % 	while for a data set with $[\xmin, \xmax] = [0,1]$ they lead to
	\begin{align*}
	 P\bigl(\{\mbox{neuron }  h_i \mbox{ is fully active}    \} \bigr) &= \Phi\Bigl( - \frac 1 {\varrho} \Bigr)\, , \\ 
	 P\bigl(\{\mbox{neuron }  h_i \mbox{ is semi-active}    \} \bigr) &=  1 - \Phi\Bigl( - \frac 1 {\varrho} \Bigr)\, ,\\
	 P\bigl(\{\mbox{neuron }  h_i \mbox{ is inactive}    \} \bigr) &= 0 \, .
	\end{align*}
	Note that for $b=0.1$  and $b=0.01$ we have $\Phi\Bigl( - \frac 1 {\varrho} \Bigr) \approx 0.5$, 
	see also Figure \ref{figure:prob-active-knots-1d}.

	Let us now consider the case $P_a := \ca U[-\a,\a]$, where $\a>0$. We define $\varrho := \frac \a {\sqrt 3 b}$
	and note that  for the method proposed by \cite{HeZhReSu15a}, we have $\s_a = \sqrt{2}$ and hence 
	$b=0.1$ again leads to $\rho \approx 14.1$ and $b=0.01$ leads to $\rho \approx 141$.
	Moreover,  the functions $f_{P_b/P_a}$, $F_{P_b/P_a}$, and $F_{P_b/P_a}^+$ are computed in Example 
	\ref{example:uniform-denominator}. 
	For a   data set with $[\xmin, \xmax] = [0,1]$,  Lemma \ref{result:one-layer-distribution-knots} 
	and Theorem \ref{result:pos-bias-better}
	then
	give
% % % 	\begin{align*}  
% % % 	 P\bigl(\{\mbox{neuron }  h_i \mbox{ is fully active}    \} \bigr) &=     
% % % 	 \begin{dcases*}
% % % 	   0 & \mbox{if}  $\varrho \leq \frac 1 {\sqrt 3}$\\
% % % 	  1 - \frac 1 {\sqrt 3\varrho} & \mbox{if} $\varrho \geq \frac 1 {\sqrt 3}$\, ,
% % % 	  \end{dcases*} \\  
% % % 	 P\bigl(\{\mbox{neuron }  h_i \mbox{ is semi-active}    \} \bigr) &= 
% % % 	 \begin{dcases*}
% % % 	   1 & \hspace*{4.1ex}\mbox{if}  $\varrho \leq \frac 1 {\sqrt 3}$\\
% % % 	  \frac 1 {\sqrt 3\varrho} & \hspace*{4.1ex}\mbox{if} $\varrho \geq \frac 1 {\sqrt 3}$\, ,
% % % 	  \end{dcases*} \\  
% % % 	 P\bigl(\{\mbox{neuron }  h_i \mbox{ is inactive}    \} \bigr) &= 0 \, ,
% % % 	\end{align*}
% % % 	while for a data set with $[\xmin, \xmax] = [0,1]$ they lead to
		\begin{align*}  
	 P\bigl(\{\mbox{neuron }  h_i \mbox{ is fully active}    \} \bigr) &=     
	 \begin{dcases*}
	   0 & \mbox{if}  $\varrho \leq \frac 1 {\sqrt 3}$\\
	  \frac 12 - \frac 1 {\sqrt {12}\varrho} & \mbox{if} $\varrho \geq \frac 1 {\sqrt 3}$\, ,
	  \end{dcases*} \\  
	 P\bigl(\{\mbox{neuron }  h_i \mbox{ is semi-active}    \} \bigr) &= 
	 \begin{dcases*}
	   1 & \hspace*{4.1ex}\mbox{if}  $\varrho \leq \frac 1 {\sqrt 3}$\\
	  \frac 12 +\frac 1 {\sqrt {12}\varrho} & \hspace*{4.1ex}\mbox{if} $\varrho \geq \frac 1 {\sqrt 3}$\, ,
	  \end{dcases*} \\  
	 P\bigl(\{\mbox{neuron }  h_i \mbox{ is inactive}    \} \bigr) &= 0 \, .
	\end{align*}
	Consequently, for $\rho \approx 14.1$ or $\rho \approx 141$ the probability of initializing a fully active neuron 
	approximately equals $0.5$ and the same is true for 
	 semi-active
	neurons, see also \ref{figure:prob-active-knots-1d}.
	Finally, %for both types of data sets 
	the distribution of each knot $x_i^*$ has the Lebesgue density
	\begin{displaymath}
	  f_{P_b/P_a}(z)  
	  =
	  \begin{dcases*}
	  0 & \mbox{if} $z\in \bigl[ -\frac 1{\sqrt 3 \varrho}, \frac 1{\sqrt 3 \varrho} \bigr]$\\
	    \frac 1{\sqrt{12} \varrho}\cdot z^{-2}  & \mbox{if} $z<-\frac 1{\sqrt 3 \varrho}$ \mbox{ or } $z> \frac 1{\sqrt 3 \varrho}$\, ,
	  \end{dcases*}
	\end{displaymath}
	and for the above mentioned values of $\rho$ the corresponding distributions are highly concentrated around $0$, see 
	Figure \ref{figure:densities-1d}.
\end{example}

\begin{figure}[t]
\begin{center}
\includegraphics[width=0.32\textwidth]{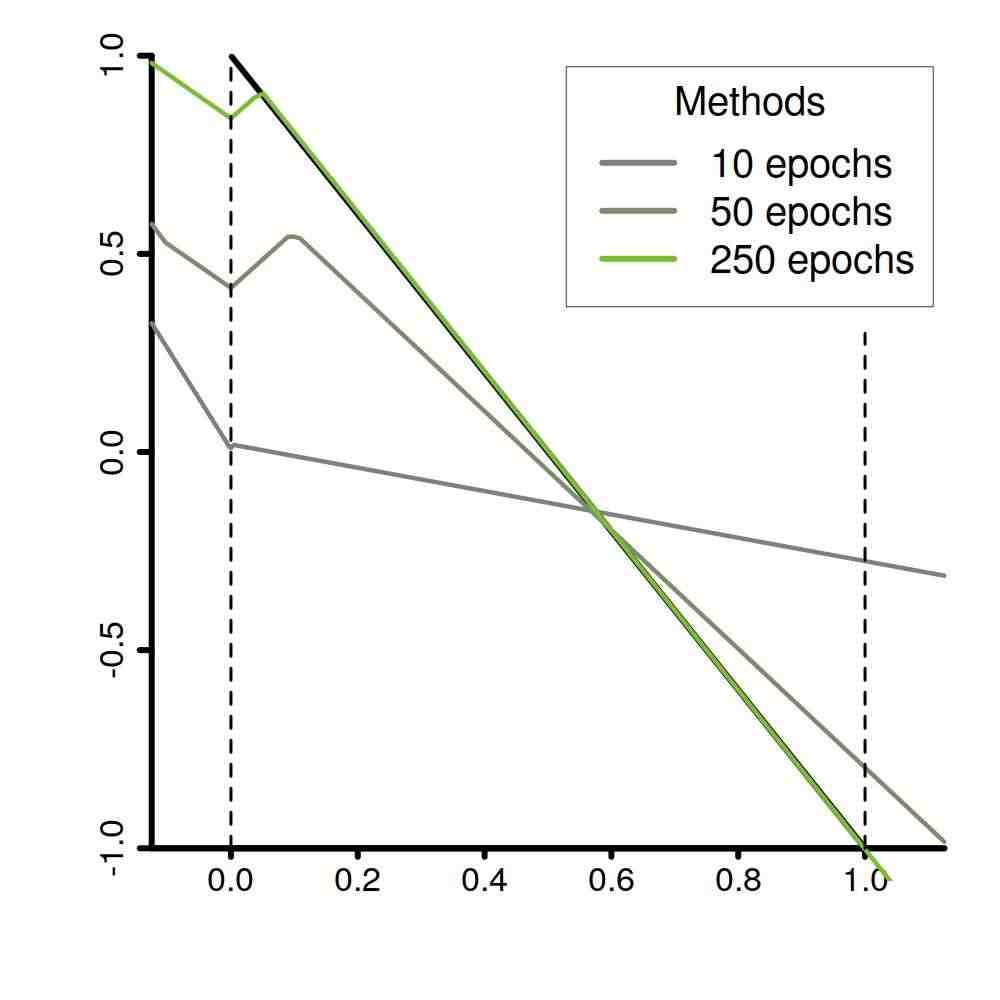}
\hspace*{-0.01\textwidth}
\includegraphics[width=0.32\textwidth]{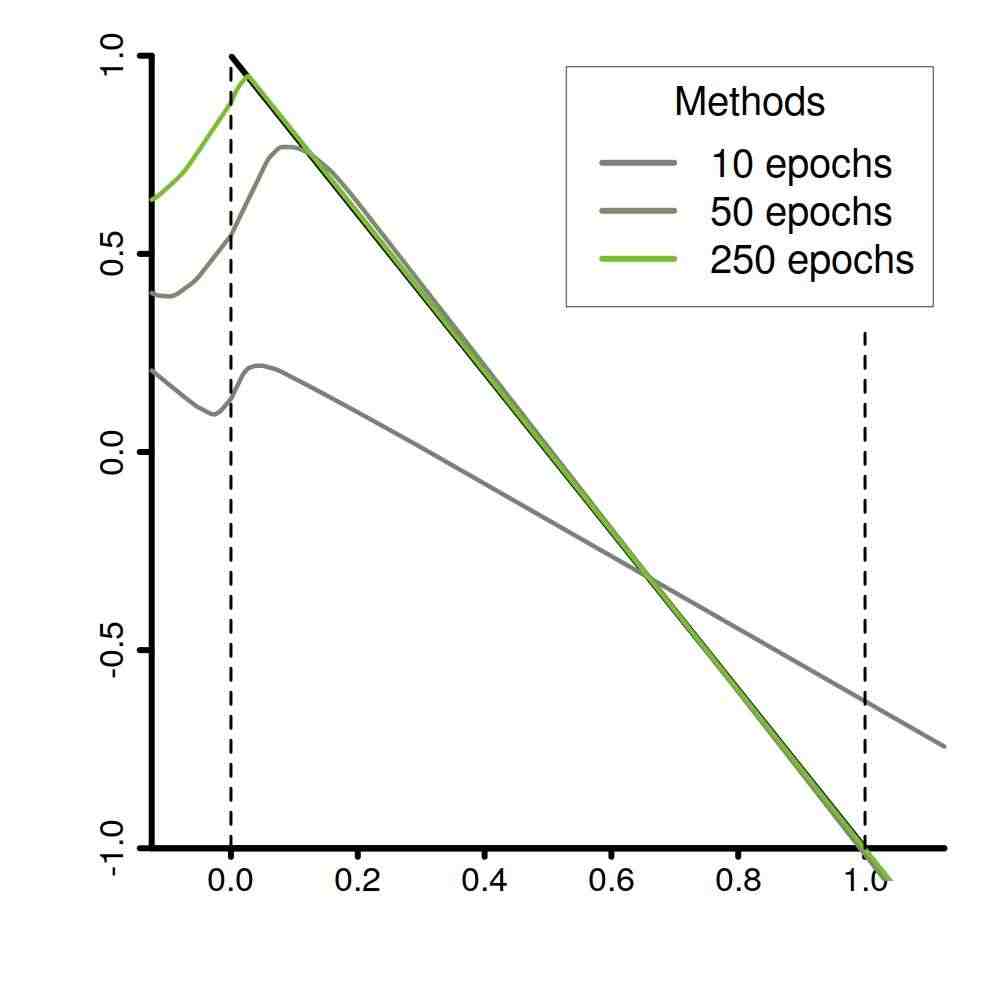}
\hspace*{-0.01\textwidth}
\includegraphics[width=0.32\textwidth]{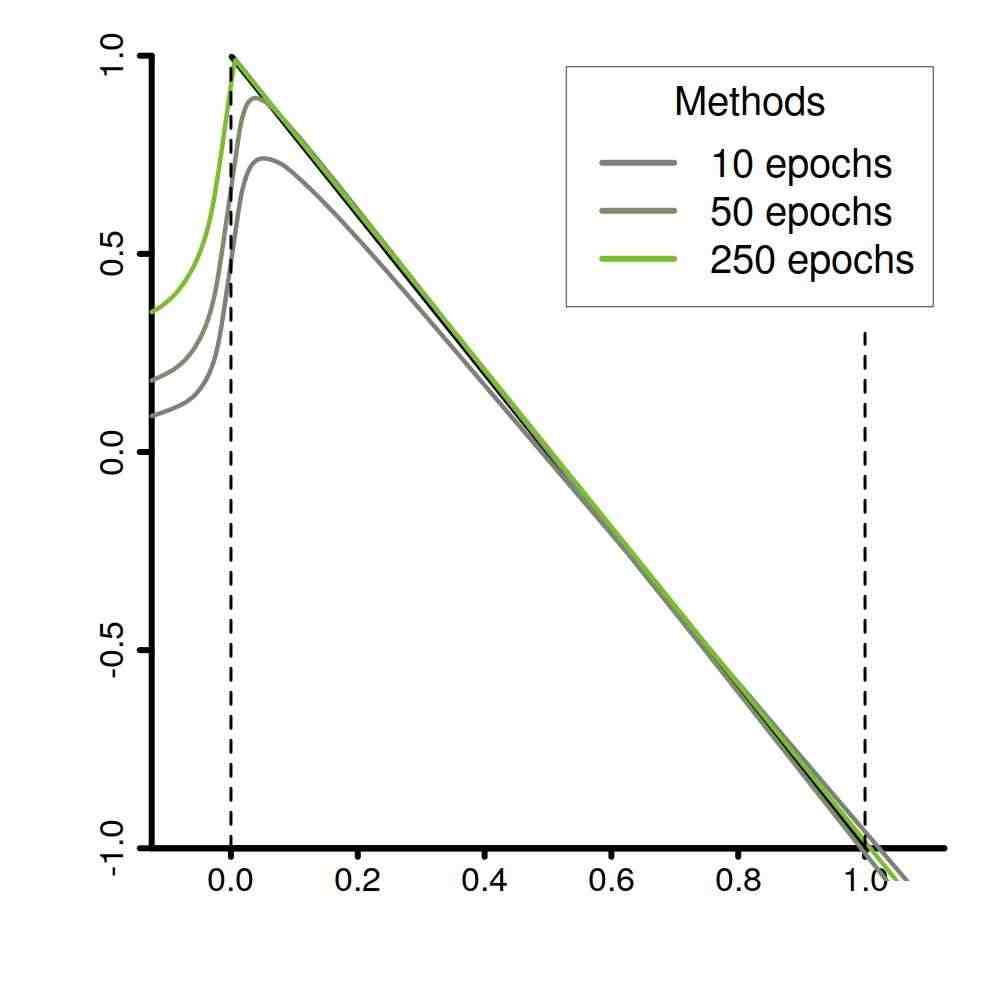}

\vspace*{-3ex}
\includegraphics[width=0.32\textwidth]{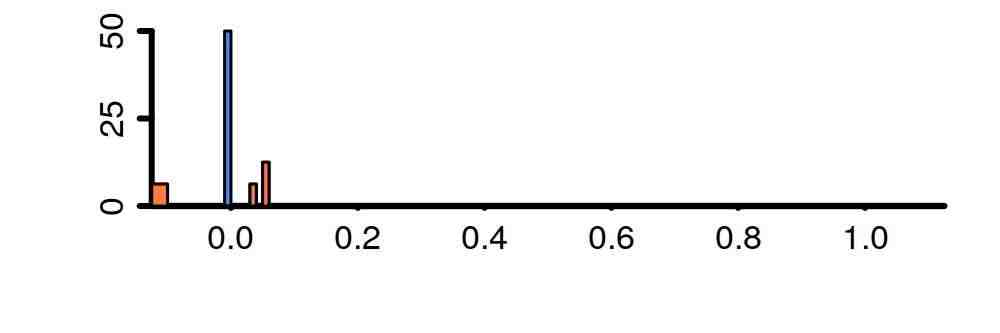}
\hspace*{-0.01\textwidth}
\includegraphics[width=0.32\textwidth]{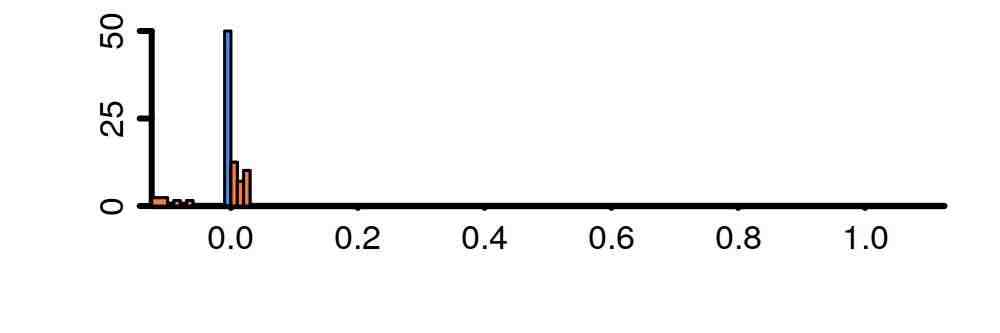}
\hspace*{-0.01\textwidth}
\includegraphics[width=0.32\textwidth]{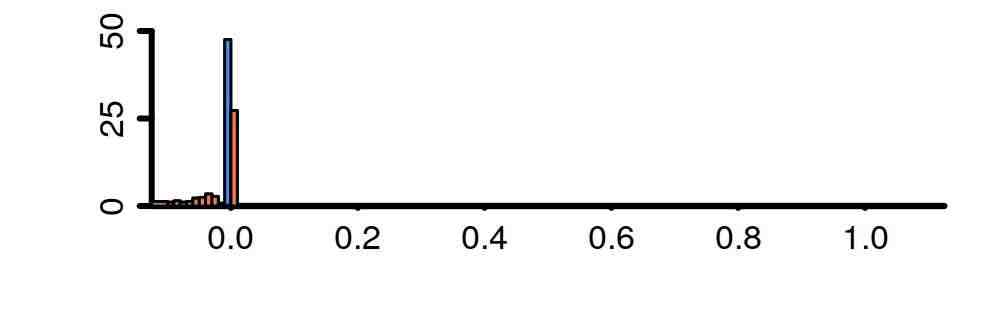}
\vspace*{-6ex}
\end{center}
\caption{Training behavior for three architectures  and the initialization strategy \heetal\ with 
zero mean normal distribution for the weights.
The biases are set to zero as in Example \ref{example:He-init-1d}.
The upper row displays how well the predictors approximate the target $f^*(t) =  2 -t$ after 10, 50, and 250 training epochs.
The histograms in the lower row indicate the distribution of knots in percent for $i\in I_+$ (orange bars) and $i\in I_-$ (blue bars on top of the orange ones).
Left: $m=16$ hidden neurons. Middle: $m=128$. Right: $m=1024$. 
In all cases, the target is not well approximated after  50 epochs.
Also, despite the fact that the target function can be represented by single hidden neuron (or even no hidden layer at all),
the already over-parameterized architecture $m=16$ exhibits some difficulties in quickly learning the target function.
Finally, the large blue bars left to zero correspond to the approximately 50 percent of dead neurons as predicted in 
Example \ref{example:He-init-1d}.
}\label{figure:random-functions-own-train-1-he}
\end{figure}

\begin{figure}[t]
\begin{center}
\includegraphics[width=0.32\textwidth]{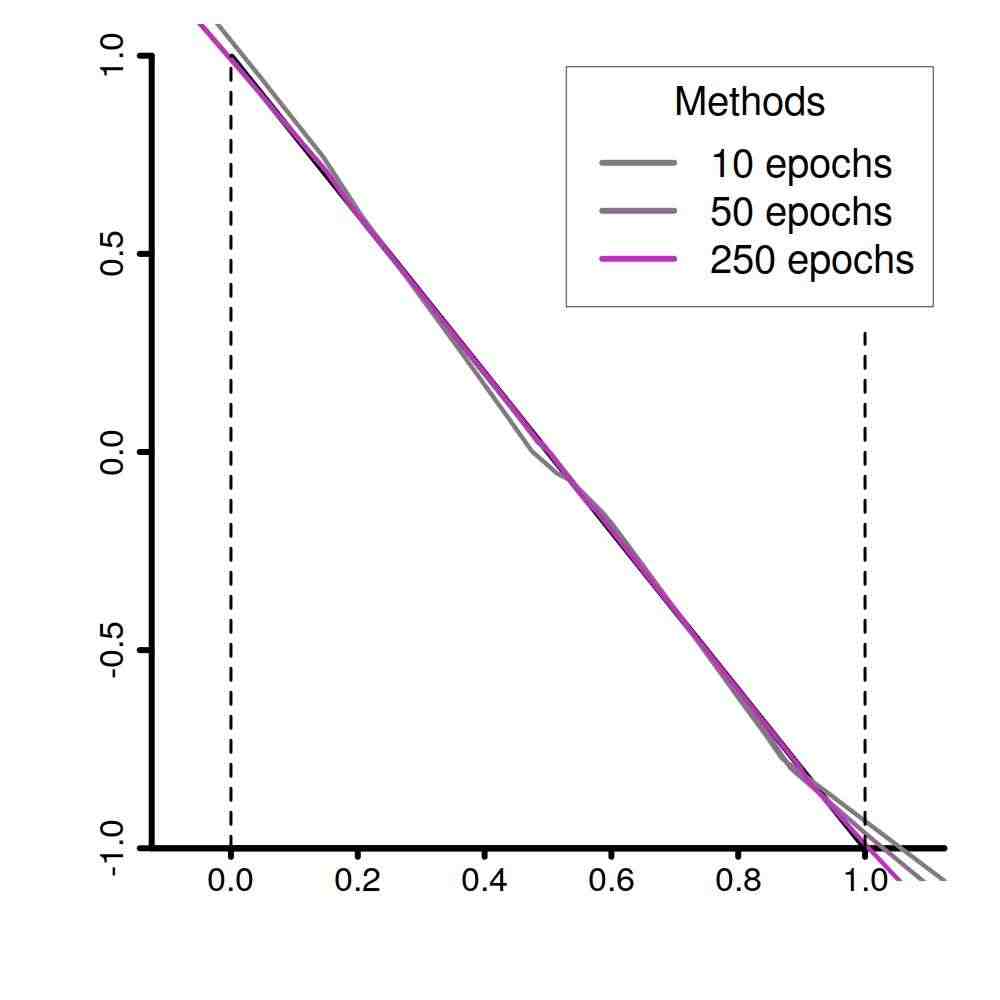}
\hspace*{-0.01\textwidth}
\includegraphics[width=0.32\textwidth]{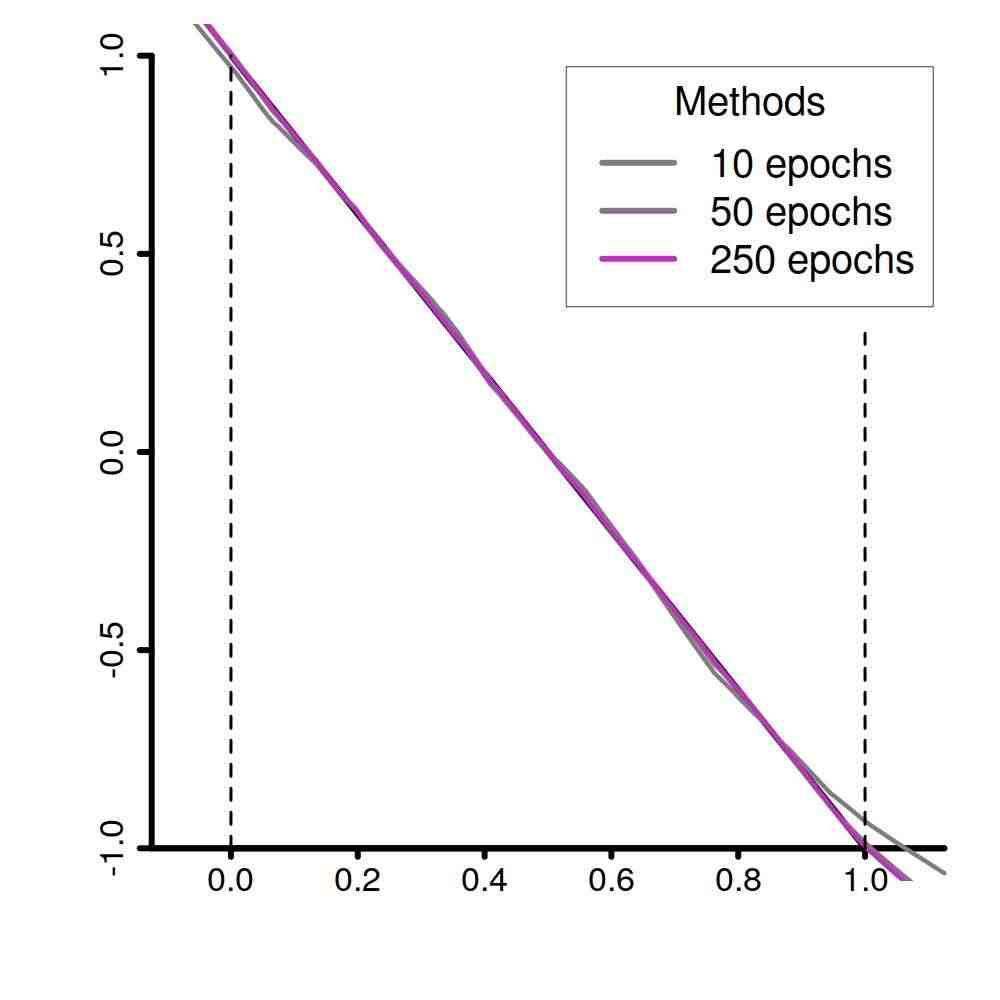}
\hspace*{-0.01\textwidth}
\includegraphics[width=0.32\textwidth]{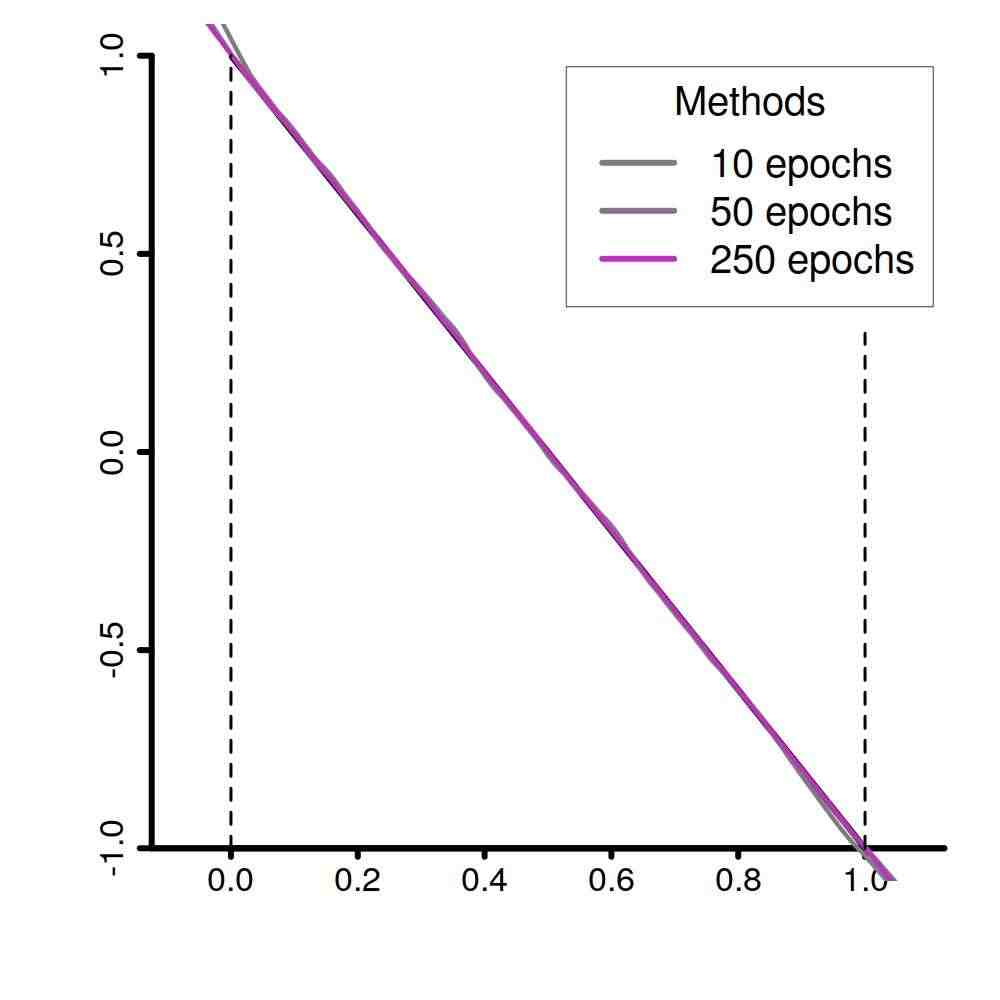}

\vspace*{-3ex}
\includegraphics[width=0.32\textwidth]{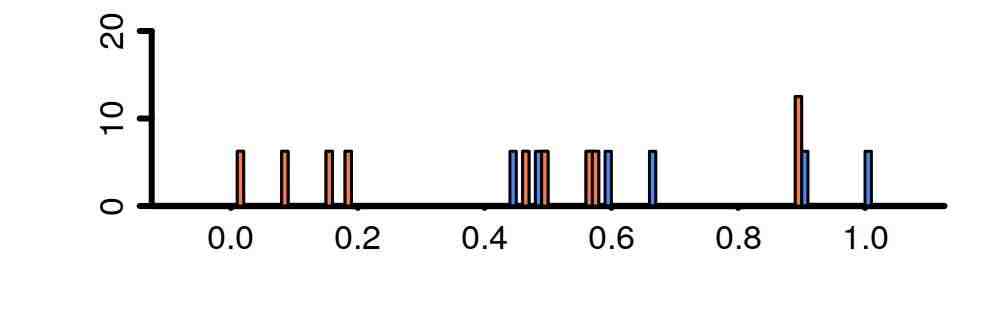}
\hspace*{-0.01\textwidth}
\includegraphics[width=0.32\textwidth]{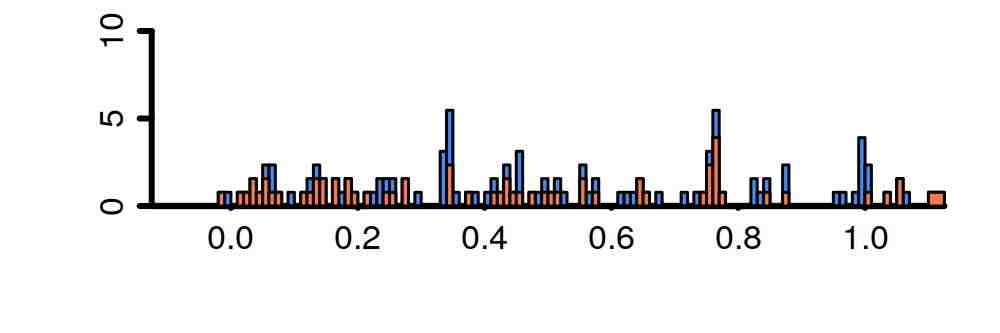}
\hspace*{-0.01\textwidth}
\includegraphics[width=0.32\textwidth]{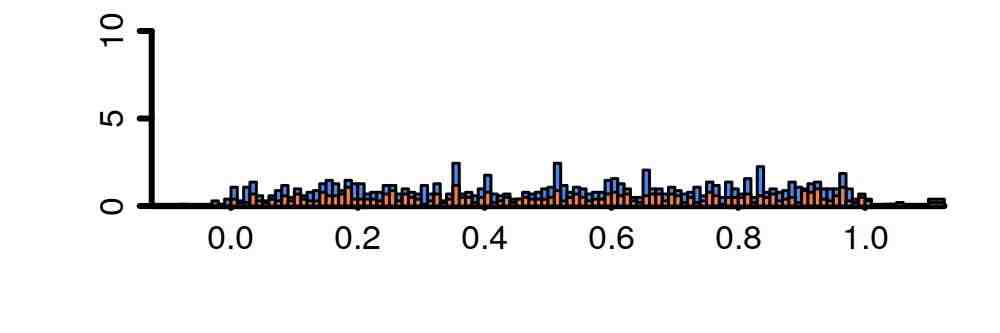}
\vspace*{-6ex}
\end{center}
\caption{Training behavior for three architectures and an initialization strategy, that 
samples the knots $x_i^*$   from $\ca U[0,1]$ and that 
initializes the weights by a zero mean normal distribution
with variances according to \heetal. The 6 graphics have a meaning analogous to 
Figure \ref{figure:random-functions-own-train-1-he}. Notice that unlike the method considered in 
Figure \ref{figure:random-functions-own-train-1-he}, the new initialization method already leads to a 
good approximation after 10 training epochs.
}\label{figure:random-functions-own-train-1-own}
\end{figure}

\begin{example}[Random Initializations]\label{example:random-init}
  Another class of possible initialization strategies initialize 
  both the weights and the biases randomly with the help of some 
  ad-hoc distributions such as the uniform or normal distribution.
  These strategies are considered in this example.
  
%  Traditionally, the parameters of neural networks were often initialized 
%  by some ad-hoc distributions such as uniform or normal distributions.
 
 Let us first investigate the case of normal distributions, that is, in the hidden layer we have
 $P_a:=\ca N(0,\s_a^2)$ and $P_b: = \ca N(0,\s_b^2)$ for some $\s_a,\s_b>0$,
 and the output layer is initialized similarly with variances $\s_w^2$ and $\s_c^2$, instead.
 Let us write $\varrho := \s_a / \s_b$ for the inverse ratio of standard deviations.
 For a given data set $D$, a combination of Lemma \ref{result:one-layer-distribution-knots}
 and Example \ref{example:normal-ratio} with \eqref{pqp-sym} and $\arctan (-t) = - \arctan (t)$ then yields
\begin{align*} 
  P\bigl(\{\mbox{neuron }  h_i \mbox{ is fully active}    \} \bigr) & =
  \frac 1 \pi \arctan (\varrho \cdot \xmax   )
  - \frac 1 \pi \arctan (\varrho \cdot \xmin  ) \, , \\  
  P\bigl(\{\mbox{neuron }  h_i \mbox{ is semi-active}    \} \bigr) &=  P\bigl(\{\mbox{neuron }  h_i \mbox{ is inactive}    \} \bigr)\\
  &=
  \frac 12 -  \frac 1 {2\pi} \arctan (\varrho \cdot \xmax   )
  + \frac 1 {2\pi} \arctan (\varrho \cdot \xmin  )\, 
\end{align*} 
 for all $i\in I$. In particular, 
% % %  if the data is scaled to $[-1,1]$, that is
% % %  $[\xmin, \xmax] = [-1,1]$, then we find
% % %  \begin{displaymath}
% % %   P\bigl(\{\mbox{neuron }  h_i \mbox{ is inactive}    \} \bigr)
% % %     = \frac 12 - 
% % %     \frac 1 \pi \arctan (\varrho  )\, ,
% % %  \end{displaymath}
% % %  and 
 if the data is scaled to $[0,1]$, 
 that is
 $[\xmin, \xmax] = [0,1]$
 then the latter probability becomes
 \begin{displaymath}
  P\bigl(\{\mbox{neuron }  h_i \mbox{ is inactive}    \} \bigr)
    = \frac 12 - 
    \frac 1 {2\pi} \arctan (\varrho  )\, .
 \end{displaymath}
 In addition, the distribution of each knot $x_i^*$ has the Lebesgue density
 \begin{displaymath}
   f_{P_b/P_a}(z) = \frac{1}{\pi} \cdot \frac{\varrho}{\varrho^2 z^2+ 1}\, , \qquad \qquad z\in \R.
\end{displaymath}

  Let us now consider the case, in which both distributions $P_a$ and $P_b$ are uniform distributions.
  We begin with the sub-case $P_b:=\ca U[0,\b]$ and $P_a:=\ca U[-\a,\a]$ for some $\a,\b>0$.
  Again, we write  $\varrho := \frac{\a}{\sqrt 3} (\frac{\b}{\sqrt 12})^{-1} =   2\a/\b$ for the inverse ratio of standard deviations.
  The formula for the cumulative distribution function  provided in  Example \ref{example:asym-uniform-ratio}  then reads as
  \begin{displaymath}
 F_{P_b/P_a}(z)
 =
 \begin{dcases*}
 - \frac 1{2\varrho z} & \mbox{if}  $z\leq-\frac 2\varrho$\\
    \frac {4+\varrho z}{8} & \mbox{if} $z\in \bigl[ -\frac 2\varrho, \frac 2\varrho  \bigr]$\\
	1- \frac 1{2\varrho z}  & \mbox{if}   $z\geq \frac 2\varrho$\, .
\end{dcases*}
\end{displaymath}
  For a data set with $[\xmin, \xmax] = [0,1]$ we consequently find by 
  Lemma \ref{result:one-layer-distribution-knots}
%   \begin{align*} 
%   P\bigl(\{\mbox{neuron }  h_i \mbox{ is fully active}    \} \bigr)
%   &=
%  \begin{dcases*}
%   \frac \varrho 4 & \mbox{if}  $\varrho \leq 2$\\
%      1- \frac 1 \varrho & \mbox{if} $\varrho \geq 2$\, ,
%     \end{dcases*}\\
%     P\bigl(\{\mbox{neuron }  h_i \mbox{ is inactive}    \} \bigr) &= 0\, .
% \end{align*} 
% Moreover, for  a data set with $[\xmin, \xmax] = [0,1]$  we obtain
  \begin{align*} 
  P\bigl(\{\mbox{neuron }  h_i \mbox{ is fully active}    \} \bigr)
 & =
 \begin{dcases*}
  \frac \varrho 8 & \mbox{if}  $\varrho \leq 2$\\
     \frac 12- \frac 1 {2\varrho} & \mbox{if} $\varrho \geq 2$\, ,
    \end{dcases*} \\
    P\bigl(\{\mbox{neuron }  h_i \mbox{ is inactive}    \} \bigr) &= 0\, .
\end{align*} 
 Finally, for both types of data sets the distribution of each knot $x_i^*$ has the Lebesgue density
 \begin{displaymath}
   f_{P_b/P_a}(z) = \frac{1}{2} \cdot \min \Bigl\{\frac \varrho 4, \frac 1 {\varrho z^2}   \Bigr\}\, , \qquad \qquad z\in \R.
\end{displaymath}
  Let us now consider the sub-case $P_b:=\ca U[-\b,\b]$ and $P_a:=\ca U[-\a,\a]$ for some $\a,\b>0$.
  Then the inverse ratio of standard deviations is   $\varrho := \a/\b$ and therefore Example \ref{example:sym-uniform-ratio}
  shows that 
    \begin{displaymath}
 F_{P_b/P_a}(z)
 =
 \begin{dcases*}
 - \frac 1{4\varrho z} & \mbox{if}  $z\leq-\frac 1\varrho$\\
    \frac {2+\varrho z}{4} & \mbox{if} $z\in \bigl[ -\frac 1\varrho, \frac 1\varrho  \bigr]$\\
	1- \frac 1{4\varrho z}  & \mbox{if}   $z\geq \frac 1\varrho$\, .
\end{dcases*}
\end{displaymath}
  For a data set with $[\xmin, \xmax] = [0,1]$ we consequently find by 
  Lemma \ref{result:one-layer-distribution-knots} that
% % %   \begin{align*} 
% % %   P\bigl(\{\mbox{neuron }  h_i \mbox{ is fully active}    \} \bigr)
% % %   &=
% % %  \begin{dcases*}
% % %   \frac \varrho 2 & \mbox{if}  $\varrho \leq 1$\\
% % %      1- \frac 1 {2\varrho} & \mbox{if} $\varrho \geq 1$\, ,
% % %     \end{dcases*}\\
% % %     P\bigl(\{\mbox{neuron }  h_i \mbox{ is inactive}    \} \bigr) &= 
% % %     \begin{dcases*}
% % %   \frac 12 -\frac \varrho 4 & \mbox{if}  $\varrho \leq 1$\\
% % %      \frac 1 {4\varrho} & \mbox{if} $\varrho \geq 1$\, ,
% % %     \end{dcases*}
% % % \end{align*}
% % % while for a data set with $[\xmin, \xmax] = [0,1]$ we obtain
  \begin{align*} 
  P\bigl(\{\mbox{neuron }  h_i \mbox{ is fully active}    \} \bigr)
  &=
 \begin{dcases*}
  \frac \varrho 4 & \mbox{if}  $\varrho \leq 1$\\
     \frac 12- \frac 1 {4\varrho} & \mbox{if} $\varrho \geq 1$\, ,
    \end{dcases*}\\
    P\bigl(\{\mbox{neuron }  h_i \mbox{ is inactive}    \} \bigr) &= 
    \begin{dcases*}
  \frac 12 -\frac \varrho 8 & \mbox{if}  $\varrho \leq 1$\\
     \frac 14 + \frac 1 {8\varrho} & \mbox{if} $\varrho \geq 1$\, .
    \end{dcases*}
\end{align*}  
 Finally,  the distribution of each knot $x_i^*$ has the Lebesgue density 
 \begin{align*}
 f_{P_b/P_a}(z) =  \frac 1 {4}  \min\Bigl\{\varrho,  \frac 1 {\varrho z^2}     \Bigr\} \, , \qquad \qquad z\in \R.
\end{align*}
\end{example}

\begin{figure}[t]
\begin{center}
\includegraphics[width=0.32\textwidth]{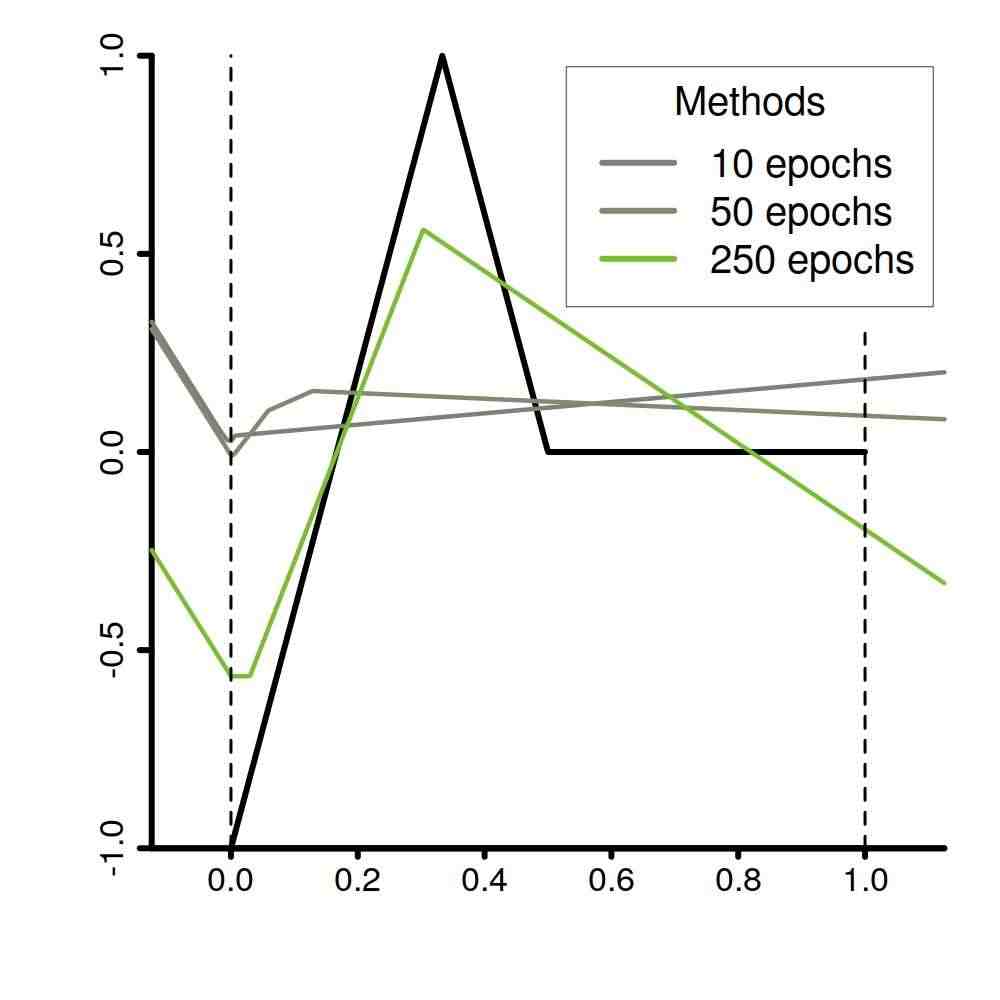}
\hspace*{-0.01\textwidth}
\includegraphics[width=0.32\textwidth]{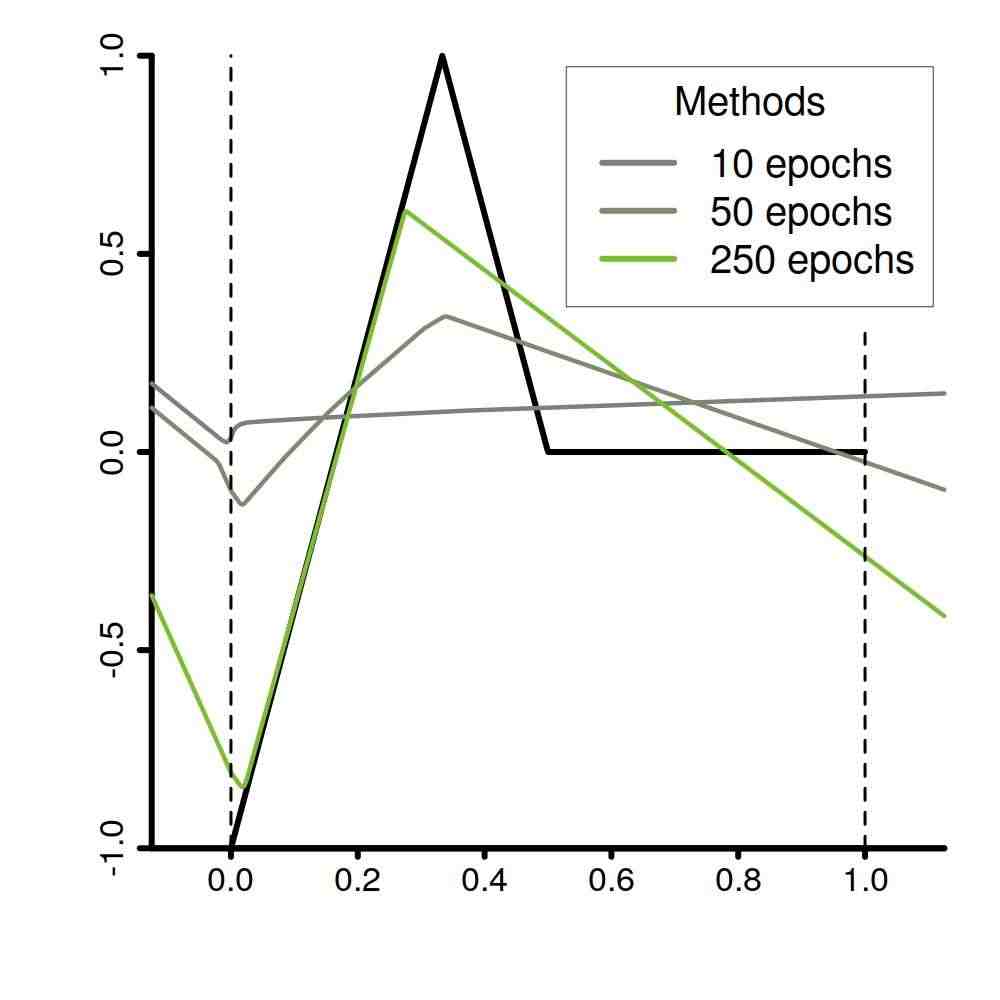}
\hspace*{-0.01\textwidth}
\includegraphics[width=0.32\textwidth]{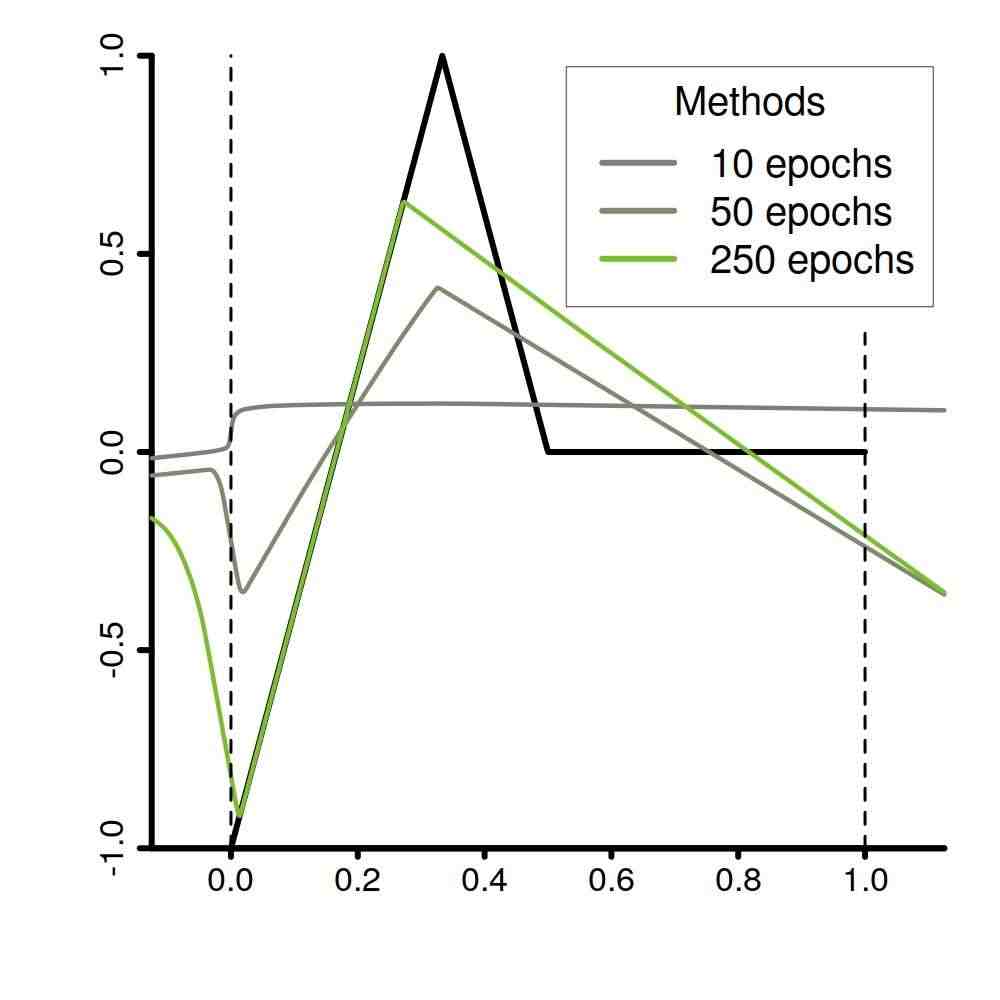}

\vspace*{-3ex}
\includegraphics[width=0.32\textwidth]{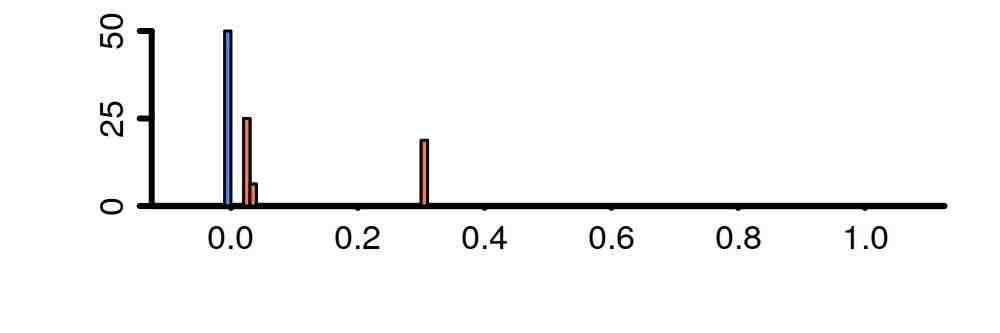}
\hspace*{-0.01\textwidth}
\includegraphics[width=0.32\textwidth]{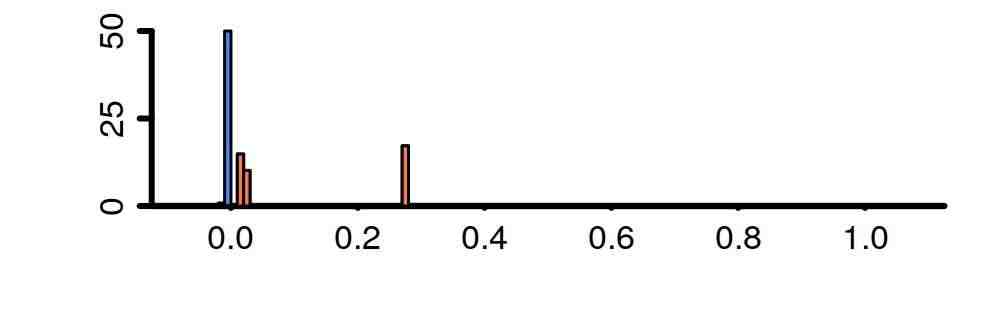}
\hspace*{-0.01\textwidth}
\includegraphics[width=0.32\textwidth]{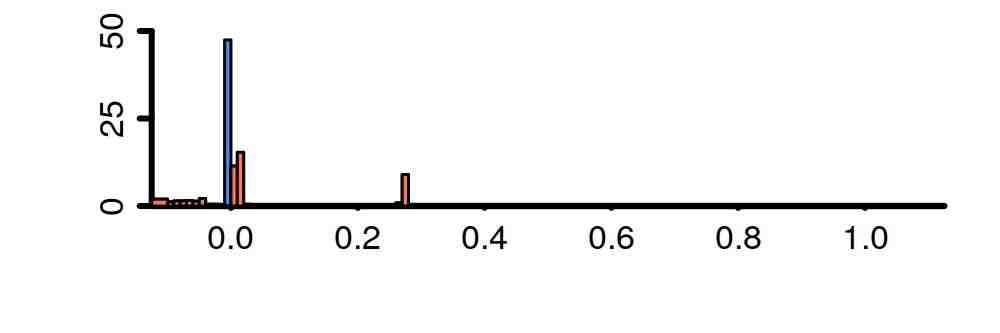}
\vspace*{-6ex}
\end{center}
\caption{Training behavior as in Figure \ref{figure:random-functions-own-train-1-he}
for the target 
function $f^*(t) =  1 - 6 \cdot |x-1/3|$, which can be represented by 2 hidden neurons with knots at $x_1^* = 1/3$
and $x_2^* = 1/2$.
Clearly, the optimizer fails to learn the target function %, even with the largest architechture $m=1024$,
within 250 epochs. %, while the new initialization method, at least for the two larger architectures,
% achieves a good approximation. 
Recall that initializing with \heetal\ places all knots at $x=0$ and 
the training algorithm apparently  has 
significant difficulties to push even a single knot   towards $x=1/3$.
% , probably since the affine linear 
% function  $g(\cdot, w,c,a,b)_{|[x_i^*, 1]}$ on the right of $x_i^*$ represents the best affine linear function 
% of $f^*_{|[x_i^*, 1]}$.
}\label{figure:random-functions-own-train-2-he}
\end{figure}

\begin{figure}[t]
\begin{center}
\includegraphics[width=0.32\textwidth]{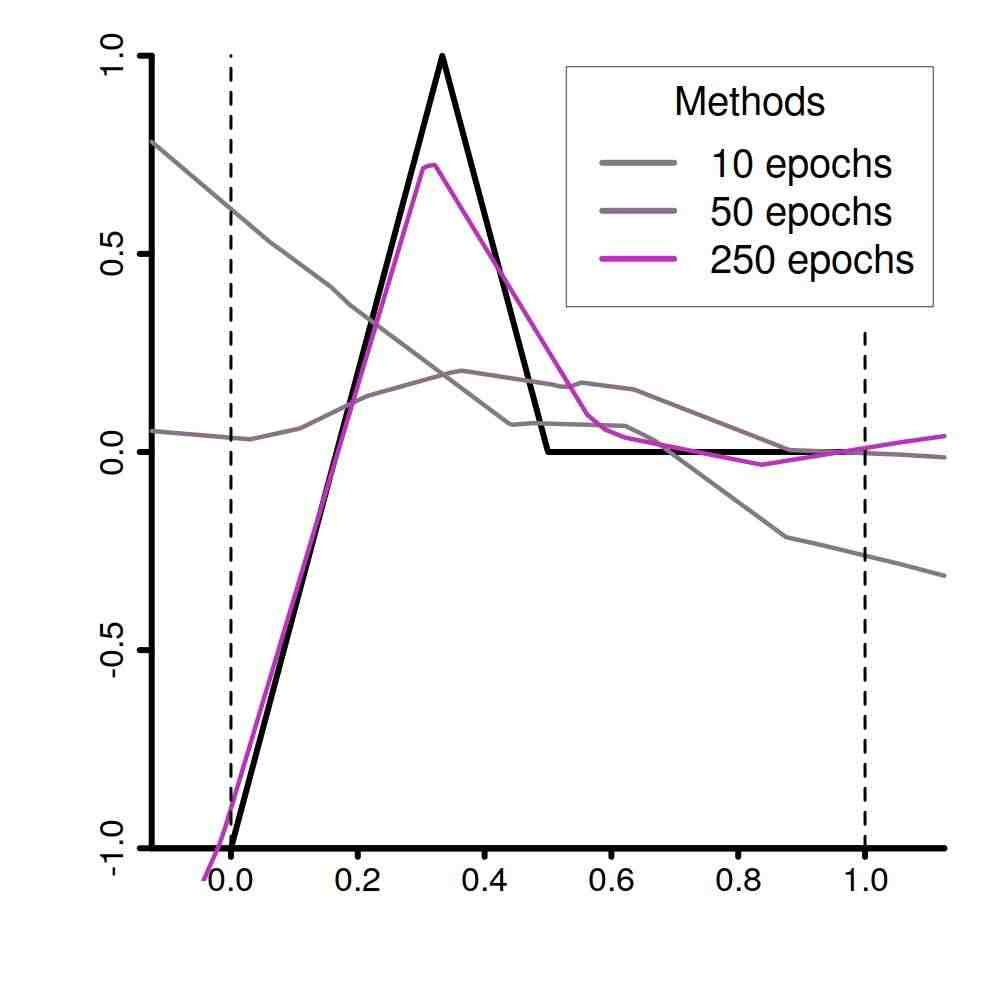}
\hspace*{-0.01\textwidth}
\includegraphics[width=0.32\textwidth]{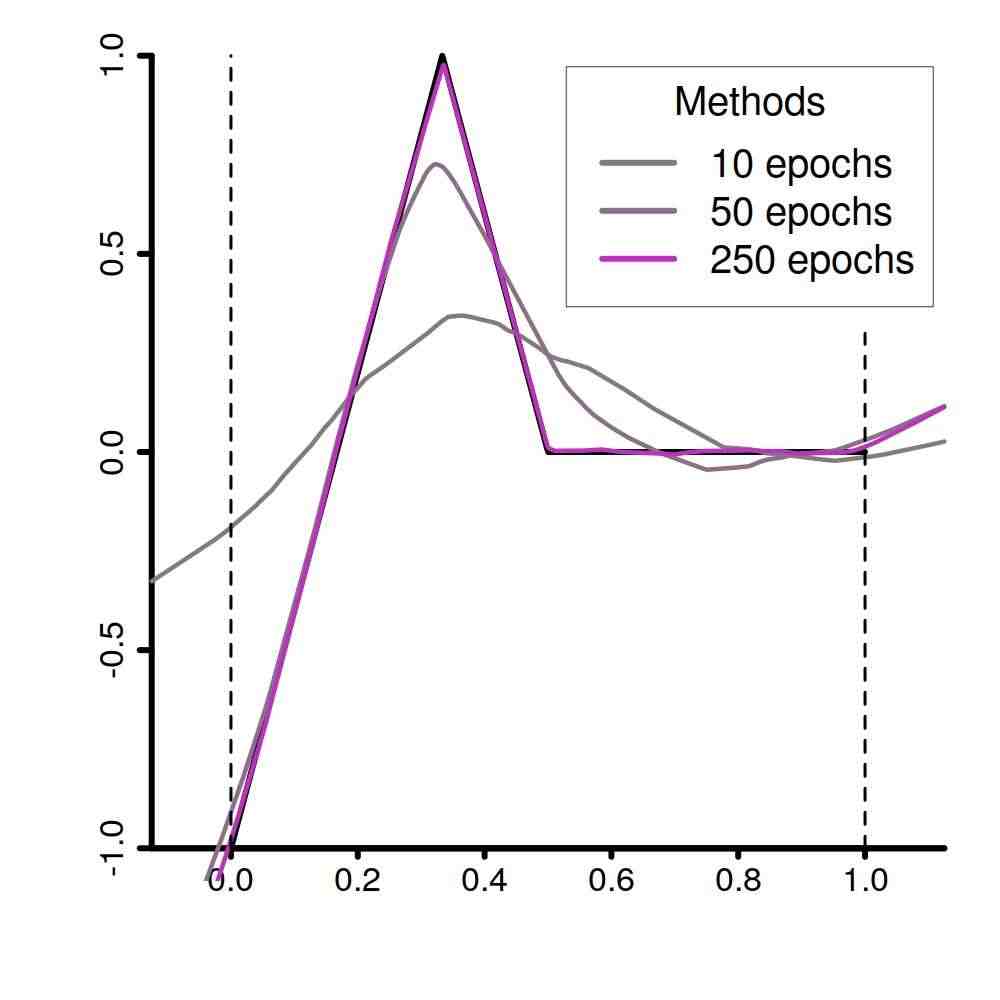}
\hspace*{-0.01\textwidth}
\includegraphics[width=0.32\textwidth]{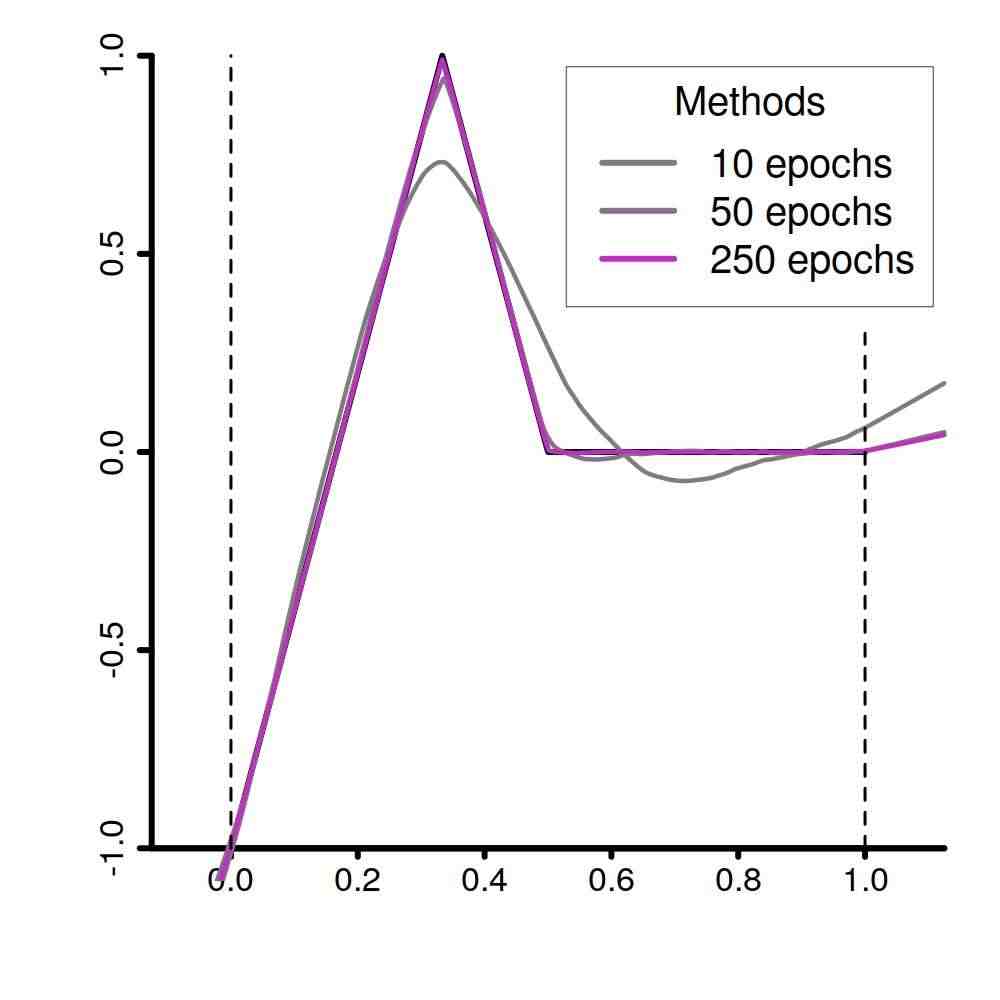}

\vspace*{-3ex}
\includegraphics[width=0.32\textwidth]{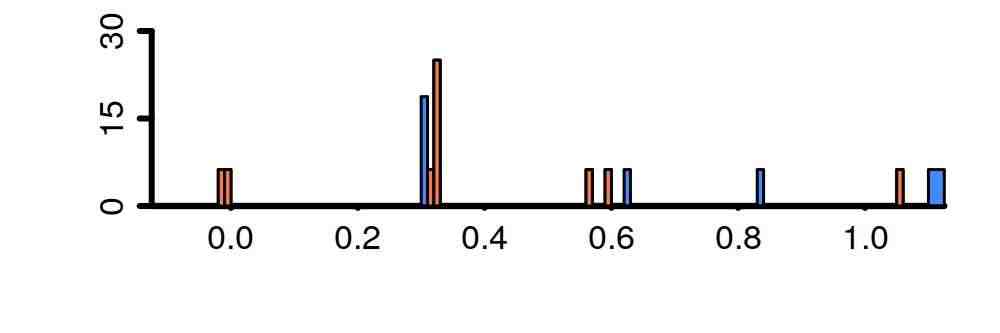}
\hspace*{-0.01\textwidth}
\includegraphics[width=0.32\textwidth]{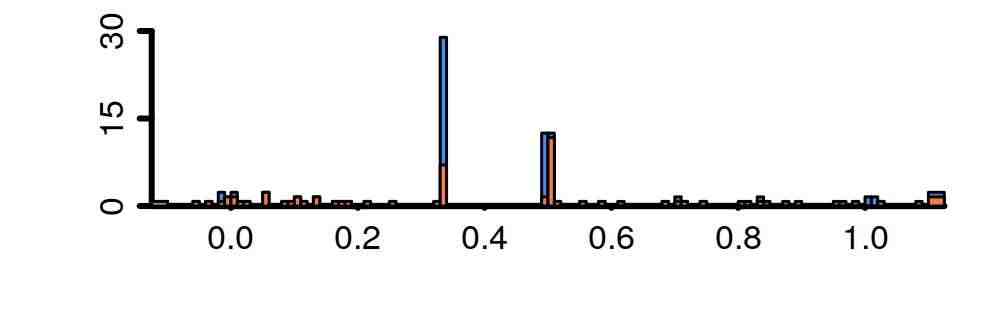}
\hspace*{-0.01\textwidth}
\includegraphics[width=0.32\textwidth]{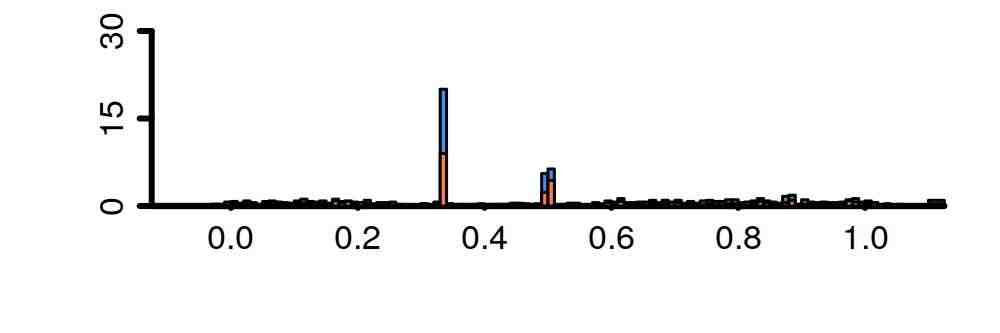}
\vspace*{-6ex}
\end{center}
\caption{Learning the function of Figure \ref{figure:random-functions-own-train-2-he} 
with the initialization method considered in Figure  \ref{figure:random-functions-own-train-1-own}.
At least for the two larger architectures, 
 the new initialization method 
achieves a good approximation 
within 250 epochs. 
}\label{figure:random-functions-own-train-2-own}
\end{figure}

\begin{figure}[t]
\begin{center}
\includegraphics[width=0.32\textwidth]{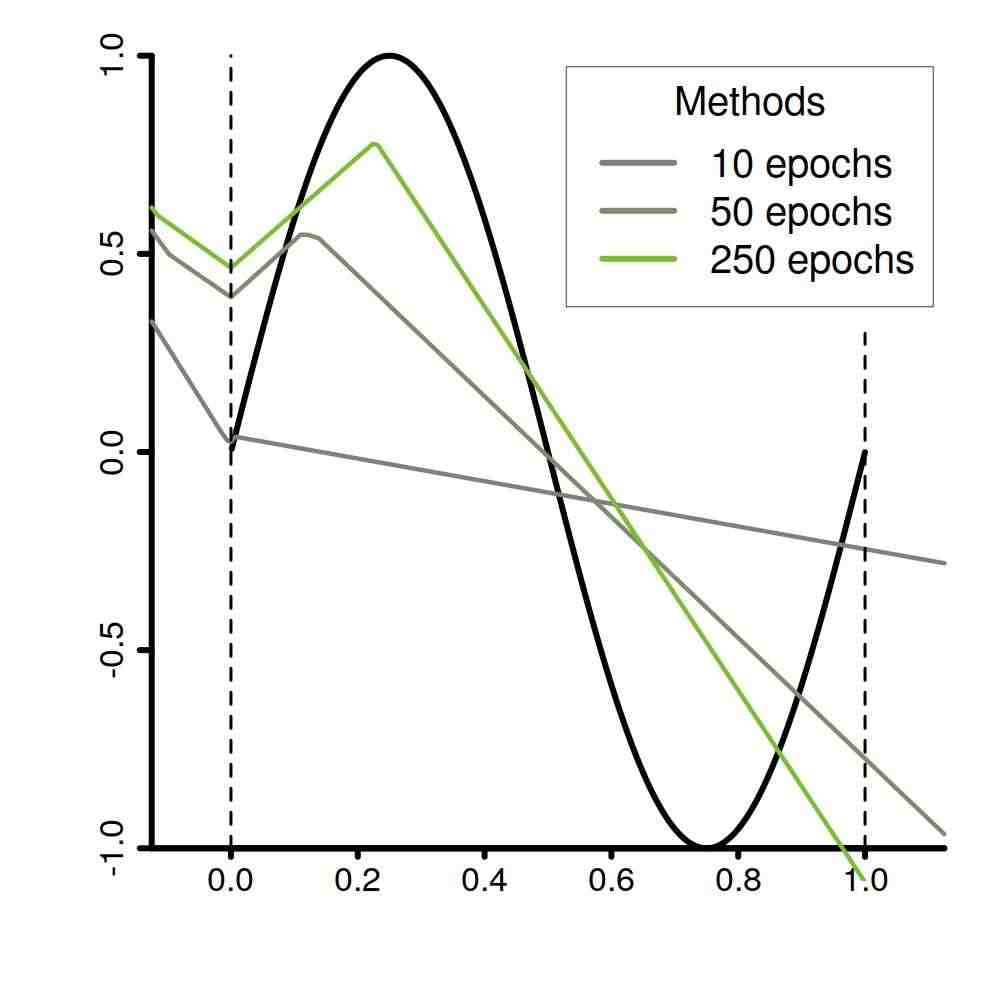}
\hspace*{-0.01\textwidth}
\includegraphics[width=0.32\textwidth]{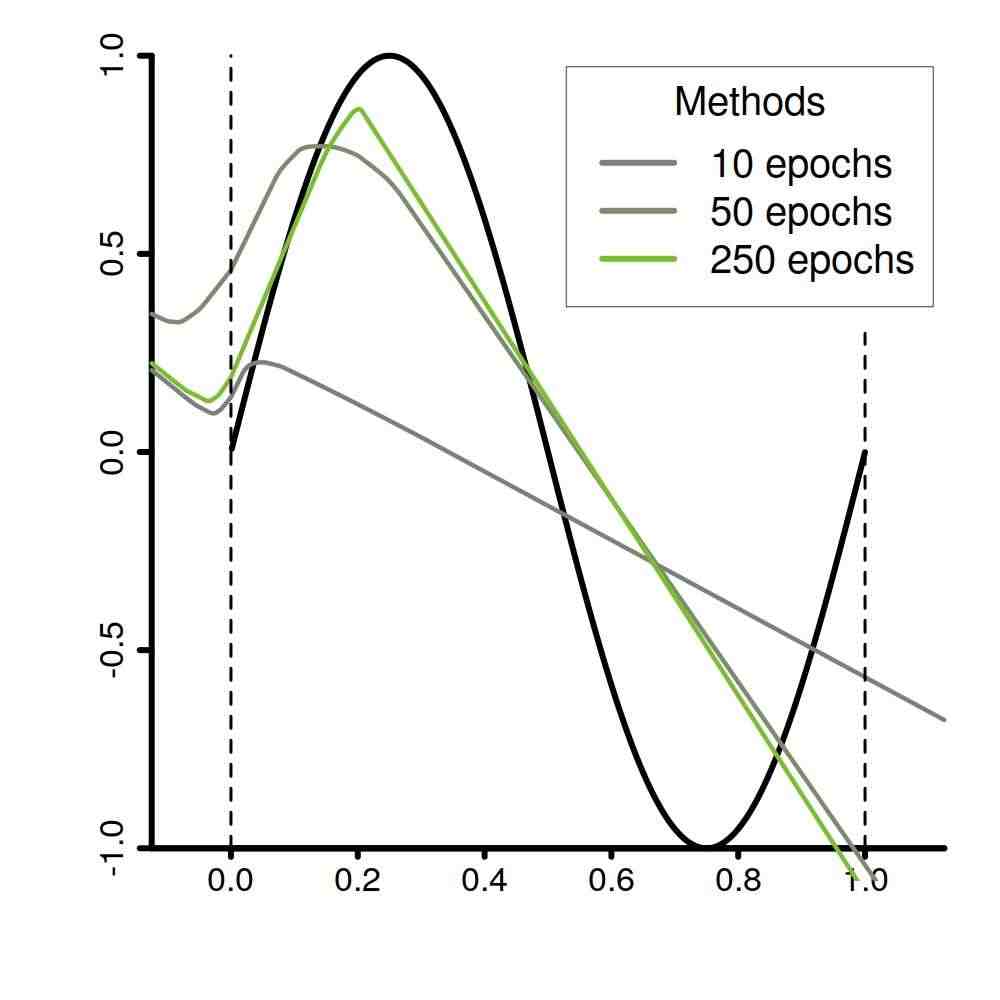}
\hspace*{-0.01\textwidth}
\includegraphics[width=0.32\textwidth]{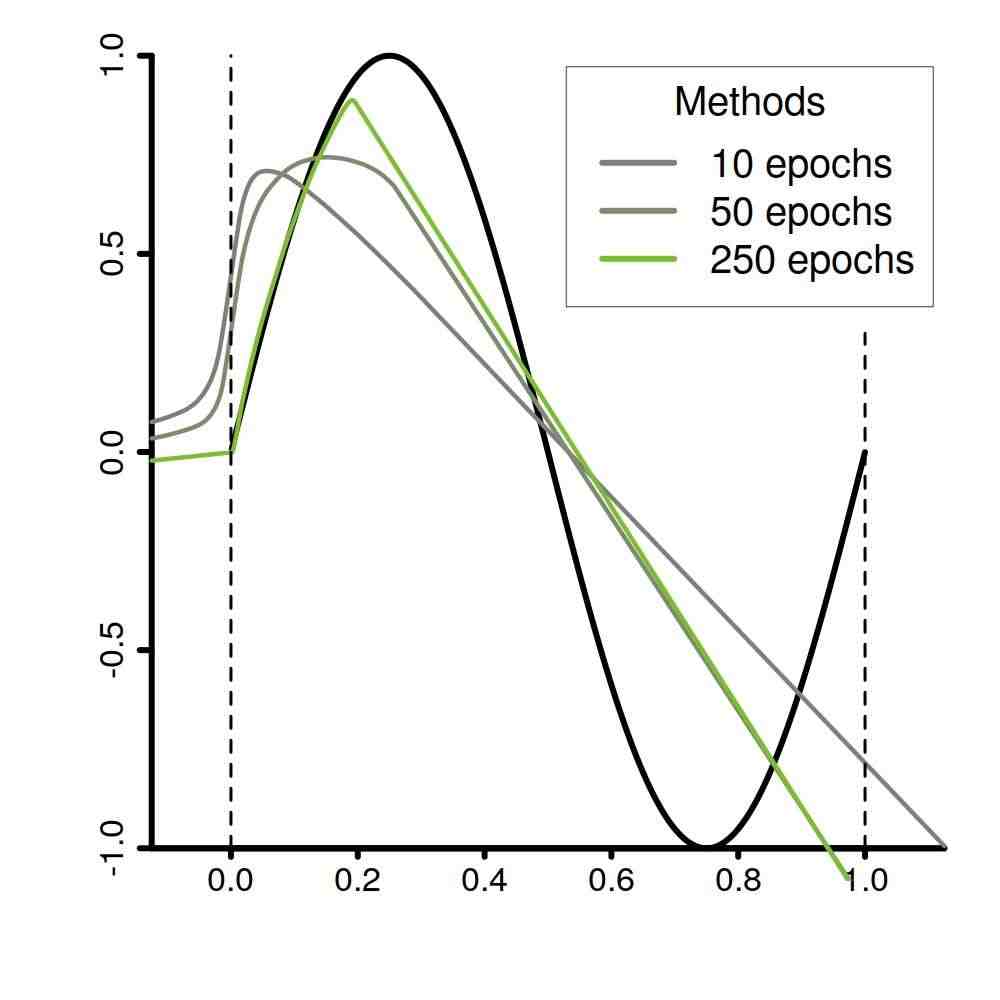}

\vspace*{-3ex}
\includegraphics[width=0.32\textwidth]{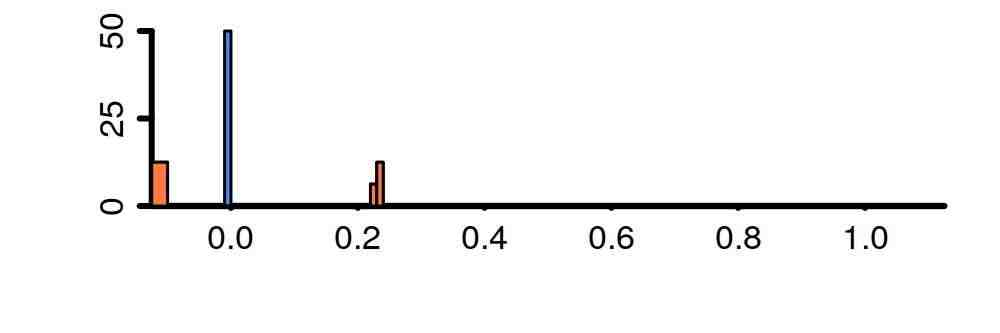}
\hspace*{-0.01\textwidth}
\includegraphics[width=0.32\textwidth]{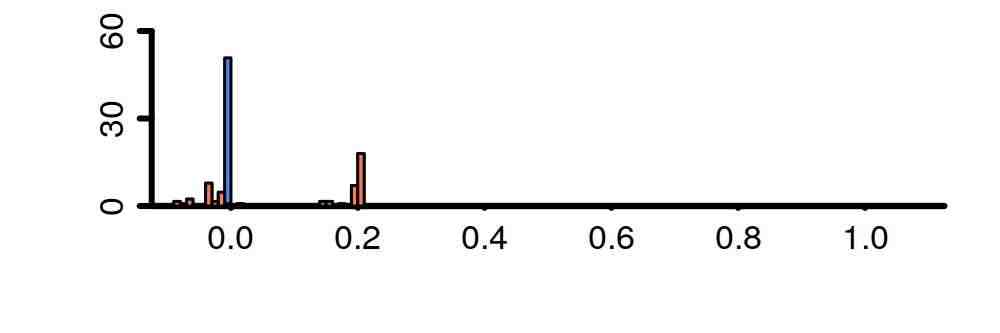}
\hspace*{-0.01\textwidth}
\includegraphics[width=0.32\textwidth]{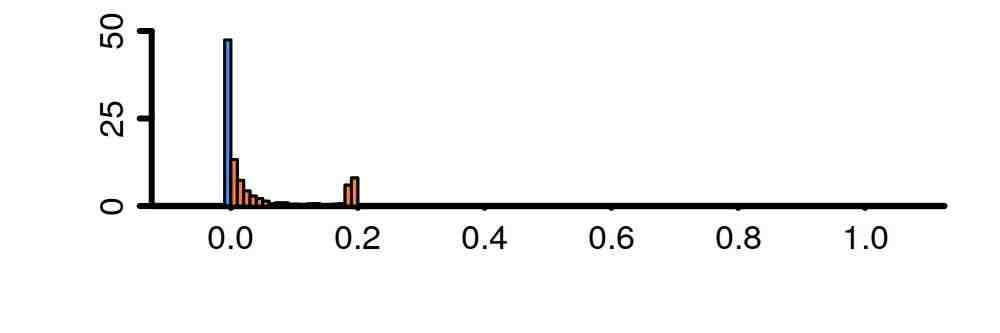}
\vspace*{-6ex}
\end{center}
\caption{Training with initialization \heetal\ as in Figure \ref{figure:random-functions-own-train-1-he}
for the target 
function $f^*(t) =  \sin(2\pi t)$.
Again,  the optimizer fails to produce meaningful approximations of $f^*$, and similar to 
Figure \ref{figure:random-functions-own-train-2-he}, the knots are not pushed beyond $1/4$. This results    in a good 
approximation on the left, but a very poor one on the right. 
}\label{figure:random-functions-own-train-3-he}
\end{figure}

\begin{figure}[t]
\begin{center}
\includegraphics[width=0.32\textwidth]{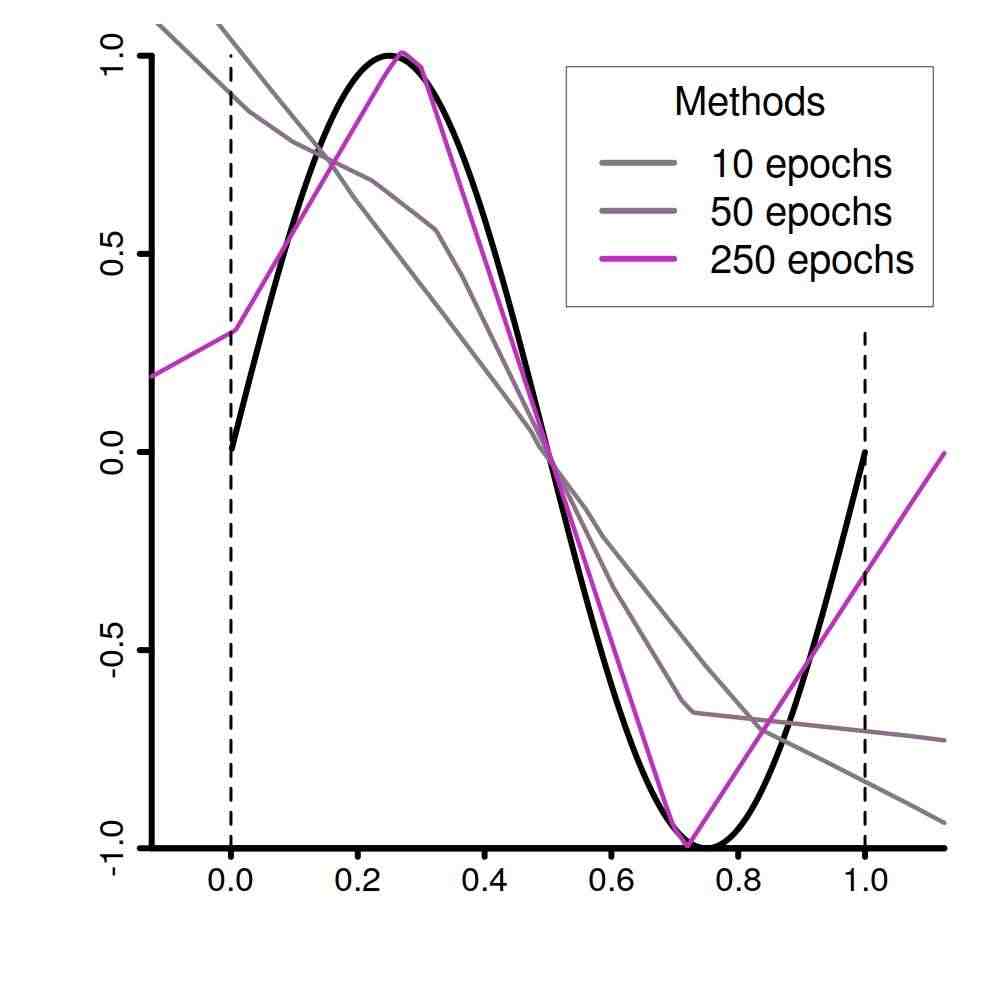}
\hspace*{-0.01\textwidth}
\includegraphics[width=0.32\textwidth]{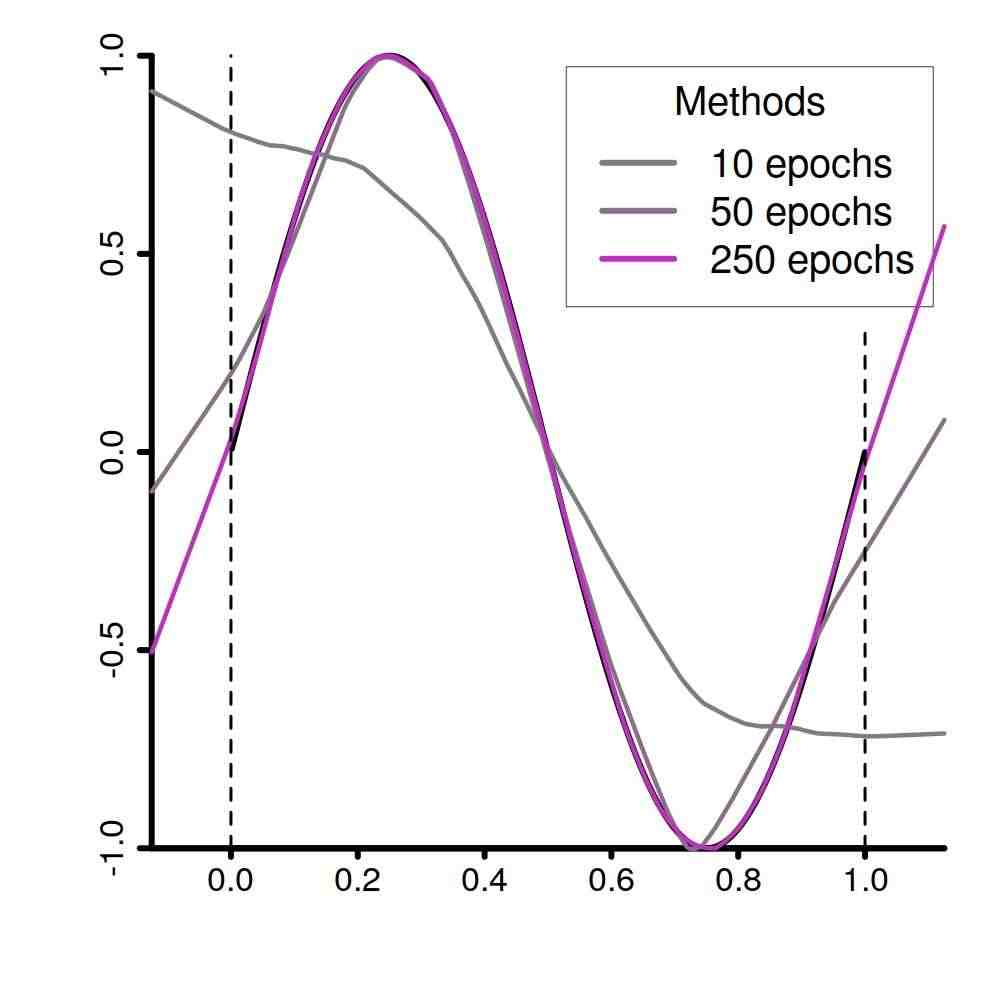}
\hspace*{-0.01\textwidth}
\includegraphics[width=0.32\textwidth]{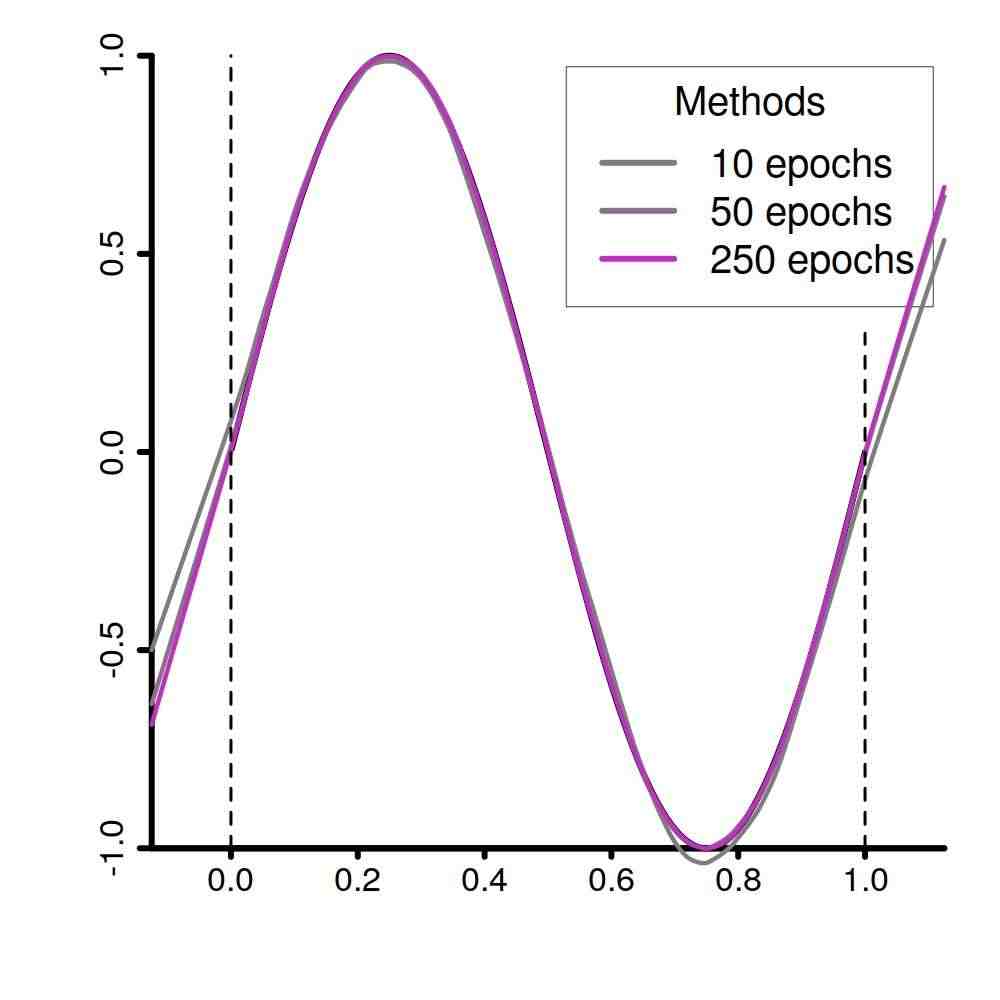}

\vspace*{-3ex}
\includegraphics[width=0.32\textwidth]{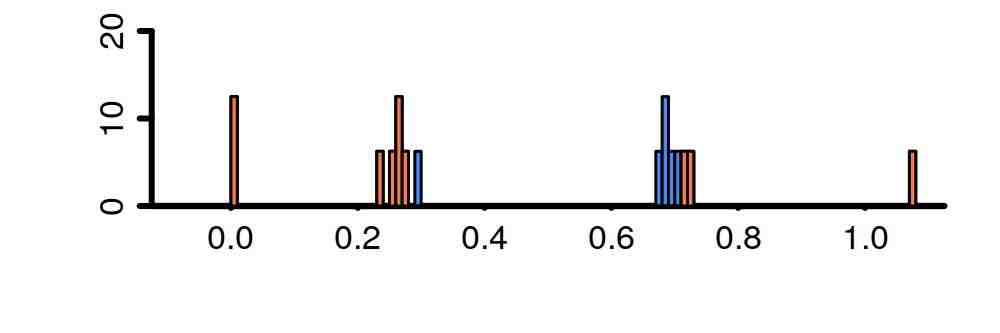}
\hspace*{-0.01\textwidth}
\includegraphics[width=0.32\textwidth]{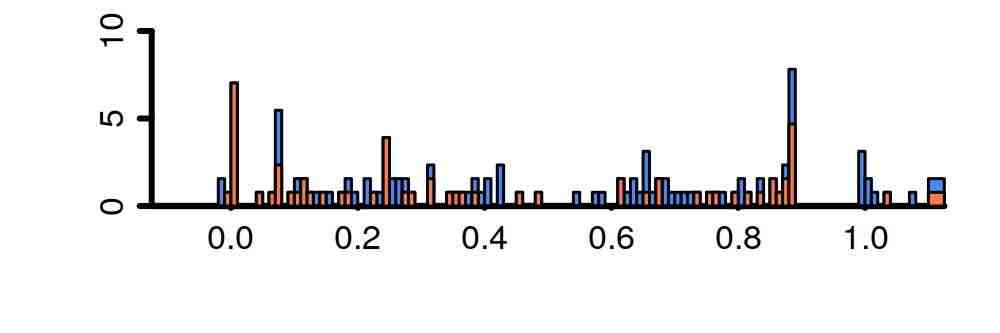}
\hspace*{-0.01\textwidth}
\includegraphics[width=0.32\textwidth]{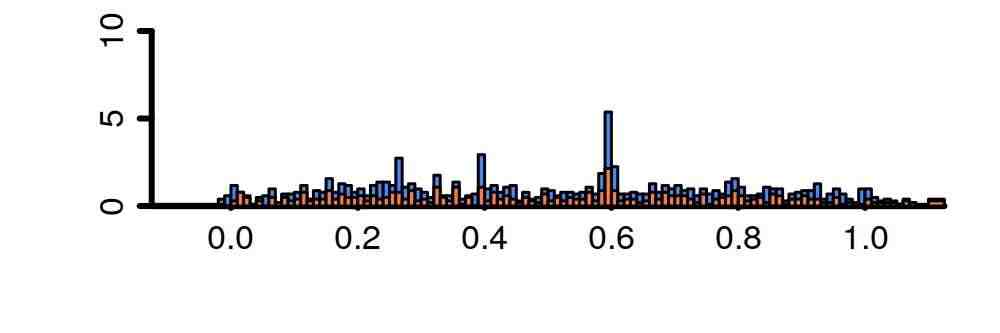}
\vspace*{-6ex}
\end{center}
\caption{The situation of Figure \ref{figure:random-functions-own-train-3-he} for the initialization
method considered in Figure \ref{figure:random-functions-own-train-1-own}.
At least for the two larger architectures the target function is well approximated, and for $m=1024$
this is almost instantly achieved.
}\label{figure:random-functions-own-train-3-own}
\end{figure}

Let us summarize our findings we made so far:
If we wish to avoid neurons to be dead right after initialization
and we also want to allow weights $a_i$ arbitrarily close to $0$,
then 
we need initialize the biases $b_i$ with strictly positive values, see Theorem \ref{result:pos-bias-better}.
However, such an approach necessarily produces semi-active neurons, too, and the only way to 
control the fraction of the latter for fixed $P_a$
is to generate small values for $b_i$, only. This, however, forces the knots $x_i^*$ to be more concentrated 
around $0$, forcing the initial function $g(\cdot, w,c,a,b)$ of our network to be almost linear 
on the data set, see 
Figure \ref{figure:random-functions-1d}.
Finally, 
in the ``limiting'' case $b_i :=0$, the function 
$g(\cdot, w,c,a,b)$
is actually linear on the data set, and no neuron is fully active. In fact, with the usual 
setting $\partial_0 = 0$, half of the initialized neurons are dead.

Now recall that the goal of the learning process is to find parameters $w,c,a$, and $b$ such that 
the resulting $g(\cdot, w,c,a,b)$ approximates the unknown target function $\fpb$ well.
For most $\fpb$, such an approximation requires the corresponding  knots  to be spread over the input interval,
which in our case is $[0,1]$. Consequently, if we force the knots to be concentrated near zero for the 
reasons discussed above, then these knots need to be significantly moved during the training phase.
This raises the question, whether such initializations really produce good starting points for the 
training process, or to phrase it differently: 
\begin{description}
   \item[Q2.] Are there other initialization strategies that ensure both a large fraction of 
   fully active neurons and a somewhat uniform distribution of the knots?
   \item[Q3.] Do such initializations produce better starting points for the training process?
\end{description}
Let us first consider \textbf{Q2}. Our discussion above showed that the conventional 
initialization strategies can only partially ensure both goals simultaneously. On the other hand,
these initialization strategies actually focus on initializing the weights and biases, whereas the location
of the knots is merely more than a side-product of this focus.
For a moment, let us therefore consider  the case, in which we begin with the distribution of the knots, instead.
For example, we could sample virtual knots $x_i^*$ according to the uniform distribution on $[0,1]$.
Using the formula $x_i^* = -b_i /a_i$, we then see that we either need to initialize $a_i$ or $b_i$. 
Moreover, the empirical success of \cite{HeZhReSu15a} suggests that initializing $a_i$
as in Example \ref{example:He-init-1d} should be kept. Following this, we would then initialize 
the biases by $b_i := - a_i x_i^*$.
Obviously, for data sets with $[\xmin,\xmax] = [0,1]$, this new initialization strategy 
almost surely  produces fully active neurons as well as uniformly distributed knots.
In other words,
both aspects of \textbf{Q2} are fully satisfied
and a comparison between Figure \ref{figure:random-functions-1d}
and Figure \ref{figure:random-functions-own} shows that the resulting initial predictors
are less biased towards a linear behavior.

Let us therefore investigate, whether the new initialization strategy also positively answers 
\textbf{Q3}.
Since later in Section \ref{sec:general} we will investigate similar initialization strategies 
in more detail, we restrict our considerations to three toy examples illustrated in Figures
\ref{figure:random-functions-own-train-1-he} to
\ref{figure:random-functions-own-train-3-own}.
These Figures show that the new initialization strategy leads in basically all considered cases to  a
faster learning of the target function than initializing with \heetal\ and zero biases does.
Moreover,  \heetal\ with zero biases seems to have serious problems when a good approximation
of the target function requires knots being located further away from $0$. As a consequence, 
some target functions could not be learned sufficiently well with this initialization method.
Based on these initial promising findings, we will generalize the new initialization method to higher dimensions.

\section{The General situation}\label{sec:general}

The goal of this section is to generalize the initialization strategy discussed at the end of Section 
\ref{sec:one-d-sit} to higher dimensions and deeper networks.
To this end, we consider throughout this section a single hidden layer within a deep architecture.
To be more precise, we assume that this hidden layer follows a layer with $d$ neurons, i.e.~$d = m_{l-1}$ and that 
the layer itself has $m$ neurons, i.e.~$m = m_l$. In particular, if the considered hidden layer is the first hidden layer,
then $d$ equals the dimension of the input space.
Moreover, to avoid notational overload, we denote the data that goes into the considered layer by $x_1,\dots, x_n$.
In particular, we have $x_j \in \R^d$, and if the considered layer is not the first hidden layer, 
the non-negativity of the ReLU-functions applied in the previous layer  actually ensures 
\begin{align}\label{data:non-negative}
 x_j \in [0,\infty)^d\, , \qquad \qquad i = 1,\dots,n.
\end{align}
To avoid a cumbersome distinction of cases, we assume in the following that \eqref{data:non-negative}
also holds for the first hidden layer, whenever the require \eqref{data:non-negative} for our results.
Now, the considered hidden layer consists of $m$ neurons of the form
\begin{align}\nonumber
 h_i:\R^d & \to [0,\infty)\\ \label{neuron-general}
 x&\mapsto  \relu{\langle a_i, x\rangle + b_i} \, ,
\end{align}
where $a_1,\dots,a_n \in \R^d$ and $b_1,\dots,b_m \in \R$ are the weight vectors and biases of these neurons.
To address \textbf{Q2}, which asks for ``a large fraction of fully active neurons and a somewhat uniform distribution
of the knots'', our first goal needs to be a translation of   ``fully active neurons'' and ``knots''.

 Let us begin with the latter notion. To this end, we note that in the one-dimensional case $d= 1$
 the knot is  defined by the equation $a_i x_i^* + b_i = 0$, and the obvious generalization to $d>1$ is 
 \begin{align*}
  x_i^* := \bigl\{ x\in \R^d: \langle a_i, x\rangle + b_i = 0\bigr\}
 \end{align*}
provided that $a_i \neq 0$. Clearly, $x_i^*$ is the affine hyperplane that separates the   two 
sets
 \begin{align*}
  \Aip &:= \{ x\in \Rd: \langle a_i,x\rangle + b_i > 0\} \\
  \Aim &:= \{ x\in \Rd: \langle a_i,x\rangle + b_i < 0\}\, .
 \end{align*}
In the following, we call $x_i^*$ the \emph{edge of the neuron} $h_i$,
and $\Aip$, $\Aim$ its \emph{region of activity} and \emph{inactivity}, respectively.
In the one-dimensional case 
the   region of activity of a neuron with $a_i>0$ is $(x_i^*, \infty)$, see \eqref{left-neuron},
while its region of inactivity is $(-\infty, x_i^*)$.  
With this information it is easy to see that the following definition 
generalizes the one-dimensional case considered in Definition \ref{def:neuron-states}.

\begin{definition}
 Let $D = (x_1,\dots,x_n)$ be a data set in $\Rd$ and
%  \begin{displaymath}
%   \ico D := \biggl\{ x\in \R^d: \exists k\geq 2,  y_j \in A, \lb_j >0 \mbox{ with }  \lb_1+\dots+\lb_k = 1 \mbox{ and } y=  \sum_{j=1}^k \lb_j y_j   \biggr\}
%  \end{displaymath}
 $h_i:\R^d\to [0,\infty)$ be a neuron of the form \eqref{neuron-general}
 with $a_i \neq 0$.
 Moreover, let
 $x_i^*, \Aip$, and $\Aim$ be as above.
% 
% with weight vector $a_i\in \Rd\setminus\{0\}$, bias $b_i\in \R$ and edge $x_i^*$.
 Then we say that  $h_i$ is:
\begin{enumerate}
 \item \emph{Fully active}, if we have $D\cap \Aip \neq \emptyset$ and $D\cap \Aim \neq \emptyset$.
 \item \emph{Semi-active}, if    $D \subset  x_i^* \cup A_i^+$ and $D\not\subset x_i^*$ hold.
 \item \emph{Inactive}, if   $D \subset  x_i^* \cup  A_i^-$ holds.
\end{enumerate}
\end{definition}

Note that each neuron with $a_i\neq 0$
is in exactly one of these states.
Our next goal is provide an alternative characterization of fully active neurons, which in the sequel 
make it possible to describe initialization strategies. To this end,
recall that the convex hull $\co A$ of a set $A\subset \R^d$ is 
the smallest convex set containing the set $A$. For a finite set $A=\{y_1,\dots,y_k\}$ we further define 
% \begin{displaymath}
%  \co A = \biggl\{ y\in \R^d: \exists  \lb_1,\dots,\lb_k \geq 0 \mbox{ with } \lb_1+\dots+\lb_k = 1 \mbox{ and } y=  \sum_{j=1}^k \lb_j y_j   \biggr\}\, .
% \end{displaymath}
% Moreover, for such $A$ we define 
\begin{displaymath}
 \ico A: = \biggl\{ y\in \R^d: \exists  \lb_1,\dots,\lb_k > 0 \mbox{ with } \lb_1+\dots+\lb_k = 1 \mbox{ and } y=  \sum_{j=1}^k \lb_j y_j   \biggr\}\, . 
\end{displaymath}
It can be shown that $\ico A$ is the interior of $\co A$ relative to the affine hull of $A$, but since
we do not need this, we skip the details. Moreover, we clearly have 
$\ico A \subset \co A$ and equality only holds if $|A| = 1$. Moreover, it is not hard to 
see that $\ico A$ is convex and that 
the closure of $\ico A$ equals $\co A$, that is 
$\overline {\ico A} = \co A$. Finally, for a data set $D = (x_1,\dots,x_n)$ we write 
$\ico D := \ico \{x_1,\dots,x_n\}$. The next lemma characterizes fully active neurons with the help of 
$\ico D$.

\begin{lemma}\label{result:ico-char}
 Let $D = (x_1,\dots,x_n)$ be a data set in $\Rd$ with $n\geq 2$ and
 $h_i:\R^d\to [0,\infty)$ be a neuron of the form \eqref{neuron-general} with $a_i\neq 0$ and edge
 $x_i^*$.
 Then the following statements are equivalent:
\begin{enumerate}
 \item The neuron $h_i$ is fully active.
 \item We have both  $x_i^* \cap \ico D\neq \emptyset$ and $\ico D\not\subset x_i^*$.
\end{enumerate}
\end{lemma}

Our next goal is to generalize Corollary \ref{result:one-layer-states},
which described how the state of a neuron influences its behavior on the data set.
Clearly, if a neuron $h_i$ is inactive, then we have  $h_i(x_j) = 0$
for all $j=1,\dots,n$, and if $h_i$ is semi-active, then 
$h_i(x_j) = \langle a_i, x_j\rangle + b_i$ for all $j=1,\dots,n$. 
Consequently, the remarks made after Corollary \ref{result:one-layer-states} remain 
valid for these types of neurons. The next lemma shows that the assertion of  
Corollary \ref{result:one-layer-states}  for fully active neurons is also true 
in the case $d>1$.

\begin{lemma}\label{result:gen-layer-states}
 Let $D = (x_1,\dots,x_n)$ be a data set in $\Rd$ with $n\geq 2$
 for which there is a $j_0\in \{1\dots,n\}$ with $x_{j_0} \in \ico D$.
 Moreover, let
 $h_i:\R^d\to [0,\infty)$ be a neuron of the form \eqref{neuron-general} with $a_i\neq 0$.
 Then the following statements are equivalent:
 \begin{enumerate}
  \item The neuron $h_i$ is fully active.
  \item The neuron $h_i$ does not behave linearly on $D$, that is, for all $\tilde a\in \Rd$, $\tilde b\in \R$, there exists a
  $j\in \{1\dots,n\}$ such that 
	\begin{displaymath}
	   h(x_j) \neq \langle \tilde a, x_j\rangle + \tilde b\, .
	\end{displaymath}
 \end{enumerate}
%  
% 
% 
%  Let $D = (x_1,\dots,x_n)$ be a data set with $x_j \in \R^d$ for all $i=1,\dots,n$, 
%  and $h_i:\R^d\to [0,\infty)$ be a neuron 
% with weight vector $a_i\in \Rd$, bias $b_i\in \R$ and edge $x_i^*$. Then the following statements are true:
%  \begin{enumerate}
%   \item If $h_i$ is fully active 
%   and there is a $j_0\in \{1\dots,n\}$ with $x_{j_0} \in \ico D$, then 
%   $h_i$ does not behave linearly on $D$, that is, for all $\tilde a\in \Rd$, $\tilde b\in \R$, there exists a
%   $j\in \{1\dots,n\}$ such that 
% 	\begin{displaymath}
% 	   h(x_j) \neq \langle \tilde a, x_j\rangle + \tilde b\, .
% 	\end{displaymath}
%  \end{enumerate}
\end{lemma}

Our next goal is to investigate initialization strategies that initialize each weight 
vector $a_i$ by some probability distribution $P_a^d$ on $\Rd$, that is, each coordinate 
of $a_i$ is independently sampled from the distribution $P_a$ on $\R$. As in the one-dimensional
case, we assume that $P_a$ is symmetric and satisfies $P_a(\{0\}) = 0$.
Obviously, the latter implies $P_a^d(\{0\}) = 0$ and some simple considerations
% using e.g., left-closed-right-open intervals in $\Rd$,  
 show that 
$P_a^d$ is symmetric in the sense of $P_a^d(A) = P_a^d(-A)$ for all measurable $A\subset \Rd$.
% We refer to   Lemma \ref{result:appendix:symm-d} for details.

In the following two remarks we investigate the size and the direction of the initialized 
weight vector, respectively. To this end, we assume that we have 
i.i.d.~random variables  $A_1,\dots,A_d$ with $A_i \sim P_a$, where $P_a$ is as above.
In other words, the random variables $A_1,\dots,A_d$ describe our 
 random initialization of a single neuron, say $h_1$.
 We additionally assume $\var A_i < \infty$ and write $A:= (A_1,\dots,A_d)$.

\begin{figure}[t]
\begin{center}
\includegraphics[width=0.32\textwidth]{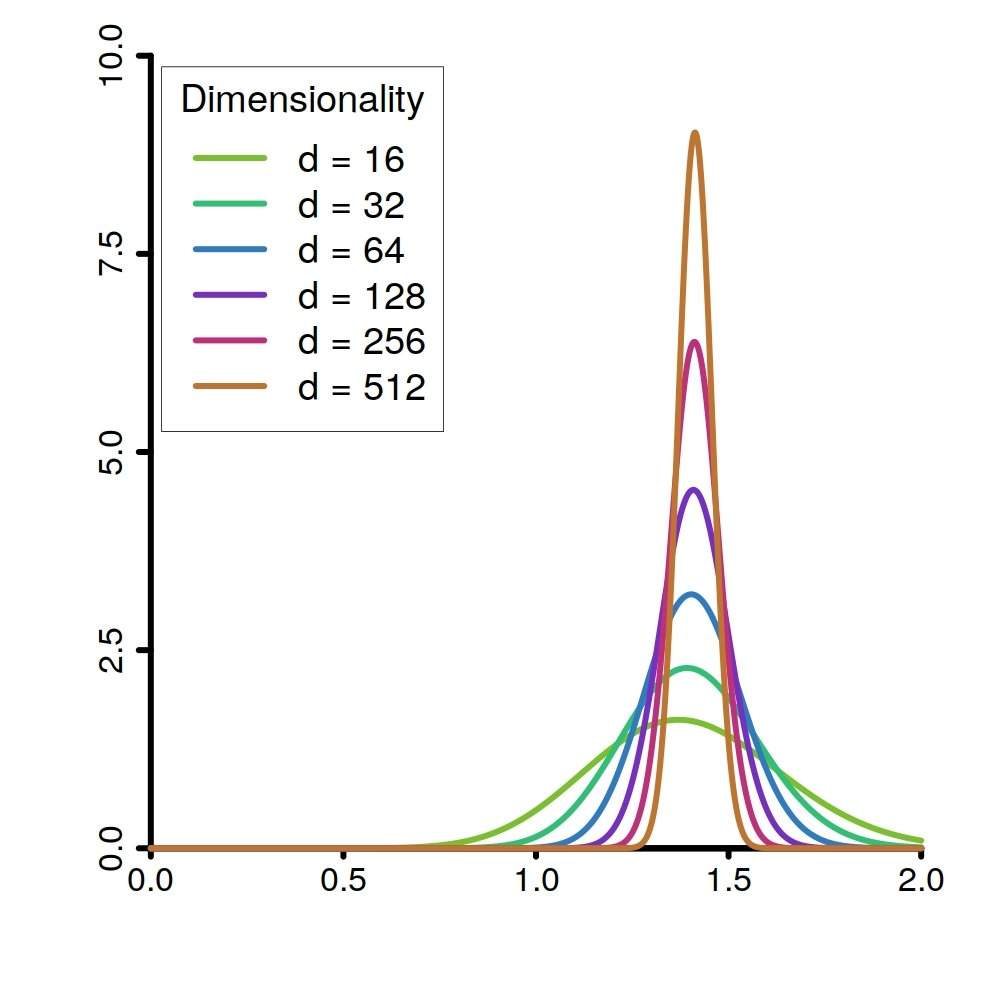}
\hspace*{-0.01\textwidth}
\includegraphics[width=0.32\textwidth]{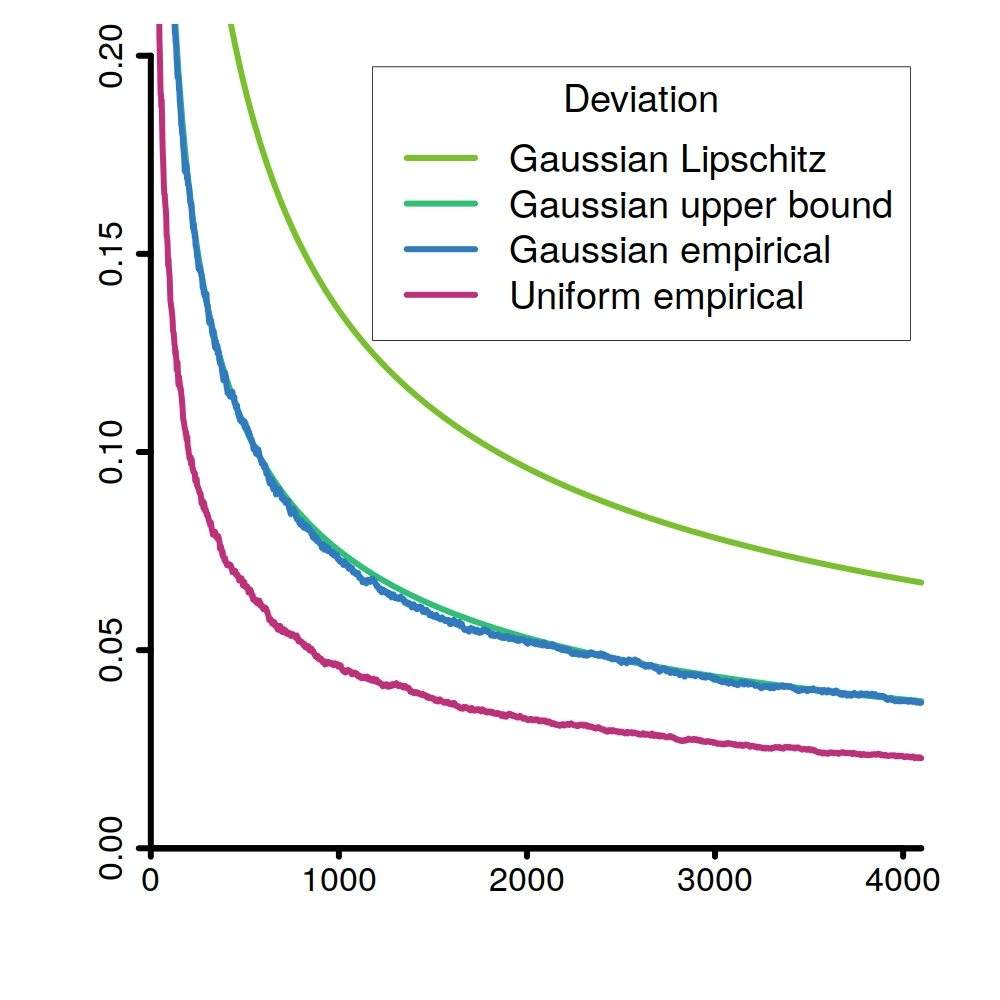}
\hspace*{-0.01\textwidth}
\includegraphics[width=0.32\textwidth]{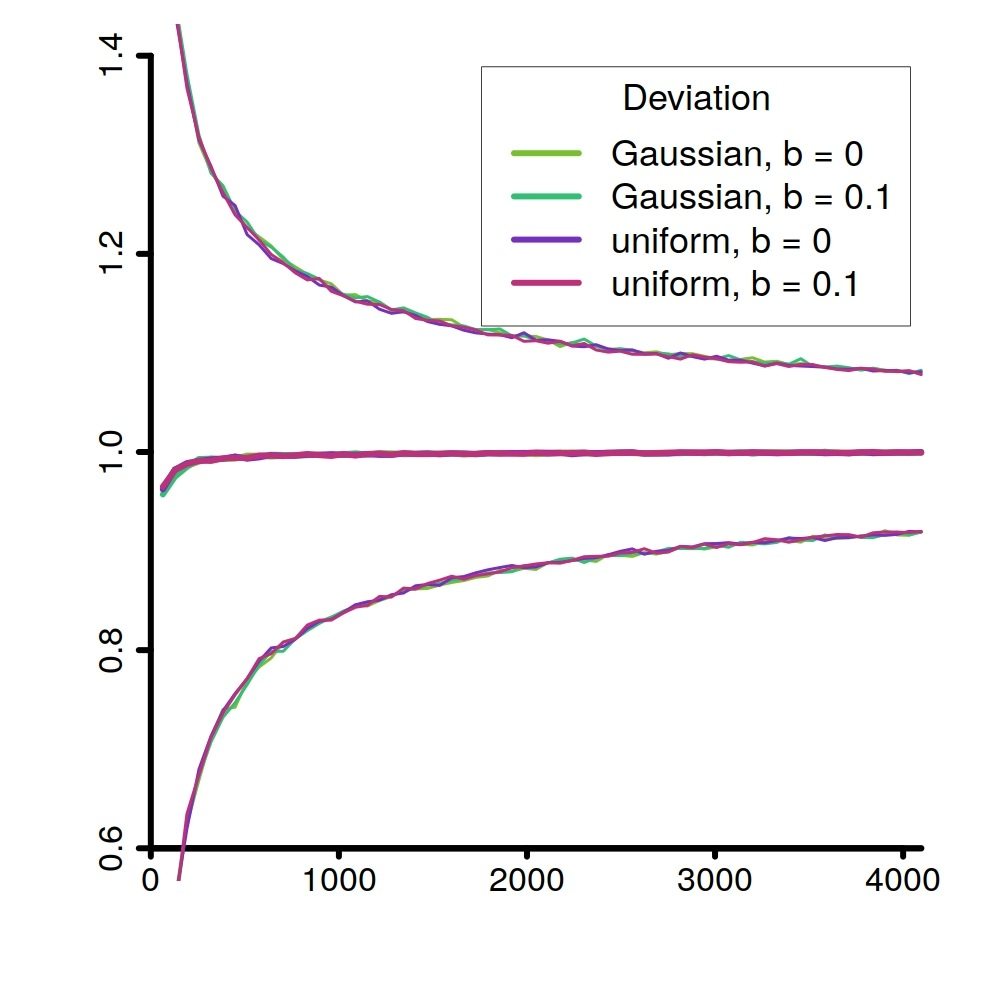}

\end{center}
\caption{Left: Densities of $\snorm A_2$, where $A=(A_1,\dots,A_d)$ is a vector of i.i.d.~random
variables with $A_i \sim \ca N(0, 2/d)$.
Middle: Plots for estimates of the smallest $\d>0$ satisfying $P( \snorm A_2    \geq  \sqrt 2 +  \d) \leq 0.01$ for $d=1,\dots,4096$. The descriptors ``Gaussian Lipschitz'' and ``Gaussian upper bound''
refer to the theoretical estimates \eqref{conc-norm-lipschitz} and 
\eqref{conc-norm-gauss}, respectively. The two
empirical estimates are based on 50.000 repetitions.
The upper bound  \eqref{conc-norm-gauss}
is on average    $2\%$ off. 
In contrast, \eqref{conc-norm-lipschitz} captures the asymptotics but is, on average, 
by a factor of about 1.8 too large.
Right: Average value and $1\%$, respectively $99\%$ percentile
of $ (\frac dm)^{1/2}{\snorm{(h_1(x), \dots, h_m(x))}_2}$ for a fixed 
input vector $x\in \R^d$ with $d=64$ and  $m= 1,\dots,4096$. The values are empirical
estimates based upon 10.000 repetitions.
}\label{figure:2d-distributions}
\end{figure}

\begin{remark}[Size of the weight vector]\label{remark:norm-weight}
%  Let $A_1,\dots,A_d$ be i.i.d.~random variables that are symmetric, with $P(\{A_i = 0\})  =0$.
%  Again we  assume, that the random variables $A_1,\dots,A_d$ describe our 
%  random initialization of a single neuron, say $h_1$. Unlike in Remark \ref{remark:pos_orthant}
%  we are now interested in the Euclidean norm of weight vector $A:=(A_1,\dots,A_d)$  
%  of $h_1$. To this end, 
In the following we investigate the size $\snorm A_2$ of the random weight vector $A$
for the initialization method \heetal.
To this end, we first note that the random variables  $Z_i :=  \frac d2  A_i^2$ are i.i.d.~with $\E Z_i = \frac d2 \var A_i = 1$,
and hence the strong law of large numbers shows that, for $d\to \infty$, we have 
\begin{displaymath}
 \snorm A_2^2 = \sum_{i=1}^d A_i^2 = \frac 2 d \sum_{i=1}^d Z_i \to 2 \qquad \qquad \mbox{almost surely.}
\end{displaymath}
In other words,  for sufficiently large $d$ we have $\snorm A_2 \approx \sqrt 2$. Under additional assumptions
on $A_i$ this approximation can be also quantified. For example, 
if we have a symmetric sub-Gaussian random variable $Y$ and assume that $A_1,\dots,A_d$ are independent copies of 
$\a Y$ with $\a^2 := \frac 2 {d \var Y}$, then Theorem \ref{result:euclid-norm-conc}
applied to $X_i := \sqrt{d/2}\cdot A_i \sim (\var Y)^{-1/2} \cdot Y$ yields 
	\begin{displaymath}
	   P\Bigl( \bigl| \,\snorm X_2 - \sqrt d \,\, \bigr| \geq t  \Bigr) \leq 2 \exp \bigl( - C_Y t^2\bigr)  \, ,
		\qquad \qquad t>0,
	\end{displaymath}
 where $C_Y$ is a constant only depending on $\var Y$ and the sub-Gaussian norm $\sgnorm Y$ of $Y$.
Using the definition of $X_i$ and $\t := C_Y t^2$, we thus find 
	\begin{align}\label{conc-norm-general}
	   P\biggl( \Bigl| \,\snorm A_2 - \sqrt 2 \,\, \Bigr| \geq \sqrt{\frac {2\t}{C_Y d}} \,\, \biggr) \leq 2 \eul^{ - \t }  
	\end{align}
for all $d\geq 1$ and $\t>0$.

 Let us  finally
 consider the specific case $A_i \sim \ca N(0, \s^2)$. 
 Combining Lemma \ref{result:expected-euclid-norm}
 with Lemma \ref{result:gautschi} we then find 
 \begin{displaymath}
  \s \sqrt {d-1/2}  \leq \E \snorm A_2 \leq  \s  \sqrt {d-1/4} 
  \qquad \mbox{ and } \qquad 
   \s^2/ 4 \leq    \var  \snorm A_2 \leq \s^2 /2\, ,
 \end{displaymath}
 and for the choice $\s^2 = 2 / d$ of the strategy \heetal\ we thus have 
 \begin{displaymath}
  \sqrt{2} \cdot \sqrt {1- \frac 1 {2d}}   \leq \E \snorm A_2 \leq  \sqrt{2} \cdot \sqrt {1- \frac 1 {4d}}
    \qquad \mbox{ and } \qquad 
   \frac 1 {2d}\leq    \var  \snorm A_2 \leq \frac 1 d\, .
 \end{displaymath}
 In particular, we have $\E \snorm A_2 < \sqrt 2$ for all $d\geq 1$, but 
$\E \snorm A_2 \to \sqrt 2$ for $d\to 0$. In fact,  even for moderate sizes of $d$ we
 actually have $\E \snorm A_2 \approx \sqrt 2$. For example, for $d=64$ the estimates guarantee
 $0.996 \cdot \sqrt 2 \leq \E \snorm A_2 \leq 0.9981\cdot \sqrt 2$.
 In addition, \eqref{conc-norm-general} can be made more explicit. For example, 
 a well-known concentration inequality for Lipschitz continuous functions acting on a standard 
 normal vector, see e.g.~\cite[Inequalities (A.5)]{Chatterjee14}, shows
 \begin{align}\label{conc-norm-lipschitz}
   	   P\biggl(  \snorm A_2    \geq  \E \snorm A_2 +  \sqrt{\frac {\t}{d}} \,\, \biggr) \leq  \eul^{ - \t/4 }  
   	   \qquad 
   	   \mbox{ and }
   	   \qquad
  	   P\biggl( \snorm A_2 \leq  \E \snorm A_2  -  \sqrt{\frac {\t}{d}} \,\, \biggr) \leq   \eul^{ - \t/4 }  \, .
 \end{align}
 However, the Lebesgue density of the random variable $\snorm A_2$ can also be explicitly computed, see Lemma \ref{result:expected-euclid-norm}
 for details, and  Figure \ref{figure:2d-distributions} shows  the shape of this density for different values of $d$.
	For example, 
	using this explicit form of the density, we see by considering 
	Equation \eqref{dev-euclid-norm} of Lemma \ref{result:expected-euclid-norm}
	 for $\d\geq -1$, $\s^2 = 2 / d$ and 
	 $s := \sqrt 2 + \d$, that  
% 	 $s := \sqrt{(1+\d)2}$, that  
	\begin{align}\label{conc-norm-gauss-1}
	 P\bigl( \snorm A_2 \geq \sqrt 2 + \d\,\bigr)  
% 	   P\bigl( \snorm A_2 \geq \sqrt 2 + \d/\sqrt 2 \,\bigr)  
% 	\leq 
%   P\bigl( \snorm A_2 \geq \sqrt{1+\d}\cdot \sqrt 2\,\bigr) 
%   = \frac {\Gamma(\frac d2, (1+\d)\frac d2)}{\Gamma(\frac d2)}\, ,
= \frac {\Gamma(\frac d2, (1+\sqrt 2\d+ \frac{\d^2}2)\cdot\frac d2)}{\Gamma(\frac d2)}\, , 
	\end{align}
	where $\G(\cdot,\cdot)$ and $\G(\cdot)$ denote the (incomplete) gamma function.
% 	and where we used $\sqrt{1 + \d} \leq 1 + \d/2$.
	Combining 
Stirling's formula \eqref{gamma-stirling} for the gamma function with 
\eqref{result:incomplete-gamma-bound-refined-upper}
% Lemma \ref{result:incomplete-gamma-bound}
we further have for $d\geq 3$ and $\a>0$:
\begin{align*}%\label{conc-norm-gauss-2}
  \frac {\Gamma(\frac d2, (1+\a)\frac d2)}{\Gamma(\frac d2)} 
\leq  \frac{2}{2+d\a} \cdot   \biggl( \frac{(1+\a) d}2 \biggr)^{d/2} \eul^{-\frac{(1+\a) d}2} 
 \sqrt{\frac{d}{\pi}} \cdot \Bigl(\frac{2\eul}{d}  \Bigr)^{\frac d2}  
 = 
\frac 1 {\sqrt\pi} \cdot  \frac{\sqrt d}{2+d\a} \cdot \biggl(\frac{ 1 + \a}{\eul^\a}  \biggr)^{\frac d2}   \, , 
\end{align*}
and applying this estimate for $\a := \sqrt 2\d+ \frac{\d^2}2$ in \eqref{conc-norm-gauss-1} gives 
\begin{align}\label{conc-norm-gauss}
P\bigl( \snorm A_2 \geq \sqrt 2 + \d\,\bigr)  
    \leq 
    \frac 1 {\sqrt\pi} \cdot  \frac{2 \sqrt d}{4+ 2 \sqrt 2 d\d+ d {\d^2}} \cdot \biggl(\frac{ 1 + \sqrt 2\d+  {\d^2}/2}{\eul^{\sqrt 2\d+  {\d^2}/2}}  \biggr)^{\frac d2}
\end{align}
Similar considerations can be made for the the probability of $\snorm A_2 \geq \sqrt 2 + \d$,
and some simple empirical experiment suggest that this probability has behavior that is 
a very similar to the one for the upper bound.
We skip the  the details but refer to Figure \ref{figure:2d-distributions} for a comparison between 
\eqref{conc-norm-lipschitz}, \eqref{conc-norm-gauss}, and empirically found bounds.
\remarkend\end{remark}

\begin{remark}[Direction of the weight vector]
%     Let $A_1,\dots,A_d$ be i.i.d.~random variables that are symmetric, with $P(\{A_i = 0\})  =0$.
%  Again we  assume, that the random variables $A_1,\dots,A_d$ describe our 
%  random initialization of a single neuron, say $h_1$. 
 Unlike in Remark \ref{remark:norm-weight}
 we are now interested in the Euclidean direction of weight vector $A:=(A_1,\dots,A_d)$  
 of $h_1$. To this end, we denote the Euclidean sphere in $\Rd$ by $\Sd$, that is 
 $\Sd := \{x\in \Rd: \snorm x_2 = 1\}$. Moreover, we write $\sd$ for the surface measure on $\Sd$.
In particular, we have, see e.g.~\cite[Beispiel §14.9]{Forster17}
\begin{align}\label{area-sphere}
   \sd (\Sd) = \frac{2 \pi^{d/2}}{\Gamma(d/2)} = d \vol_d(B_{\ell_2^d})\, ,
\end{align}
and it is well-known that by normalizing $\sd$ we obtain the uniform distribution on $\Sd$.

Let us first consider the case  $A_i \sim \ca N(0, \s^2)$ for some $\s>0$.
Then it is well-known, see e.g.~\cite[page 227]{Devroye86}, that the normalized vector 
$  A/\snorm A_2$ is uniformly distributed on $\Sd$.
Consequently,  all orientations of the 
 hyperplanes described by the weight vector $A$ are equally likely.

Let us now consider the case $A_i \sim \ca U[-\a,\a]$ for some $\a>0$. Then 
$f := (2\a)^{-d} \eins_{[-\a,\a]^d}$ is the Lebesgue density of distribution of $A:=(A_1,\dots,A_d)$, and
Theorem \ref{result:projection-measure} shows that the $\sd$-density of the distribution 
of the normalized vector $A/\snorm A_2$ is given by 
		\begin{displaymath}
		   h(\xi) = \int_0^\infty f(r \xi) \, r^{d-1}\, dr \, , \qquad \qquad \xi \in \Sd.
		\end{displaymath}
Now observe that we have $r\xi \in [-\a,\a]^d$ if and only if $r \inorm \xi \leq \a$, and
hence we obtain %$\eins_{[-\a,\a]^d}(r\xi) = \eins_{[0, \a / \inorm \xi]}(r)$.
\begin{displaymath}
    h(\xi) 
= (2\a)^{-d} \int_0^\infty \eins_{[-\a,\a]^d}(r\xi) \, r^{d-1}\, dr  
= (2\a)^{-d} \int_0^{\a / \inorm \xi}   r^{d-1}\, dr
= \frac 1 {d \, 2^d \, \snorm \xi_\infty^d}\, , \qquad \qquad \xi \in \Sd.
\end{displaymath}
In particular, the distribution of $A/\snorm A_2$ is independent of $\a$ and does \emph{not} 
equal the uniform distribution on $\Sd$. In fact, since we have $d^{-1/2} \leq \inorm \xi \leq 1$ for all 
$\xi \in \Sd$, we find 
\begin{displaymath}
   \frac 1 {d \, 2^d} \leq h(\xi) \leq \frac {d^{d/2}} {d \, 2^d}
\end{displaymath}
and both the lower and the upper bound are attained. In fact, for 
disjoint $J_+, J_-\in \{1,\dots,d\}$ and 
\begin{displaymath}
 \xi_{J_+, J_-} :=  \frac 1 {\sqrt l} \biggl(\,\sum_{j\in J_+} e_j -  \sum_{j\in J_-} e_j\biggr)\, ,
\end{displaymath}
where $l := |J_+\cup J_-|$, we have both 
$\snorm {\xi_{J_+, J_-} }_2 = 1$ and $\inorm {\xi_{J_+, J_-} } = l^{-1/2}$, and hence the above formula reduces to 
\begin{align}\label{xi-direction}
   h(\xi_{J_+, J_-} ) = \frac {l^{d/2}} {d \, 2^d}\, .
\end{align}
For $l=1$, respectively $l=d$, the lower and upper bound are thus attained. Let us investigate the relation between
$h$ and the uniform distribution on $\Sd$ in a bit more detail. To this end, let $g$ be the density of the uniform
distribution with respect to $\sd$. Equation \eqref{area-sphere} then shows 
\begin{displaymath}
   g(\xi) =  \frac{\Gamma(d/2)}{2 \pi^{d/2}}\, , \qquad \qquad \xi\in \Sd.
\end{displaymath}
Now using 
Stirling's formula \eqref{gamma-stirling} for the gamma function
we have 
\begin{displaymath}
   \Gamma(d/2) = 2 \sqrt{\frac{\pi}{d}} \cdot \Bigl(\frac{d}{2\eul}  \Bigr)^{d/2} \cdot \eul^{\mu(d)}\, ,
\end{displaymath}
where $\mu(d)$ satisfies $0<\mu(d) < \frac 1{6d}$. Consequently, $\xi\in \Sd$ satisfies $g(\xi) = h(\xi)$
if and only if 
\begin{displaymath}
   \frac 1 {d \, 2^d \, \snorm \xi_\infty^d} =  \sqrt{\frac{\pi}{d}} \cdot \Bigl(\frac{d}{2\eul \pi}  \Bigr)^{d/2} \cdot \eul^{\mu(d)}\, , 
\end{displaymath}
and the latter is equivalent to 
\begin{align*}
   \snorm \xi_\infty 
= \Biggl(\frac 1 {d \, 2^d } \cdot  \sqrt{\frac{d}{\pi}} \cdot \Bigl(\frac{2\eul \pi} {d} \Bigr)^{d/2} \cdot \eul^{-\mu(d)}\Biggr)^{1/d}
 = \sqrt{\frac{\eul \pi} {2} } \cdot  (\pi d)^{-\frac 1 {2d}} \cdot  \eul^{-\nu(d)}   \cdot \frac 1 {\sqrt d}\, ,
\end{align*}
where $\nu(d)$ satisfies $0<\nu(d) < \frac 1{6d^2}$. Now, some numerical calculations show 
$\sqrt{\frac{\eul \pi} {2} }\approx 2.066365676$ and it is well known that 
$(\pi d)^{-\frac 1 {2d}} \cdot  \eul^{-\nu(d)} \leq 1$ for all $d\geq 1$ and
$(\pi d)^{-\frac 1 {2d}} \cdot  \eul^{-\nu(d)} \to 1$ for $d\to \infty$. In fact, four our purposes, this convergence is 
somewhat fast, for example for $d\geq 86$, respectively $d\geq 1024$, we already have 
\begin{displaymath}
   \sqrt{\frac{\eul \pi} {2} } \cdot  (\pi d)^{-\frac 1 {2d}} \cdot  \eul^{-\nu(d)}  > 2
\qquad \mbox{ and } \qquad
\sqrt{\frac{\eul \pi} {2} } \cdot  (\pi d)^{-\frac 1 {2d}} \cdot  \eul^{-\nu(d)}   >
2.05823276  \, .
\end{displaymath}
If $d\geq 86$ and $\inorm \xi\leq 2/\sqrt d$, we thus find $h(\xi) > g(\xi)$, and consequently
such directions $\xi$ are preferred when sampling $A_i \sim \ca U[-\a,\a]$ instead of sampling
$A_i \sim \ca N(0, \s^2)$. Conversely, for all $d\geq 1$ our calculations above  show that 
$\inorm \xi\geq 2.06636568/\sqrt d$ implies $h(\xi) < g(\xi)$ and hence such directions $\xi$
are disrated by sampling $A_i \sim \ca U[-\a,\a]$ compared to the sampling
$A_i \sim \ca N(0, \s^2)$.

In particular, if the previous layer was sufficiently wide in the sense of $d\geq 86$, then the 
directions $\xi_{J_+, J_-}$ given by \eqref{xi-direction} are preferred if 
$l \geq d/4$
and disrated if $l \leq d/4.2698672$.

% These simple calculation already indicate that this initialization strategy strongly favors directions of $A$
% that are along the ``diagonals'' $\R \cdot (\e_1 e_1 + \dots + \e_d e_d)$, where $\e_i = \pm 1$.
% Indeed, the likelihood for \emph{one} such  direction is $d^{d/2}$-times larger than that for a direction 
% $\e_i e_i$
% along one
% axis $\R e_i$. Moreover, there are $2^d$ such diagonal directions, but only $2d$ directions along an axis.
\remarkend\end{remark}

\begin{remark}[Size of the output vector]
 In this remark, we again assume that the weights of neuron $h_i$ 
 are initialized by a realization of the vector $A_i=(A_{i,1},\dots,A_{i,d})$.
 In addition, we consider an input sample $x=(x_1,\dots,x_d)$ and first ask for the distribution of the
 size of the initial output
 $(h_1(x), \dots, h_m(x))$. To be more precise, we have   
 \begin{displaymath}
 h_i (x) =  \biggl| \sum_{k=1}^d A_{i,k} x_k + b_i \biggr|_+\, , \qquad \qquad i=1,\dots, m,
 \end{displaymath}
    where  we  assume that there is a $b\in \R$ with  $b_i=b$ for all $i=1,\dots,m$, and we are interested in the distribution of $\snorm{(h_1(x), \dots, h_m(x))}_2$. To this end let us fix 
      i.i.d.~symmetric random variables $A_{i,k}$ with $\var A_{i,k} = \s^2$.
%     $A_{i,k}\sim \ca N(0,\s^2)$. 
    Then, the random variables $Y_i := \sum_{k=1}^d A_{i,k} x_k + b$ are 
    i.i.d.~with $\E Y_i= b$ and $\var Y_i = \s^2 \snorm x_2^2$.
    Moreover, we find 
        \begin{align*}
     \E \snorm{(h_1(x), \dots, h_m(x))}_2^2
     = \E \sum_{i=1}^m \biggl| \sum_{k=1}^d A_{i,k} x_k + b\biggr|_+^2 = \sum_{i=1}^m \E \relu{Y_i}^2 = m \E \relu{Y_1}^2 \, ,
%      =  \frac {m   \snorm x_2^2}{d} \, .
    \end{align*}
    and for $m\to \infty$, the strong law of large numbers gives
        \begin{align*}
     \frac{\snorm{(h_1(x), \dots, h_m(x))}_2^2} m 
     = \frac 1m\sum_{i=1}^m \biggl| \sum_{k=1}^d A_{i,k} x_k + b\biggr|_+^2 
     =  \frac 1m\sum_{i=1}^m  \relu{Y_i}^2 
     \to \E \relu{Y_1}^2  
    \end{align*}
    almost surely. In particular, for $b= 0$ 
    Lemma \ref{result:half-sym-distribution} shows that 
    $2\E \relu{Y_i}^2 =   \E Y_i^2 =  {\s^2 \snorm x_2^2}$,
    and for the choice $\s^2 = 2/d$ of the strategy 
    \heetal~we thus obtain 
    \begin{align*}
     \frac{\snorm{(h_1(x), \dots, h_m(x))}_2^2} m 
     \to \E \relu{Y_1}^2  = \frac {  \snorm x_2^2}{d}\, .
    \end{align*}
   With high probability, see Figure \ref{figure:2d-distributions} for some 
   empirical estimates,
   we consequently 
    have
    \begin{displaymath}
     \frac{\snorm{(h_1(x), \dots, h_m(x))}_2} {\sqrt m} \approx  \frac { \snorm x_2}{\sqrt d}\, .
    \end{displaymath}
   Note that if we define the \emph{normalized} Euclidean norm 
    on $\R^k$ by $\nsnorm x_2 := \frac {\snorm x_2}{\sqrt k}$, then the above approximation
    reads as $\nsnorm{(h_1(x), \dots, h_m(x))}_2  \approx \nsnorm x_2$.
    In other words, the size of the output of the layer is approximately equal to the 
    size of its input, if both are measured in $\nsnorm\cdot_2$.
%     
%     
%     that is,  the \emph{normalized} size of the output of the layer is approximately
%     equal to the \emph{normalized} size of its input.
    Clearly, this approximate equality remains unchanged by compositions of several layers, in other words
    the normalized output of sample $x_j$ at the $l$-th layer is approximately equal to the normalized 
    norm of $x_j$ at the input layer. 
    
    To investigate the case $b\neq 0$, we restrict our considerations to the case $A_{i,k}\sim \ca N(0,\s^2)$.
    Our previous considerations then show that $Y_i \sim \ca N(b,\t^2)$, where $\t^2 = \s^2 \snorm x_2^2$.
%     For the computation of $\E \relu{Y_i}^2$ 
%     we  write $f(t) := \relu t^2$ for $t\in \R$. Then $f$ is invertible 
%     on $[0,\infty)$ and we have $(f^{-1})'(t) = \frac 12 t^{-1/2}$ for all $t>0$.
    This yields 
        \begin{align*}
        \E \relu{Y_1}^2
     &= \frac 1 {\sqrt{2\pi \t^2}} \int_{0}^{\infty}  \eul^{-\frac{(s-b)^2}{2\t^2}} \, s^2 \, ds \\
     &= \frac 1 {\sqrt{2\pi \t^2}} \int_{-b}^{\infty}  \eul^{-\frac{s^2}{2\t^2}} \, (s + b)^2 \, ds \\
     &= \frac 1 {\sqrt{2\pi \t^2}} \int_{-b}^{\infty}  \eul^{-\frac{s^2}{2\t^2}} \, s^2   \, ds
     + \frac {2b} {\sqrt{2\pi \t^2}} \int_{-b}^{\infty}  \eul^{-\frac{s^2}{2\t^2}} \, s  \, ds
     + \frac {  b^2} {\sqrt{2\pi \t^2}} \int_{-b}^{\infty}  \eul^{-\frac{s^2}{2\t^2}} \,    ds\\
     & =  \frac {2\t^2} {\sqrt{\pi}} \int_{-\frac b{\sqrt 2\t}}^{\infty}  \eul^{- s^2} \, s^2   \, ds
     + 2b \sqrt{\frac{2\t^2}{\pi}} \int_{-\frac b{\sqrt 2\t}}^{\infty}  \eul^{- {s^2}} \, s  \, ds
     + \frac {  b^2} {\sqrt{ \pi }} \int_{-\frac b{\sqrt 2\t}}^{\infty}  \eul^{- s^2} \,    ds\, .
%      &= \frac 1 {2 \sqrt{2\pi \t^2}} \int_0^y  \eul^{-\frac{t}{2\t^2}} \cdot t^{-1/2}   \, dt \\
%       &= \frac 1 {2 \sqrt{\pi}} \int_0^{\frac y {2\t^2}}  \eul^{-t} \cdot t^{-1/2}   \, dt \\
%       & = \frac 12 \cdot  \frac {\G(\frac 1 2, \frac y {2\t^2})}{\G(\frac 12)} 
    \end{align*}
    Now, for $c>0$ we have 
    \begin{align*}
     \int_{-c}^{\infty}  \eul^{- {s^2} } \, s^2   \, ds 
     = 
     \int_{0}^{c}  \eul^{- {s^2} } \, s^2   \, ds 
     + \int_{0}^{\infty}  \eul^{- {s^2} } \, s^2   \, ds 
     &= \frac 12 \int_{0}^{c^2}  \eul^{- s} \, s^{1/2}   \, ds 
     + \frac 12\int_{0}^{\infty}  \eul^{- s} \, s^{1/2}    \, ds \\ 
     &= \G\Bigl(\frac 32\Bigr) - \frac 12 \G\Bigl(\frac 32, c^2\Bigr) \\
     & = \frac{\sqrt \pi}{2}   - \frac 14 \G\Bigl(\frac 12, c^2\Bigr) - \frac {c\, \eul^{-c^2} }2 \, ,
    \end{align*}
    where in the last step we used the well known identities $\G(x+1) = x\G(x)$,
    and $\G(1/2) = \sqrt \pi$, as well as the recurrence formula $\G(a+1,x) = a\G(a,x) + \eul^{-x}x^a$
    of the incomplete gamma function, see e.g.~\cite[Lemma A.1.1]{StCh08}.
    Moreover, we have
    \begin{align*}
     \int_{-c}^{\infty}  \eul^{- {s^2} } \, s   \, ds 
     = 
     -\int_{0}^{c}  \eul^{- {s^2} } \, s   \, ds 
     + \int_{0}^{\infty}  \eul^{- {s^2} } \, s   \, ds 
     =- \frac 12 \int_{0}^{c^2}  \eul^{- s}     \, ds 
     + \frac 12\int_{0}^{\infty}  \eul^{- s}     \, ds 
     = \frac12 \eul^{-c^2}
    \end{align*}
    and 
      \begin{align*}
     \int_{-c}^{\infty}  \eul^{- {s^2} }    \, ds 
     = 
     \int_{-\infty}^{c}  \eul^{- {s^2} }    \, ds 
    = \sqrt \pi \, \Phi\bigl(  {\sqrt 2} \cdot  c\bigr)\, , 
    \end{align*}  
    the $\Phi$ denotes the cumulative distribution function of $\ca N(0,1)$.
    By combining these equations for $c := \frac b{\sqrt 2\, \t}$, that is, 
    $c^2 = \frac {b^2}{  2\t^2}$, we obtain 
    \begin{align*}
     \E \relu{Y_1}^2
     &= \frac {2\t^2} {\sqrt{\pi}} \cdot  \biggl( \frac{\sqrt \pi}{2}   - \frac 14 \G\Bigl(\frac 12, \frac {b^2}{  2\t^2} \Bigr) - \frac {b\, \eul^{-{\frac {b^2}{  2\t^2}} }}{2 \sqrt 2 \t}   \biggr)
     + 2b \cdot \sqrt{\frac{2\t^2}{\pi}} \cdot \frac12 \cdot \eul^{-\frac {b^2}{  2\t^2}}
     +  \frac {  b^2} {\sqrt{ \pi }} \cdot \sqrt \pi \cdot \Phi\Bigl( \frac b{\t}\Bigr) \\
     & = \t^2 -
       \frac {\t^2} {2\sqrt{\pi}} \cdot     \G\Bigl(\frac 12, \frac {b^2}{  2\t^2} \Bigr) 
       -   {\t}  \cdot \frac {b\, \eul^{-{\frac {b^2}{  2\t^2}} }}{\sqrt {2 \pi}}   
     +2{\t}  \cdot \frac {b\, \eul^{-{\frac {b^2}{  2\t^2}} }}{\sqrt {2 \pi}}  
     +    b^2  \cdot \Phi\Bigl( \frac b{\t}\Bigr) \\
     & = \t^2 -
       \frac {\t^2} {2\sqrt{\pi}} \cdot     \G\Bigl(\frac 12, \frac {b^2}{  2\t^2} \Bigr) 
     +{\t}  \cdot \frac {b\, \eul^{-{\frac {b^2}{  2\t^2}} }}{\sqrt {2 \pi}}  
     +    b^2  \cdot \Phi\Bigl( \frac b{\t}\Bigr) \, .
    \end{align*}
    Now using  $\t^2 = \s^2 \snorm x_2^2$ and restricting our considerations to the strategy
    \heetal, that is 
    $\s^2 = \frac 2d$, we find
    \begin{align*}
     \E \relu{Y_1}^2
     &=  \frac{\snorm x_2^2}d \biggl( 2 -
       \frac {1} {\sqrt{\pi}} \cdot     \G\Bigl(\frac 12, \frac {b^2\, d}{  4\snorm x_2^2} \Bigr) \biggr)
     + \frac{\  \snorm x_2}{\sqrt d}   \cdot \frac {b}{\sqrt \pi} \cdot  \exp\Bigl( -\frac {b^2\, d}{  4\snorm x_2^2} \Bigr)
     +    b^2  \cdot \Phi\Bigl( \frac {b \, \sqrt d} {\sqrt 2\snorm x_2}\Bigr) \\
     &=  \nsnorm x_2^2  \cdot  \biggl( 2 -
       \frac {1} {\sqrt{\pi}} \cdot     \G\Bigl(\frac 12, \frac {b^2}{  4\nsnorm x_2^2} \Bigr) \biggr)
     +  \nsnorm x_2    \cdot \frac {b}{\sqrt \pi} \cdot  \exp\Bigl( -\frac {b^2 }{  4\, \nsnorm x_2^2} \Bigr)
     +    b^2  \cdot \Phi\Bigl( \frac {b } {\sqrt 2\, \nsnorm x_2}\Bigr)\\
     & = :\Psi(\nsnorm x_2 , b) \, .
     \end{align*}
     To obtain an intuitive understanding of this result, assume for a moment, that 
     the previous layer is actually the input layer, and that the 
     data was normalized during pre-processing, e.g.~to $[-1,1]^d$ or 
     $[0,1]^d$. 
     Then we have $\nsnorm x_2 \leq 1$ and 
     for $\nsnorm x_2 = 0$ we easily find $(\E \relu{Y_1}^2)^{1/2} = \sqrt{\Psi(0, b)} = b$.
     Moreover, for 
     e.g.~$b=0.1$
     some numerical calculations 
     show that $\nsnorm x_2\mapsto \sqrt{\Psi(\nsnorm x_2, b)} - \nsnorm x_2$ is monotonically decreasing
     on $[0,1]$ with $\sqrt{\Psi(1, b)} - 1 \approx 0.057323$.
     Consequently, such a moderate choice of $b>0$ does not lead to output vectors 
     whose normalized norm is significantly larger than $1$. For larger values of $b$, however,
     the influences may be more pronounced. For example, for $\nsnorm x_2 = 1$ and $b\to \infty$,
     the function $\sqrt{\Psi(1, b)}$ behaves like $\sqrt{2+b^2}$, that is, like $b$.
%     
%     
%     
%     
%     and 
%         \begin{align*}
%      \int_{-c}^{\infty}  \eul^{- {s^2} } \, s   \, ds 
%      = 
%      -\int_{0}^{c}  \eul^{- {s^2} } \, s   \, ds 
%      + \int_{0}^{\infty}  \eul^{- {s^2} } \, s   \, ds 
%      =- \frac 12 \int_{0}^{c^2}  \eul^{- s}     \, ds 
%      + \frac 12\int_{0}^{\infty}  \eul^{- s}     \, ds 
%      = \frac12 \eul^{-c^2}\, ,
%     \end{align*}
%     
%     
%     
%     For $y>0$ this 
%     gives 
%     \begin{align*}
%       P\bigl( 0< \relu {Y_i}^2 \leq y\bigr) 
%      =  P\bigl( f^{-1}(0)<  Y_i  \leq f^{-1} (y)\bigr) 
%      &= \frac 1 {\sqrt{2\pi \t^2}} \int_{f^{-1}(0)}^{f^{-1}(y)}  \eul^{-\frac{s^2}{2\t^2}} \, ds \\
%      &= \frac 1 {2 \sqrt{2\pi \t^2}} \int_0^y  \eul^{-\frac{t}{2\t^2}} \cdot t^{-1/2}   \, dt \\
%       &= \frac 1 {2 \sqrt{\pi}} \int_0^{\frac y {2\t^2}}  \eul^{-t} \cdot t^{-1/2}   \, dt \\
%       & = \frac 12 \cdot  \frac {\G(\frac 1 2, \frac y {2\t^2})}{\G(\frac 12)} 
%     \end{align*}
% 
%     
%     Habe die folgende ``version'' der substitutionsregel von rechts nach links benutzt
%     \begin{displaymath}
%      \int_a^b h(\phi^{-1}(t)) (\phi^{-1})'(t) dt = \int_{\phi^{-1}(a)}^{\phi^{-1}(b)} h(s) ds
%     \end{displaymath}
\remarkend\end{remark}

Our next goal is to investigate the effect of different initialization strategies for the 
offsets. 
We begin with the zero-bias initialization, that is, each $b_i$ is set to $b_i = 0$.
Note that in this case, $x_i^*$ is almost surely a linear subspace with $\dim x_i^* = d-1$, and this 
observation will significantly simplify our considerations below. 
In these considerations, we will require the dual cone of a set $A\subset \Rd$, which is defined 
by 
\begin{displaymath}
 A^\star := \bigl\{ y\in \Rd: \langle y, x\rangle \geq 0 \mbox{ for all } x\in A\bigr\}\, .
\end{displaymath}
Some properties of this and other geometric set construction are summarized in Appendix \ref{appendix:geometry}.
For now, we only recall that $A^\star$ is always a convex, closed cone, and that 
$A\subset B$ implies $B^\star \subset A^\star$.
Now assume that the neuron $h_i$ is inactive, that is $D\subset x_i^* \cup \Aim$. Since $b_i = 0$, this is equivalent
to 
\begin{displaymath}
 \langle a_i, x_j \rangle \leq 0\, , \qquad \qquad x_j \in D\, ,
\end{displaymath}
and the latter condition means $-a_i \in D^\star$. A similar consideration for semi-active 
neurons together with some considerations dealing with the condition $D\not\subset x_i^*$
leads to the following result, which is shown in Subsection \ref{subsec:proof-general}.

\begin{theorem}\label{result:zero-bias-general}
  Let $D = (x_1,\dots,x_n)$ be a data set in $\Rd$ in  which there exists a sample $x_j \neq 0$.
  Moreover, let $P_a$ be  a symmetric distribution on $\R$ that is Lebesgue absolutely continuous
  and let 
 $h_i:\R^d\to [0,\infty)$ be a neuron of the form \eqref{neuron-general}.  If 
    $a_i$ is sampled from 
    $P_a^d$ and  $b_i =0$, 
     then we have 
    \begin{align*}
     P_a^d\bigl(\{ \mbox{ neuron } h_i \mbox{ is inactive }   \}  \bigr) 
     &= P_a^d\bigl(\{ \mbox{ neuron }  h_i \mbox{ is semi-active }   \}  \bigr) = P_a^d(D^\star)\\
      P_a^d\bigl(\{ \mbox{ neuron }  h_i \mbox{ is fully active }   \}  \bigr) &= 1 - 2P_a^d(D^\star)\, .
    \end{align*}
\end{theorem}

To illustrate this result, let us recall from the beginning of this section that we are mostly 
interested in data sets $D\subset [0,\infty)^d$. Now assume that the conical hull $\coni D$ of $D$,
that is, the smallest convex cone that contains $D$, satisfies 
\begin{align}\label{D-cone-condition}
 \coni D = [0,\infty)^d\, .
\end{align}
Using some properties listed in  Appendix \ref{appendix:geometry}, we then have 
$D^\star = (\coni D)^\star = ([0,\infty)^d)^\star = [0,\infty)^d$, and $P_a(\{0\}) = 0$ together with 
the symmetry
of $P_a$ then yields 
\begin{equation}\label{prob-inactive-general}
 P_a^d\bigl(\{ \mbox{ neuron } h_i \mbox{ is inactive }   \}  \bigr)  
 = P_a^d(D^\star) 
 = P_a^d \bigl([0,\infty)^d\bigr) = 2^{-d}\, .
\end{equation}
In other words, if \eqref{D-cone-condition} is satisfied, then even for moderate  sizes
$d= m_{l-1}$
of the previous layer we can essentially ignore the problem of initializing 
a neuron into an inactive or semi-active state. 
On the other hand, Lemma \ref{result:full-cone-char} shows that 
\eqref{D-cone-condition} is satisfied if and only if the data set contains, modulo positive 
constants, all vectors of the standard ONB of $\Rd$. In other words, for each $k=1,\dots,d$, 
there needs to be a sample $x_{j_k}$ 
whose precursor in the previous layer \emph{only} falls 
into the region of activity of the $k$-neuron. Unfortunately,  estimating
the probability of such events is rather complicated as the following remark,
which describes the transformation of the data set by a single, randomly initialized neuron,
shows.

\begin{figure}[t]
\begin{center}
\includegraphics[width=0.20\textwidth]{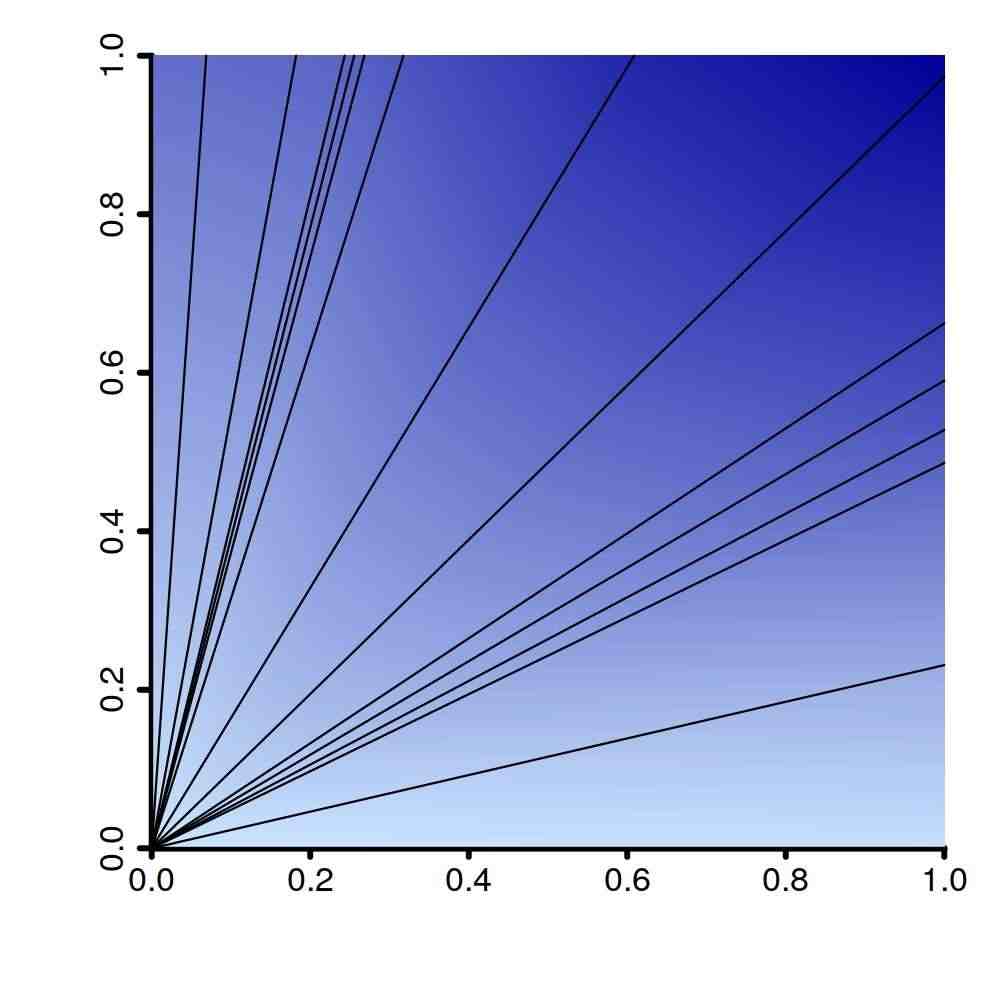}
\hspace*{-0.02\textwidth}
\includegraphics[width=0.20\textwidth]{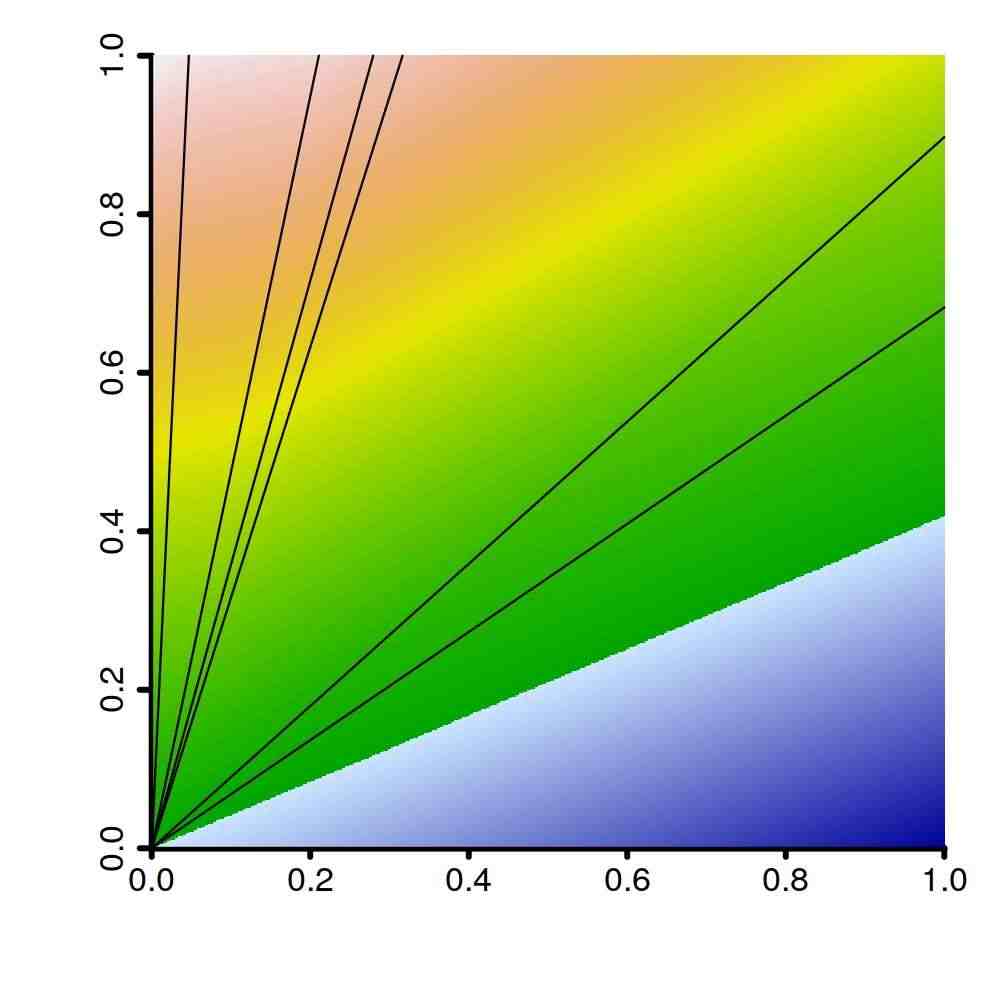}
\hspace*{-0.02\textwidth}
\includegraphics[width=0.20\textwidth]{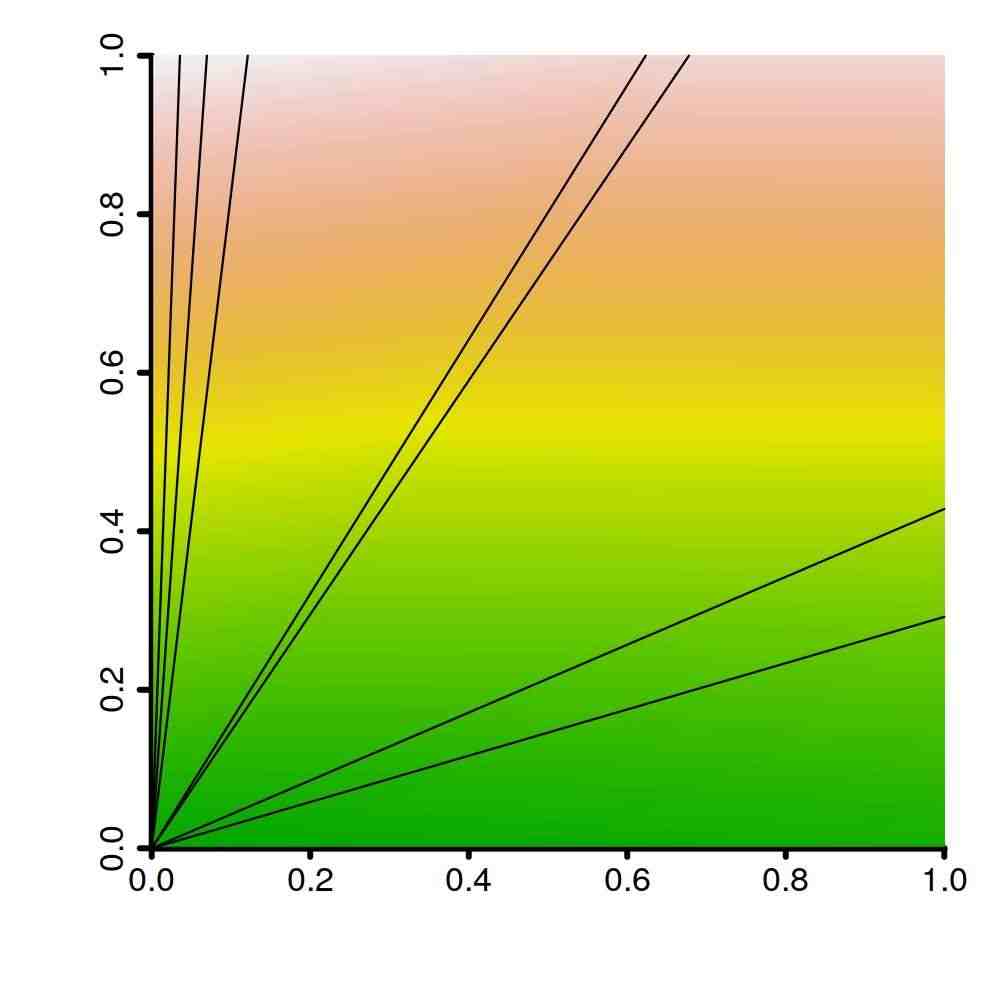}
\hspace*{-0.02\textwidth}
\includegraphics[width=0.20\textwidth]{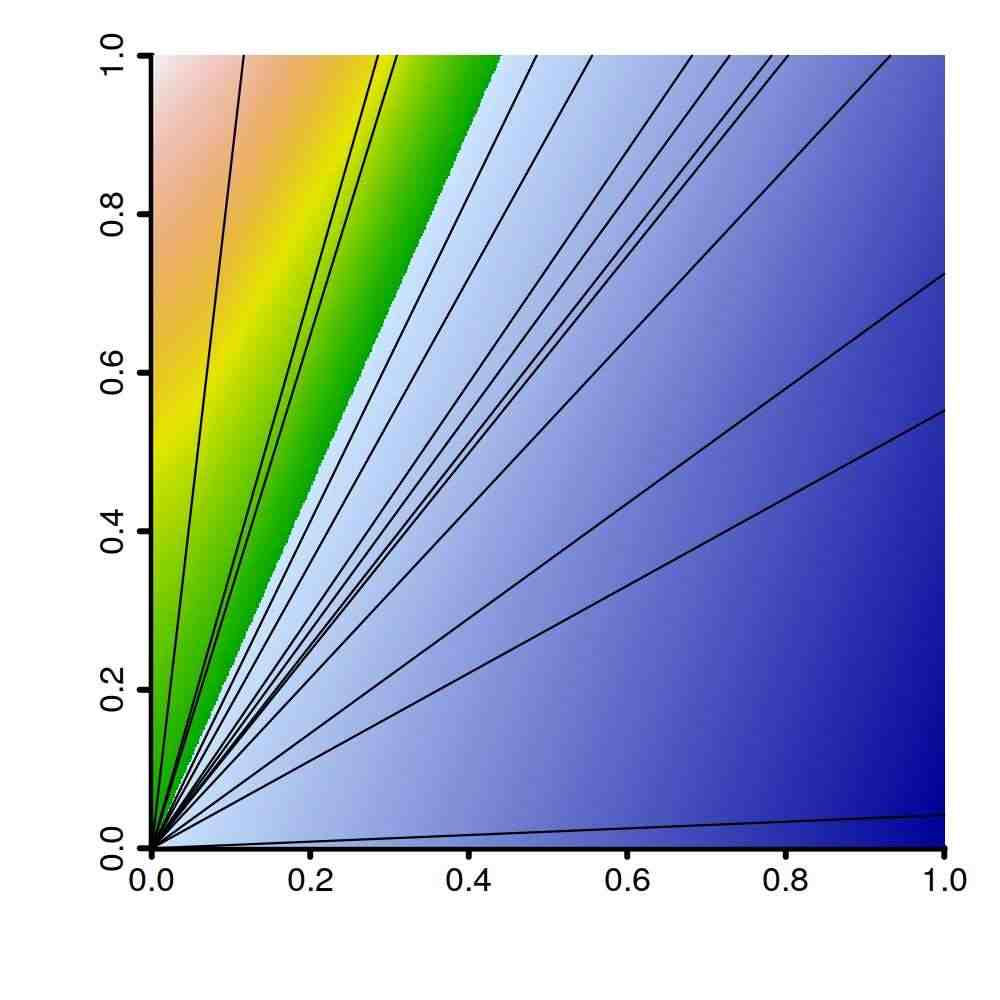}
\hspace*{-0.02\textwidth}
\includegraphics[width=0.20\textwidth]{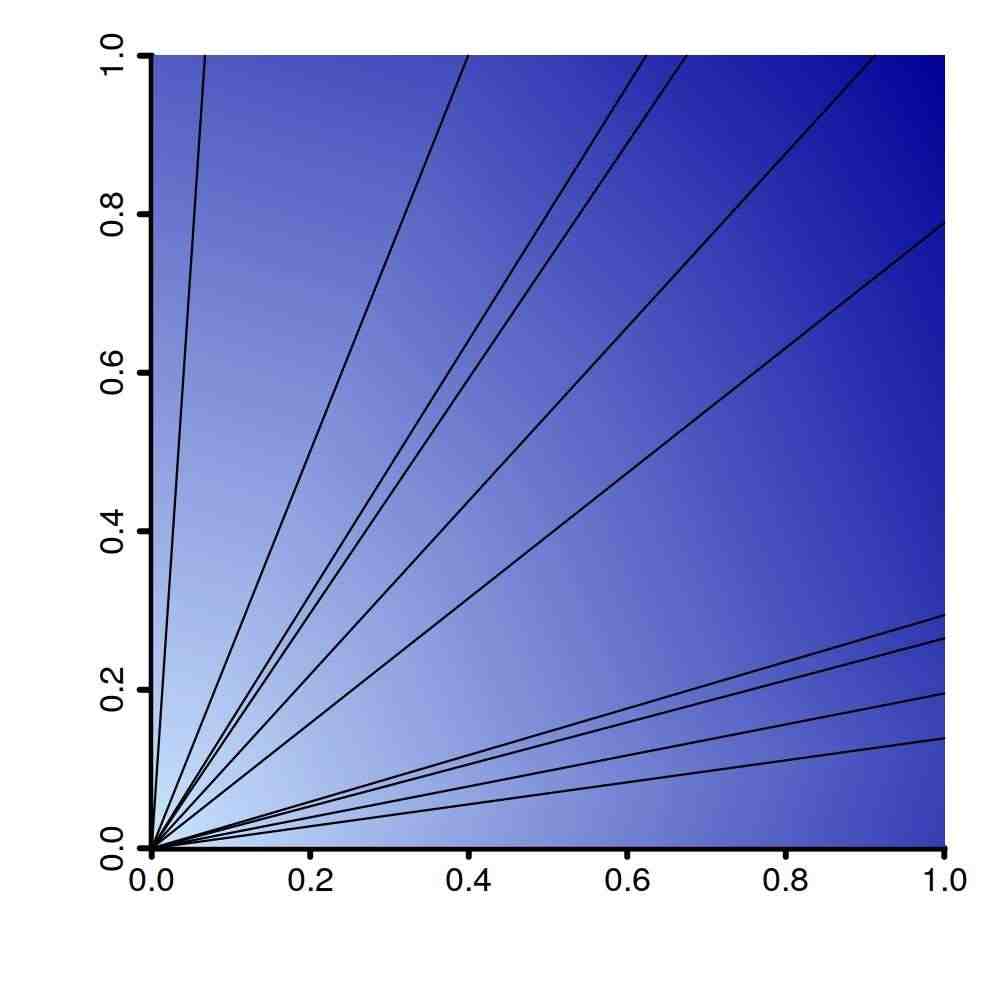}

\includegraphics[width=0.20\textwidth]{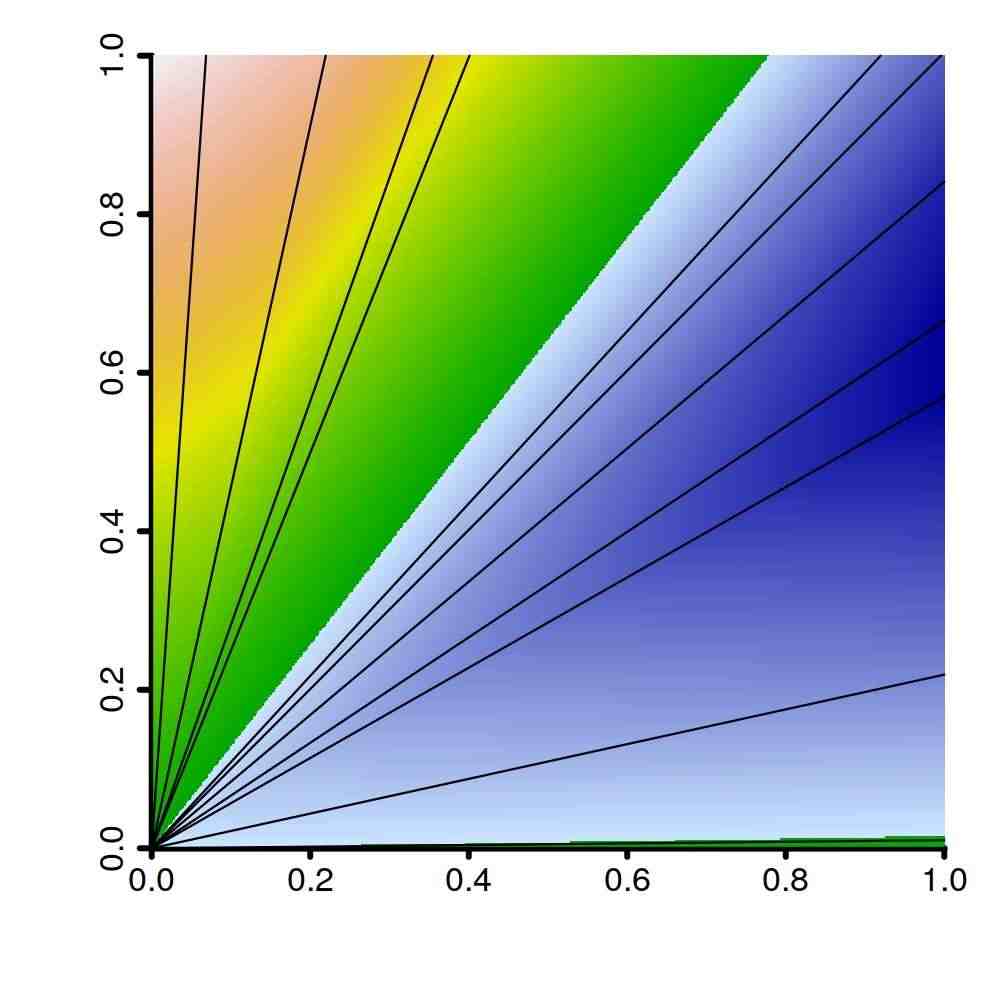}
\hspace*{-0.02\textwidth}
\includegraphics[width=0.20\textwidth]{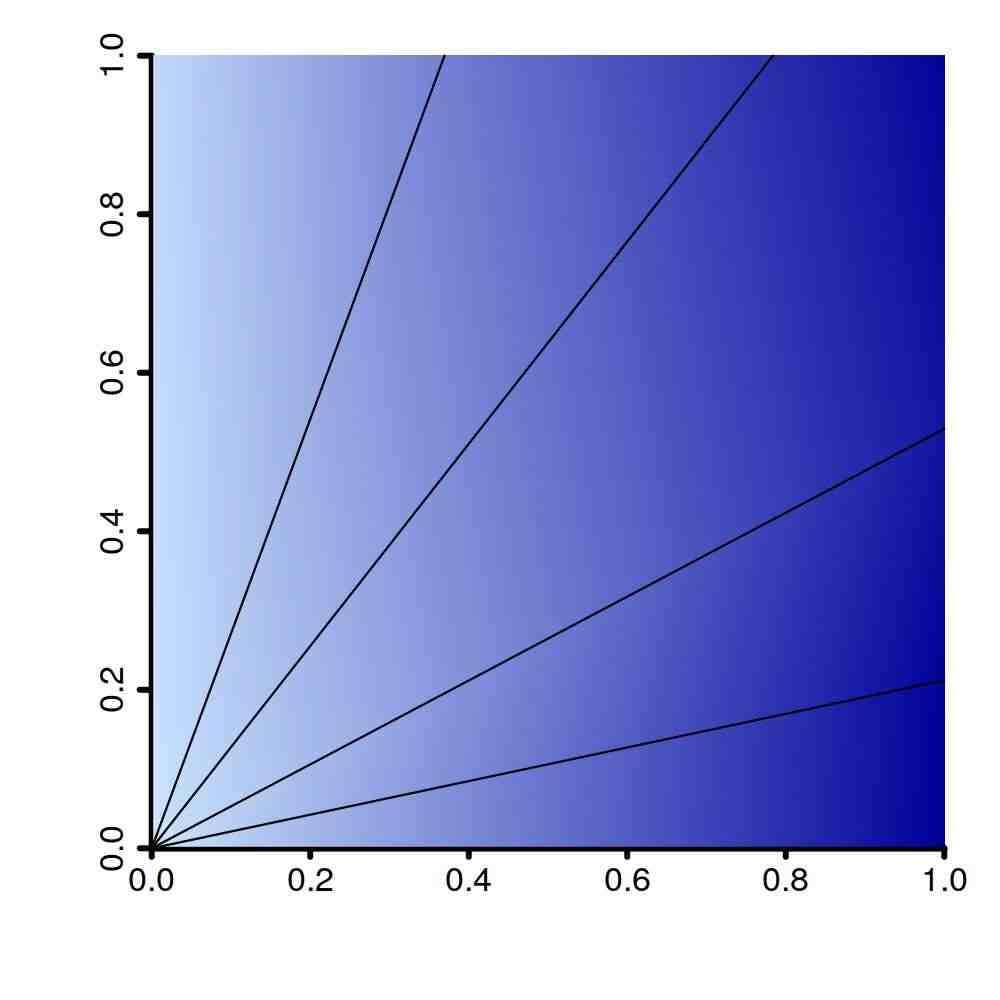}
\hspace*{-0.02\textwidth}
\includegraphics[width=0.20\textwidth]{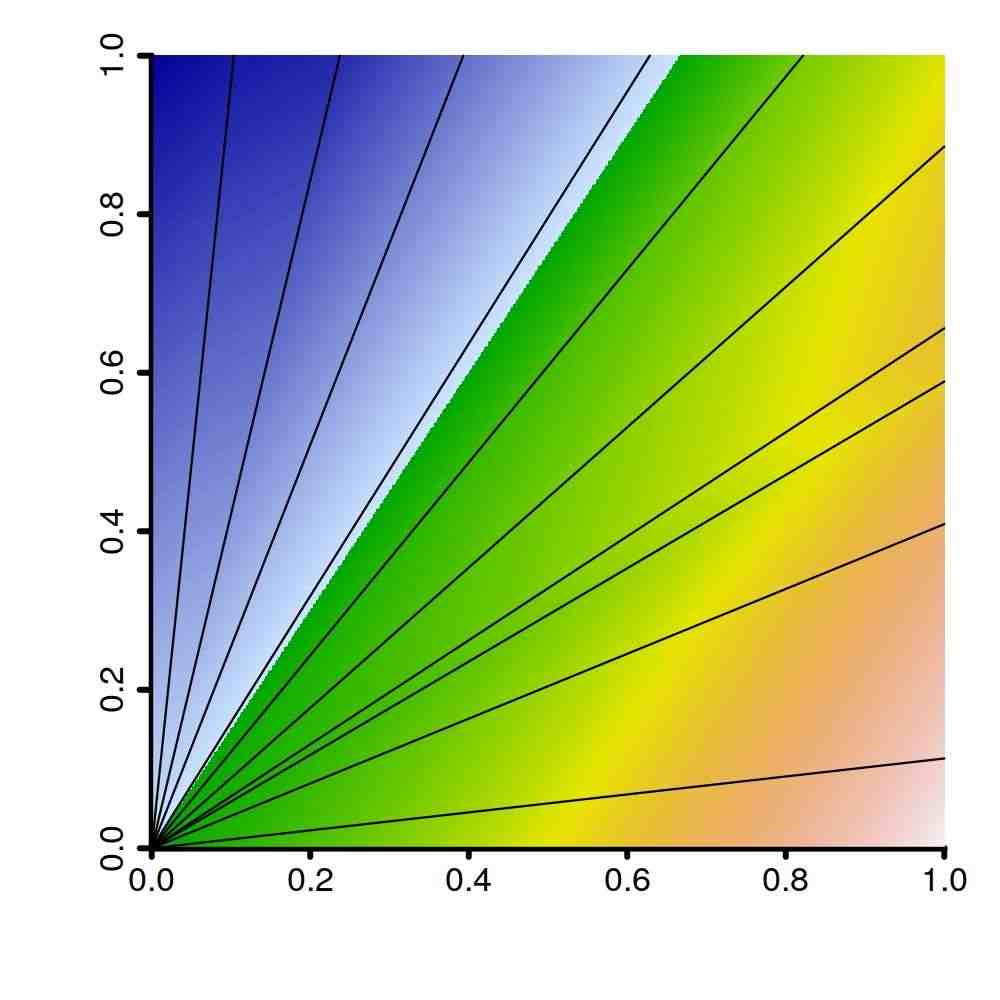}
\hspace*{-0.02\textwidth}
\includegraphics[width=0.20\textwidth]{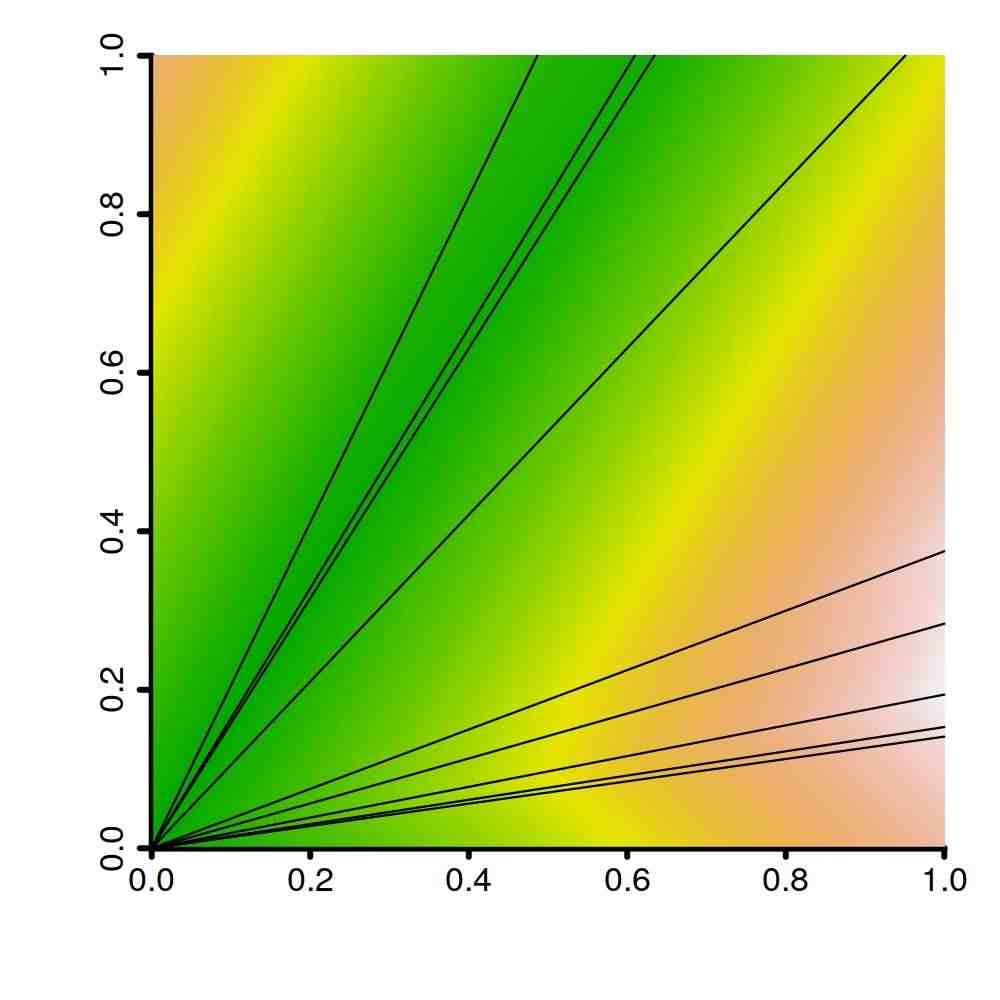}
\hspace*{-0.02\textwidth}
\includegraphics[width=0.20\textwidth]{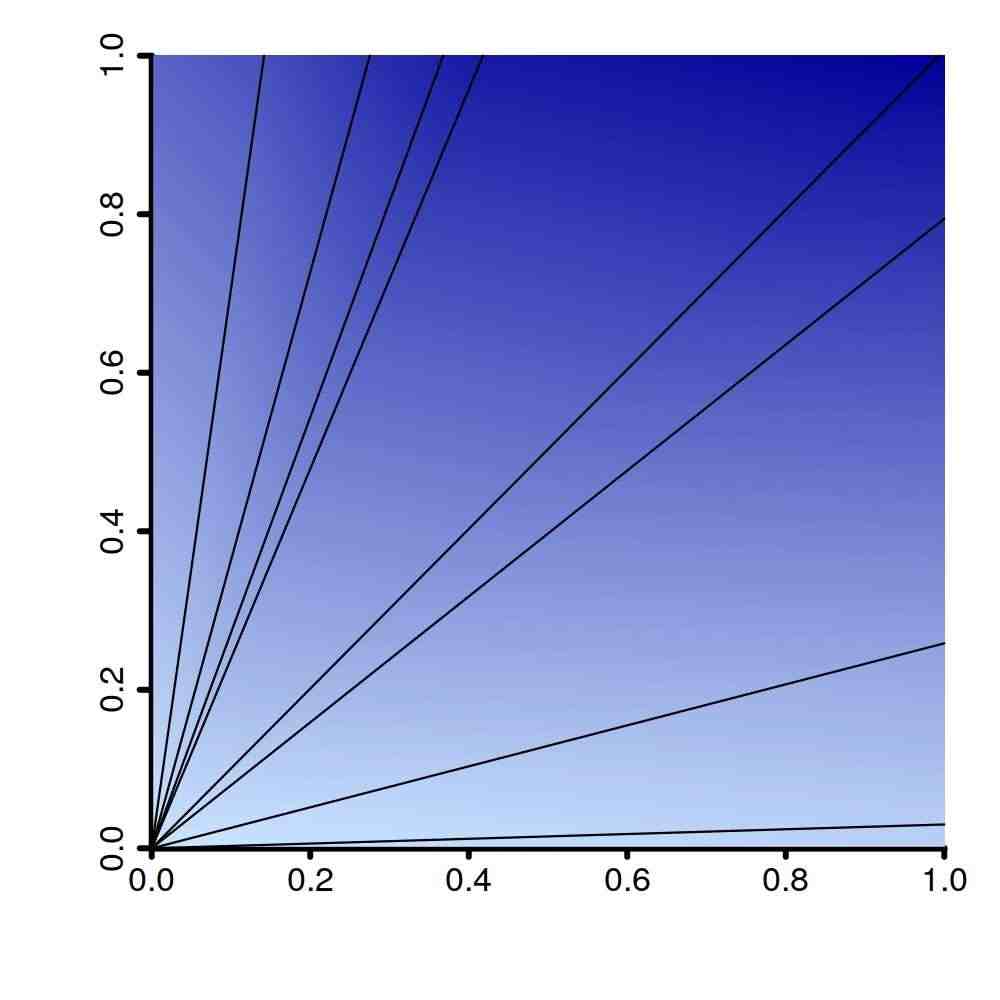}
\end{center}
\caption{Ten randomly initialized predictors of a neural network with one layer of 20 hidden neurons. The weights were sampled
from $\ca N(0,\s^2)$ with $\s^2 = 2/d = 1$ according to \heetal~and the biases are set to zero. The color scheme is similar to those of 
geographic maps. Namely, 
blue colors indicate negative outputs and darker blues correspond to smaller values. 
Conversely, green colors indicate small positive values and yellow, brown, and white colors correspond to 
larger values. The black lines show the edges $x_i^*$ of the neurons.}\label{figure:random-functions-he-2d}
\end{figure}

\begin{remark}[Functions with zero bias]\label{remark:zero-bias-functions}
	Recall, that a function $f:\Rd \to \R^m$ is   positively homogeneous, if for all $\a>0$ and all $x\in \Rd$ we have 
\begin{displaymath}
   f(\a x) = \a f(x)\, .
\end{displaymath}
	We will now show that if we initialize all biases of our network of arbitrary depth and width 
with $0$, then the resulting 
	function represented by the entire network is  positively homogeneous.
	We begin by showing that an arbitrary hidden layer $H_l:\Rd \to \R^m$ is positively homogeneous.
		To this end,  let  $h_1,\dots,h_m: \Rd \to \R$ be the neurons of the hidden layer. 
		Since they are initialized with  $b_i = 0$, we then have 
		\begin{displaymath}
		   h_i(x) = \relu{\langle a_i, x \rangle}\, , \qquad \qquad x\in \Rd\, .
		\end{displaymath}
		Combining the examples of  positively homogeneous functions listed in  Appendix \ref{appendix:function-classes}
		with Lemma \ref{result:pos-homo-permanent}, we easily see that each $h_i$ is positively homogeneous
		and another application of  Lemma \ref{result:pos-homo-permanent} then shows that $H_l = (h_1,\dots, h_m):\Rd\to \R^m$
		is also positively homogeneous. Moreover, Lemma \ref{result:pos-homo-permanent} further recalls that 
		the composition of positively homogeneous is positively homogeneous, and therefore, the composition
    of all hidden layers is positively homogeneous. Finally, the output layer is linear and thus positively homogeneous,
		so that another application of Lemma \ref{result:pos-homo-permanent} shows that the
	function represented by the entire network is  positively homogeneous. Figure \ref{figure:random-functions-he-2d}
	presents a few such random functions.

		We have already seen in the one-dimensional case that a zero-bias-initialization
   leads to a very restrictive function class on e.g.~$[0,1]$, namely linear functions. Obviously,
   such functions cannot approximate a nonlinear continuous function arbitrarily well. Now, in the general case,
   our network with zero biases is able to represent more general functions, namely positively homogeneous, continuous functions.
   There could thus be some hope that such a network is able to approximate suitably large classes of functions.
   Unfortunately, this is not true. Indeed, Corollary \ref{result:pos-homo-closed-cx} shows that 
   for every compact $X\subset \Rd$ and every continuous function $g\in C(X)$ that is \emph{not}
   positively homogeneous there is 
   an $\e>0$ such that 
   \begin{displaymath}
    \inorm {g - f} \geq \e
   \end{displaymath}
    for all functions $f:X\to \R$ that can be represented by an \emph{arbitrary} 
    network with ReLU-activation functions. Moreover, Corollary \ref{result:pos-homo-closed-lp}
    shows that the same result remains valid if we replace $C(X)$ with its norm $\inorm\cdot$ by 
    $\Lx pP$ and $\snorm\cdot_{\Lx pP}$, where $p\in [1,\infty)$ and $P$ is an arbitrary probability measure 
    on $\Rd$ provided that the target function $g$ does not $P$-almost surely coincide with a 
   positively homogeneous function. 
    Consequently, considering ReLU-networks without bias violates any sort of universal approximation property
    in a very strong sense, and initializing ReLU-networks with zero biases requires 
    updating the biases during training for basically all interesting target functions.
\remarkend\end{remark}

Our next goal is to investigate the effects of non-zero bias initialization strategies. We begin 
by presenting the following lemma that considers deterministic initializations of the bias.

\begin{lemma}\label{result:nonzero-bias-lemma}
   Let $D = (x_1,\dots,x_n)$ be a data set in $\Rd$,
 $P_a$ be  a symmetric distribution on $\R$ that is Lebesgue absolutely continuous, and 
$b_-,  b_+\in \R$ with $b_- < b_+$. Moreover, 
    let 
 $h_i:\R^d\to [0,\infty)$ be a neuron of the form \eqref{neuron-general}.  If its weight
    $a_i$ is sampled from 
    $P_a^d$ and its bias is initialized by either $b_-$ or $b_+$
     then we have 
		\begin{align*}
     P_a^d\otimes \d_{b_-}\bigl(\{ \mbox{ neuron } h_i \mbox{ is inactive }   \}  \bigr) 
     &\geq P_a^d\otimes \d_{b_+}\bigl(\{ \mbox{ neuron } h_i \mbox{ is inactive }   \}  \bigr) \\
			P_a^d\otimes \d_{b_-}\bigl(\{ \mbox{ neuron } h_i \mbox{ is semi-active }   \}  \bigr) 
     &\leq P_a^d\otimes \d_{b_+}\bigl(\{ \mbox{ neuron } h_i \mbox{ is semi-active }   \}  \bigr) \, .
		\end{align*}
\end{lemma}

\begin{figure}[t]
\begin{center}
\includegraphics[width=0.20\textwidth]{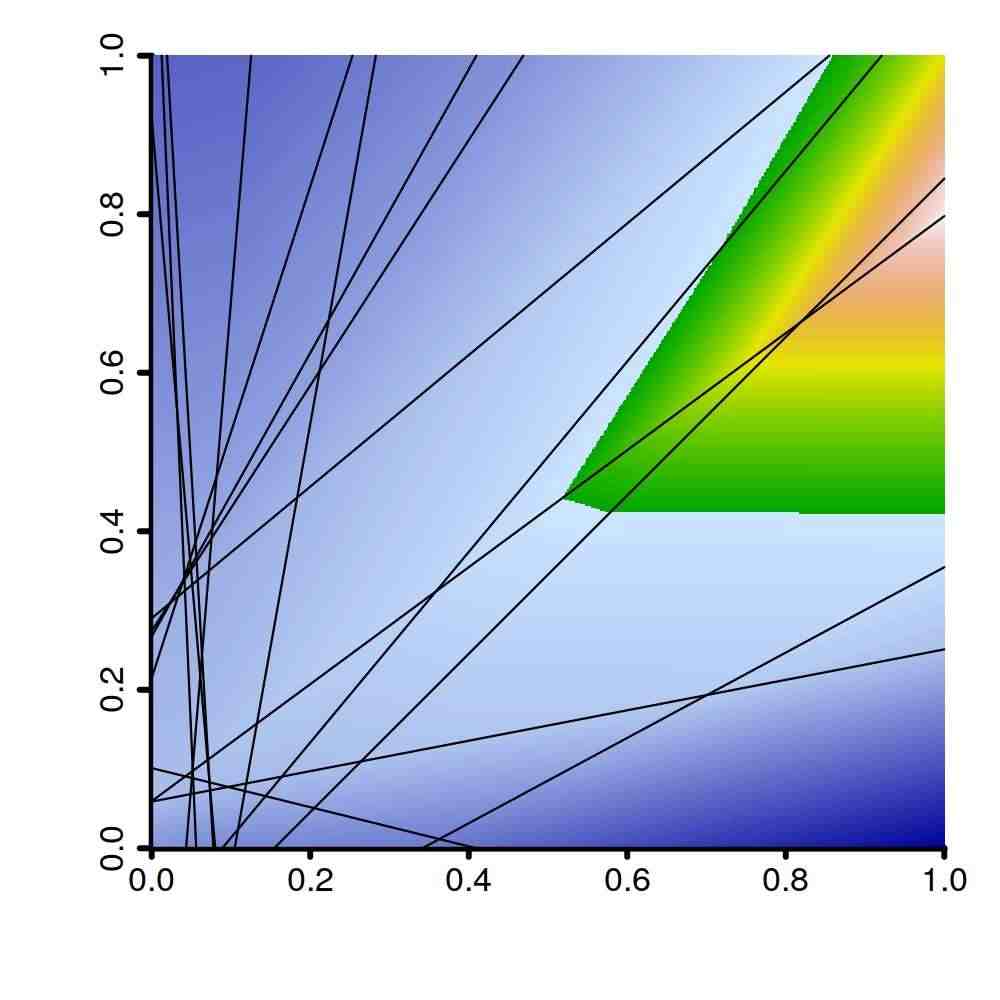}
\hspace*{-0.02\textwidth}
\includegraphics[width=0.20\textwidth]{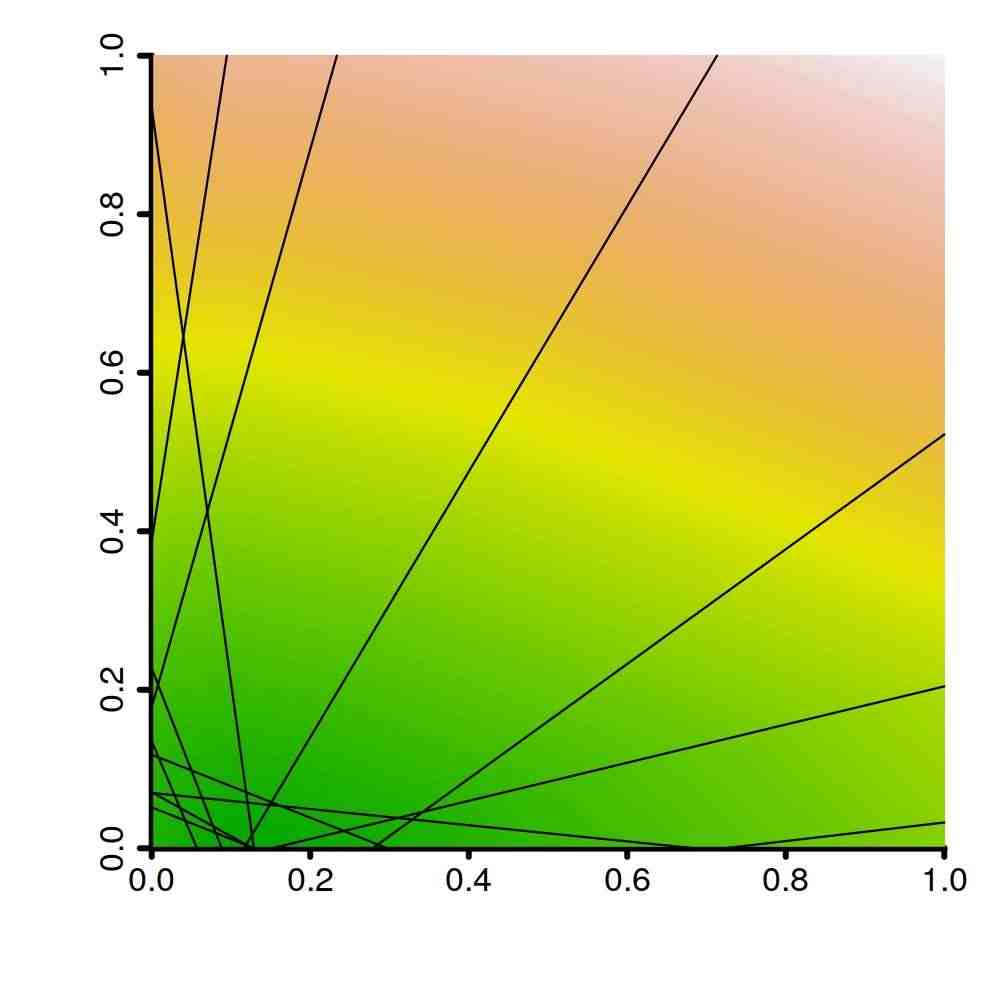}
\hspace*{-0.02\textwidth}
\includegraphics[width=0.20\textwidth]{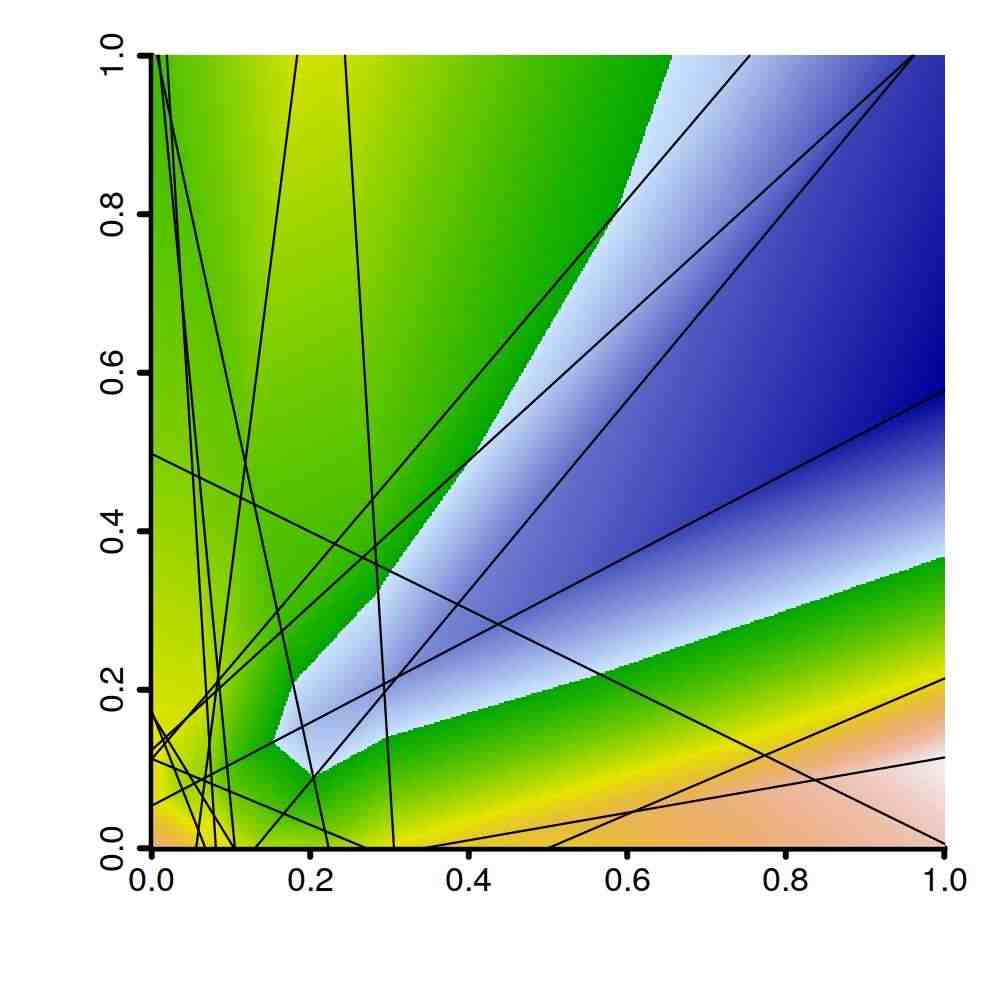}
\hspace*{-0.02\textwidth}
\includegraphics[width=0.20\textwidth]{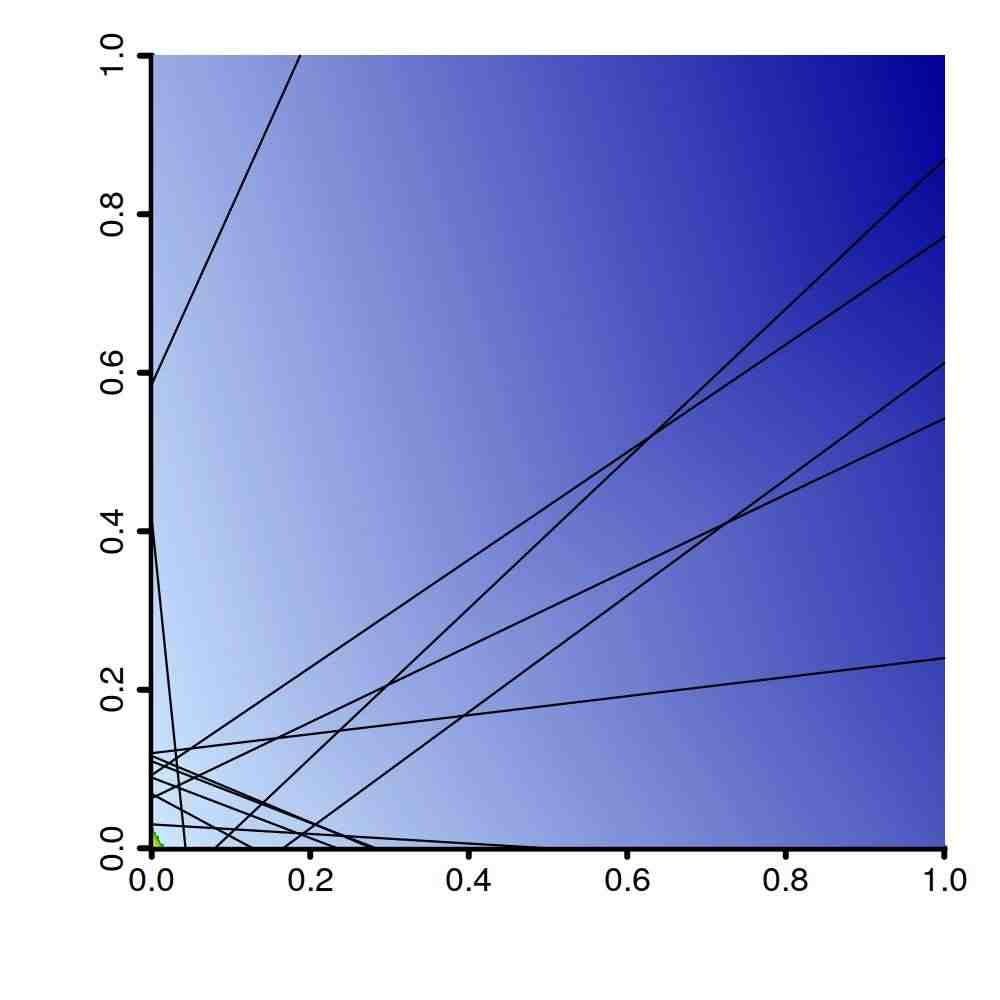}
\hspace*{-0.02\textwidth}
\includegraphics[width=0.20\textwidth]{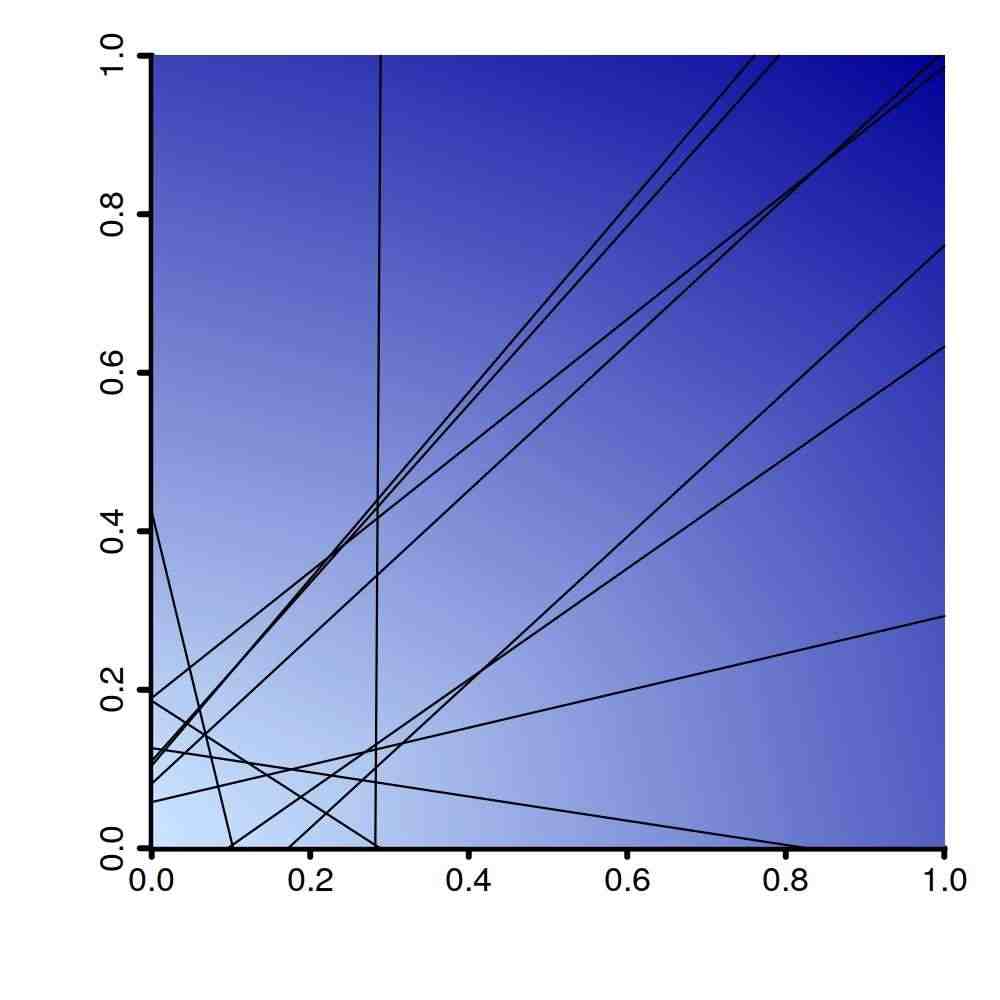}

\includegraphics[width=0.20\textwidth]{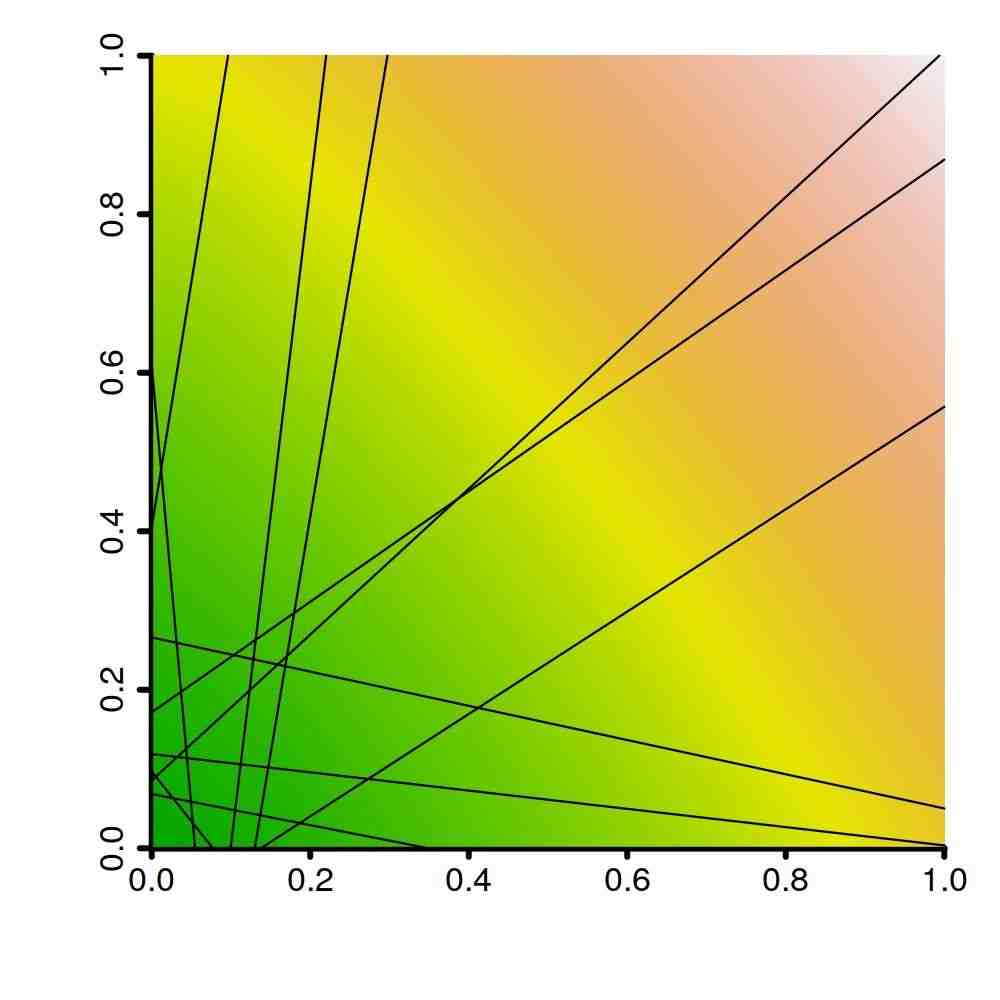}
\hspace*{-0.02\textwidth}
\includegraphics[width=0.20\textwidth]{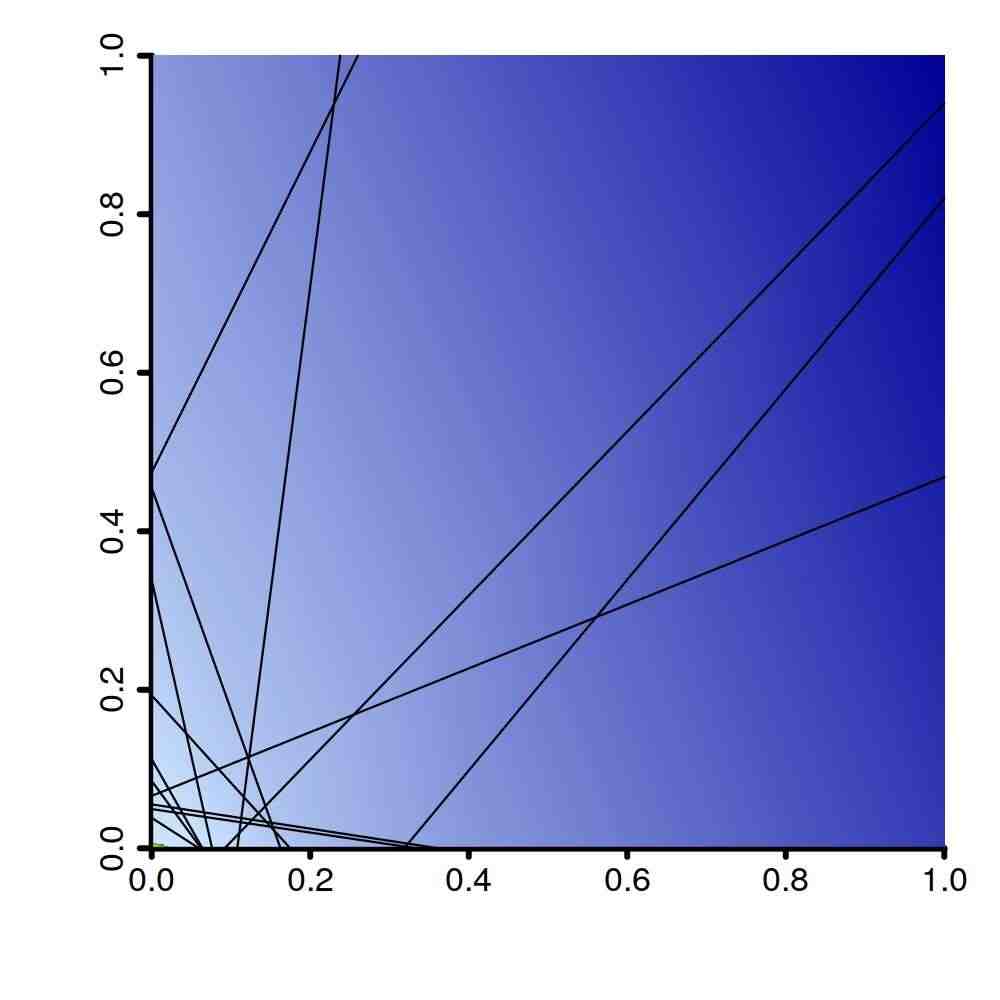}
\hspace*{-0.02\textwidth}
\includegraphics[width=0.20\textwidth]{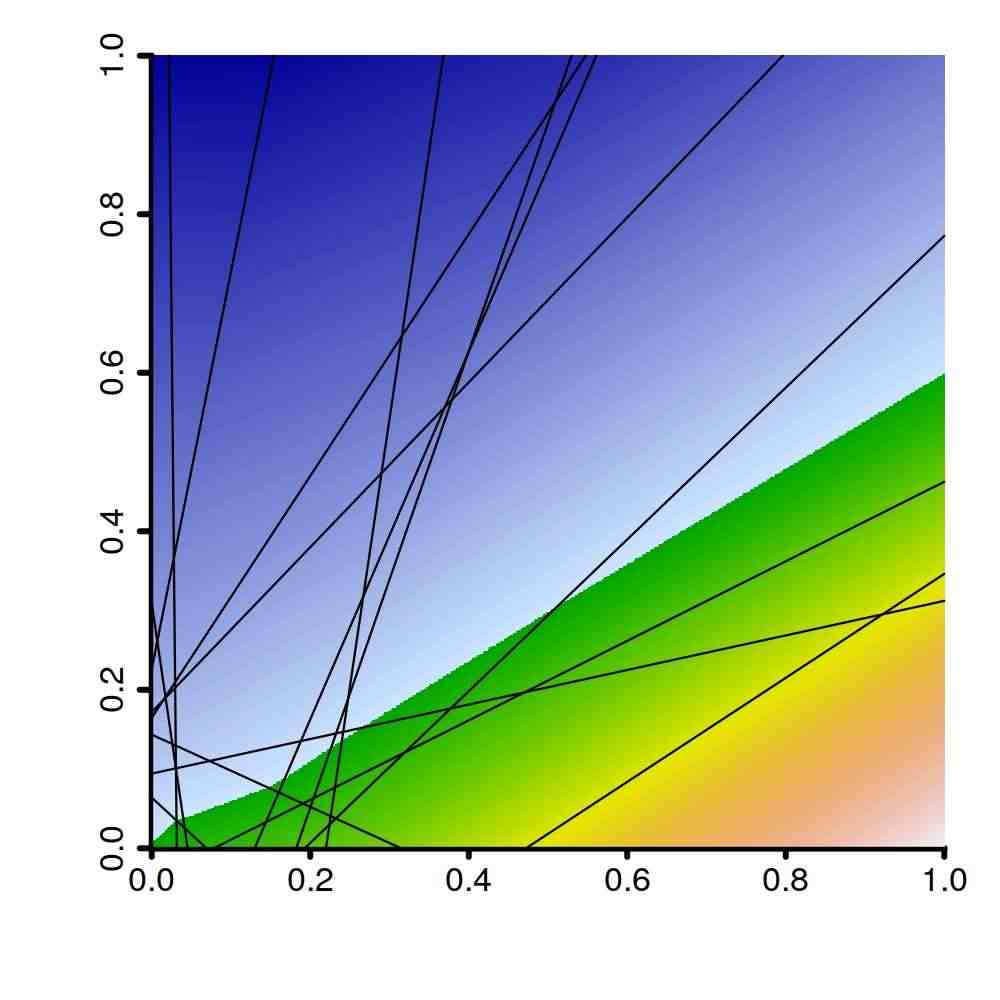}
\hspace*{-0.02\textwidth}
\includegraphics[width=0.20\textwidth]{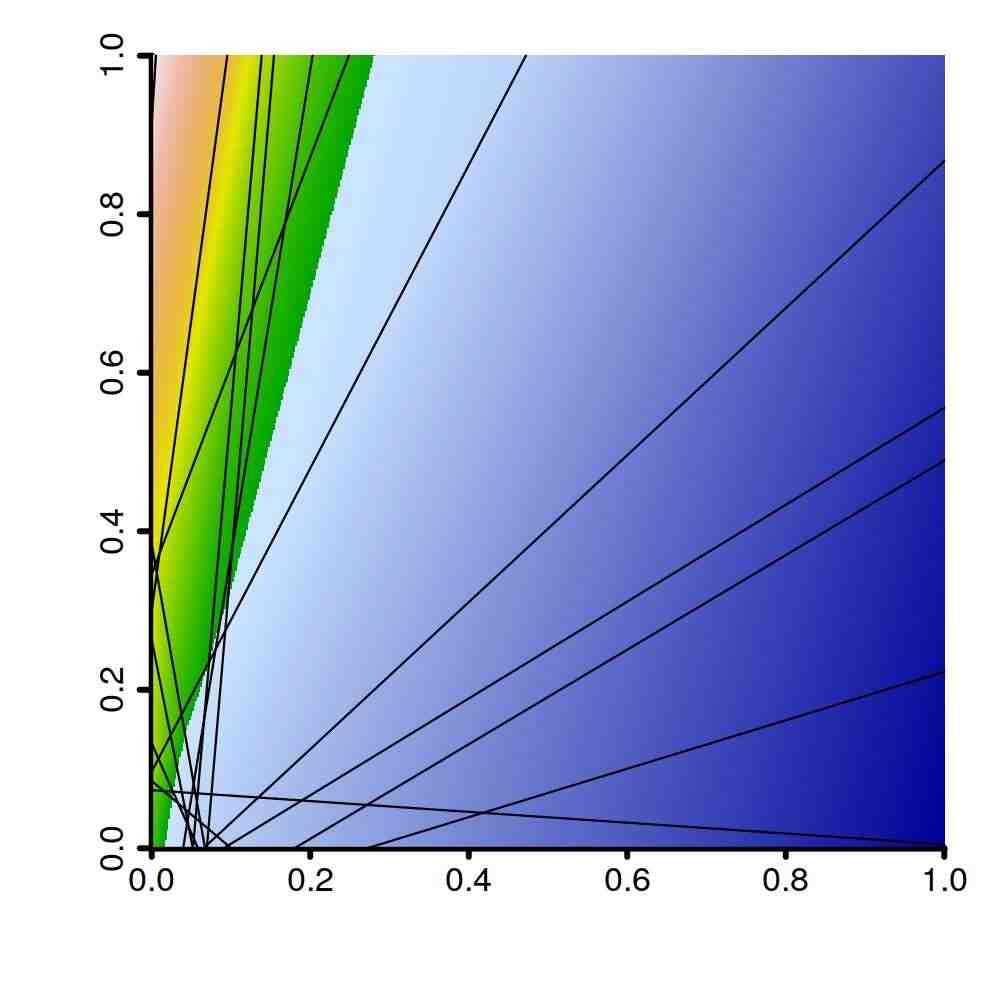}
\hspace*{-0.02\textwidth}
\includegraphics[width=0.20\textwidth]{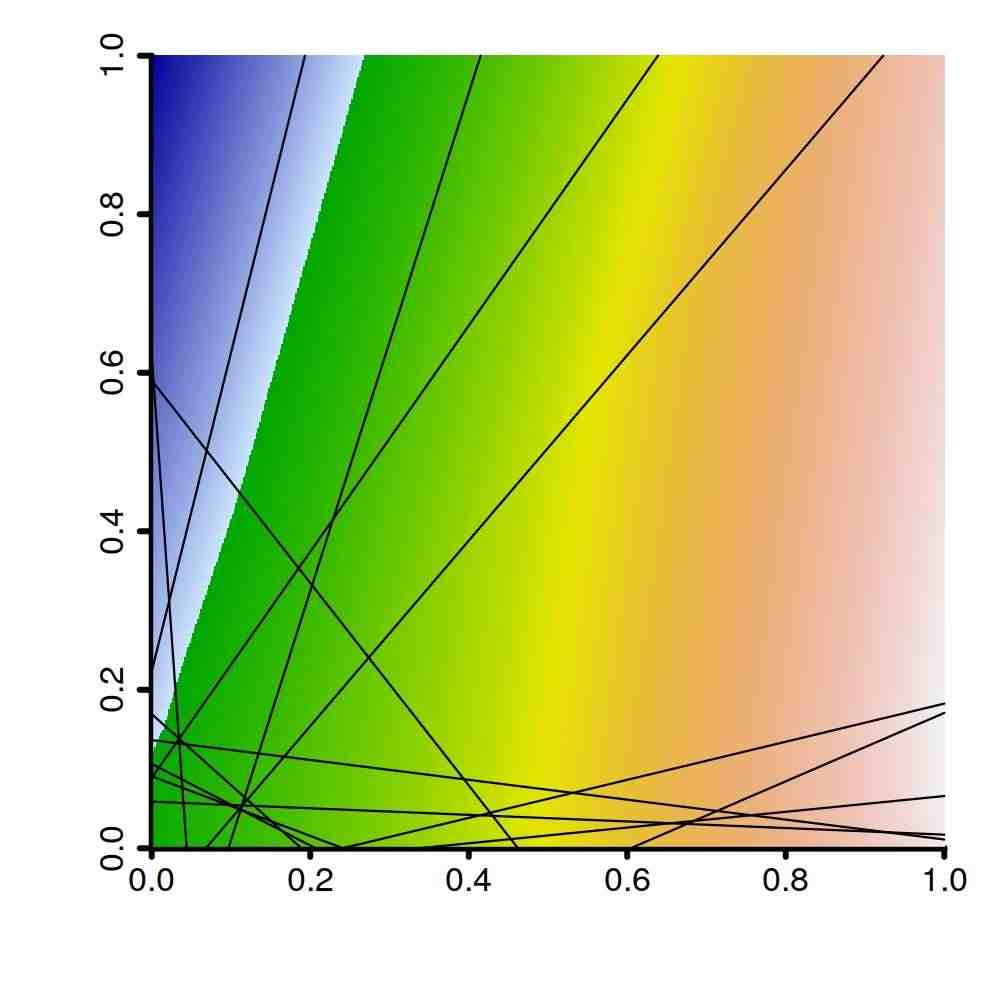}
\end{center}
\caption{Ten randomly initialized predictors of a neural network with one layer of 20 hidden neurons. The weights were sampled
from $\ca N(0,\s^2)$ with $\s^2 = 1$ according to \heetal~and the biases are set to $b=0.1$.
The color scheme equals that of Figure \ref{figure:random-functions-he-2d}.
Compared to the zero-bias-initialization depicted in Figure \ref{figure:random-functions-he-2d} we see that 
the edges no longer contain the origin, but most edges are still in the vicinity of the origin.}
\label{figure:random-functions-nd-2d}
\end{figure}

\begin{remark}[Functions with non-zero bias]\label{remark:nonzero-bias-functions}
By applying Lemma \ref{result:nonzero-bias-lemma} in the case $b_+ = 0$, we see that, compared to 
a zero-bias initialization,
the probability of obtaining an inactive neuron increases if we choose a negative deterministic bias.
Similarly, by considering $b_- = 0$, we observe that the
probability of obtaining an inactive neuron decreases when we choose a positive deterministic bias.
This may explain the fact that some popular initialization heuristics 
prefer a positive deterministic bias, 
but to the best of our knowledge, there is no  initialization heuristic
described in the   literature
that chooses a negative deterministic bias.
Finally recall that in the one-dimensional case, 
Theorem \ref{result:pos-bias-better} provided a significantly stronger result, if $0 \in \co D$.
In fact, one could also reproduce Theorem \ref{result:pos-bias-better}  for $d>1$ if $0 \in \co D$.
However, we are mostly interested in data sets $D$ contained in $[0,\infty)^d$, and for such $D$,
the condition $0 \in \co D$ is equivalent to $x_{j_0}=0$ for some $j_0\in \{1,\dots,\}$.
In other words, there needs to be at least one sample that is mapped to $0$ by all 
neurons of the previous layer. So far, it is unclear to us, how likely this situation occurs, and hence
we omitted the generalization of  Theorem \ref{result:pos-bias-better}  to the case $d>1$.

Another consequence of Lemma \ref{result:nonzero-bias-lemma} is that using deterministic 
bias initialization we cannot simultaneously decrease the probabilities of inactive and semi-active neurons.
This is in alignment with the one-dimensional situation described in Theorem \ref{result:pos-bias-better}.

Now recall that the distance of the hyperplane $x_i^*$ can be computed by $|b_i| /\snorm{a_i}_2$. Moreover, 
we have already seen in Remark \ref{remark:norm-weight} that e.g.~the initialization strategy \heetal\
results in $\snorm{a_i}_2 \approx \sqrt 2$ with high probability. Consequently, the distance of $x_i^*$
concentrates around $|b_i|/\sqrt 2$ with high probability. For the usual choices  $b_i = 0.1$ and $b_i = 0.01$,
this shows that most hyperplanes are very close to the origin.  Figure \ref{figure:random-functions-nd-2d} illustrates this
in the case $d=2$.
\remarkend\end{remark}

\begin{figure}[t]
\begin{center}
\includegraphics[width=0.20\textwidth]{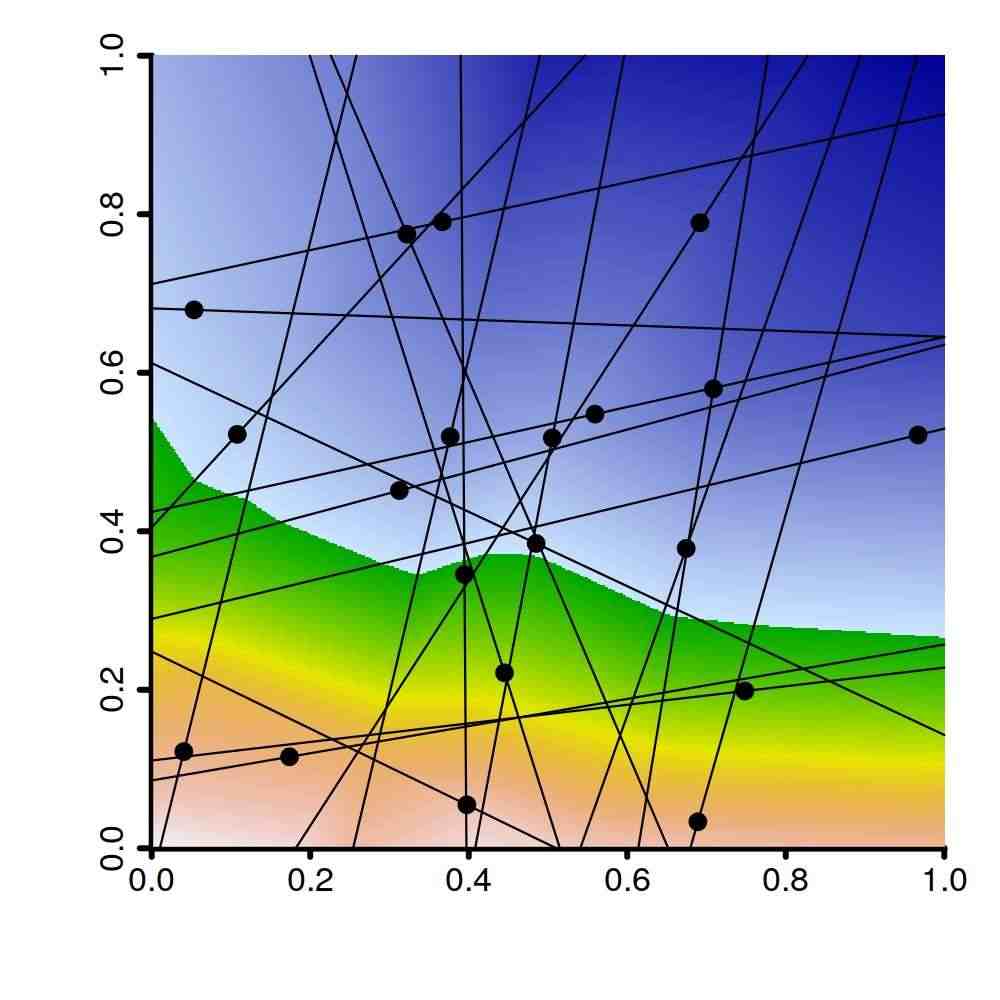}
\hspace*{-0.02\textwidth}
\includegraphics[width=0.20\textwidth]{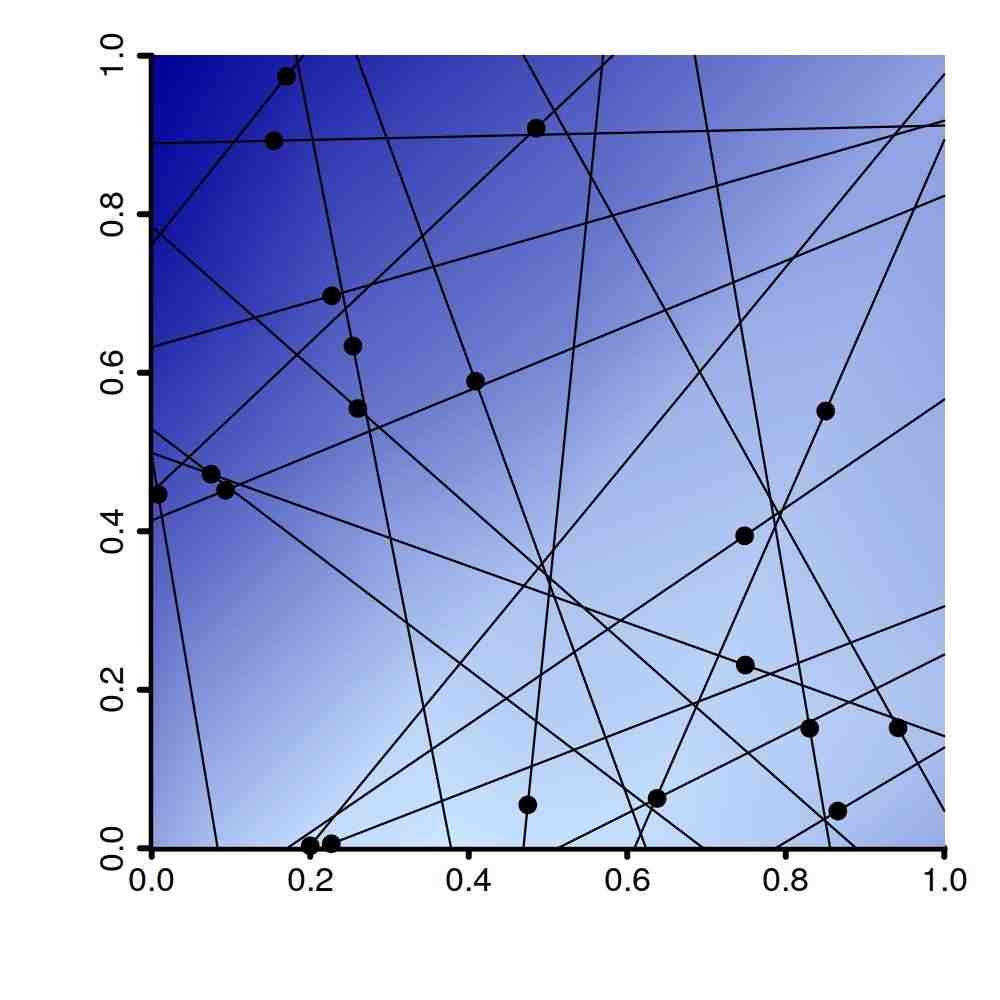}
\hspace*{-0.02\textwidth}
\includegraphics[width=0.20\textwidth]{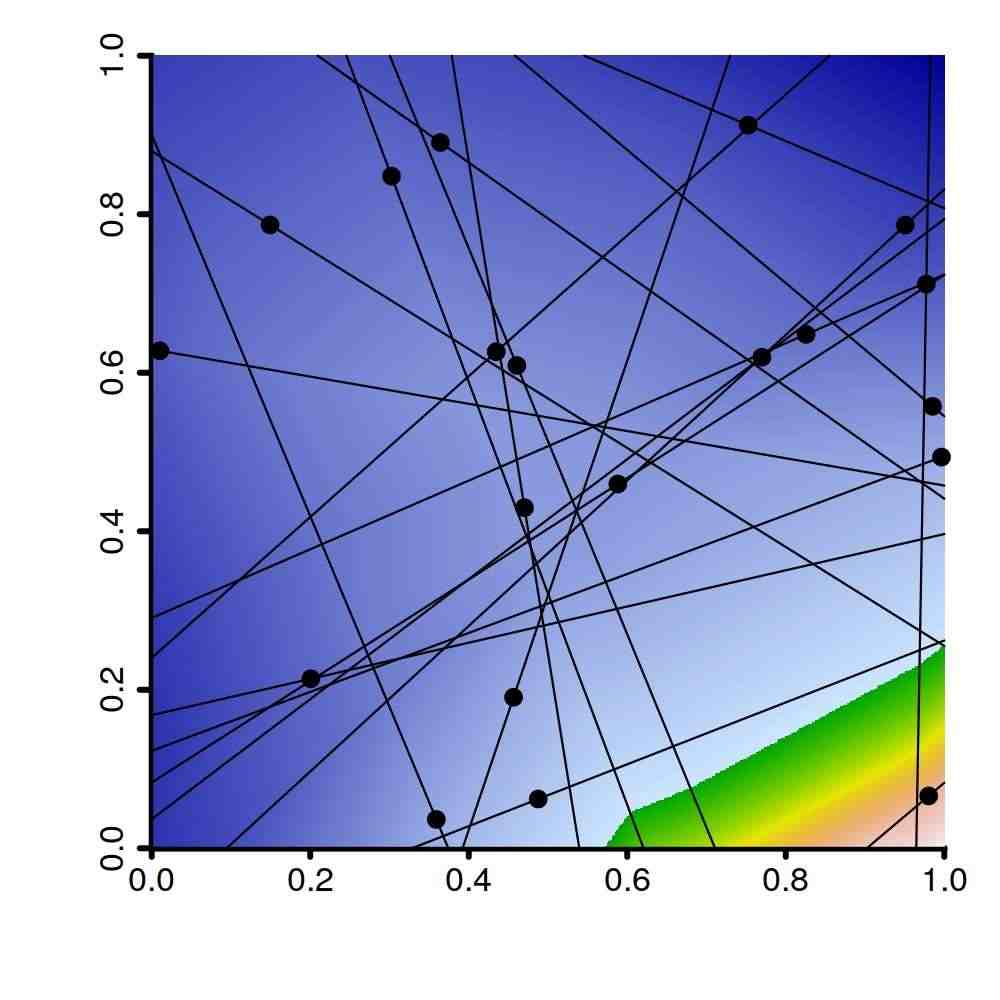}
\hspace*{-0.02\textwidth}
\includegraphics[width=0.20\textwidth]{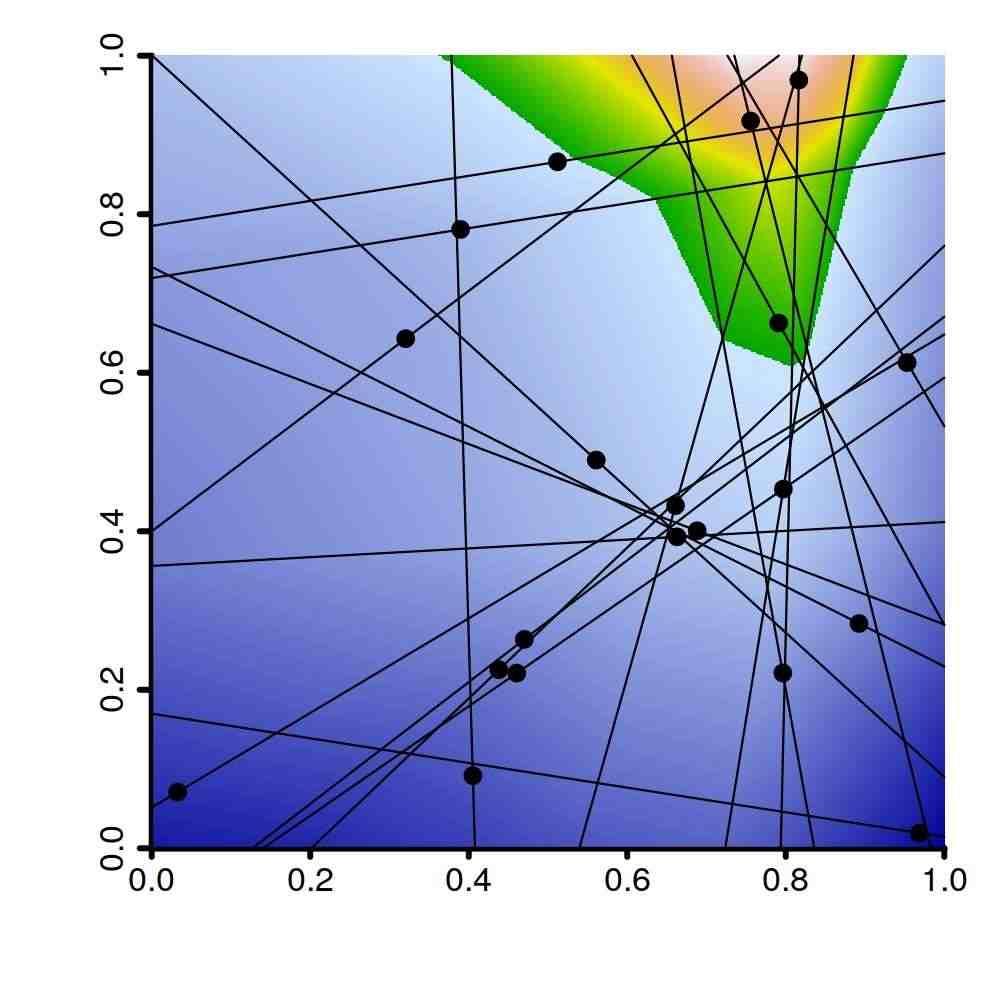}
\hspace*{-0.02\textwidth}
\includegraphics[width=0.20\textwidth]{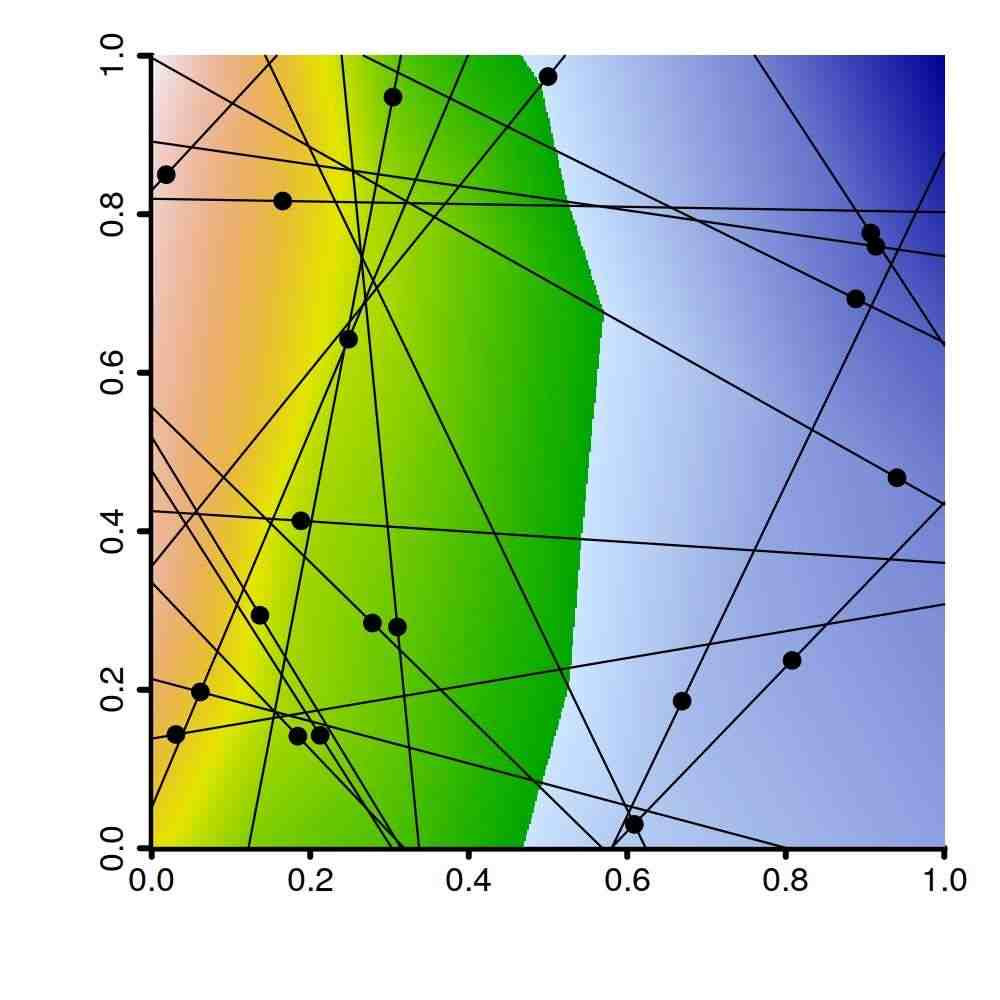}

\includegraphics[width=0.20\textwidth]{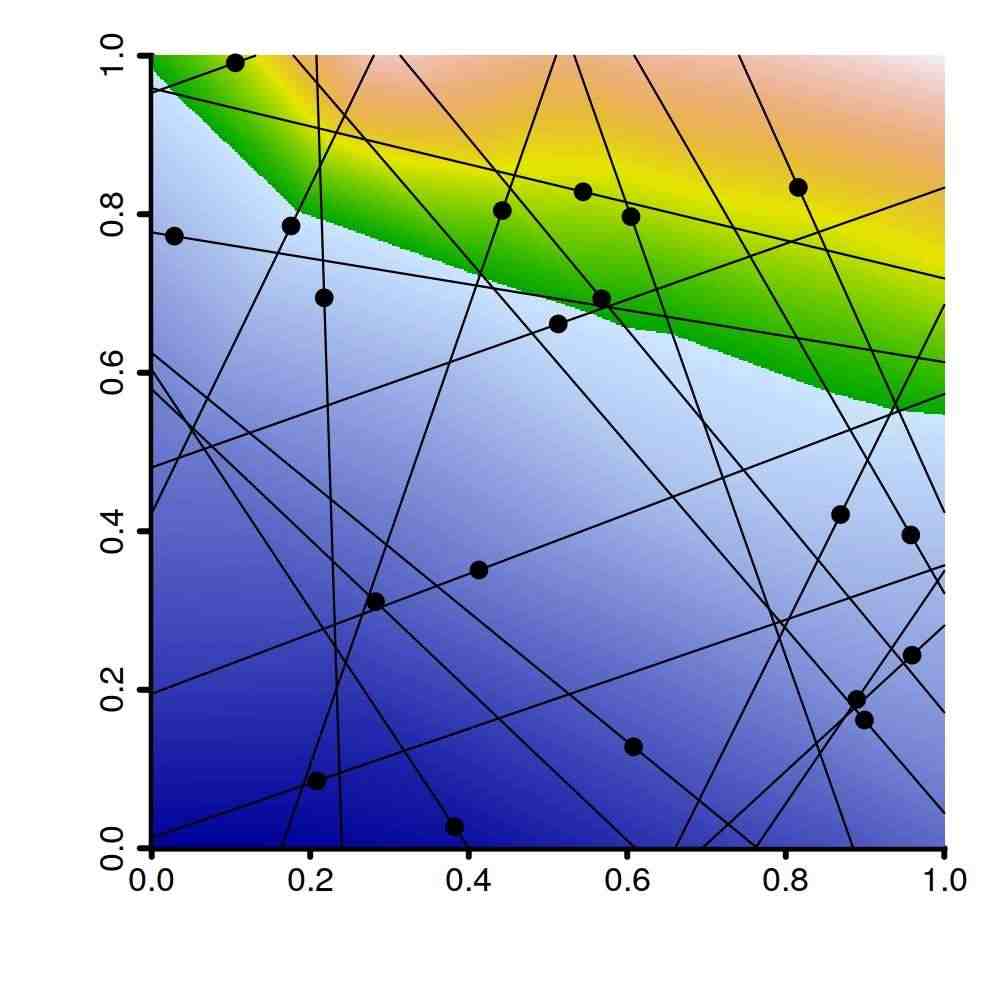}
\hspace*{-0.02\textwidth}
\includegraphics[width=0.20\textwidth]{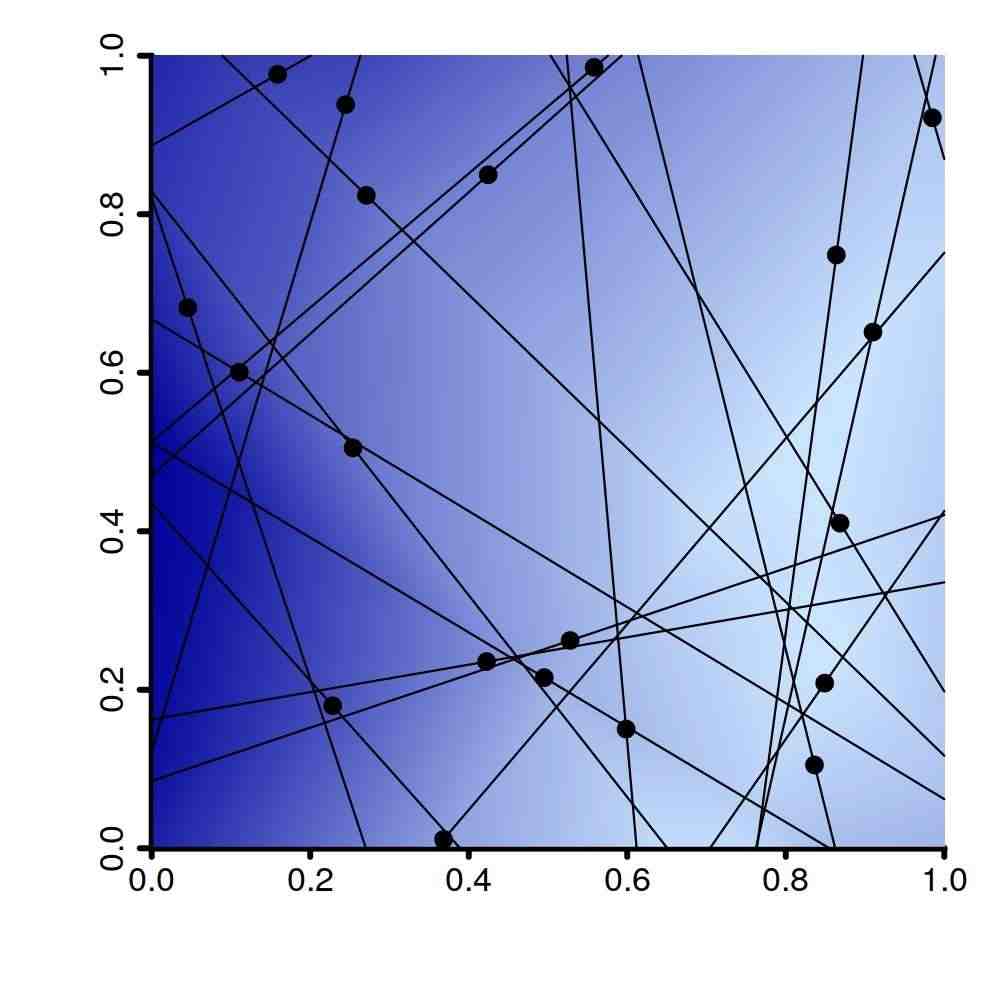}
\hspace*{-0.02\textwidth}
\includegraphics[width=0.20\textwidth]{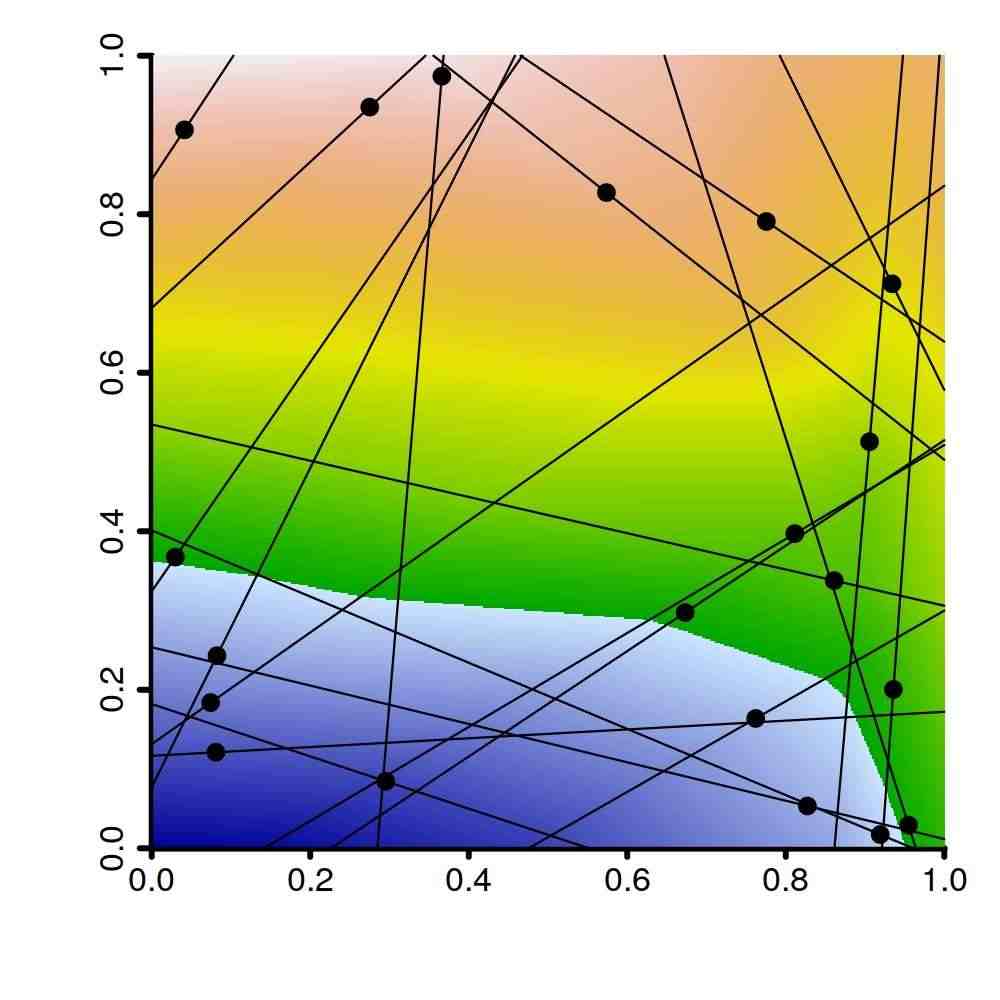}
\hspace*{-0.02\textwidth}
\includegraphics[width=0.20\textwidth]{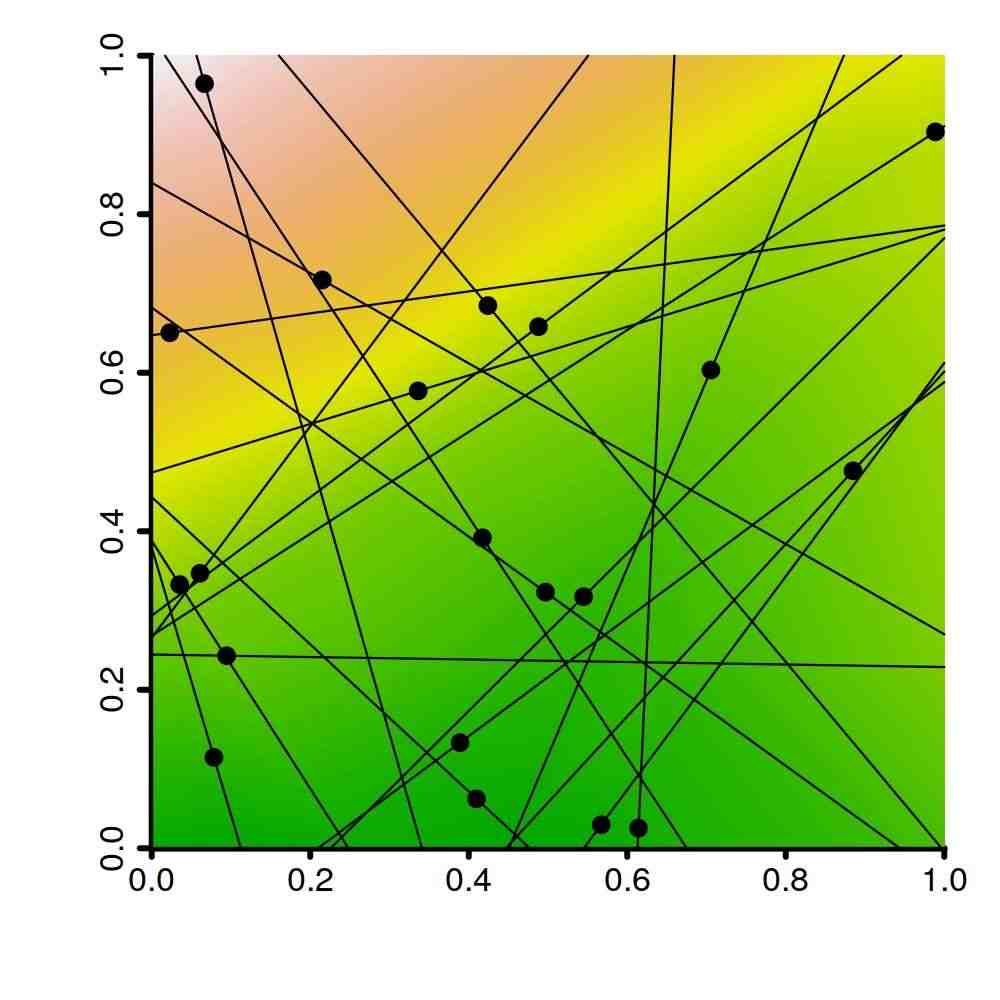}
\hspace*{-0.02\textwidth}
\includegraphics[width=0.20\textwidth]{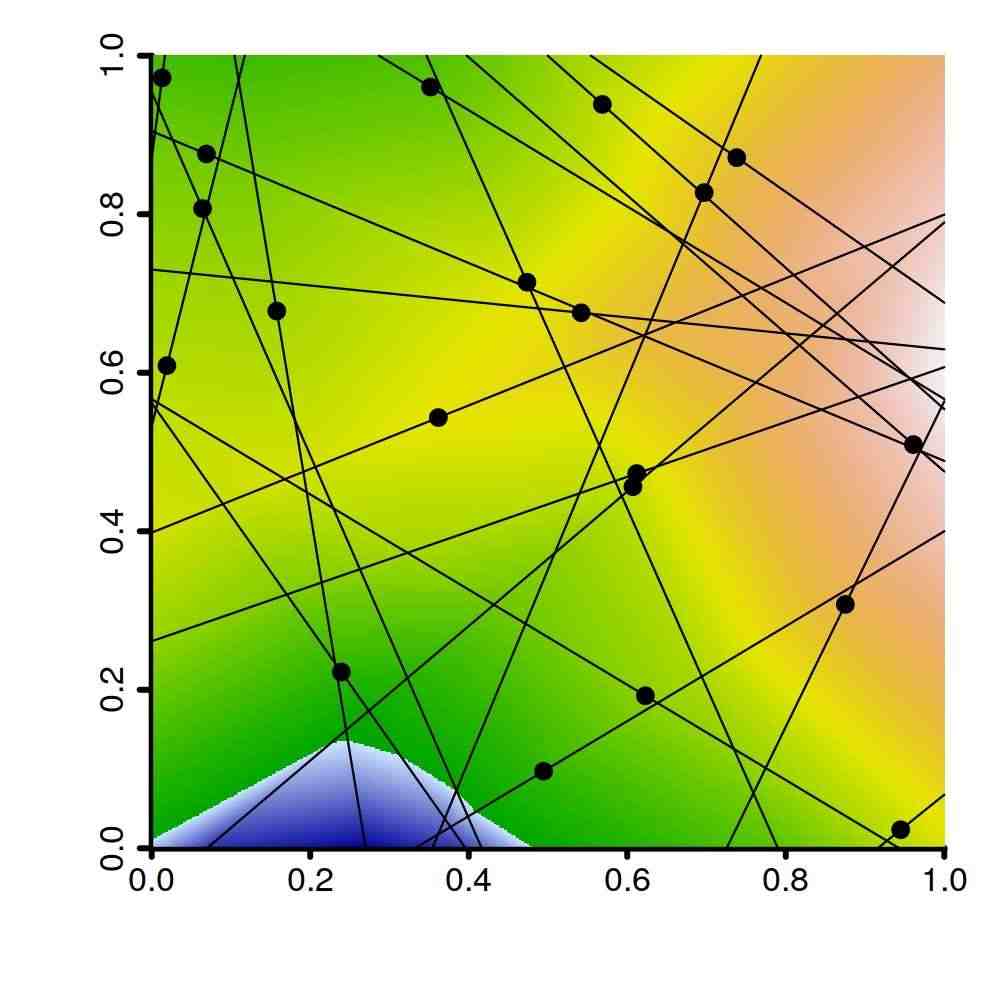}
\end{center}
\caption{Ten randomly initialized predictors of a neural network with one layer of 20 hidden neurons. The weight vectors 
were uniformly sampled from $\mathbb{S}^1$ and the biases we determined by $b_i := -\langle a_i, x_i^\star\rangle$,
where the points $x_i^\star$ depicted as black spots were sampled from  $\ca U[0,1]^2$.
The color scheme equals that of Figure \ref{figure:random-functions-he-2d}.
Compared to the initializations shown in Figures \ref{figure:random-functions-he-2d} and
\ref{figure:random-functions-nd-2d}, we see that the edges are longer in 
 the vicinity of the origin.}\label{figure:random-functions-ms-2d}
\end{figure}

The final goal of this section is to develop an initialization strategy for the offsets 
that addresses Question \textbf{Q2}. To this end, let us quickly summarize our findings that relate to \textbf{Q2}.
\begin{itemize}
 \item Lemma \ref{result:ico-char} essentially shows that the edge $x_i^*$ of a fully active neurons $h_i$ 
 (needs to) intersect the convex hull of the data.
 \item For the zero-bias initialization, Theorem \ref{result:zero-bias-general} exactly computes
 the probability of initializing a neuron in an inactive, semi-active, or fully active state respectively. 
 Unfortunately, the key quantity $P_a^d(D^\star)$ for these computations 
 depends on the unknown random geometry of the data. Under some ideal assumptions 
 on the data \eqref{D-cone-condition},
 however, the probability of an inactive neuron, may be negligible, see \eqref{prob-inactive-general}.
 \item Deterministic, non-zero bias initializations change the probability of inactive neurons, and 
 Lemma \ref{result:nonzero-bias-lemma} shows that larger values for the bias are preferable.
 \item Initializing all biases with zero forces the 
  initial function represented by the network to be positively homogeneous as discussed in Remark
  \ref{remark:zero-bias-functions}. Such functions have, 
  independent of the network width and depth, 
  very bad approximations properties.
  \item Small positive initial values for the biases create functions that are in general not 
  positively homogeneous, but at each layer, the edges of the neurons remain in the vicinity of $0$
  as discussed in Remark \ref{remark:nonzero-bias-functions}. As a result, the initial function 
  represented by the network is close to a positively homogeneous function.
\end{itemize}
In summary, the probability of inactive neurons highly depends, unlike in the one-dimensional case, 
on the geometry of the data, and therefore empirical investigations seem to be suitable 
to determine, if too many inactive neurons are actually created.
Moreover, initializing the biases with either zero or a small positive value leads to functions
with restricted approximation properties. Whether this hinders the training process needs to be investigated 
empirically, too. To this end, however, we first need to develop an alternative initialization strategy.
In view of our findings above, such a new strategy should ensure that 
\emph{a)} each edge $x_i^*$ intersects the convex hull of the data;
and
\emph{b)} the edges are not concentrated in the vicinity of the origin. 
One way to ensure both conditions is to (randomly) pick
a point $x_i^\star \in \ico D$  for each neuron $h_i$ and to initialize
the bias by $b_i:= - \langle a_i, x_i^\star\rangle$, where the weight vector 
$a_i \in \R^d$ of $h_i$ is initialized by a common strategy such as \heetal.
Indeed, a simple calculation shows $x_i^\star \in x_i^*$, and the distance 
of $x_i^*$ to the origin is given by 
\begin{displaymath}
 d(x_i^*, 0) = \frac{|\langle a_i, x_i^\star\rangle|}{\snorm {a_i}_2}\, .
\end{displaymath}
We refer to Figure \ref{figure:random-functions-ms-2d} for some illustrations 
in the case of $\ico D = (0,1)^2$.
It thus remains to develop methods for picking $x_i^\star\in  \ico D$.  
One such method would be to use the uniform distribution on the set $\ico D$.
Unfortunately, however, this choice would   require to 
find all extreme points of $\ico D$, which is, even for moderate values of   $n$ and $d$,  prohibitive.
For this reason, we consider cheap ``approximations'' of this approach. Namely, we 
first pick $N$ random samples $x_{j_1}, \dots,x_{j_N}$ from $D$, and then choose
$x_i^\star$ according to the uniform distribution on $\ico \{x_{j_1}, \dots,x_{j_N}\}$.
For computational reasons, $N$ should be small, and in our experiments reported in the following
section we therefore consider both fixed $N=5$, denoted by \scalingname{hull +5} in the experiments, and randomly chosen $N\sim \ca U(\{1,\dots,5\})$, denoted by \scalingname{hull -5}.

Moreover note that with the new strategy discussed so far, the bias 
$b_i:= - \langle a_i, x_i^\star\rangle$ may be significantly larger than $0$ and therefore we also 
investigate alternative scalings for the distribution from which the weights $a_i$ are 
initialized. These include a scaling called \scalingname{sphere} that first uses 
the normal distribution to generate the entries of a weight vector $a_i$, and then 
normalizes this weight vector with respect to the Euclidean norm. As a result,
each weight vector is uniformly sampled from $\Sd$, where $d$ is the input dimension 
of the initialized neuron. A second scaling called \scalingname{ball} multiplies the 
weight vector obtained by \scalingname{sphere} by another random number $R\sim\ca U[0,2]$.
As a result the weight vector of \scalingname{ball} is an an element of the ball with radius
2 and its expected norm equals 1.

\section{Experiments}\label{sec:experiments}

In this section we present some experiments assessing the quality of the new initialization
method and comparing it to the standard approach \heetal.
Let us begin be briefly describing the key aspects of our experiments.

\descriptpar{Data}
We downloaded all data sets from the UCI repository, that have between 2,500 
and 50,000 samples of dimension not exceeding 1,000, that were labeled as 
classification or regression task, and whose description made it straightforward to 
convert the original data set into a numeric .csv format. 
During this conversion, rows with missing values were removed, and we kept only those
data sets that still had at least 2,500 samples.
Since we were only interested in 
regression and binary classification, we extracted the largest two classes from 
the multi-class data sets and only kept the resulting binary classification data 
set if it still had at least 2,500 samples.
Some data sets are labeled both 
as regression and classification data sets, in which case we used them for both. 
Also, some data sets contained different versions, and since we were hesitating 
to choose one, we used them all. Altogether this resulted in 40 for regression and 61 data sets for binary classification.
Tables \ref{regressiondatacharacteristics} and \ref{logdatacharacteristics}
summarize key characteristics of these data sets.
Finally, we collected some data sets from other sources to 
conduct some in-front experiments for the identification of the most promising 
variants of the new initialization strategy introduced at the end of Section 
\ref{sec:general}.

\descriptpar{Hardware and Software}
We had seven desktops with varying hardware at our disposal: one with a 
GTX Titan, one with both a GTX 1060 and a GTX 1070, one with two GTX 1080,
one with a GTX 1060, and three with a GTX 1080. Except the desktop with the single 
GTX 1060, all desktops had 64GB RAM, and the first four desktops were running
\softwarename{Tensorflow 1.4}, while the 3 identical computers were running \softwarename{Tensorflow 1.10}. 
All computers were solely used for the experiments to ensure that the timing 
is as exact as possible.

\descriptpar{Initial Experiments for Exploration}
So far we used the least squares loss for the regression-type data sets 
and the logistic loss for the classification-type data sets. 
For the least squares loss we initially considered, besides the scalings \scalingname{sphere}
and \scalingname{ball}, some other but similar scalings, too. However, since these 
showed inferior performance on some initial, less structured experiments on the 
additional data, we abandoned these alternatives quickly. 
As a result of these initial experiments we decided to only consider the 
variants \scalingname{sphere hull -5}, \scalingname{sphere hull +5}, \scalingname{ball hull -5},
and \scalingname{ball hull +5} in all subsequent experiments. 
However, considering all four alternatives in the experiments would have been too
expensive, and in addition, it would have changed the character of the experiments 
from the validation of one initialization method to an exploration of different 
initialization methods. To pick one of the four variants for each loss function, 
we thus conducted 
structured experiments on the additional data sets.

\begin{table}[t]
 \begin{center}
  \begin{tabular}{|r|r|l|}
  \hline
 Architecture Number &  Depth & Widths \\ \hline
1&    2&256 -- 128\\ \hline
2&    2&512 -- 256\\ \hline
3&    2&1024  -- 512\\ \hline
4&    3&512 -- 256 -- 128 \\ \hline
5&    3&1024  -- 512 -- 256 \\ \hline
6&    3&2048 -- 1024  -- 512 \\ \hline
7&    4&512 -- 256 -- 128 -- 64 \\ \hline
8&    4&1024  -- 512 -- 256 -- 128 \\ \hline
9&    4&2048 -- 1024  -- 512 -- 256 \\ \hline
10&    8&512  -- 512 -- 256 -- 256 -- 128 -- 128 -- 64 -- 64 \\ \hline
11&    8&1024 -- 1024  -- 512  -- 512 -- 256 -- 256 -- 128 -- 128 \\ \hline
12&    8&2048 -- 2048 -- 1024 -- 1024  -- 512  -- 512 -- 256 -- 256 \\ \hline
\end{tabular}
\caption{Considered network architectures. Each number in the right column
stands for the width of one layer, and the first hidden layer corresponds to the most left number.}\label{tab:architectures}
 \end{center}
\end{table}

\descriptpar{Main Experiments}
Every data set we used from the UCI repository was randomly split into 
$60\%$ samples for training, $20\%$ samples for validation, and 
$20\%$ samples for testing. On the training samples we trained
networks of twelve architectures with depth varying between 2 and 8, see Table \ref{tab:architectures}
for details. All methods and architectures received the same splitting of the data sets.

The optimization of the network parameters was performed by
the function \softwarename{AdamOptimizer} provided by
\softwarename{Tensorflow}. The optimizer was run with its default values
and a batch size of 128. After $k$ batches, we computed both the validation and the test error, where 
\begin{displaymath}
 k = \max\Bigl\{\Bigl\lfloor  \frac {\lceil n/128\rceil} {10} \Bigr\rfloor , 5\Bigr\}\, 
\end{displaymath}
and $n$ is the size of the training set.
Consequently, for  training sets with $n < 7680$ we checked the validation error after five batches, whereas for larger
training sets we waited for more than 5 batches. 
We kept training until the validation error did not decrease for 15 epochs,
but a post analysis 
of the training log data suggested that 5 epochs would have sufficed. For this reasons, all 
experimental results we report are actually based on a patience of 5 epochs, which is possible, because we 
computed the test error whenever we computed the validation error. All timings, however, do not 
include the time needed for computing the test error. 

The training described so far yields a pair of validation and test error for each architecture, that is, 12 pairs 
altogether.
We then chose the pair with the smallest validation error and saved the corresponding test error. 
This entire procedure was repeated 50 times with different random splits, and the errors reported are the 
average test errors over these 50 repetitions. More precisely,
the reported for each method on the $i$-th data set is
\begin{align}\label{ate}
 \ate_i (\mbox{method}) := \frac 1 {50} \sum_{j=1}^{50} \te_{i,j} (\mbox{method}) \, ,
\end{align}
where $\te_{i,j} (\mbox{method})$ denotes either the classification error or the root mean squared error
of the considered  method on the $j$-th split of the  $i$-th data set.

\begin{figure}[h]
\begin{center}
\includegraphics[width=0.32\textwidth]{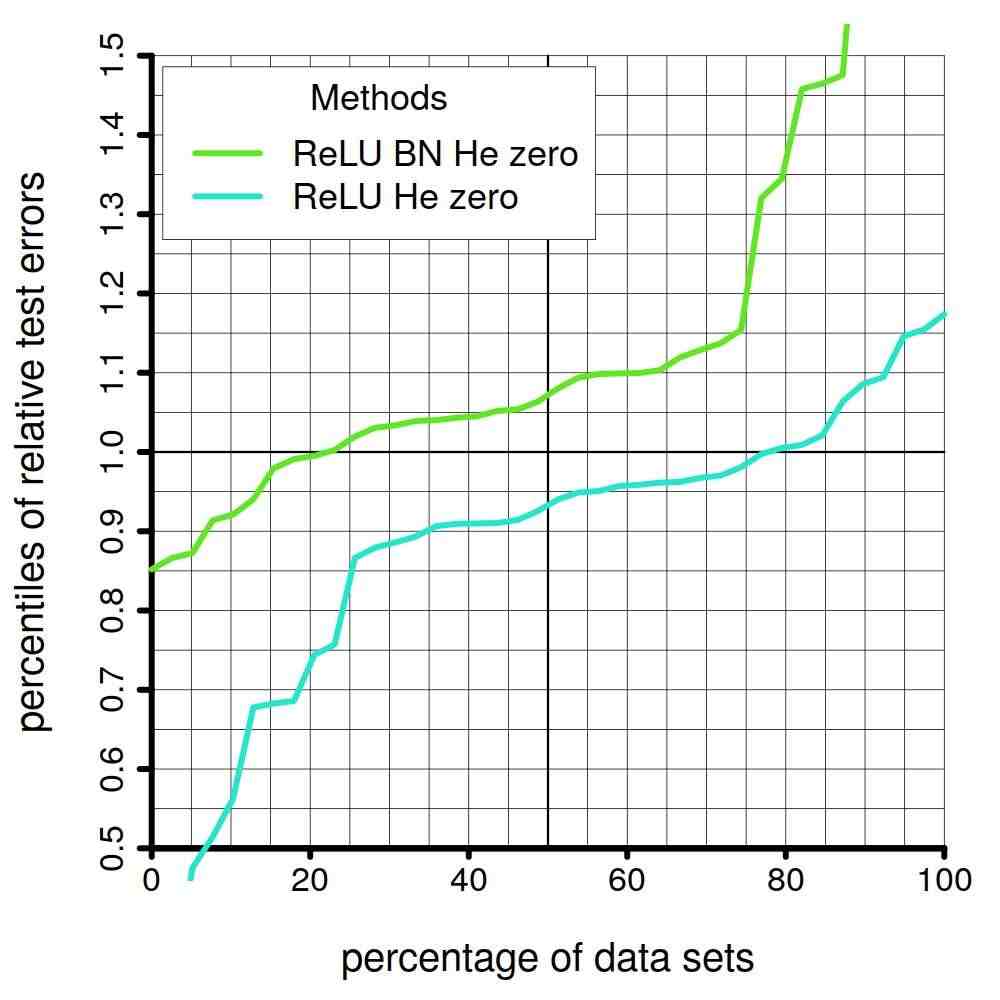}
\hspace*{-0.01\textwidth}
\includegraphics[width=0.32\textwidth]{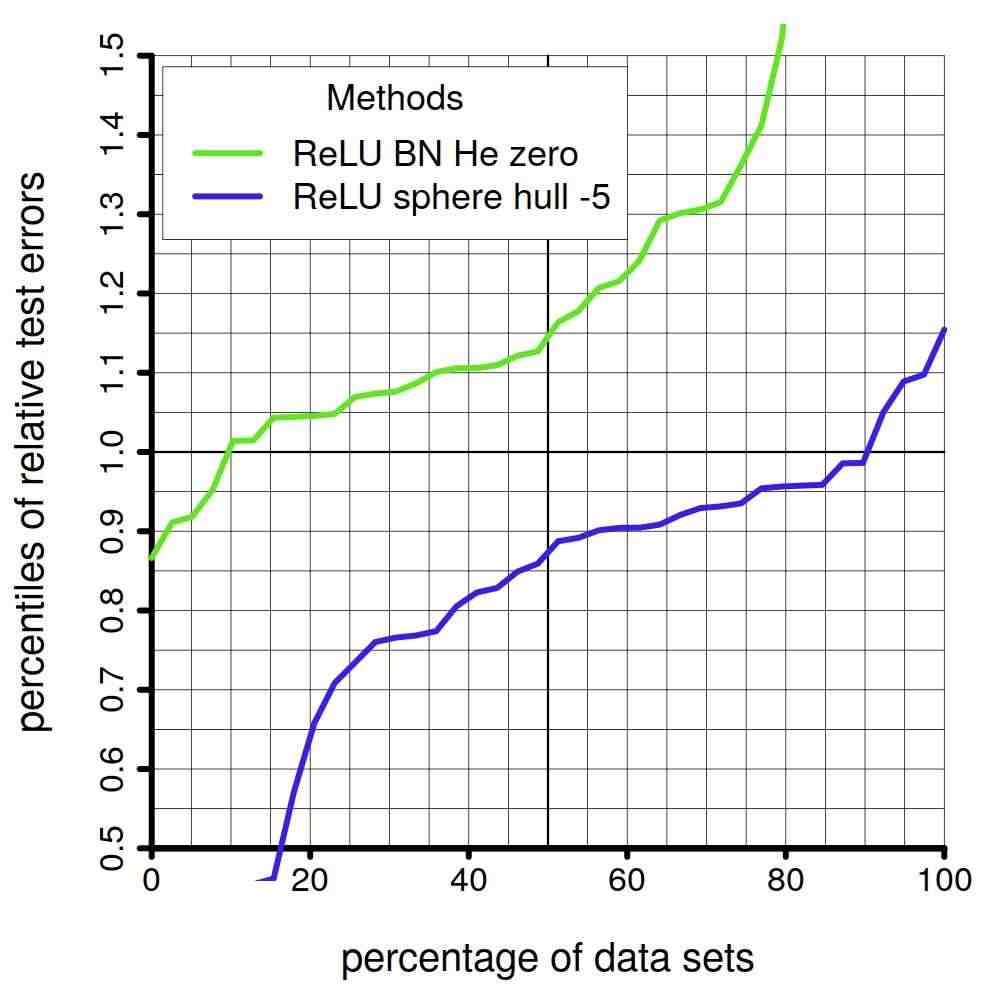}
\hspace*{-0.01\textwidth}
\includegraphics[width=0.32\textwidth]{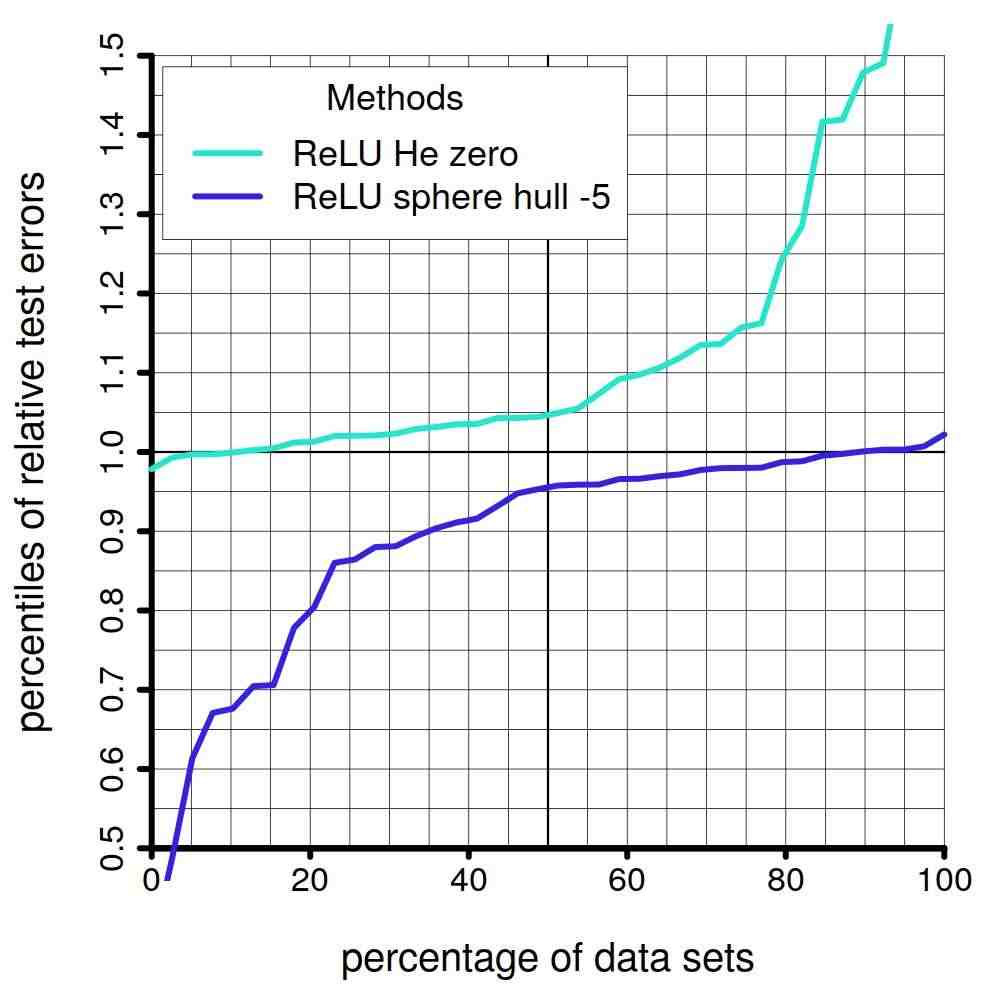}

\includegraphics[width=0.32\textwidth]{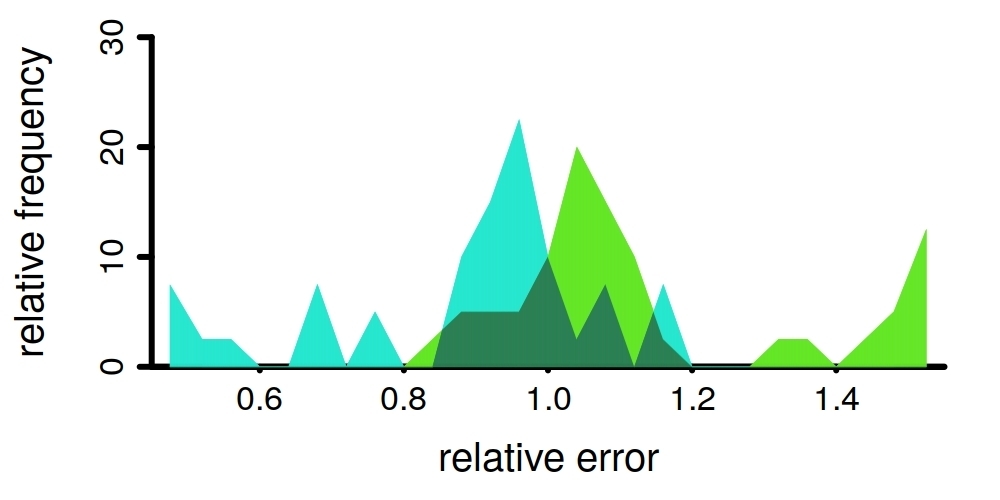}
\hspace*{-0.01\textwidth}
\includegraphics[width=0.32\textwidth]{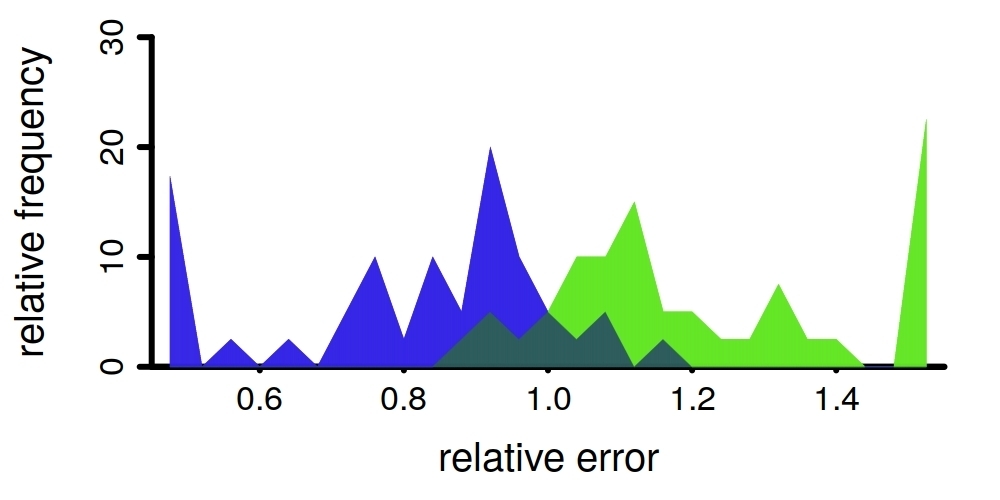}
\hspace*{-0.01\textwidth}
\includegraphics[width=0.32\textwidth]{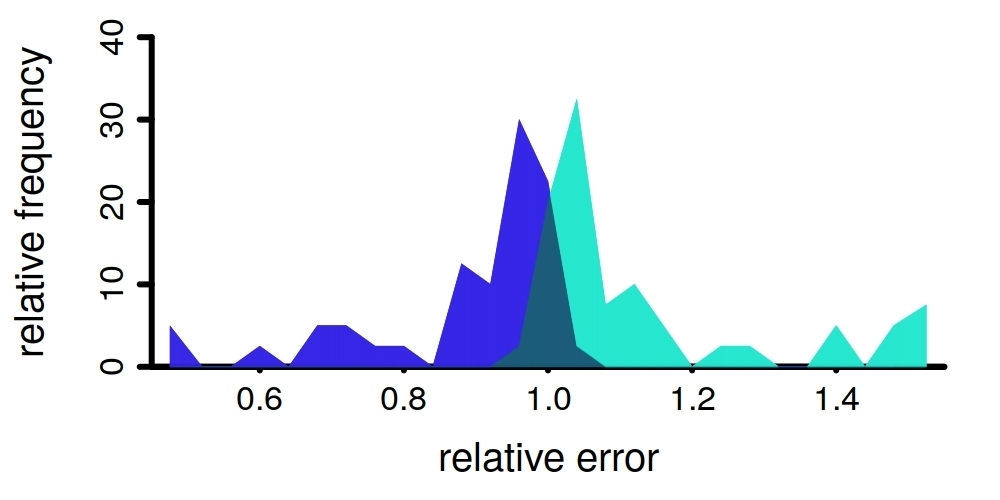}
\vspace*{-4ex}
\end{center}
\caption{Pairwise comparisons of old and new methods for regression with ReLU activation function in terms of test errors.
In the first row, each column displays the empirical percentile functions of the ``observations'' 
$y_i := \rate_i(\mbox{colored method}, \mbox{other method})$ for  $i=1,\dots,40$,
where $\rate_i(\mbox{colored method}, \mbox{other method})$
denotes the modification \eqref{safe-error-ratio} of the relative error
$\ate_i(\mbox{colored method}) / \ate_i(\mbox{other method})$ and
where $\ate_i$ denotes the usual average test error, see \eqref{ate}.
Note that the colored method is better than the other method on the $i$-th data set, if and only if the
$y_i < 1$. Consequently, colored methods, whose percentile functions stay significantly below 1 for a large range on the horizontal,
achieve   significantly better results on a large portion of the data sets compared to the competing method. For example,
the first column shows that \scalingname{ReLU He zero} outperforms 
\scalingname{ReLU BN He zero} by at least $10\%$ on about $35\%$ percent of the data sets, while conversely 
\scalingname{ReLU BN He zero} is only able to outperform \scalingname{ReLU He zero} by at least $10\%$ on about $6\%$ of the 
data sets. Similarly, the   behavior of the percentile above $1$ describes to which extend and  on how many data sets the colored method 
was outperformed by the competing method. In general, colored 
methods with small percentiles on the left of the diagram achieve 
significant gains over the competing method, whereas colored methods with small percentiles on the right do not suffer from corresponding
significant losses. 
The second row of each column displays a density estimate of the distributions of 
$\tilde y_1,\dots,\tilde y_{40}$,
where for reasons of presentation we considered the clipped values
$\tilde y_i := \max\{0.45, \min\{1.55, y_i\}\}$. The density estimate is based on histograms with a bin width of $0.04$.
A method, whose density estimate has a significant portion on $(0,1]$, achieves corresponding gains against its 
competitor, while a significant portion on $[1,\infty)$ stands for corresponding losses.
The first column shows that \scalingname{ReLU He zero} significantly outperforms \scalingname{ReLU BN He zero}, 
while the
second and third column show that our new strategy 
\scalingname{ReLU sphere hull -5}
clearly outperforms both 
\scalingname{ReLU BN He zero} and \scalingname{ReLU He zero}.
}\label{figure:first-comp-relu-regress}
\end{figure}

Besides our methods we also considered some baseline methods in the experiments. To describe them,
we write \scalingname{ReLU} if the network uses the ReLU activation function and 
\scalingname{SeLU} for Self-Normalizing Neural Networks proposed in \cite{KlUnMaHo17a}.
Moreover, weight initialization according to \heetal~with normal distributions is denoted by \scalingname{He},
and the modification for SeLUs proposed in \cite{KlUnMaHo17a} is denoted by \scalingname{SNN}.
Initializing  the bias to zero is indicated by \scalingname{zero}, and if 
batch normalization is used in the ReLU networks we additionally write \scalingname{BN}.
Now, for the classification tasks we considered \scalingname{ReLU BN He zero},
\scalingname{ReLU He zero}, and \scalingname{SeLU SNN zero} as baseline methods,
\scalingname{ReLU ball hull +5} and \scalingname{SeLU ball hull -5}  as our new methods for the two types 
of activation functions, as well as 
\scalingname{ReLU He hull -5} for illustrating the differences between 
\scalingname{ReLU He zero} and \scalingname{ReLU ball hull +5}.
Similarly, for the regression tasks we considered 
\scalingname{ReLU BN He zero},
\scalingname{ReLU He zero}, and \scalingname{SeLU SNN zero} as baseline methods
and \scalingname{ReLU sphere hull -5} and \scalingname{SeLU ball hull -5} as new methods.

In summary, each initialization strategy required 600 training  runs for each data set from the UCI repository,
that is, 37,800 runs for the classification data sets and 24,000 runs for the regression data sets.
For the classification task, we considered 6 different methods, so that in summary 225,600 networks 
were trained, whereas for the regression tasks, we have only considered 5 different methods so far, which results
in another 120,000 networks. Together the log files comprise almost 20GB of data, which can potentially be
used for further investigations, and the entire experiments took between 4 and 5 months.

\begin{figure}[t]
\begin{center}
\includegraphics[width=0.32\textwidth]{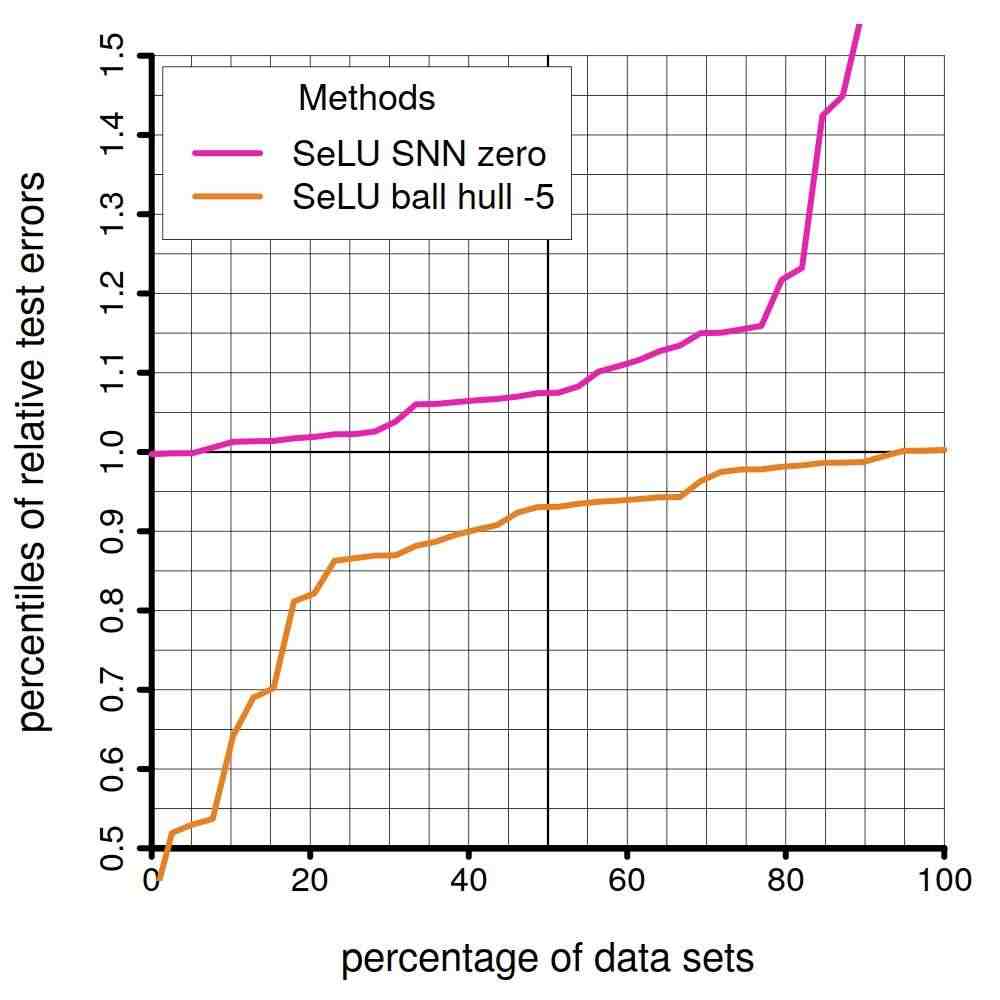}
\hspace*{-0.01\textwidth}
\includegraphics[width=0.32\textwidth]{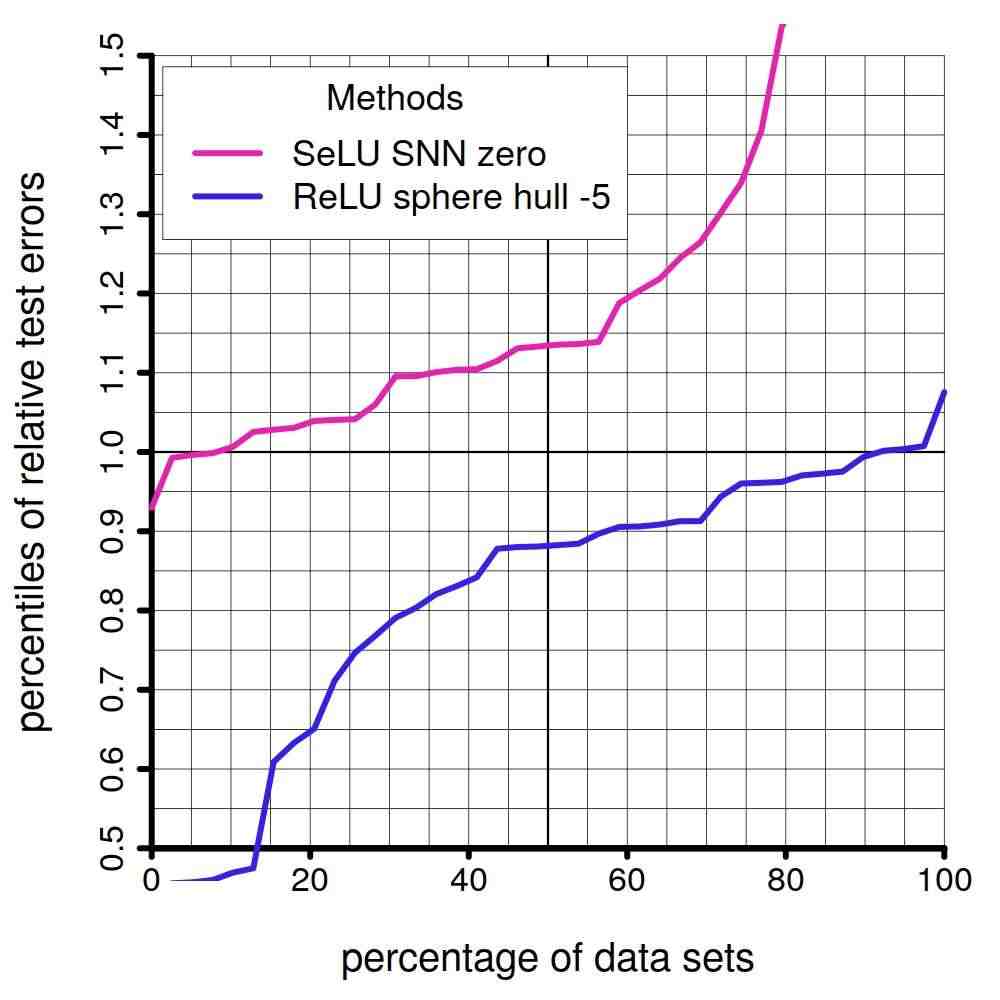}
\hspace*{-0.01\textwidth}
\includegraphics[width=0.32\textwidth]{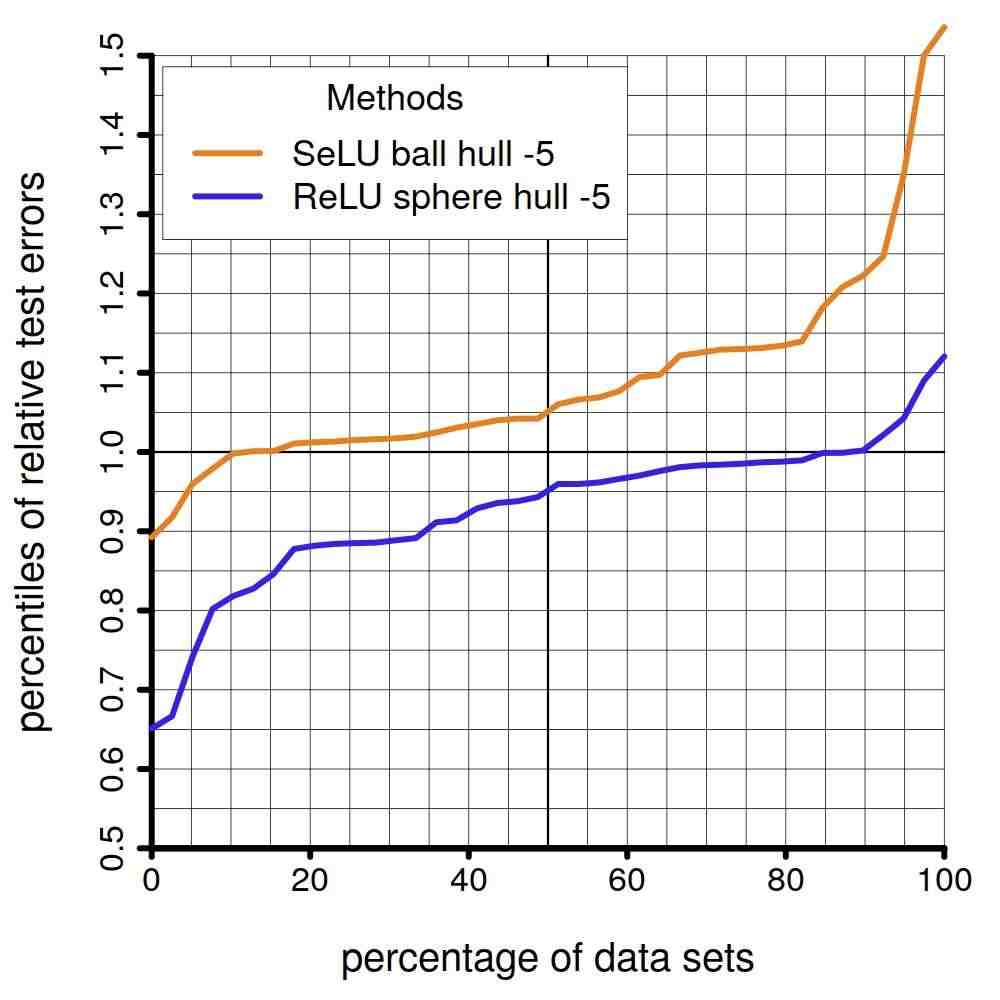}

\includegraphics[width=0.32\textwidth]{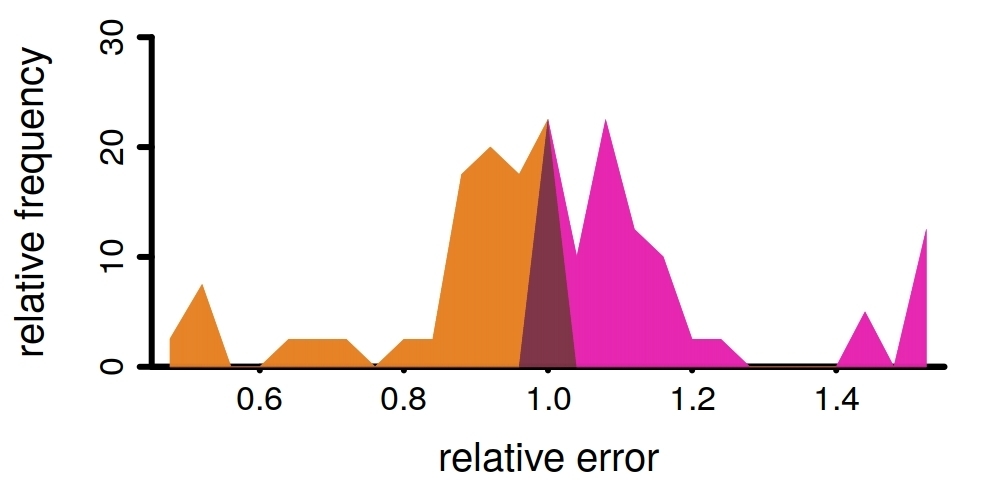}
\hspace*{-0.01\textwidth}
\includegraphics[width=0.32\textwidth]{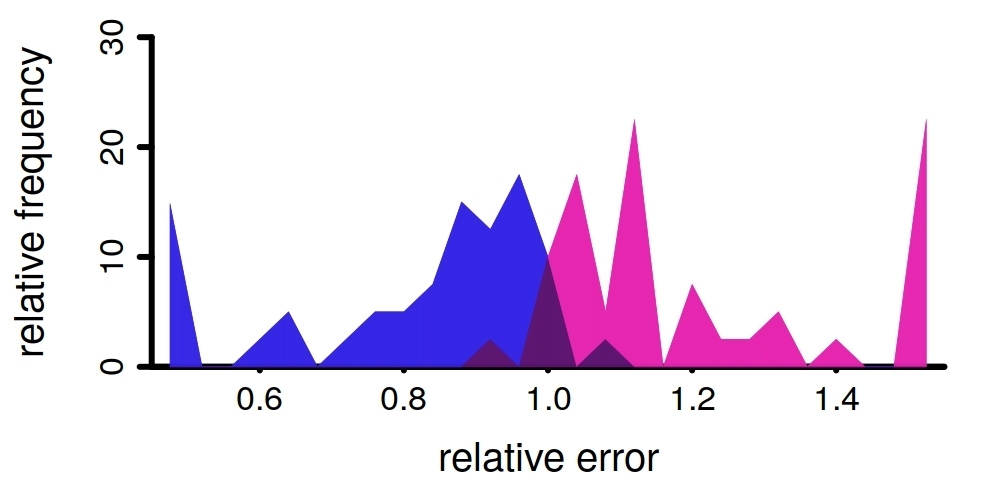}
\hspace*{-0.01\textwidth}
\includegraphics[width=0.32\textwidth]{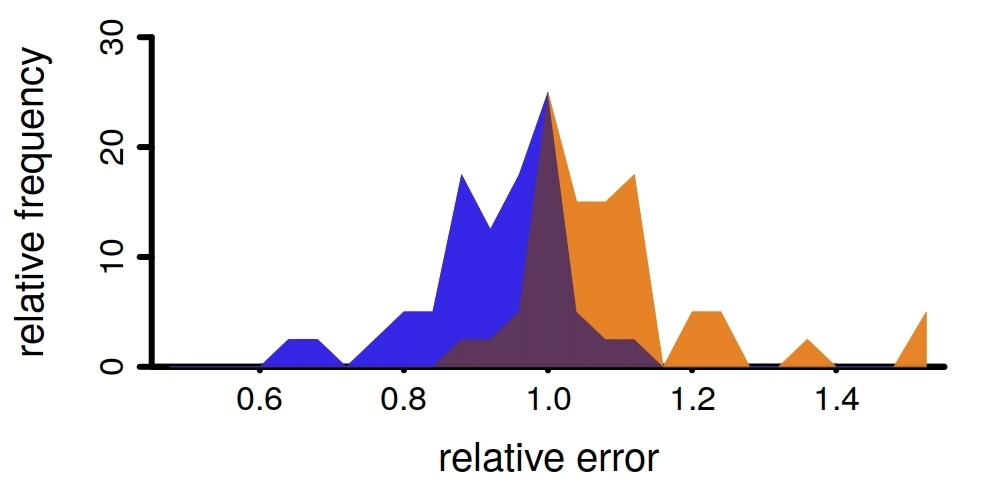}
\vspace*{-4ex}
\end{center}
\caption{Pairwise comparisons of old and new methods for regression with self-normalizing activation function
 in terms of test errors.
The graphics have the same interpretation as the ones in Figure \ref{figure:first-comp-relu-regress}. 
The first column shows that the original \scalingname{SeLU SNN zero}
proposed in \cite{KlUnMaHo17a} is clearly outperformed by our new initialization strategy for self-normalizing 
neural networks, namely \scalingname{SeLU ball hull -5}. The second column shows that 
\scalingname{SeLU SNN zero} is also outperformed by \scalingname{ReLU ball hull -5}, and the third
column shows that \scalingname{ReLU ball hull -5} also outperforms 
 \scalingname{SeLU ball hull -5}.}\label{figure:first-comp-selu-regress}
\end{figure}

\begin{figure}[t]
\begin{center}
\includegraphics[width=0.32\textwidth]{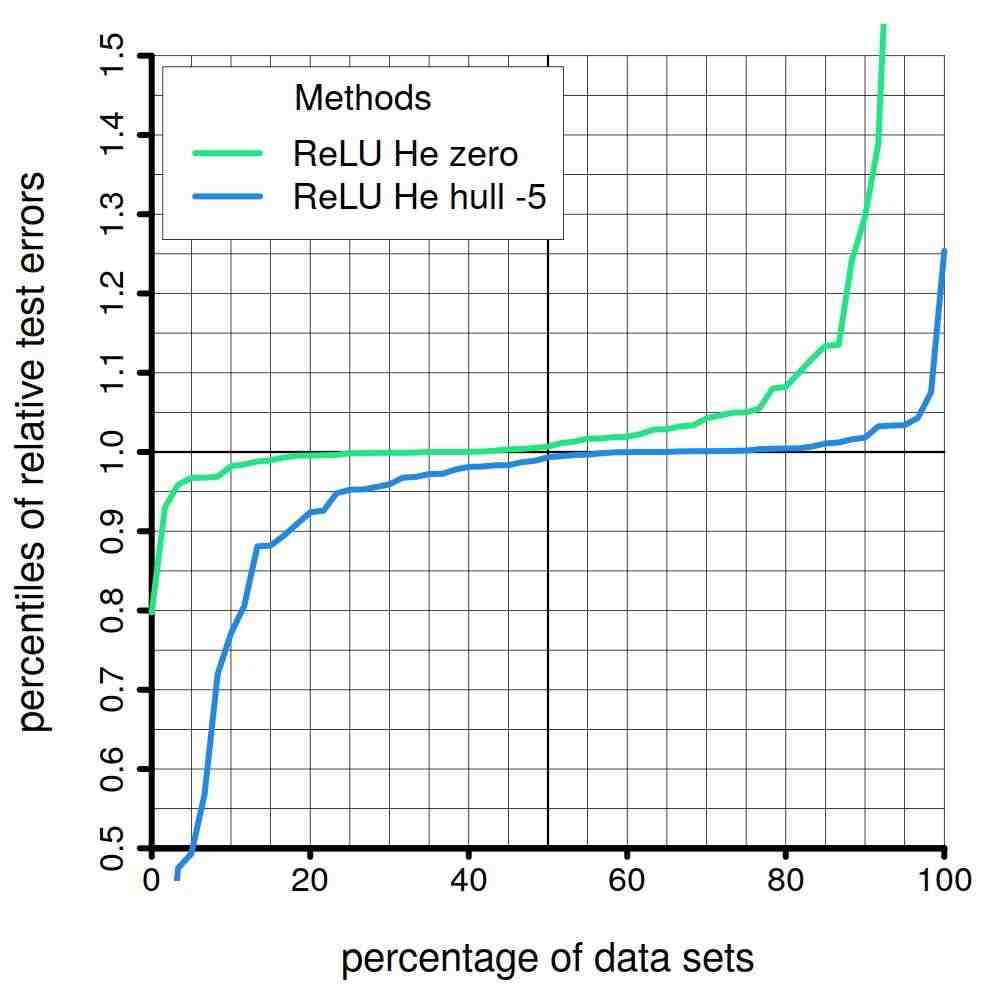}
\hspace*{-0.01\textwidth}
\includegraphics[width=0.32\textwidth]{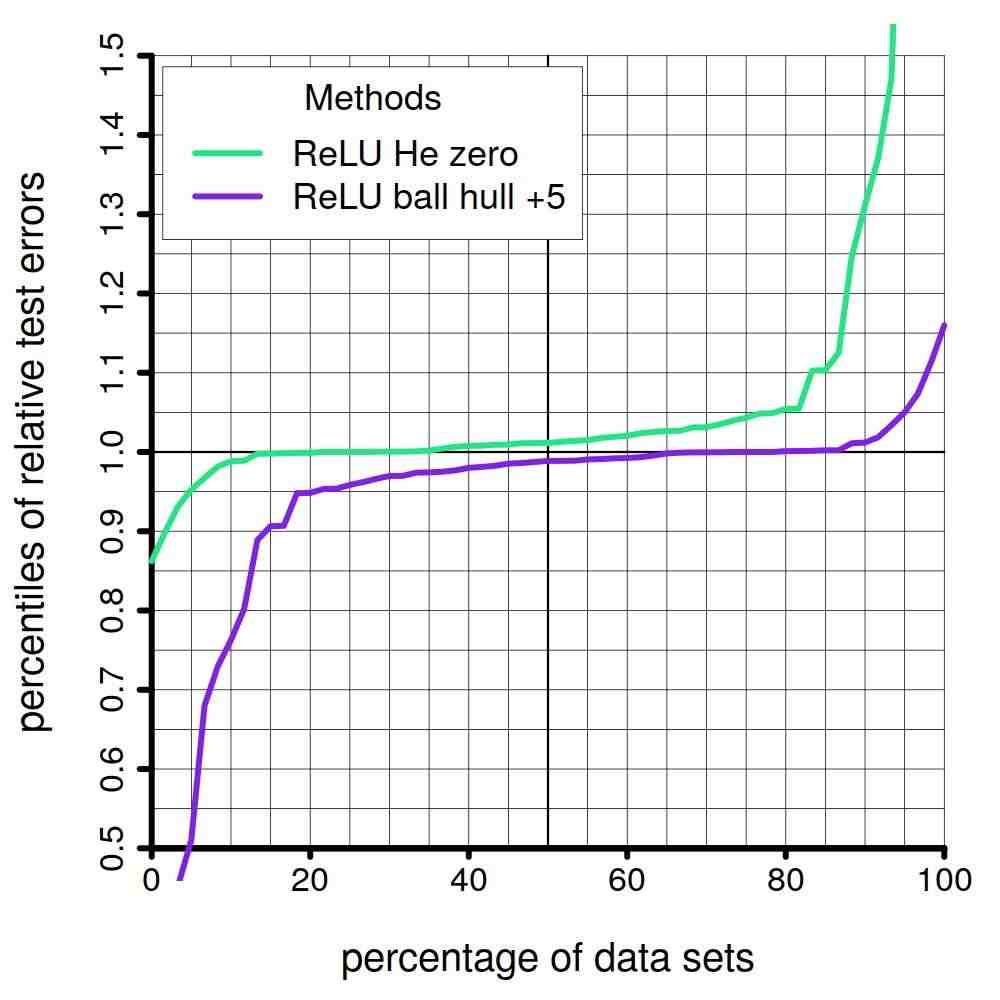}
\hspace*{-0.01\textwidth}
\includegraphics[width=0.32\textwidth]{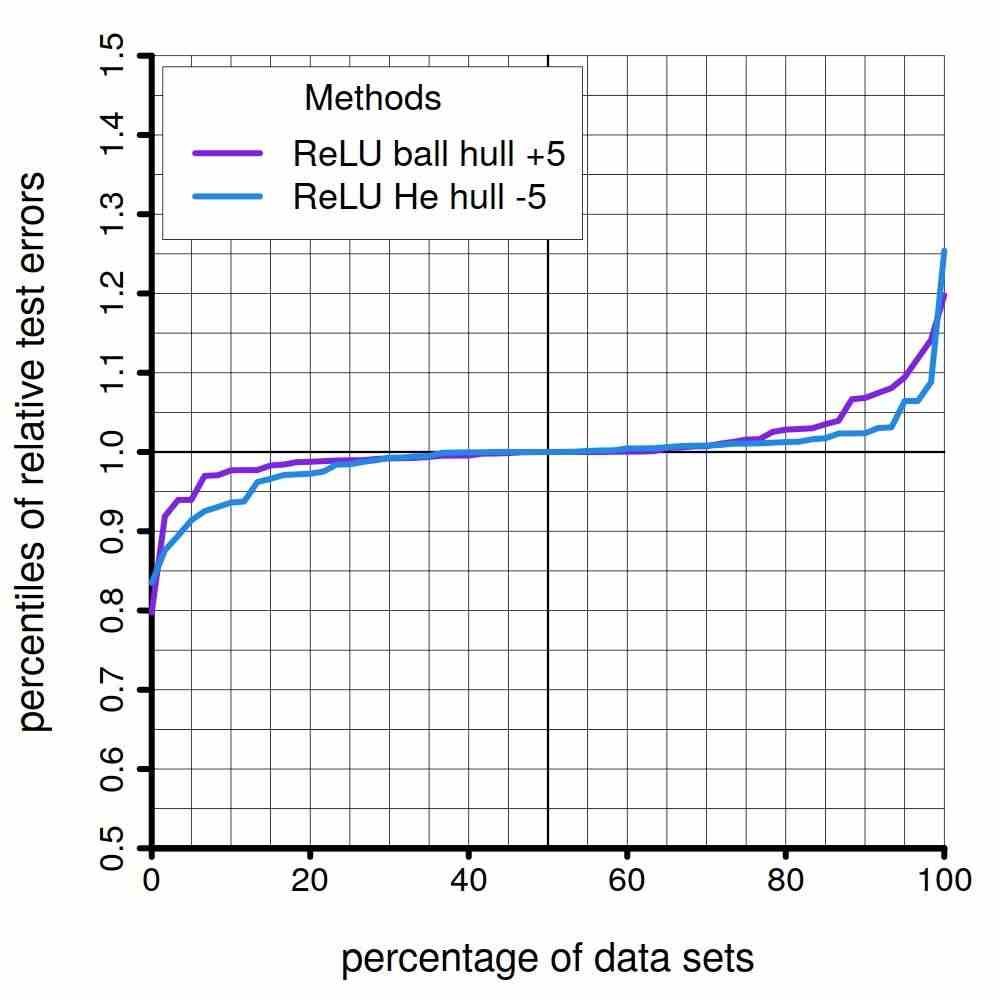}

\includegraphics[width=0.32\textwidth]{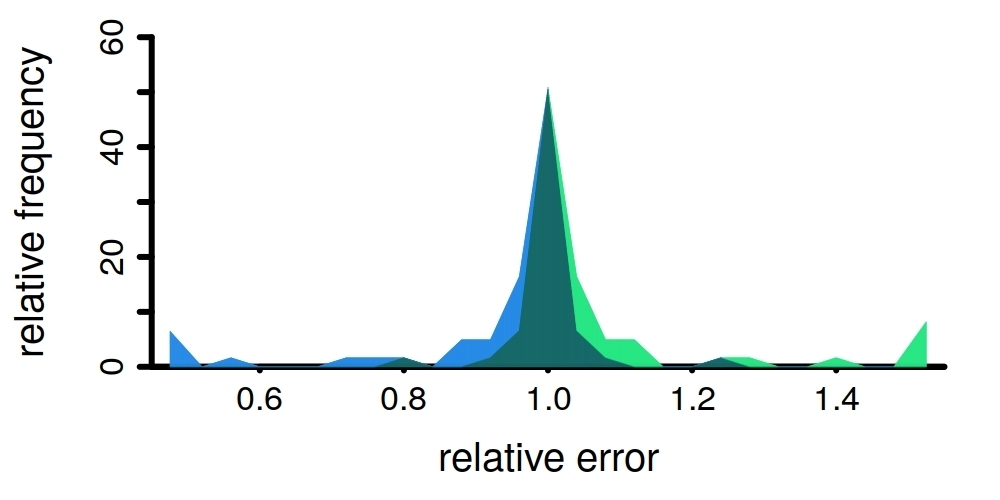}
\hspace*{-0.01\textwidth}
\includegraphics[width=0.32\textwidth]{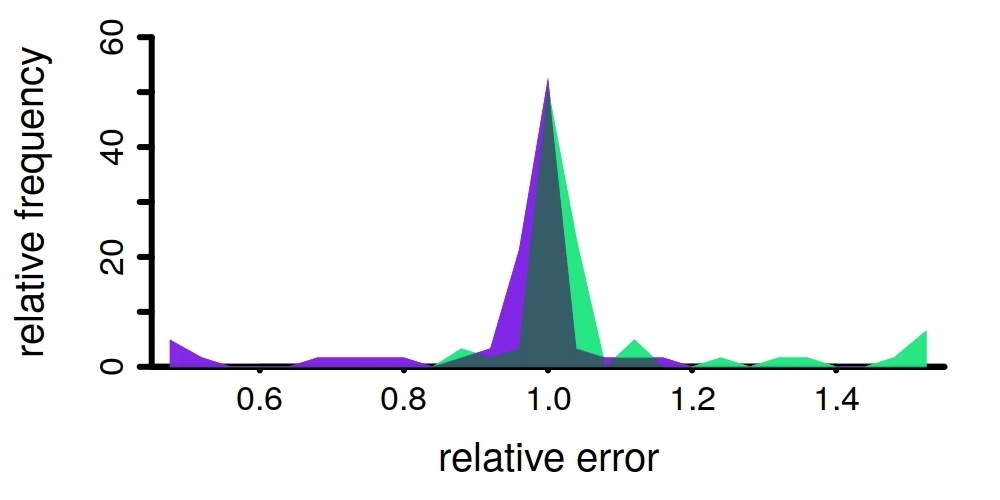}
\hspace*{-0.01\textwidth}
\includegraphics[width=0.32\textwidth]{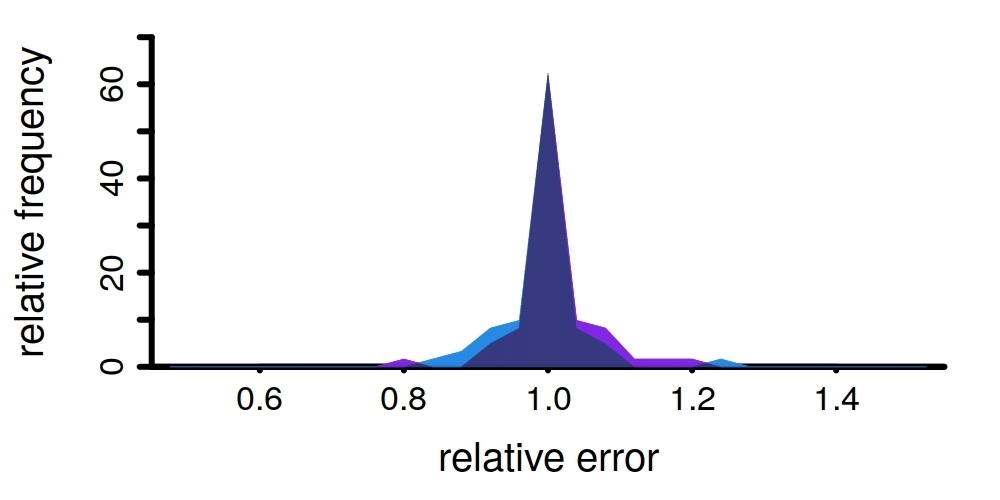}
\vspace*{-4ex}
\end{center}
\caption{Pairwise comparisons of old and new methods without batch normalization for binary classification
 in terms of test errors.
The graphics have the same interpretation as the ones in Figure \ref{figure:first-comp-relu-regress}, but 
this time we used the results on the 61 binary classification data sets 
from the UCI repository.
The figures  show that \scalingname{ReLU He zero}  is outperformed by both 
 \scalingname{ReLU ball hull +5} and  \scalingname{ReLU He hull -5}, and in the direct comparison of the latter two methods,
 \scalingname{ReLU He hull -5} has a slight edge over  \scalingname{ReLU ball hull +5}.
Also note that for each pairwise comparison there is a considerable fraction of data sets, on which both methods   
achieve an essentially equal performance. In fact, the central peak in the density estimates is always located in the 
bin $[0.98, 1.02]$, and the graphics thus show that between $50\%$ and $60\%$ of the data sets fall into this bin. In other words,
on  $50\%$ to $60\%$ of the data sets, the difference between the considered two methods is minimal in the sense of 
$ \rate_i(\mbox{colored method}, \mbox{other method}) \in [0.98, 1.02]$.
}\label{figure:first-comp-he}
\end{figure}

\descriptpar{Aspects of the Analysis}
It is common knowledge, that in many cases the (average) test errors 
greatly vary over different data sets, and that this variation is mostly due to
difference in the data sets. This phenomenon also occurred in our experiments:
In the regression case reported in Table \ref{regressionAllunlimpatfivecverrorAvA}, for example, all methods 
achieved an 
average test error of about $0.027$ on the data set 
\datasetname{online-news-popularity}, while on the data set
\datasetname{skill-craft}, the average test errors were around $1.0$.
Similarly, in the classification case reported in Table \ref{logAllunlimpatfivecverrorAvA}, all methods achieved zero
test error on \datasetname{mushroom}, while on 
\datasetname{wine-quality-all}, the test errors of all methods
were around $0.29$. For this reason, one often considers either the 
rank of each method on a fixed data set, or the relative errors, e.g.
\begin{align}\label{plain-error-ratio}
 \ate (\mbox{Method 1}) / \ate (\mbox{Method 2})
\end{align}
for each data set. In the following, we report both, but mostly with the following 
modifications: 
\begin{enumerate}
 \item There are several data sets, on which 
$\ate (\mbox{Method X}) = 0$ or $\ate (\mbox{Method X}) \approx 0$ for several methods X, see Table
\ref{logAllunlimpatfivecverrorAvA}.
For such data sets, the plain ratio   \eqref{plain-error-ratio} is either not defined, or may be 
highly misleading, and for this reason, we call the  modification 
\begin{align}\label{safe-error-ratio}
 \rate_i (\mbox{Method 1, Method 2}) := \frac{\ate_i (\mbox{Method 1}) + 0.0001}{\ate_i (\mbox{Method 2}) + 0.0001}
\end{align}
the \emph{relative average test error} of Method 1 compared to Method 2 on the $i$-th data set.
In the following, relative errors always refer to 
$\rate$ instead of \eqref{plain-error-ratio}.
\item There are also several data sets, on which most of the methods performed not exactly equally, but 
at least essentially equally. For example, in Table \ref{logAllunlimpatfivecverrorAvA}
we see that on the data set \datasetname{polish-companies-bankruptcy-2year},
three methods achieved either an average test error of  $.03932$ or $.03933$. Note that this 
data set contains 10173 samples, and hence about 2035 samples are used for testing. 
If we have two predictors that only differ on exactly one test sample, then the resulting 
test error differs by $1/2035 \approx 0.00049$. All smaller differences in the average 
test errors are therefore 
a result of averaging over 50 runs. To be more precise, a simple calculation ignoring possible  rounding errors 
in the average test errors
shows that 
the method achieving an average test error of $.03932$ predicted exactly one sample in exactly one of the 
fifty runs better than the methods achieving an average test error of $.03933$. We do not believe that 
such a small difference should result in different rankings of the methods, in particular, 
since these small differences may also result from aspects not related to the considered methods, 
e.g.~an unfortunate pick of the architecture based on the validation error. For this reason, 
we considered the following adjustment: If Method 1   performed worse than Method 2 on the $i$-th data set,
that is $\rate_i (\mbox{Method 1, Method 2}) > 1$, but we also  have 
\begin{align}\label{not-sig-worse}
   \rate_i (\mbox{Method 1, Method 2}) \leq 1.001
\end{align}
then Method 1 was viewed to have the same performance as Method 2. As a result, there are several data sets, 
in particular for the classification case, in which more than one method is considered best, even if these
methods have different average test errors, see Table \ref{logAllunlimpatfivecverrorAvA}. Moreover, 
to apply this notion of equal performance to ranking, we proceeded as follows on each data set: First we sorted the methods
according to their average test errors, and assigned them a temporary rank according to their position in the sorted list. 
Then we adjusted these temporary ranks by iteratively going from the best to the worst method. More precisely, we assigned 
all methods that did not achieve the best average test error, but that achieved \eqref{not-sig-worse} also the 
$\rank = 1$. Then we applied the same procedure to the remaining methods and so on. Finally, to ensure that 
the adjusted ranks of the considered $M$ methods sum up to standard value $M(M+1)/2$, we applied 
\softwarename{R}'s \softwarename{rank} function to the adjusted ranks
with the default ``average'' method for ties. In the Tables \ref{regressionAllunlimpatfivecverrorAvA}
and \ref{logAllunlimpatfivecverrorAvA} we report both, a ``usual'' or ``raw'' ranking that 
ignores the situation
\eqref{not-sig-worse} as well as the adjusted ranking described above. 
On the regression data sets, both types of ranking led to almost identical average rankings, which is 
not surprising since the situation \eqref{not-sig-worse} does not occur very often
in the regression case.
On the 
classification data sets, the two types of rankings led to more pronounced differences, yet the largest 
difference of both rankings was an average raw rank of $3.811$ compared to an 
average adjusted rank of $3.885$. Moreover, the ordering of the 6 considered methods 
with respect to the average (adjusted) rank did not change.
Thus it seems fair to say that both types of ranking  led to essentially the same 
results.
\end{enumerate}

\begin{figure}[t]
\begin{center}
\includegraphics[width=0.32\textwidth]{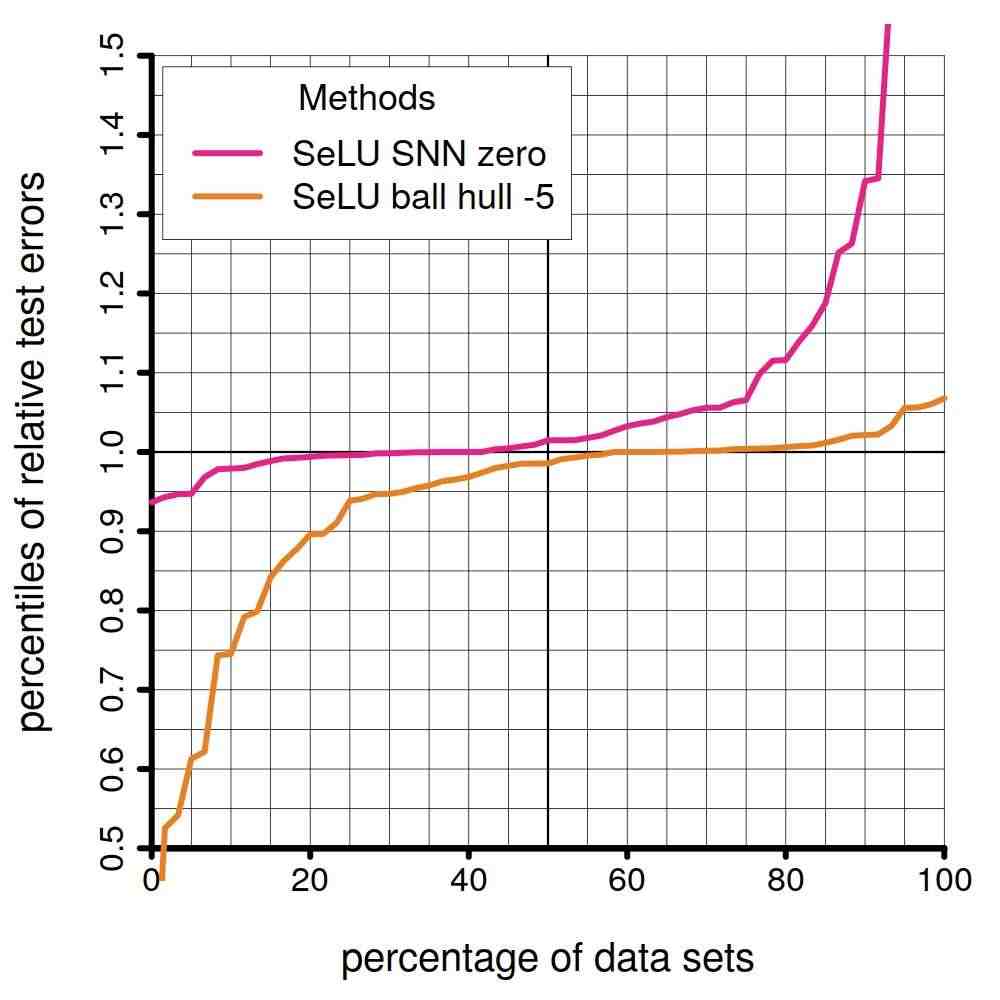}
\hspace*{-0.01\textwidth}
\includegraphics[width=0.32\textwidth]{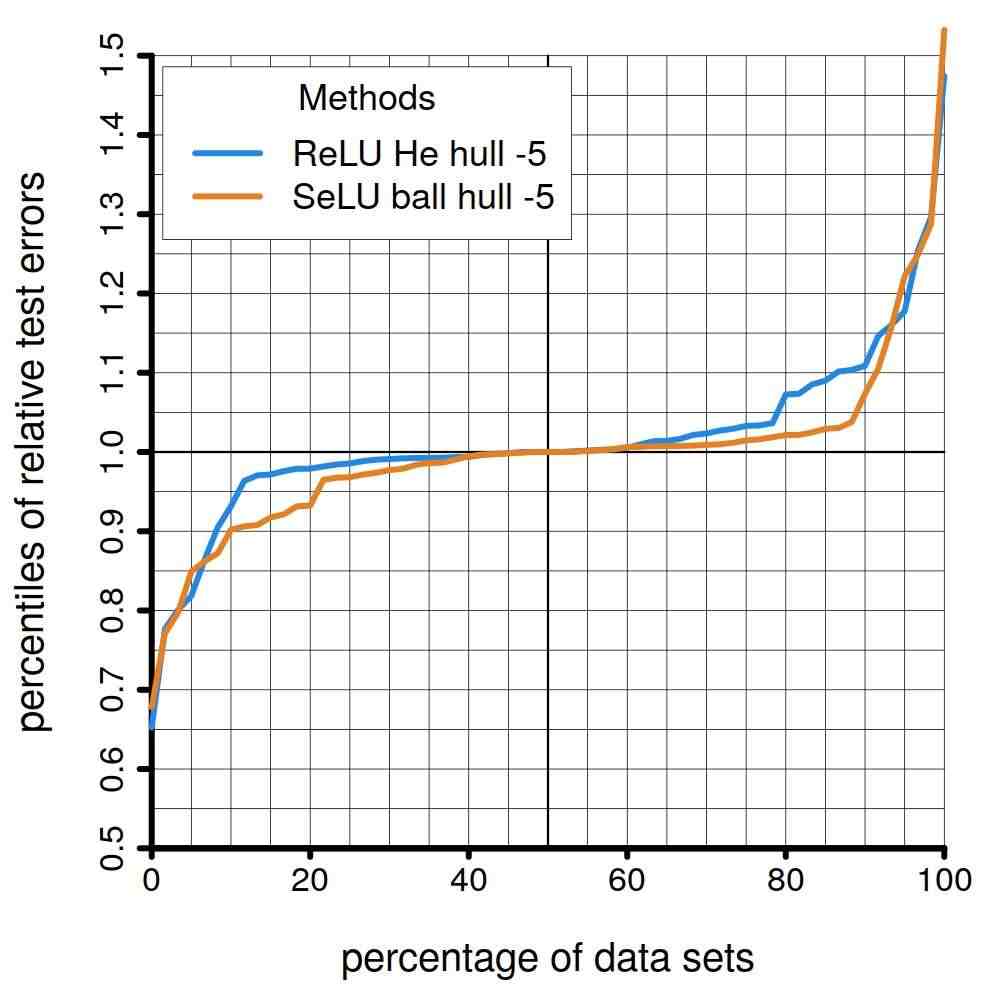}
\hspace*{-0.01\textwidth}
\includegraphics[width=0.32\textwidth]{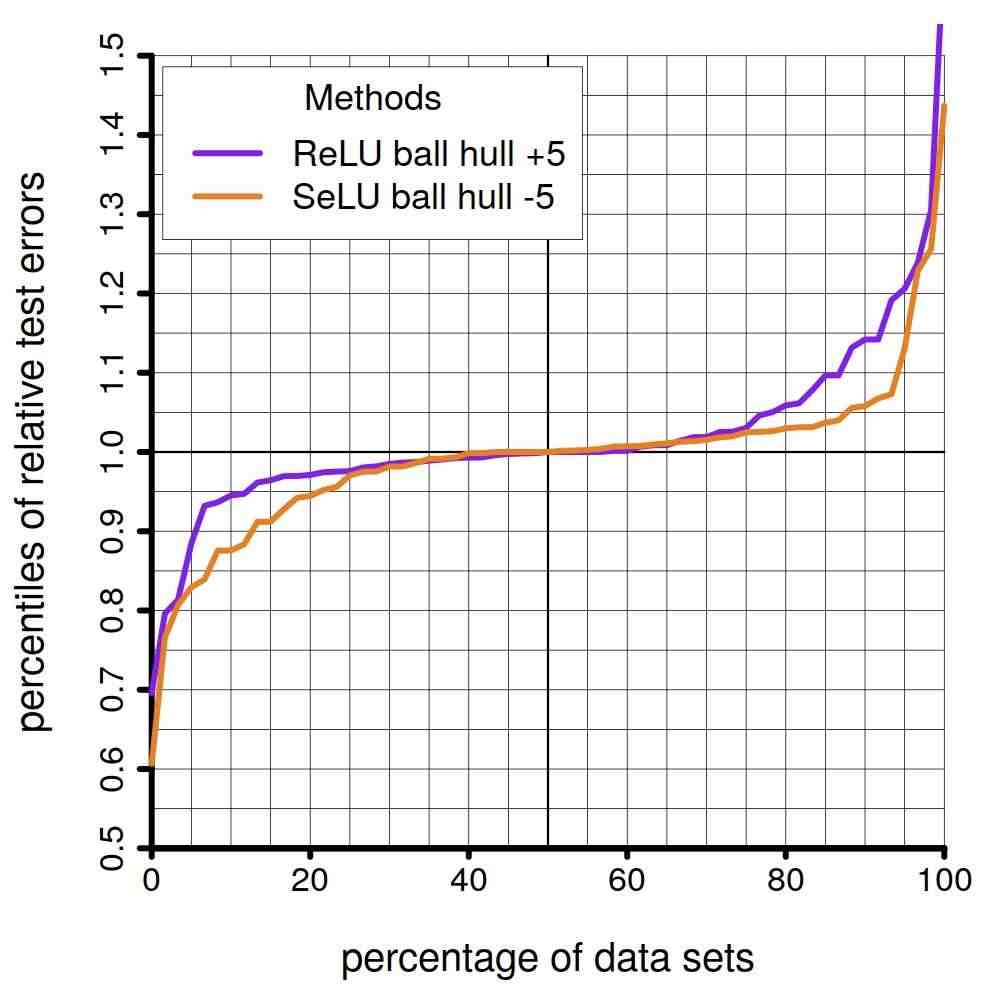}

\includegraphics[width=0.32\textwidth]{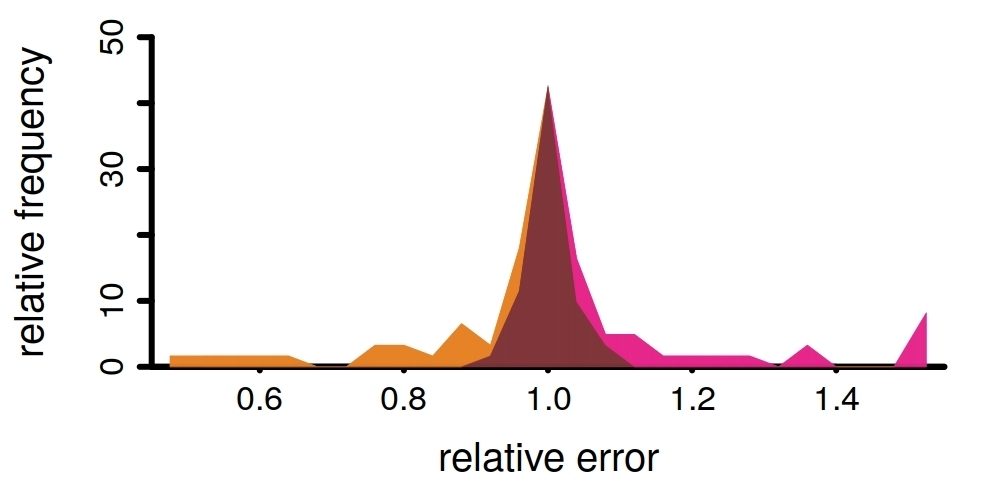}
\hspace*{-0.01\textwidth}
\includegraphics[width=0.32\textwidth]{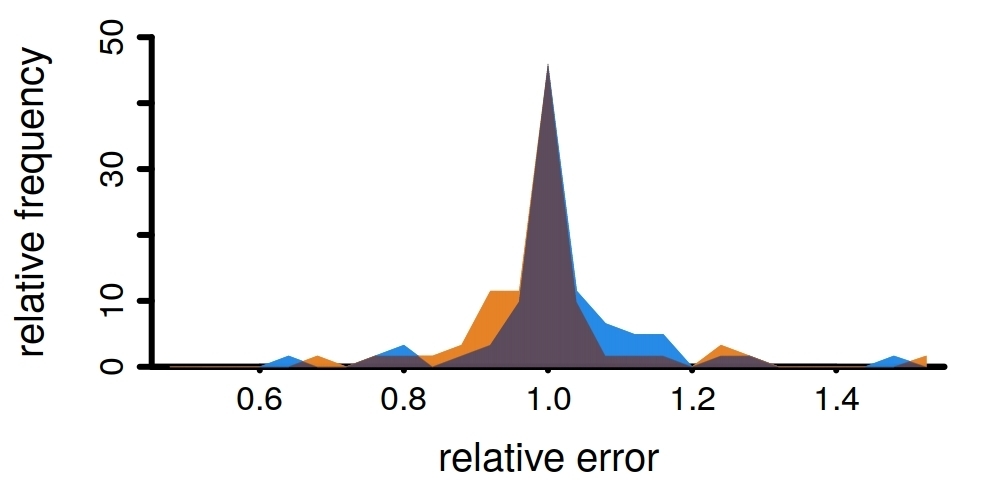}
\hspace*{-0.01\textwidth}
\includegraphics[width=0.32\textwidth]{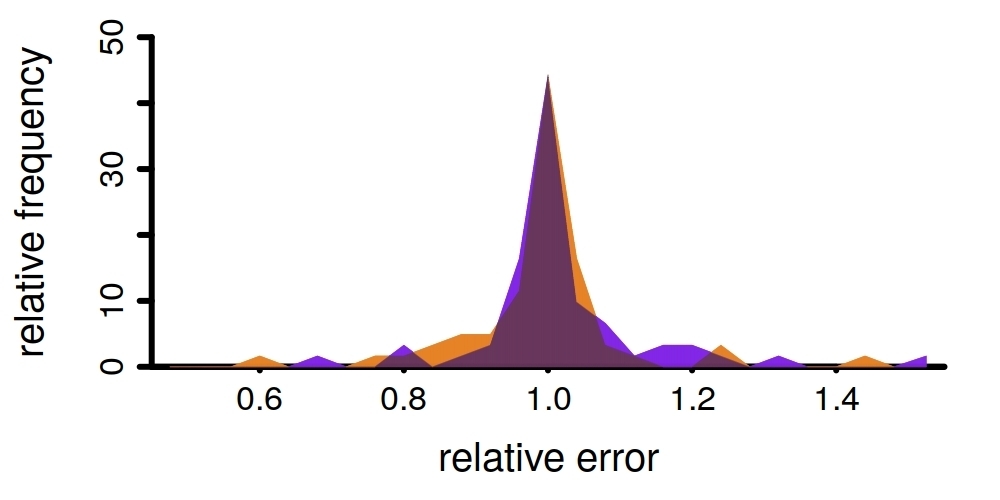}
\vspace*{-4ex}
\end{center}
\caption{Pairwise comparisons for binary classification methods with self-normalizing activation function 
 in terms of test errors.
 The graphics have the same interpretation as the ones in Figure \ref{figure:first-comp-he}. The first column shows that the original \scalingname{SeLU SNN zero}
proposed in \cite{KlUnMaHo17a} is clearly outperformed by our new initialization strategy for self-normalizing 
neural networks, namely \scalingname{SeLU ball hull -5}. The second and third column show that 
\scalingname{SeLU ball hull -5} slightly outperforms both  
\scalingname{ReLU He hull -5}  and \scalingname{ReLU ball hull +5}. Not surprisingly,
analogous comparisons between \scalingname{SeLU SNN zero} and \scalingname{ReLU ball hull +5}, respectively 
\scalingname{ReLU He hull -5}, which are not displayed here for brevity's sake, show that \scalingname{SeLU SNN zero}  is also 
outperformed by the latter two methods. 
}\label{figure:first-comp-selu}
\end{figure}

\begin{figure}[t]
\begin{center}
\includegraphics[width=0.32\textwidth]{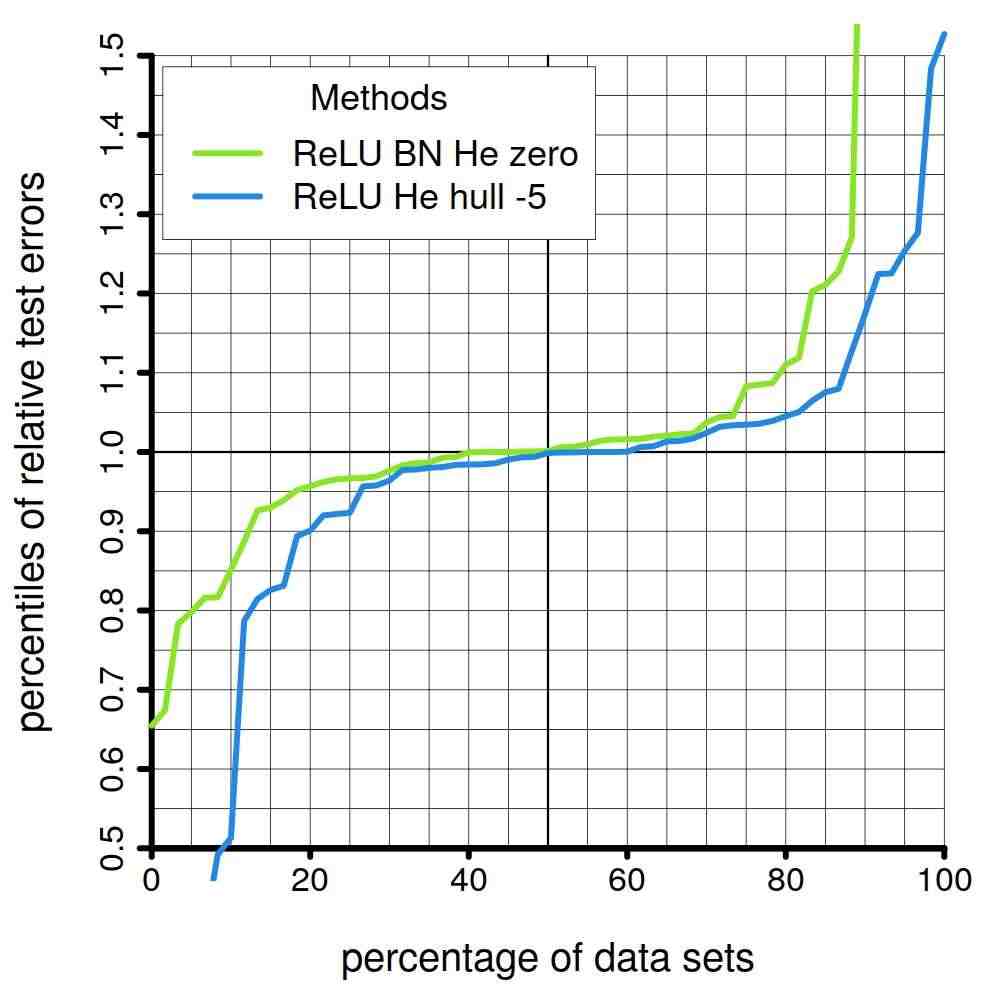}
\hspace*{-0.01\textwidth}
\includegraphics[width=0.32\textwidth]{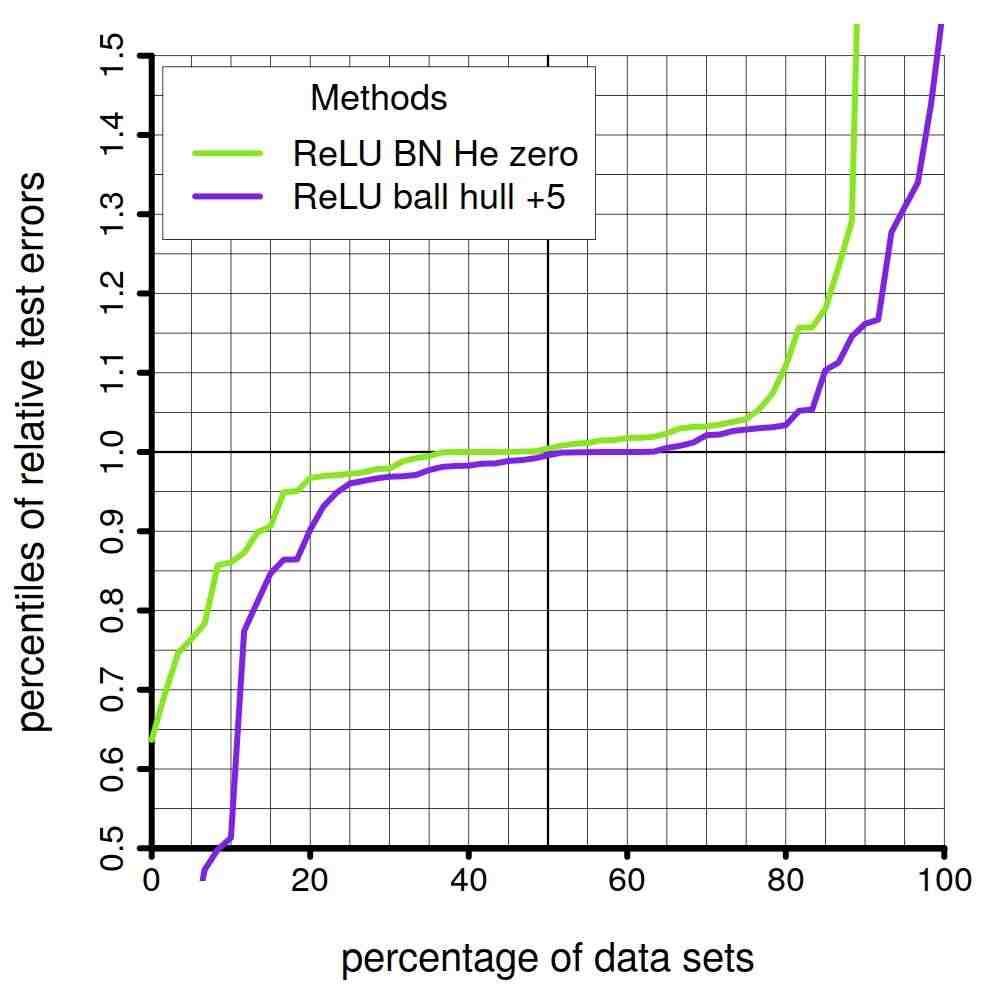}
\hspace*{-0.01\textwidth}
\includegraphics[width=0.32\textwidth]{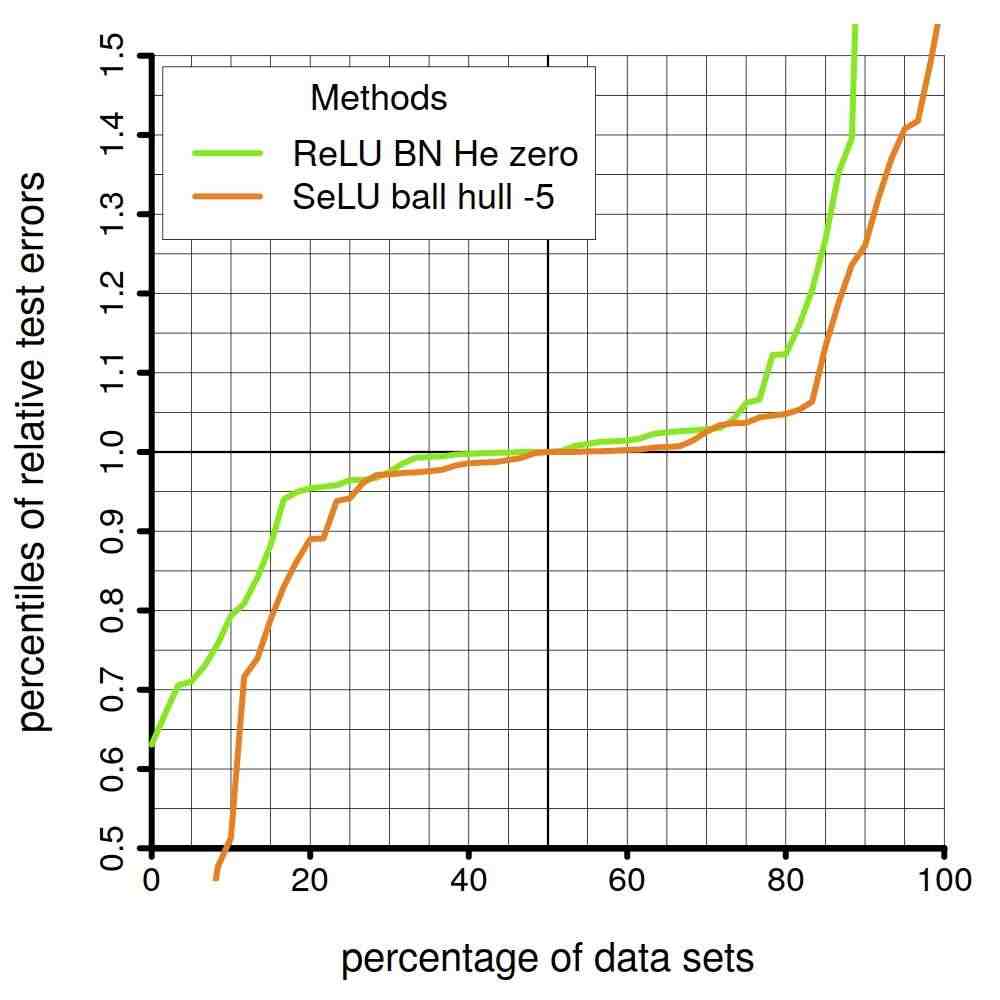}

\includegraphics[width=0.32\textwidth]{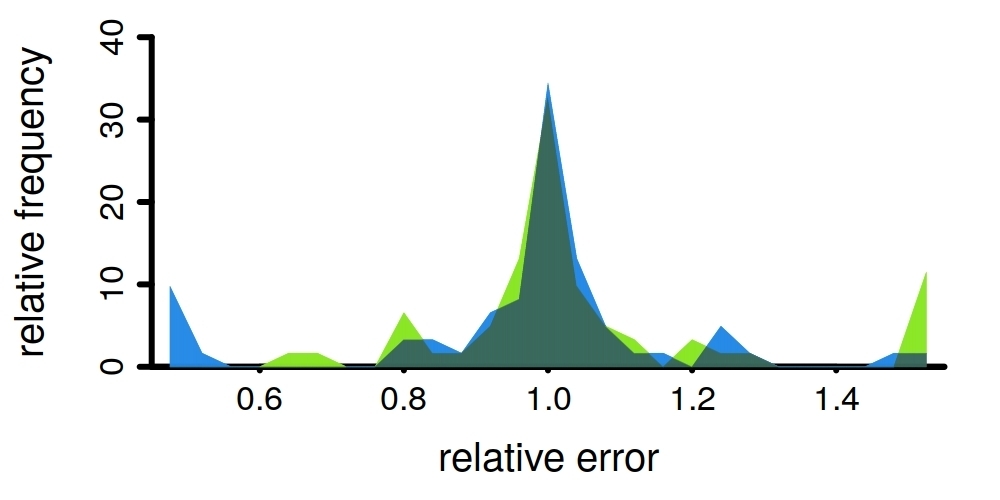}
\hspace*{-0.01\textwidth}
\includegraphics[width=0.32\textwidth]{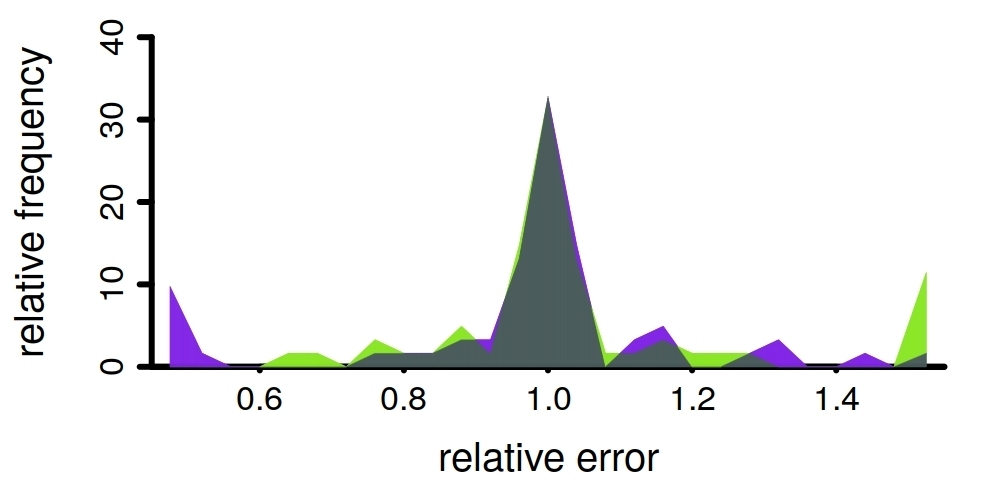}
\hspace*{-0.01\textwidth}
\includegraphics[width=0.32\textwidth]{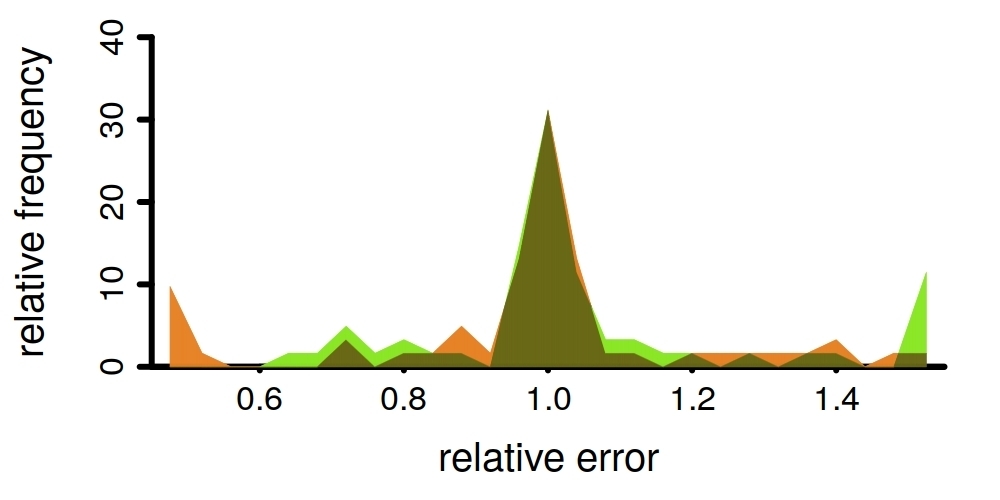}
\vspace*{-4ex}
\end{center}
\caption{Pairwise comparisons against batch normalization with standard initialization
 in terms of test errors on the binary classification data sets.
The three 
columns, which have the same meaning as in Figure \ref{figure:first-comp-he}, show that ReLUs with batch normalization 
and standard initialization are outperformed by all three  new initialization strategies. 
In particular, all three new strategies achieve at least a $10\%$ gain on at least $20\%$ of the data sets.
Conversely, \scalingname{ReLU BN He zero} only achieves a
little less than a 5\% gain over its competitors on at least $20\%$ of the data sets.}\label{figure:first-comp-bn}
\end{figure}

Tables of the form of Table \ref{regressionAllunlimpatfivecverrorAvA} and Table \ref{logAllunlimpatfivecverrorAvA}
are certainly the most common way of reporting experimental results in the machine learning community.
However, in most cases significantly less data sets are considered and in such cases, 
tables together with some simple statistics such as average rank are still comprehensible as a whole.
For more extended experiments, however, this may change. 
For example,  Table \ref{logAllunlimpatfivecverrorAvA} reports 366 average test errors, and even 
by highlighting the best and worst average test errors with the help of a color code, 
it is still rather difficult to draw conclusions from 
Table \ref{logAllunlimpatfivecverrorAvA}. Indeed, a full understanding of the performance of 
different methods requires, besides rankings and an emphasis on best and worst behavior, also 
an understanding of the distribution of relative average test errors. To be more specific, consider 
the results on the data set \datasetname{avila} reported in Table 
\ref{logAllunlimpatfivecverrorAvA}. Here, the method \scalingname{ReLU ball hull -5} scores third, while
\scalingname{SeLU SNN zero} scores fourth. Consequently, neither of the two methods are 
highlighted in \ref{logAllunlimpatfivecverrorAvA} and their ranking on this data set does not
substantially influence their average ranking. Nonetheless, their performance drastically differs
since \scalingname{ReLU ball hull -5} achieves an average test error of $0.10157$, while
\scalingname{SeLU SNN zero} only achieves an average test error of $0.14361$. Consequently, we have 
\begin{displaymath}
   \rate_6 (\mbox{\scalingname{ReLU ball hull -5}, \scalingname{SeLU SNN zero}}) \approx 1.413\, ,
\end{displaymath}
that is, on \datasetname{avila}, the average test error of \scalingname{SeLU SNN zero}  is more than $40\%$ worse than that of 
\scalingname{ReLU ball hull -5}. Of course, all this information is contained in Table \ref{logAllunlimpatfivecverrorAvA},
but it requires at least substantial effort to extract and comprehend this information.
For this reason, we also display pairwise comparisons of selected methods with the help of 
percentile functions on the relative average test errors. We refer to Figure \ref{figure:first-comp-relu-regress}
for a detailed explanation of these graphics and to 
Figures 
\ref{figure:first-comp-selu-regress},
\ref{figure:first-comp-he},
\ref{figure:first-comp-selu}, and
\ref{figure:first-comp-bn}
for further pairwise comparisons.

\descriptpar{Findings}
Let us now have a look at some of the results to assess the quality of the new initialization strategies.
To this end, we focus on the following aspects:
\begin{enumerate}
 \item Average test errors 
 \item Training costs in number of iterations and training time
 \item Influence of the considered architectures on the test errors
\end{enumerate}

\emph{i).}  Let us first consider average test errors. In the regression case, Table \ref{regressionAllunlimpatfivecverrorAvA}
immediately shows that the two new methods \scalingname{ReLU sphere hull -5} and \scalingname{SeLU ball hull -5}
are ranked first and second, and that  \scalingname{ReLU sphere hull -5}, which is ranked first, actually
achieves the best average test error of all methods
on $75\%$ of the data sets. In most cases, these test errors are statistically significant better than the second best 
test errors.
Moreover, Figure \ref{figure:first-comp-relu-regress} shows that the new initialization strategy \scalingname{ReLU sphere hull -5}
outperforms both \scalingname{ReLU BN He zero} and \scalingname{ReLU He zero} on around $90\%$ of the data sets, and on 
a considerable number of data sets, the gains achieved by \scalingname{ReLU sphere hull -5} is very substantial.
Finally, Figure \ref{figure:first-comp-selu-regress} shows that self-normalizing networks with standard
initialization, that is \scalingname{SeLU SNN zero}, are almost uniformly outperformed by both of the 
new initialization strategies, i.e.~\scalingname{ReLU sphere hull -5} and \scalingname{SeLU ball hull -5}. This figure 
further shows that \scalingname{ReLU sphere hull -5}  outperforms \scalingname{SeLU ball hull -5} on around $90\%$ of the data sets.

\begin{figure}[t]
\begin{center}
\includegraphics[width=0.32\textwidth]{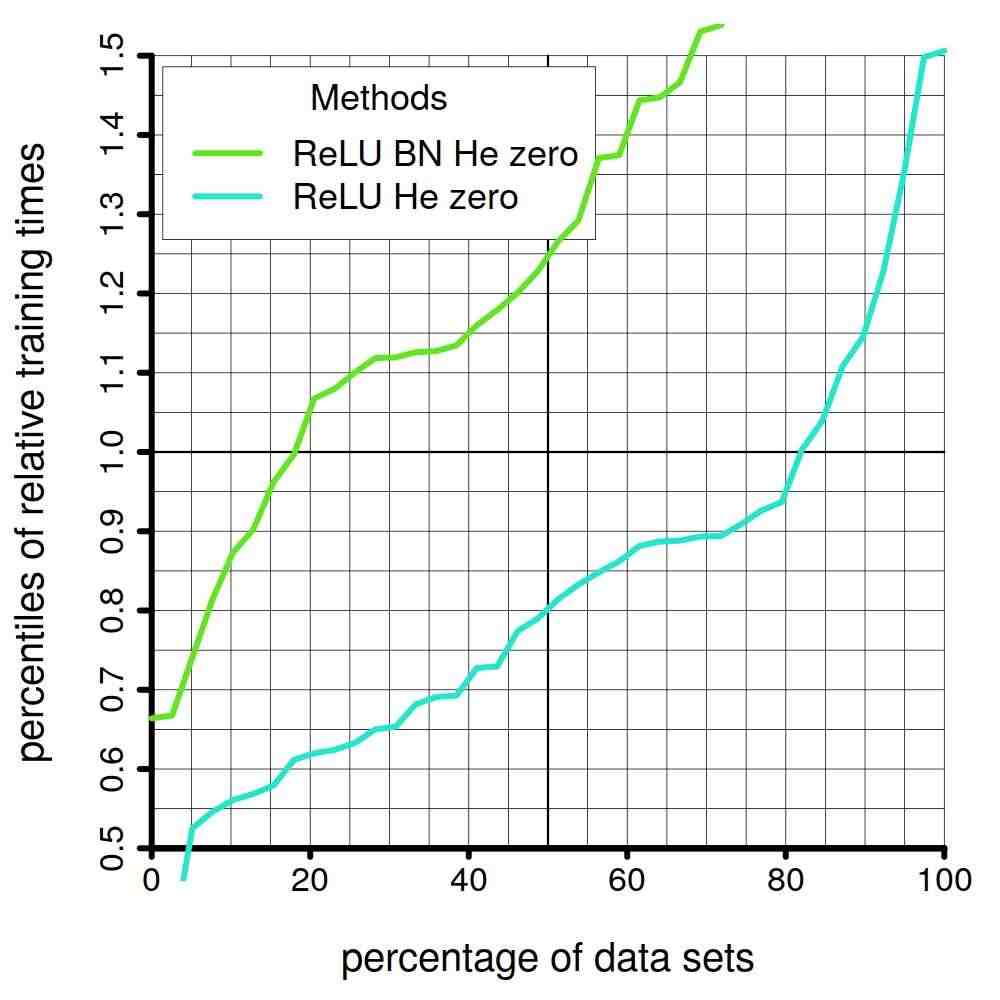}
\hspace*{-0.01\textwidth}
\includegraphics[width=0.32\textwidth]{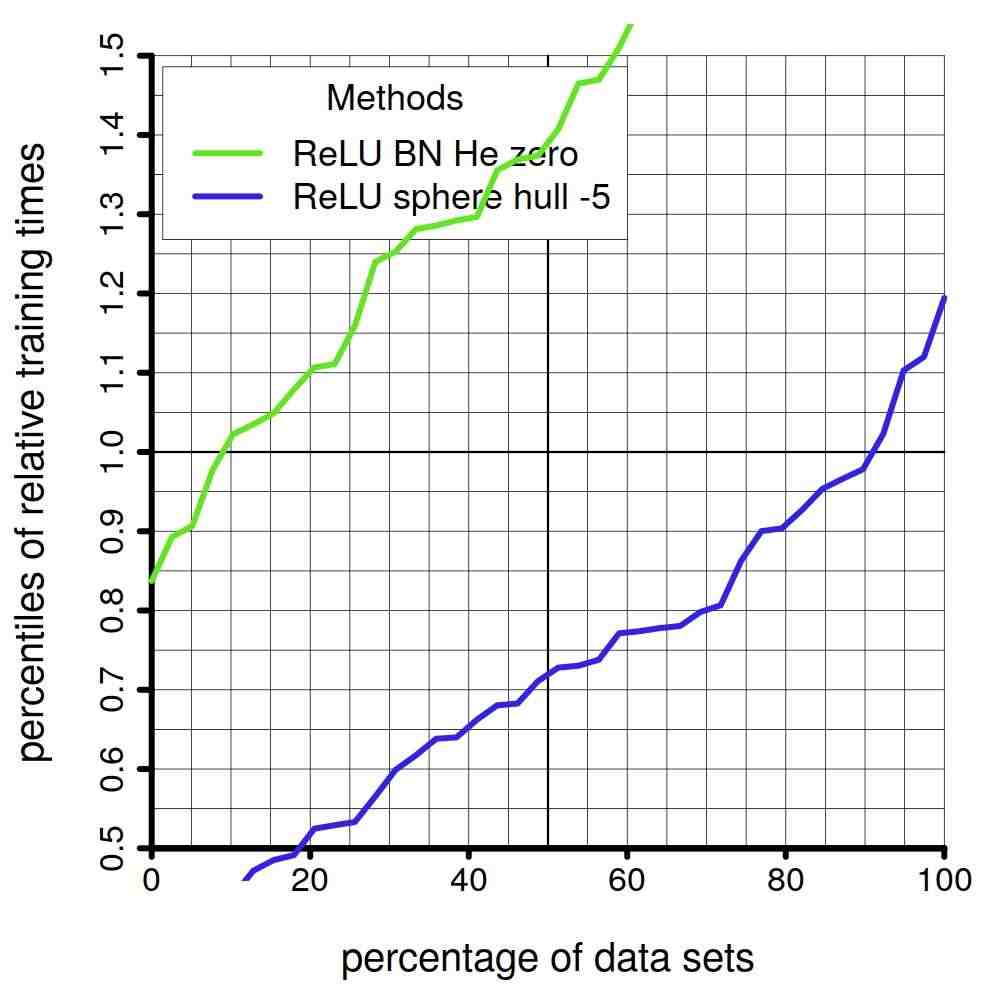}
\hspace*{-0.01\textwidth}
\includegraphics[width=0.32\textwidth]{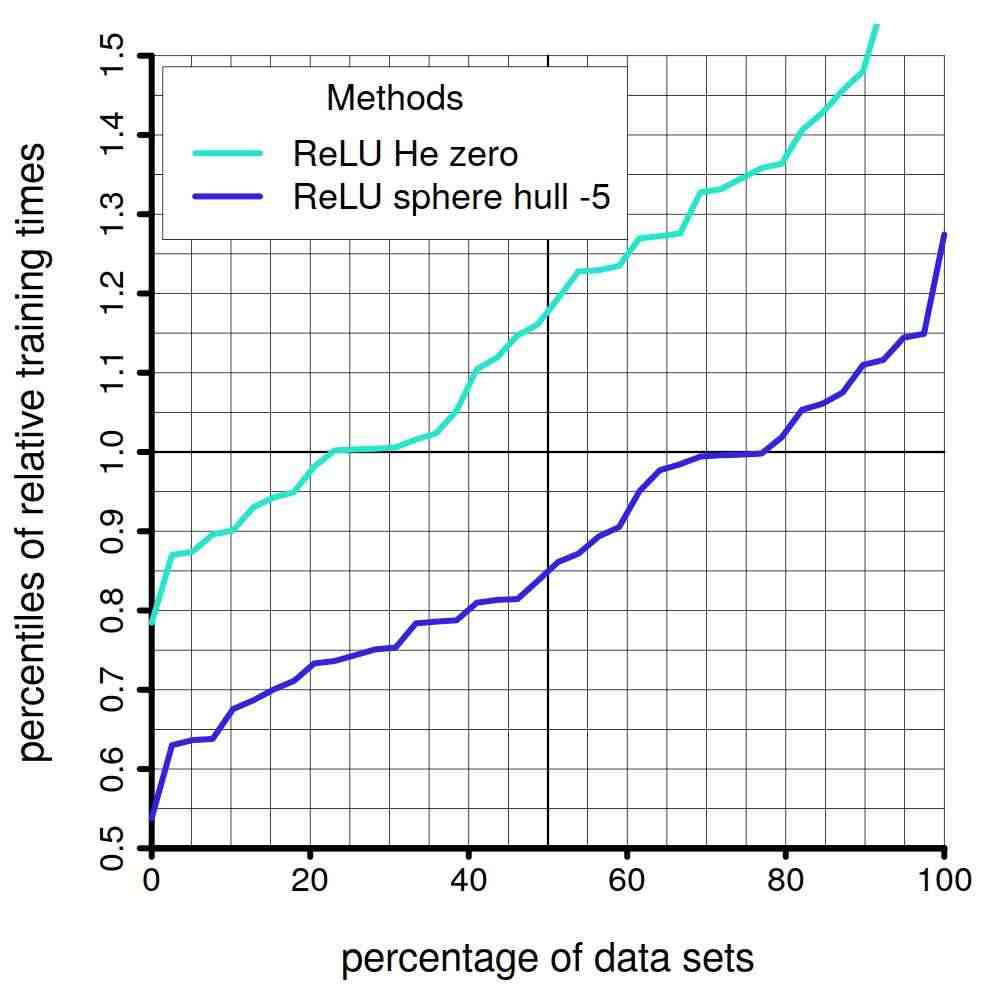}

\includegraphics[width=0.32\textwidth]{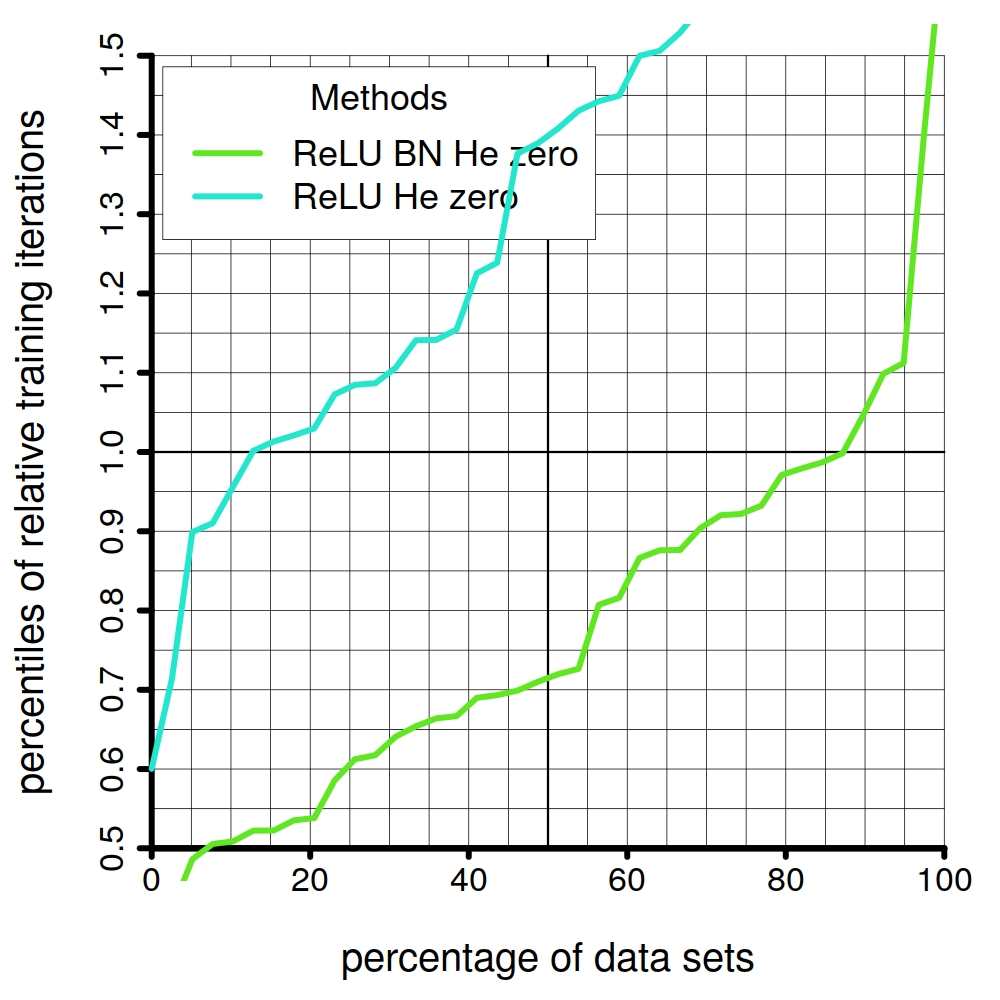}
\hspace*{-0.01\textwidth}
\includegraphics[width=0.32\textwidth]{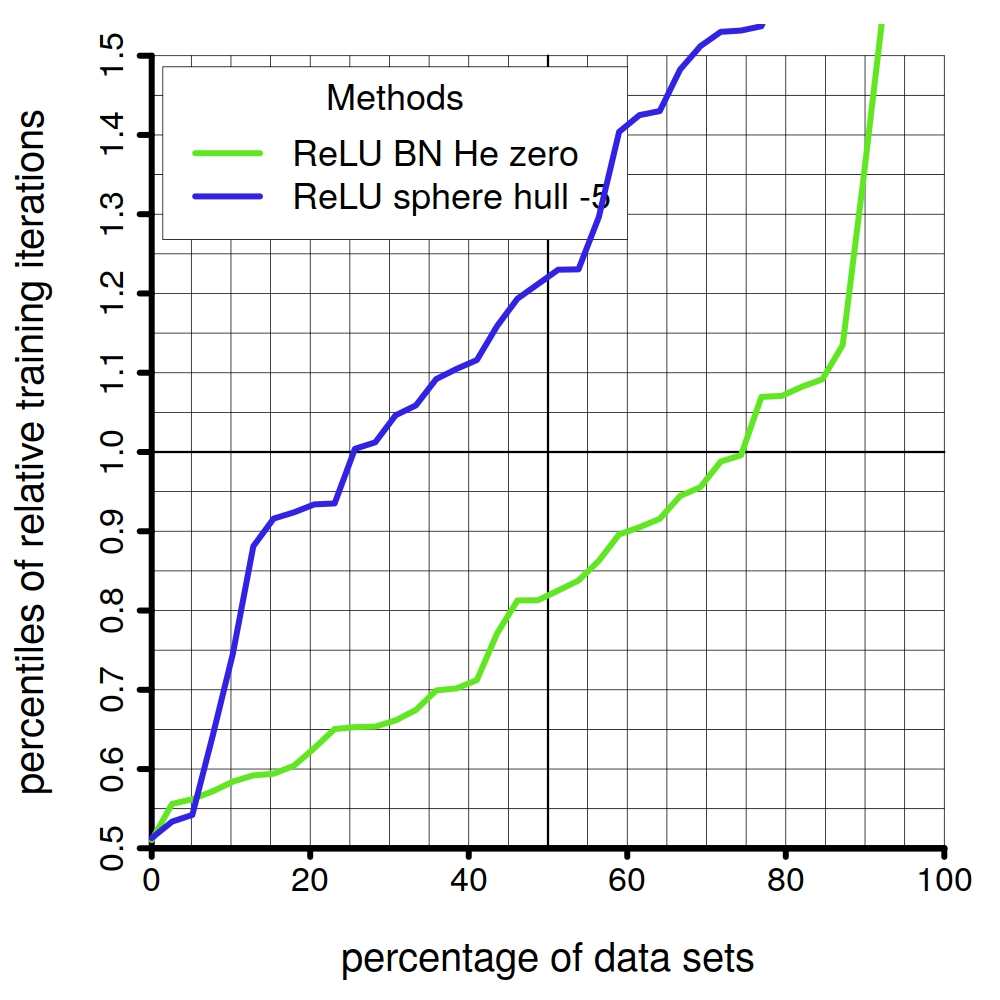}
\hspace*{-0.01\textwidth}
\includegraphics[width=0.32\textwidth]{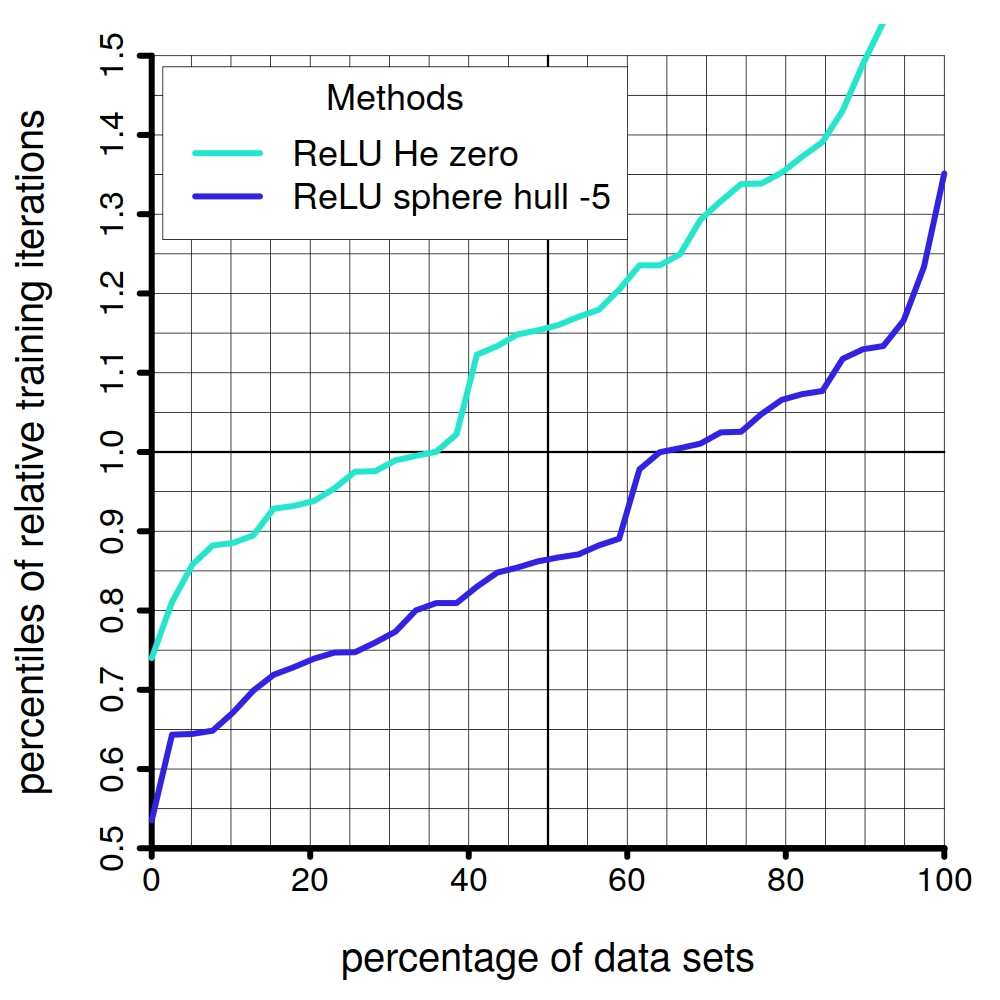}
\vspace*{-4ex}
\end{center}
\caption{Pairwise comparisons of old and new methods for regression with ReLU activation functions in terms of computational resources.
The first row displays the percentile curves of the ``observations''
$y_i = \att_i(\mbox{colored method})/\att_i(\mbox{other method})$ for  $i=1,\dots,40$,
where $\att_i$ denotes the average training time of the considered method on the $i$-th data set.
The second row displays the percentile curves of the ``observations''
$y_i = \ati_i(\mbox{colored method})/\ati_i(\mbox{other method})$ for  $i=1,\dots,40$,
where $\ati_i$ denotes the average number of training iterations of the considered method on the $i$-th data set.
Consequently, the second columns shows that
\scalingname{ReLU sphere hull -5} was faster than \scalingname{ReLU BN He zero} on more than $90\%$ 
of the data sets, despite the fact that \scalingname{ReLU BN He zero} required less training iterations
than \scalingname{ReLU sphere hull -5} on $75\%$ of the data sets.
}\label{figure:first-comp-relu-regress-train-resourc}
\end{figure}

\begin{figure}[t]
\begin{center}
\includegraphics[width=0.32\textwidth]{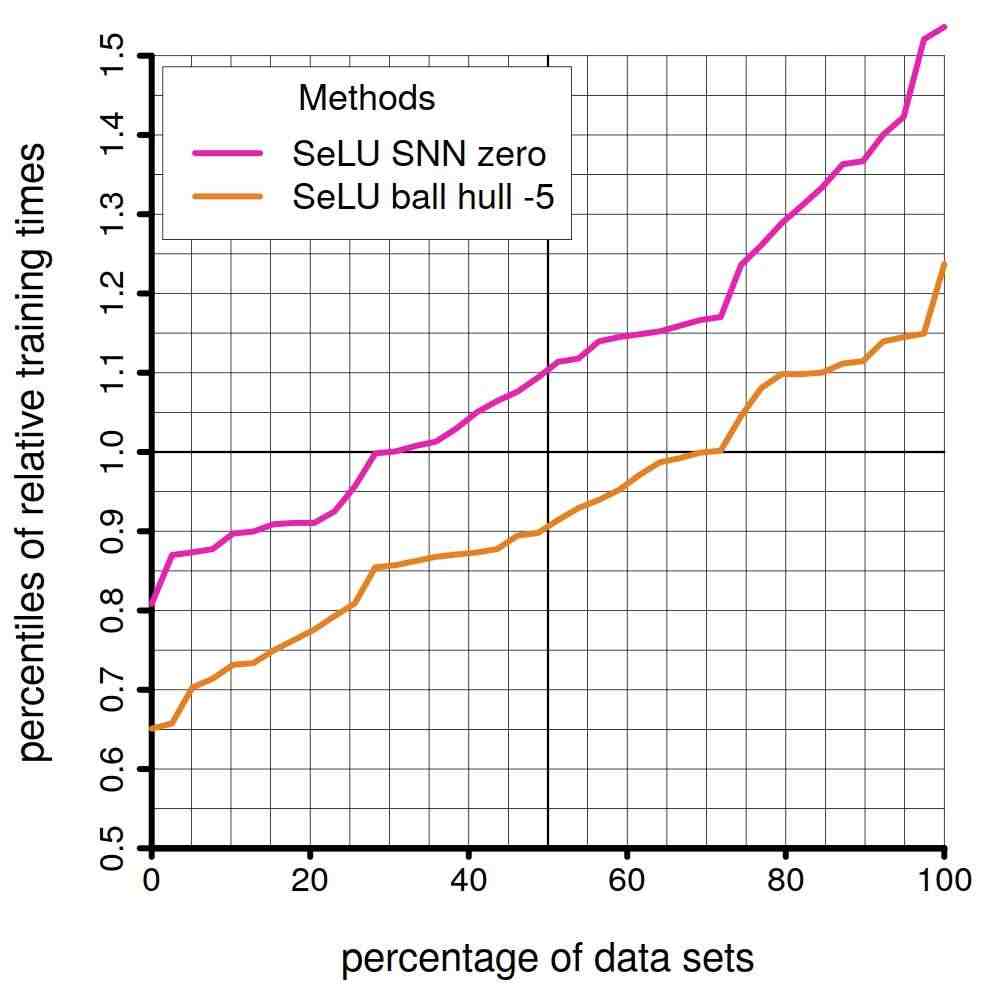}
\hspace*{-0.01\textwidth}
\includegraphics[width=0.32\textwidth]{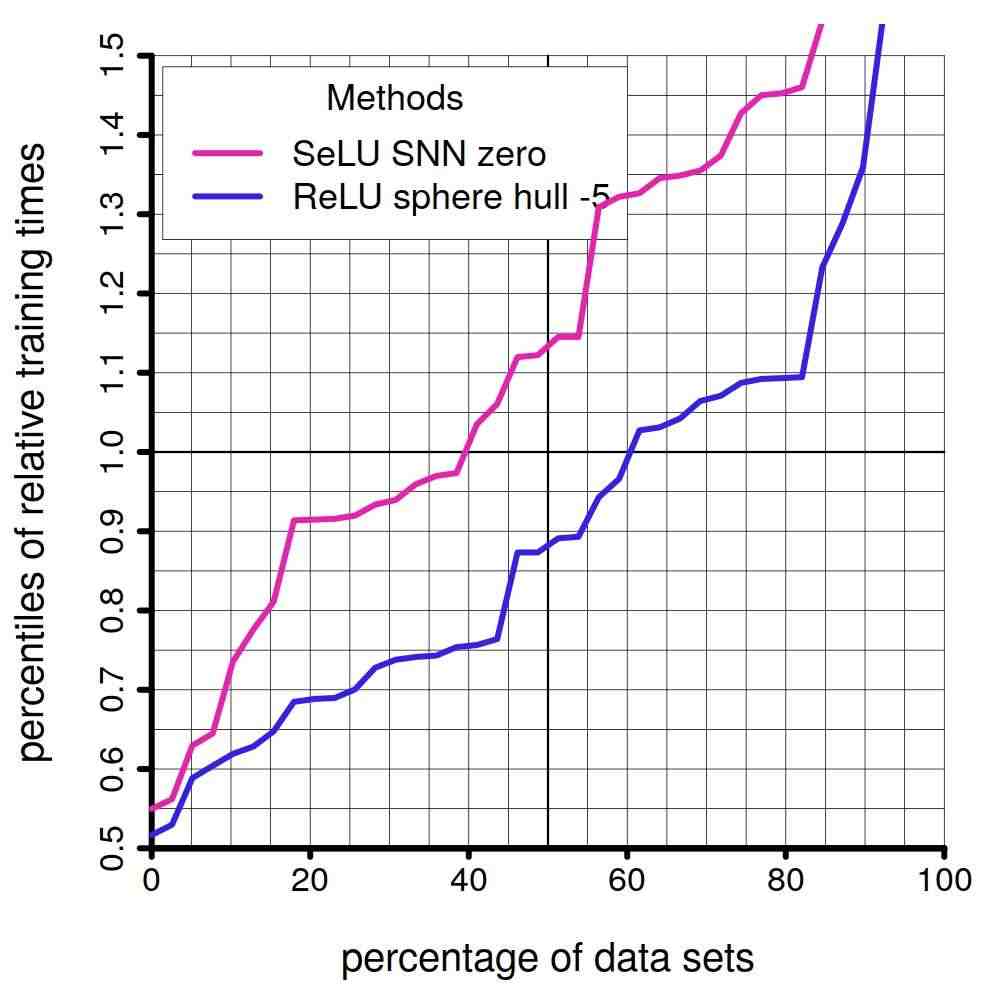}
\hspace*{-0.01\textwidth}
\includegraphics[width=0.32\textwidth]{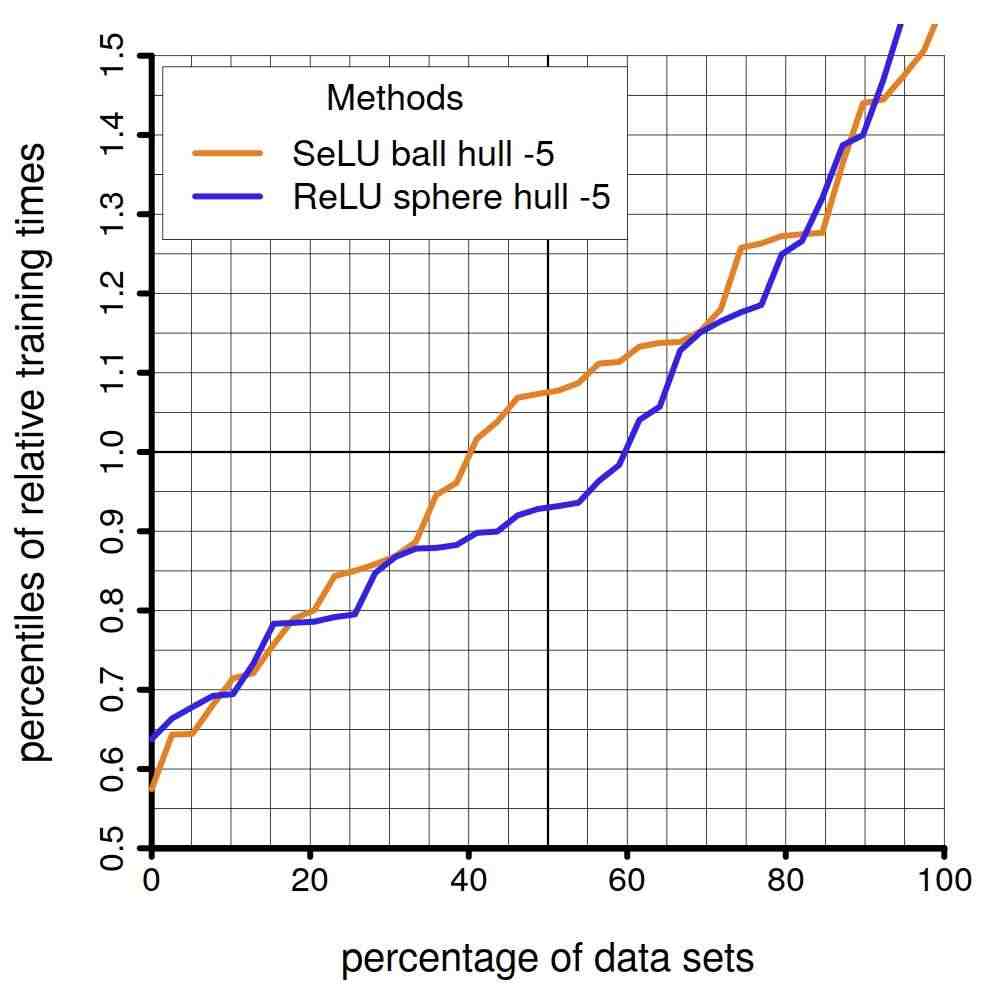}

\includegraphics[width=0.32\textwidth]{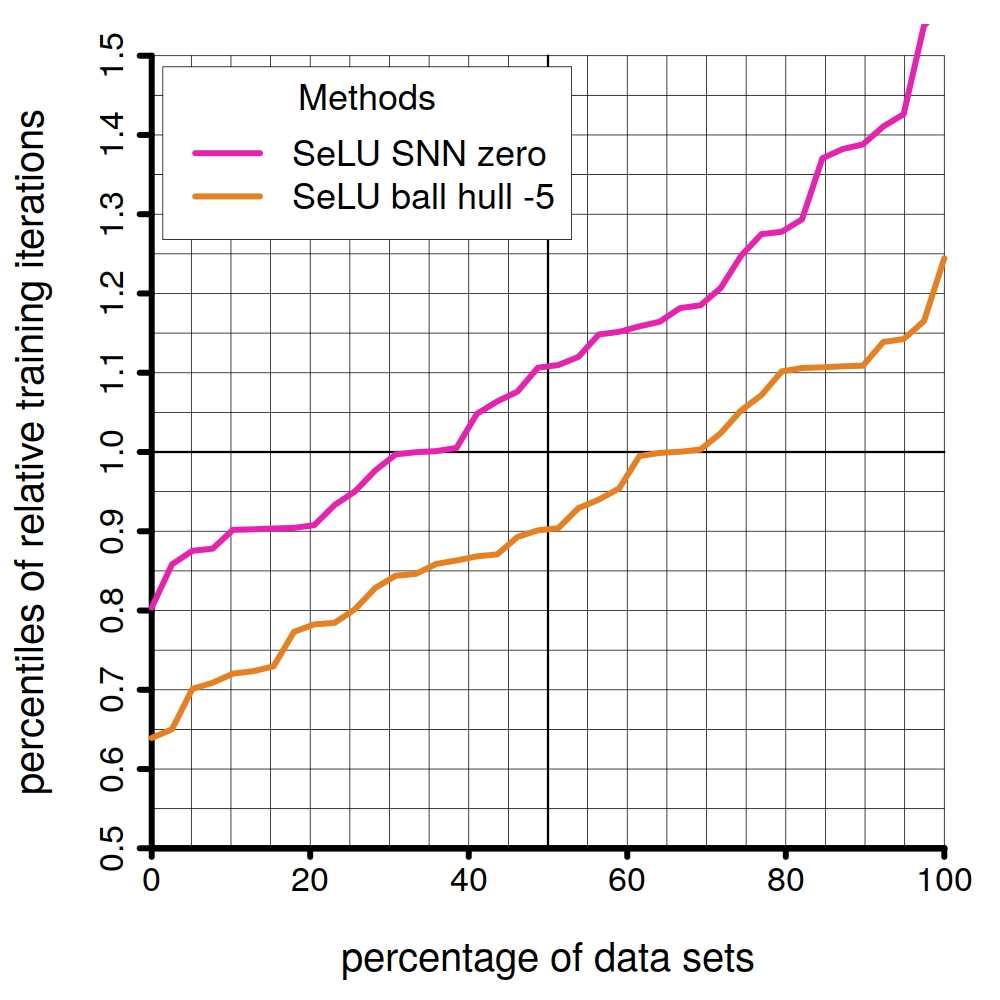}
\hspace*{-0.01\textwidth}
\includegraphics[width=0.32\textwidth]{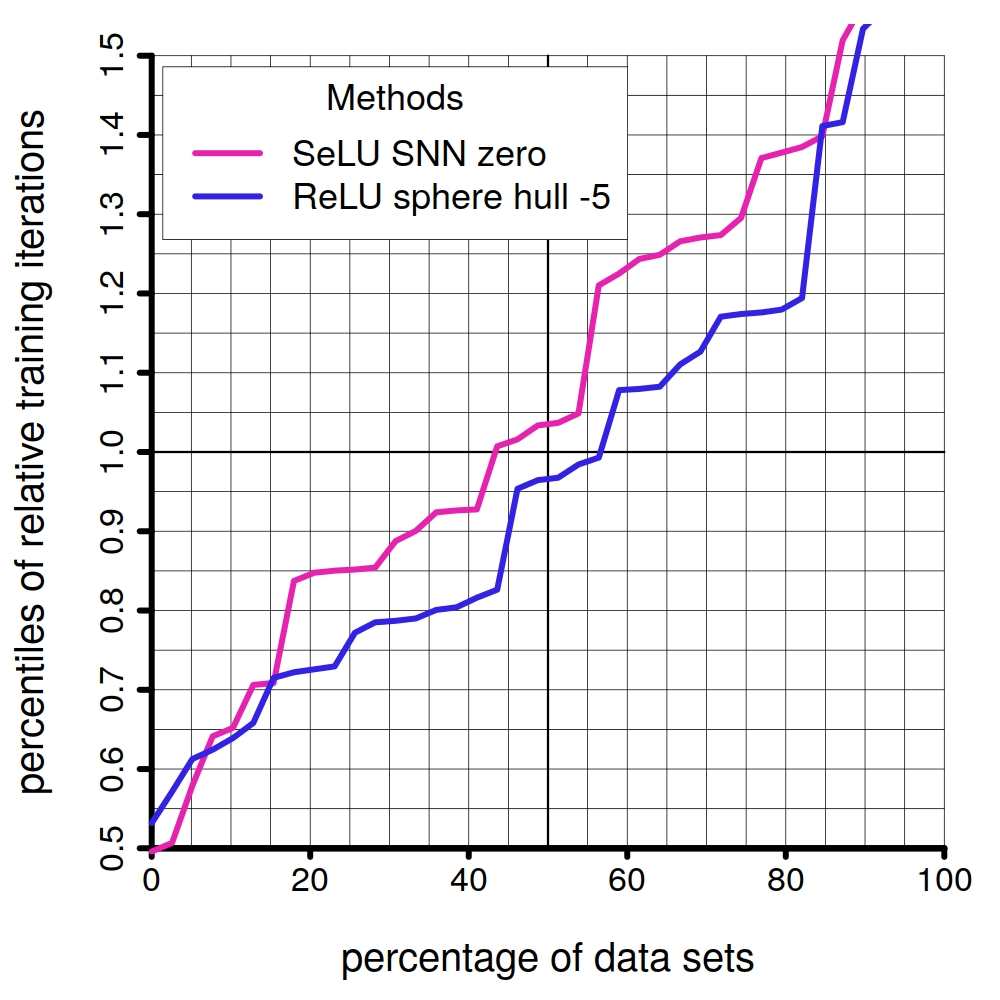}
\hspace*{-0.01\textwidth}
\includegraphics[width=0.32\textwidth]{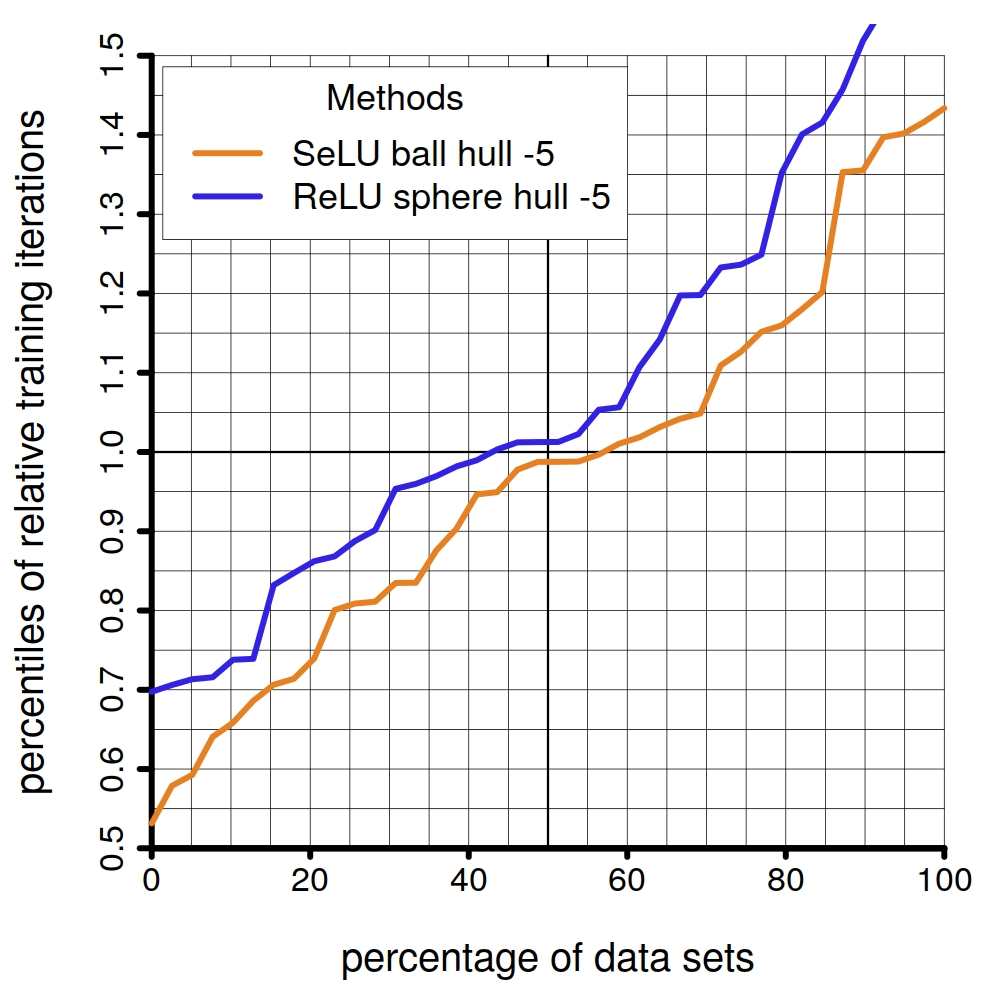}
\vspace*{-4ex}
\end{center}
\caption{Pairwise comparisons of old and new methods for regression with self-normalizing activation function
in terms of computational resources. The graphics, which have the same meaning as those in 
Figure \ref{figure:first-comp-relu-regress-train-resourc}, show that our new initialization strategy
\scalingname{SeLU ball hull -5} leads to substantially faster training of self-normalizing networks, but it is not 
as fast as 
\scalingname{ReLU sphere hull -5} despite the fact that it requires less iterations than 
\scalingname{ReLU sphere hull -5}.
}\label{figure:first-comp-selu-regress-train-resourc}
\end{figure}

For the classification data sets, Table \ref{logAllunlimpatfivecverrorAvA} shows that 
our new three initialization strategies \scalingname{ReLU He hull -5}, \scalingname{ReLU ball hull +5},
and \scalingname{SeLU ball hull -5}   achieve the second, first, and third rank, respectively. 
However, a closer look reveals that unlike in the regression case, the situation is a bit more diffuse.
For example, \scalingname{ReLU BN He zero}, which is ranked fourth, achieves the adjusted first rank on around 
$40\%$ of the data sets, whereas  the three new methods are only ranked first on around $20\%$, $25\%$, and $30\%$ 
of the data sets, respectively. To better understand the situation let us therefore consider Figures 
\ref{figure:first-comp-he},  \ref{figure:first-comp-selu}, and \ref{figure:first-comp-bn}.
For example, Figure \ref{figure:first-comp-he} shows that on more than $80\%$ of the data sets we have 
\begin{displaymath}
 0.97 \leq \rate(\scalingname{ReLU He hull -5}, \scalingname{ReLU ball hull +5}) \leq 1.03\, .
\end{displaymath}
Therefore, these two methods have a very similar performance on the vast majority of data sets.
In comparison, Figure \ref{figure:first-comp-bn} shows that we have
\begin{displaymath}
 0.97 \leq \rate(\scalingname{ReLU BN He zero}, \scalingname{ReLU He hull -5}) \leq 1.03\, 
\end{displaymath}
 and
\begin{displaymath}
 0.97 \leq \rate(\scalingname{ReLU BN He zero}, \scalingname{ReLU ball hull +5}) \leq 1.03\, .
\end{displaymath}
on around on around $40\%$ of the data sets, only. 
In this respect note that  on \datasetname{human-activity-smartphone}, \datasetname{mushroom}, and
\datasetname{smartphone-human-activity-postural},
that is on $5\%$ of the data sets,
almost all 
the methods  achieved   zero average test errors,
while on \datasetname{insurance-benchmark}, on the data sets \datasetname{polish-companies-bankruptcy-1year} to 
\datasetname{polish-companies-bankruptcy-5year}, and on \datasetname{seismic-bumps}, 
\datasetname{thyroid-all-hypo}, and \datasetname{thyroid-dis}, that is on around $15\%$ of all data sets,
all tested methods, as well as SVMs tested as a sanity check, were not able to outperform the naive classifier 
that simply predicts all new labels by the majority of the labels found in the training set, see Tables
\ref{logAllunlimpatfivecverrorAvA} and \ref{logdatacharacteristics}. In other words, around $20\%$ of the considered 
data sets were either particularly simple or hard to learn from and on these data sets 
one can  expect most classification methods to perform very similarly.
To sum up this discussion,
 we conclude that 
\scalingname{ReLU He hull -5} and \scalingname{ReLU ball hull +5} win or loose in most cases together, whereas
\scalingname{ReLU BN He zero} exhibits strengths and weaknesses that are rather different from the aforementioned 
new initialization strategies. Consequently, if one is willing to consider two initialization strategies 
during the selection phase, it seems to be more beneficial to consider one of the new initialization strategies plus
\scalingname{ReLU BN He zero} instead of considering the two new initialization strategies. 
Finally, Figure \ref{figure:first-comp-selu} shows that for self-normalizing networks 
the new initialization strategy \scalingname{SeLU ball hull -5} substantially outperforms the standard
initialization strategy \scalingname{SeLU SNN zero}. Figure \ref{figure:first-comp-selu}
further shows that \scalingname{SeLU ball hull -5}  slightly outperforms both 
\scalingname{ReLU He hull -5} and \scalingname{ReLU ball hull +5}.
In this sense, \scalingname{SeLU ball hull -5} can be viewed as the best performing method, while
in terms of  raw and adjusted ranking it is only placed third. 
\emph{In any case, 
whether it is in terms of ranking  ranking or of pairwise comparisons 
with the help of percentiles of $\rate_i(\cdot, \cdot)$, all three new initialization strategies 
clearly outperform the standard initialization strategies.}

\begin{figure}[t]
\begin{center}
\includegraphics[width=0.32\textwidth]{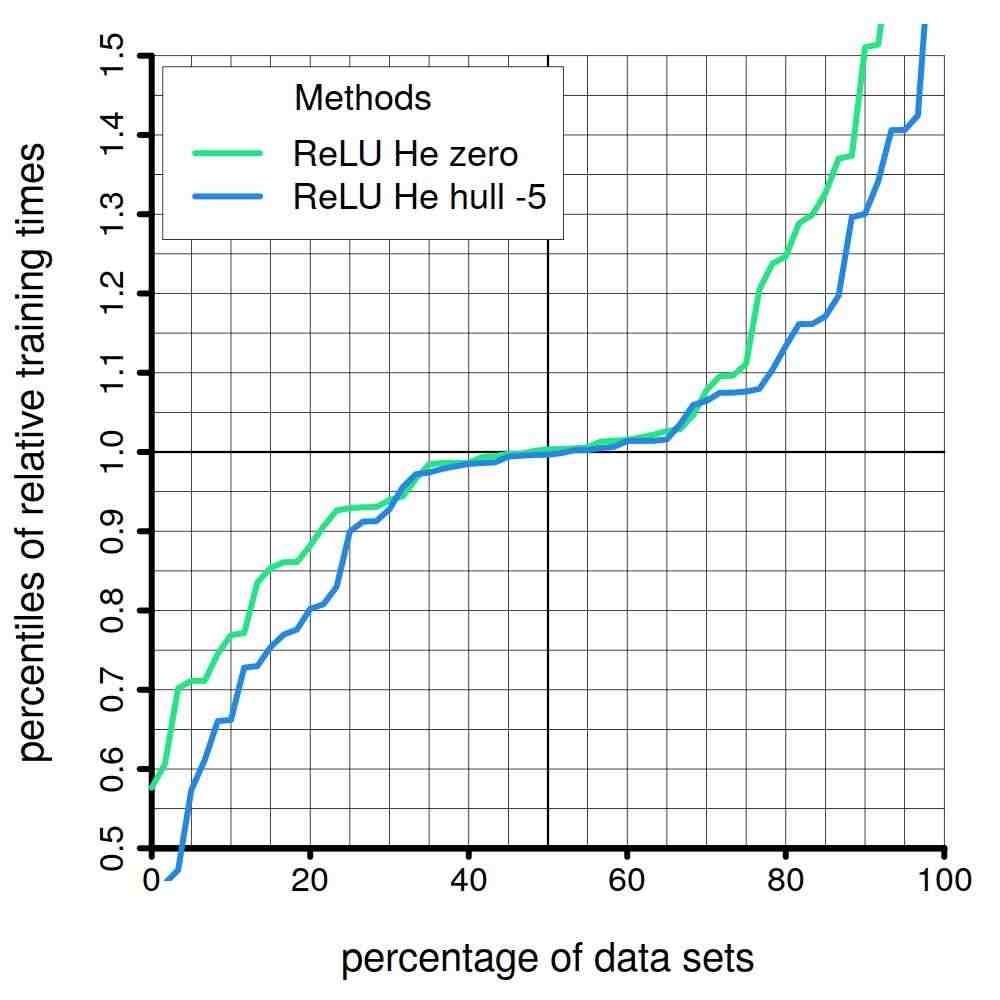}
\hspace*{-0.01\textwidth}
\includegraphics[width=0.32\textwidth]{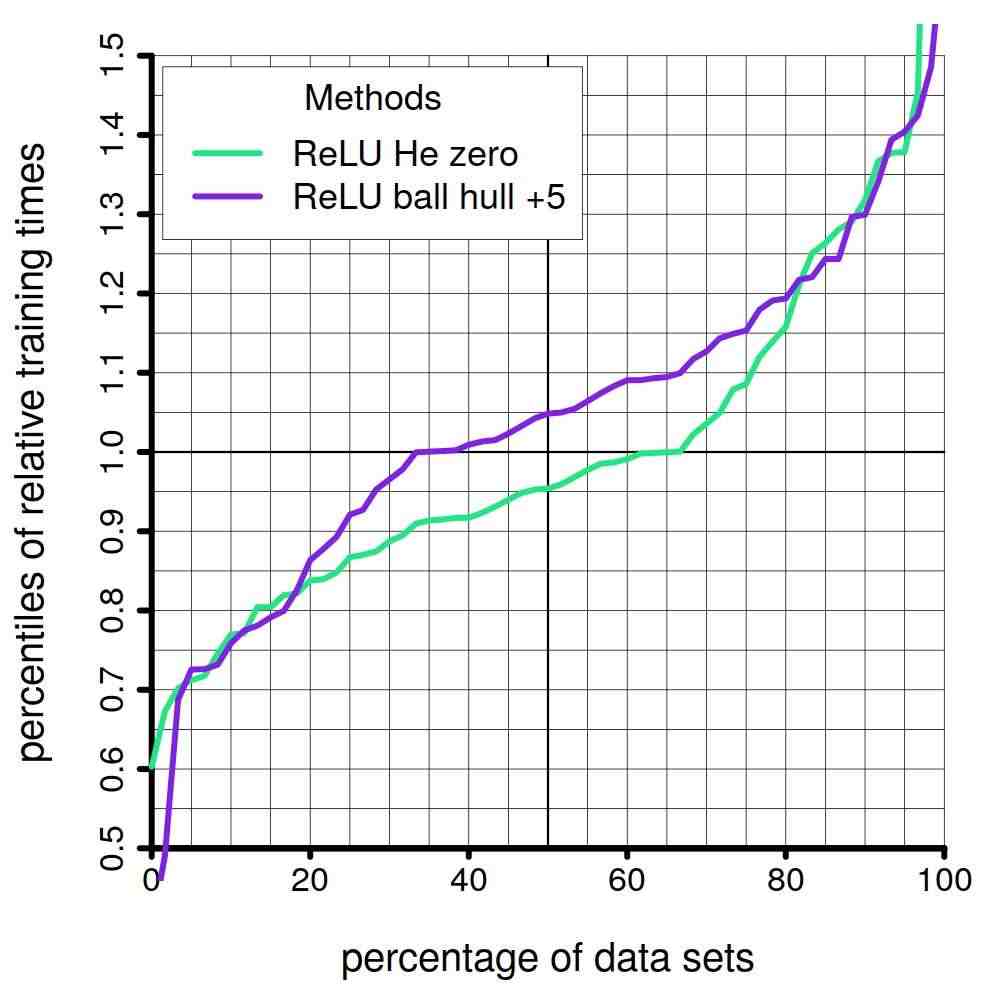}
\hspace*{-0.01\textwidth}
\includegraphics[width=0.32\textwidth]{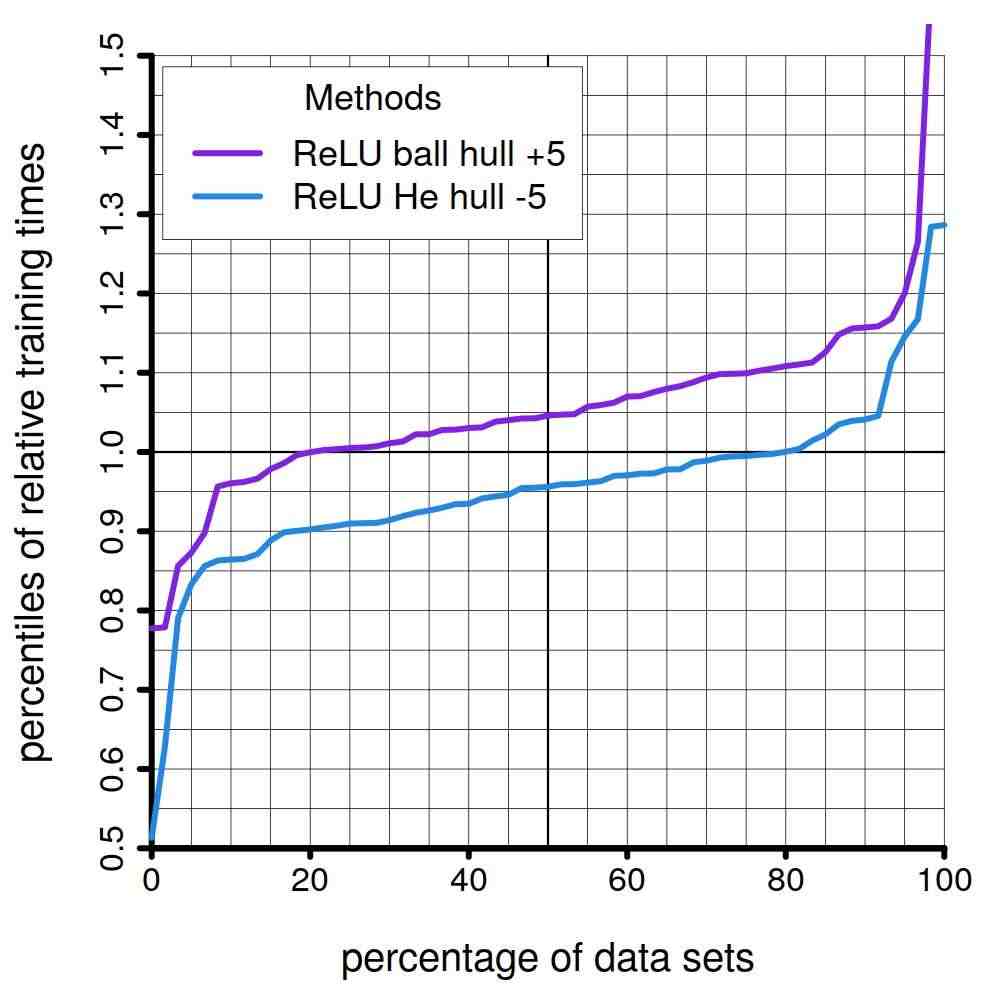}

\includegraphics[width=0.32\textwidth]{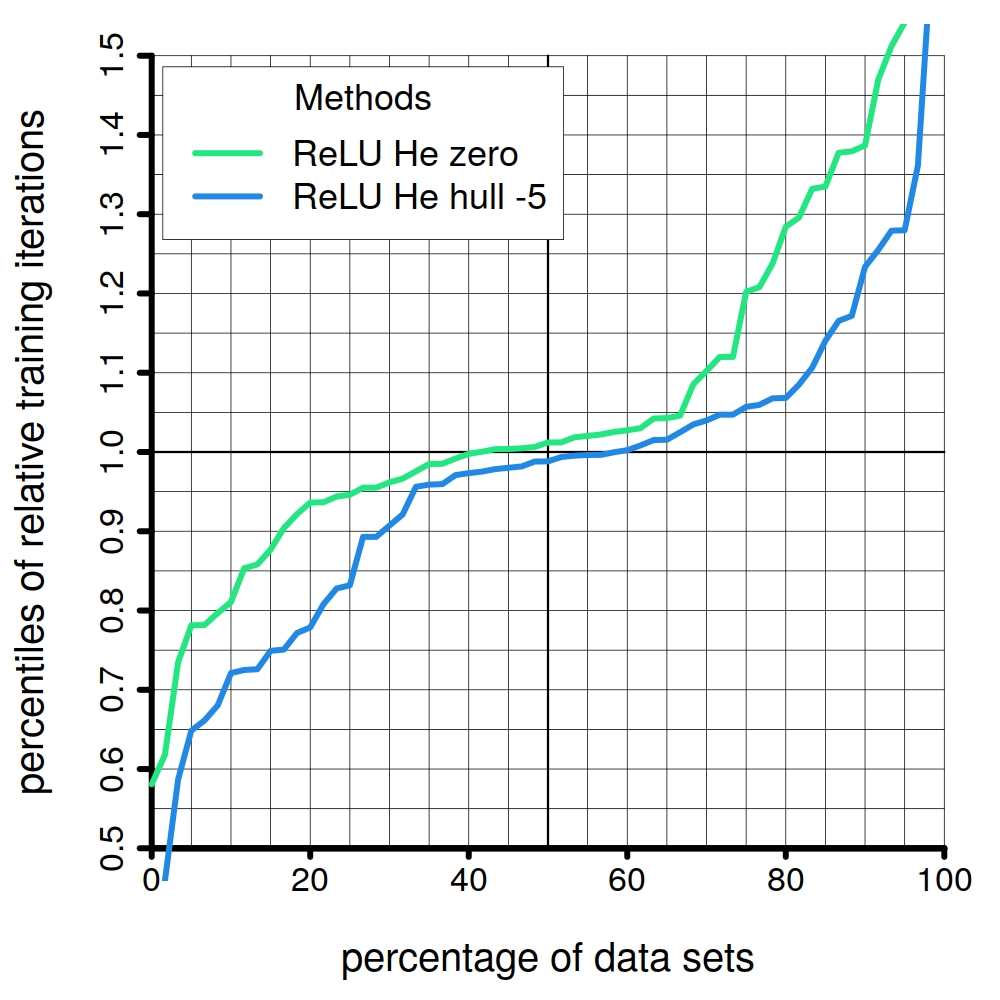}
\hspace*{-0.01\textwidth}
\includegraphics[width=0.32\textwidth]{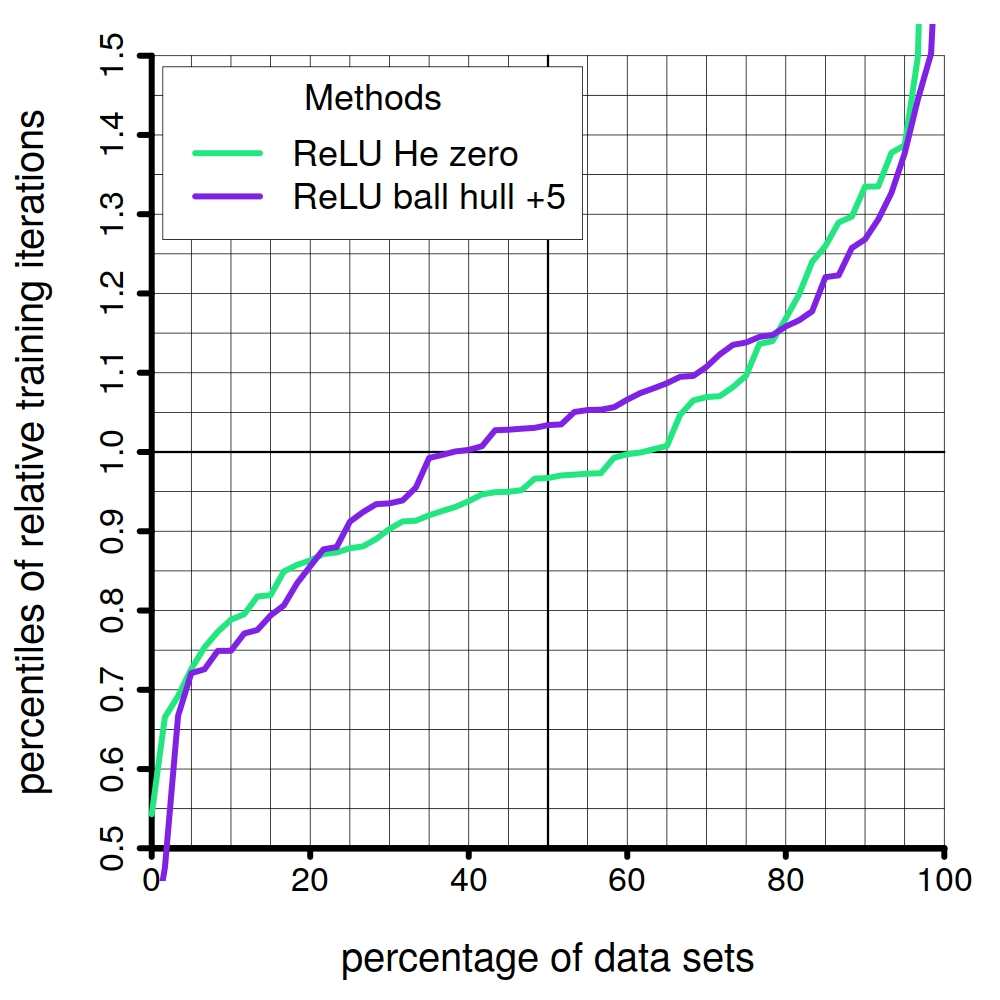}
\hspace*{-0.01\textwidth}
\includegraphics[width=0.32\textwidth]{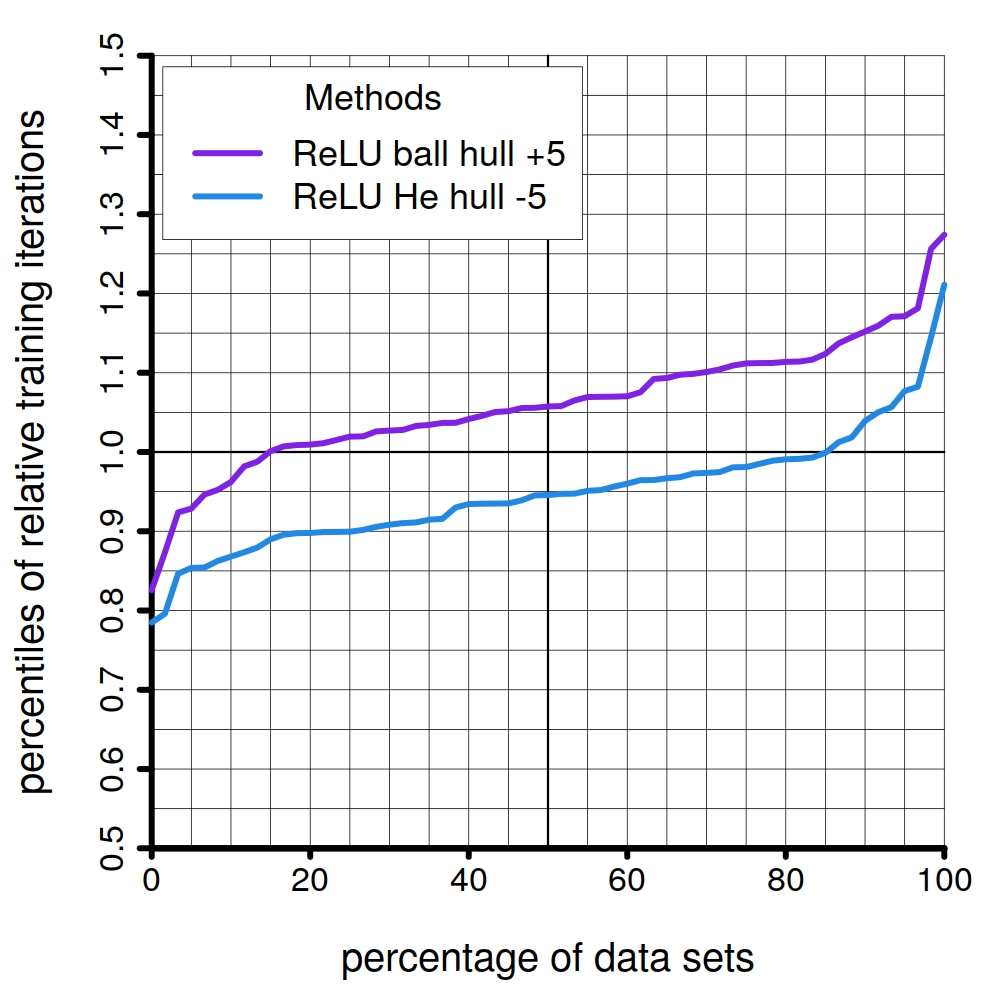}
\vspace*{-4ex}
\end{center}
\caption{Pairwise comparisons of old and new methods without batch normalization for binary classification in terms of computational resources. The graphics, which have the same meaning as those in 
Figure \ref{figure:first-comp-relu-regress-train-resourc}, show that our new initialization strategy
\scalingname{ReLU He hull -5} leads to faster training compared to 
standard \scalingname{ReLU He zero} and the new
\scalingname{ReLU ball hull +5}.
}\label{figure:first-comp-he-train-resourc}
\end{figure}

\begin{figure}[t]
\begin{center}
\includegraphics[width=0.32\textwidth]{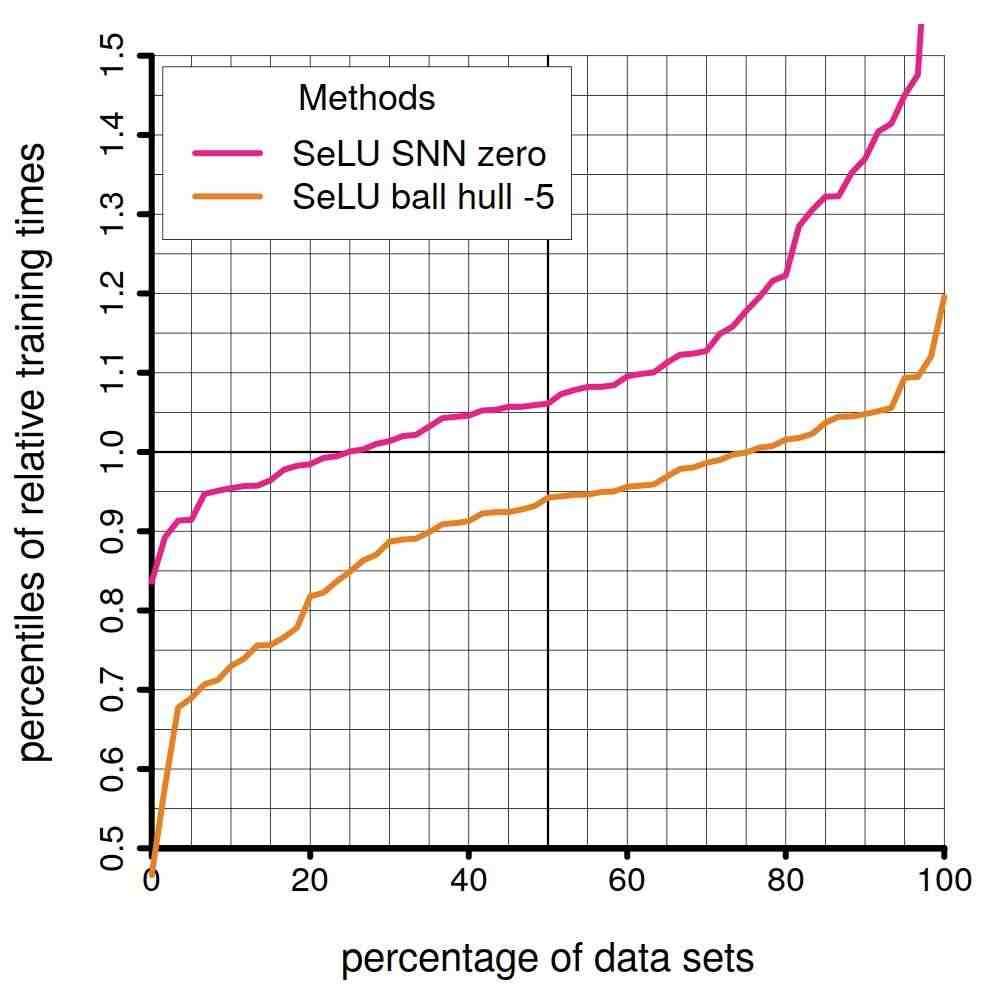}
\hspace*{-0.01\textwidth}
\includegraphics[width=0.32\textwidth]{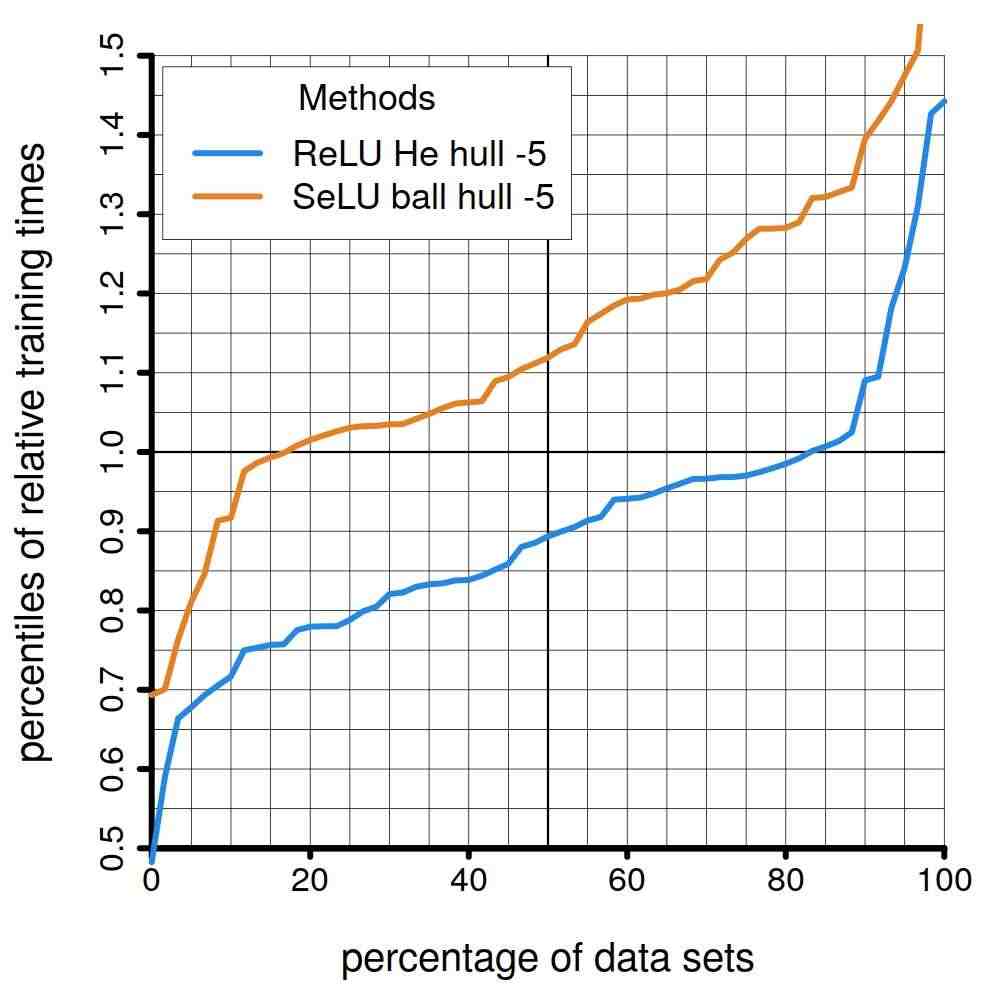}
\hspace*{-0.01\textwidth}
\includegraphics[width=0.32\textwidth]{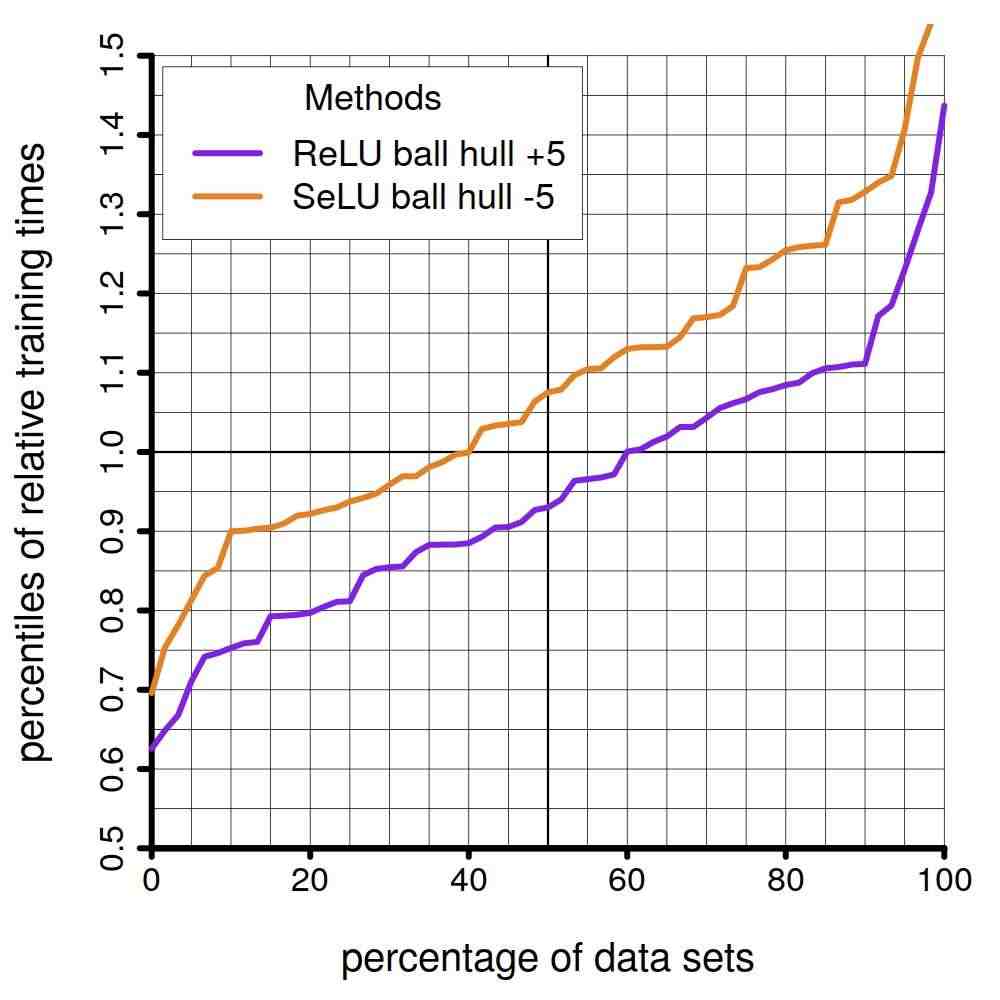}

\includegraphics[width=0.32\textwidth]{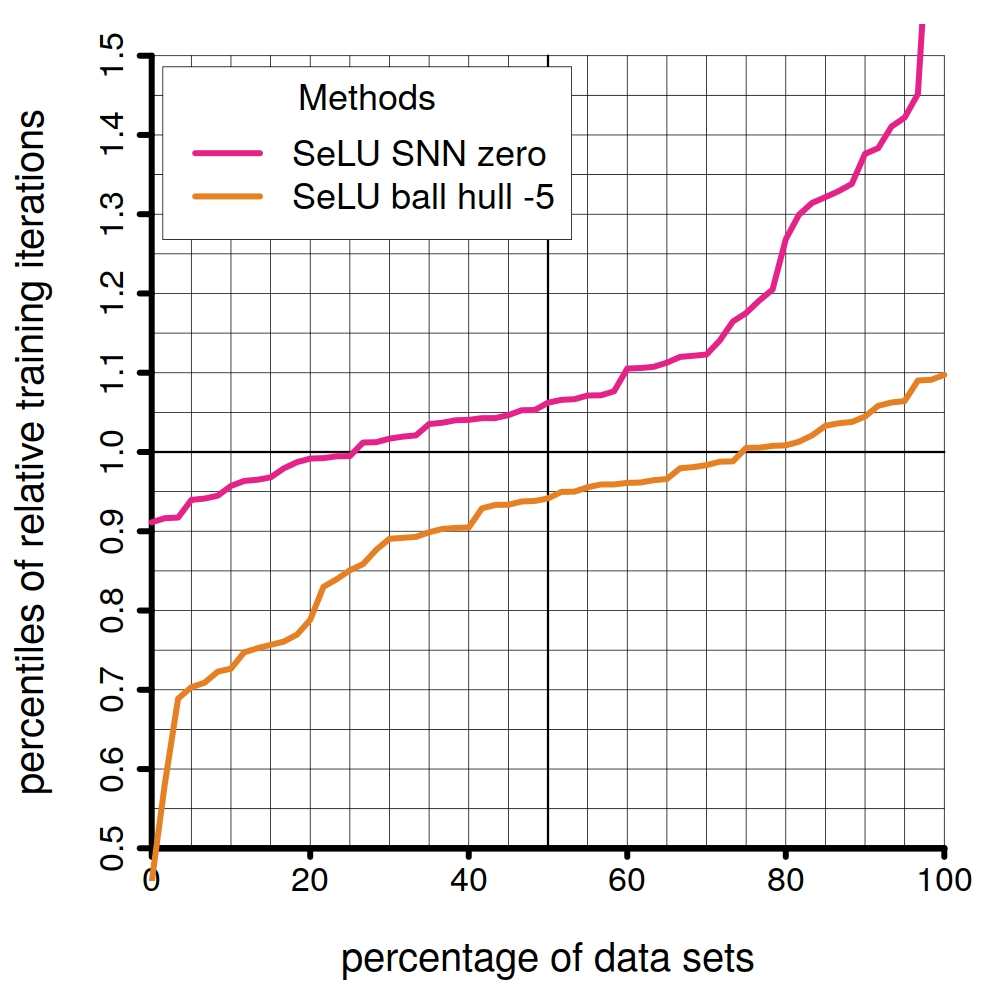}
\hspace*{-0.01\textwidth}
\includegraphics[width=0.32\textwidth]{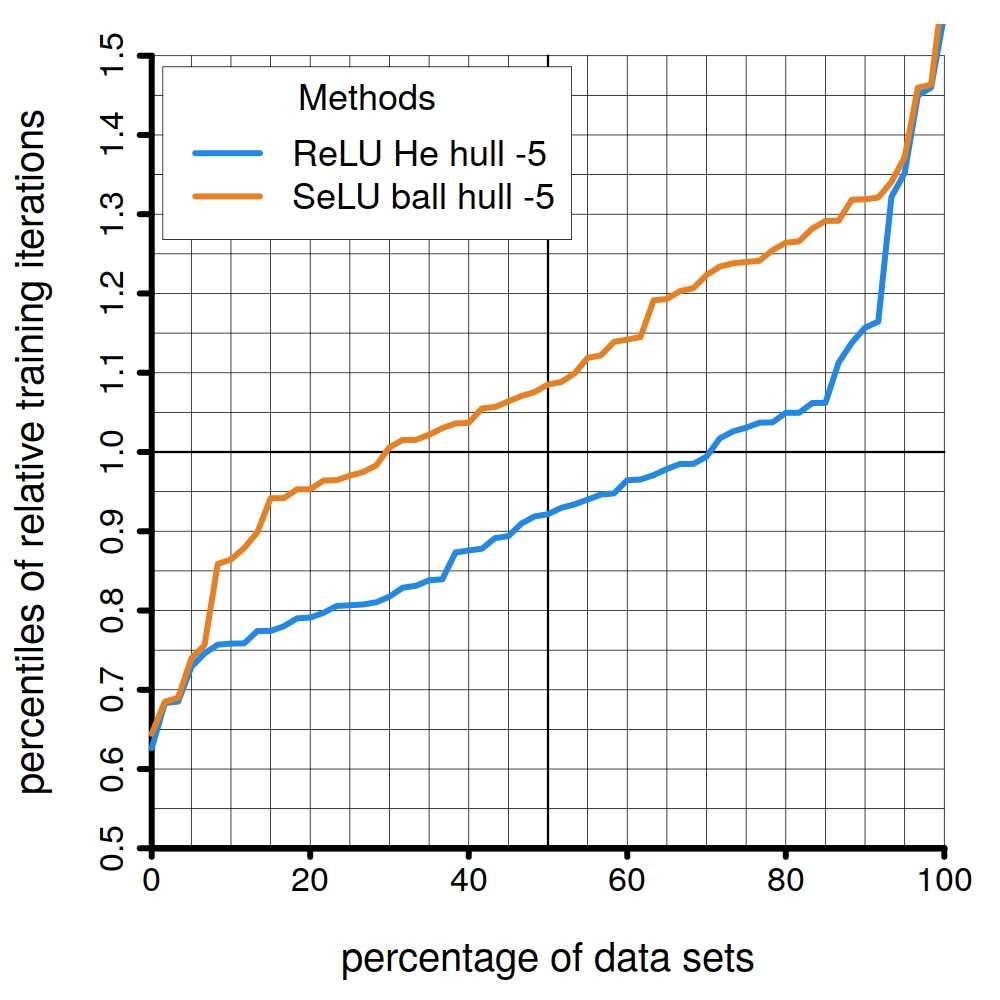}
\hspace*{-0.01\textwidth}
\includegraphics[width=0.32\textwidth]{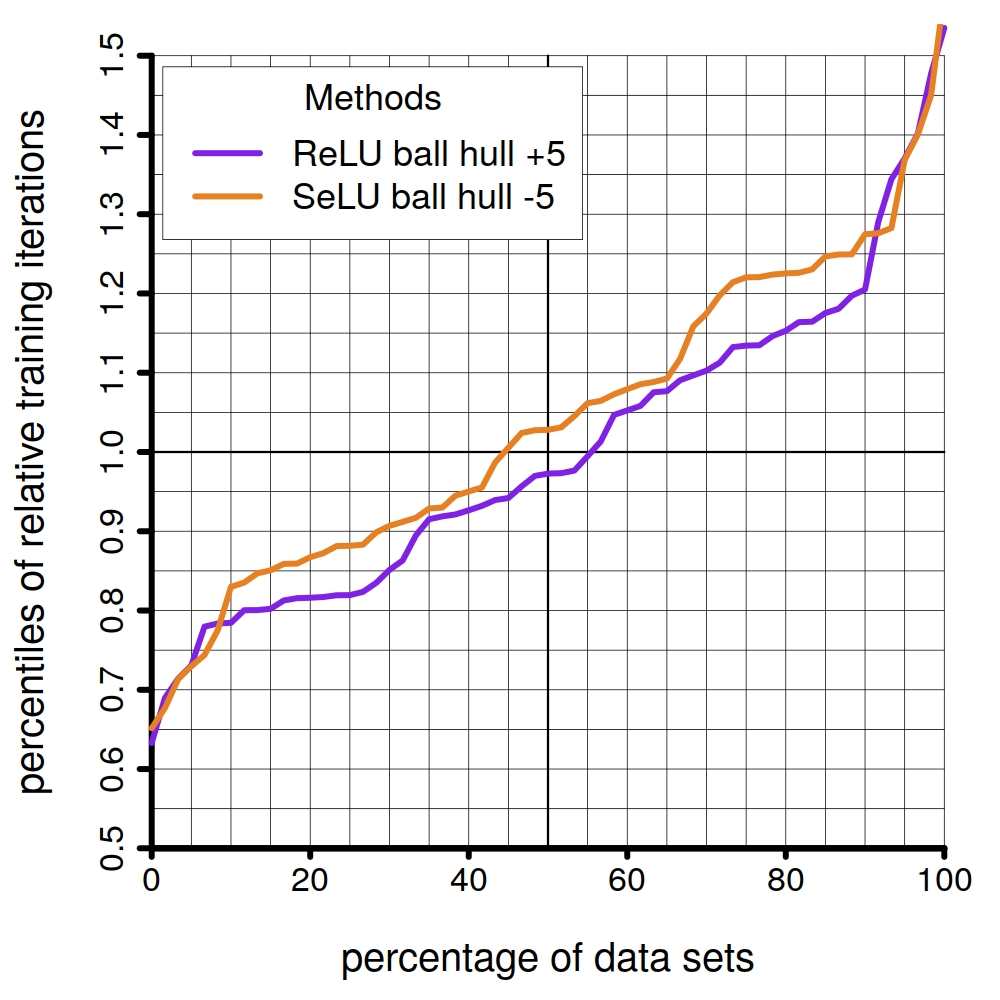}
\vspace*{-4ex}
\end{center}
\caption{Pairwise comparisons for methods with self-normalizing activation function for binary classification
in terms of computational resources. The graphics, which have the same meaning as those in 
Figure \ref{figure:first-comp-relu-regress-train-resourc}, show that our new initialization strategy
\scalingname{SeLU ball hull -5} leads to substantially faster training compared to 
standard \scalingname{SeLU SNN zero}. However, \scalingname{SeLU SNN zero} is slower than 
\scalingname{ReLU He hull -5} and 
\scalingname{ReLU ball hull +5}.
}\label{figure:first-comp-selu-train_resourc}
\end{figure}

\begin{figure}[t]
\begin{center}
\includegraphics[width=0.32\textwidth]{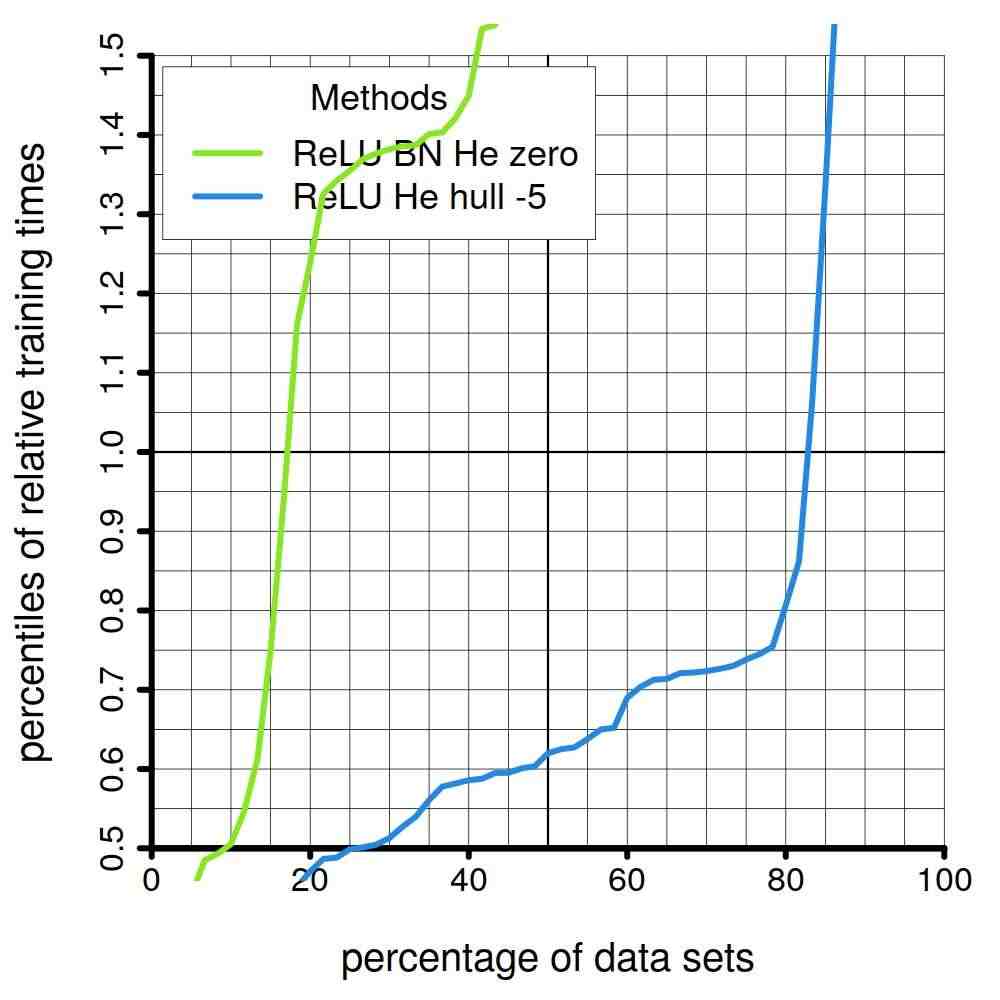}
\hspace*{-0.01\textwidth}
\includegraphics[width=0.32\textwidth]{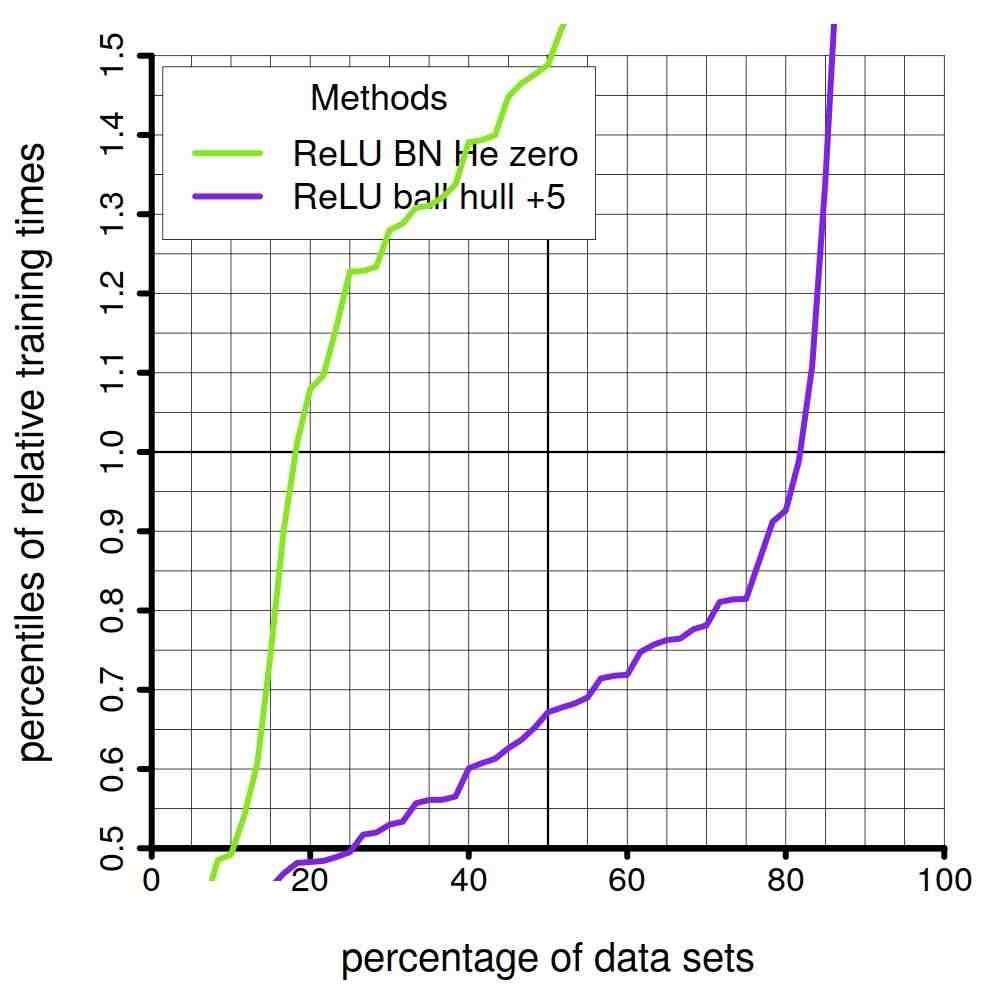}
\hspace*{-0.01\textwidth}
\includegraphics[width=0.32\textwidth]{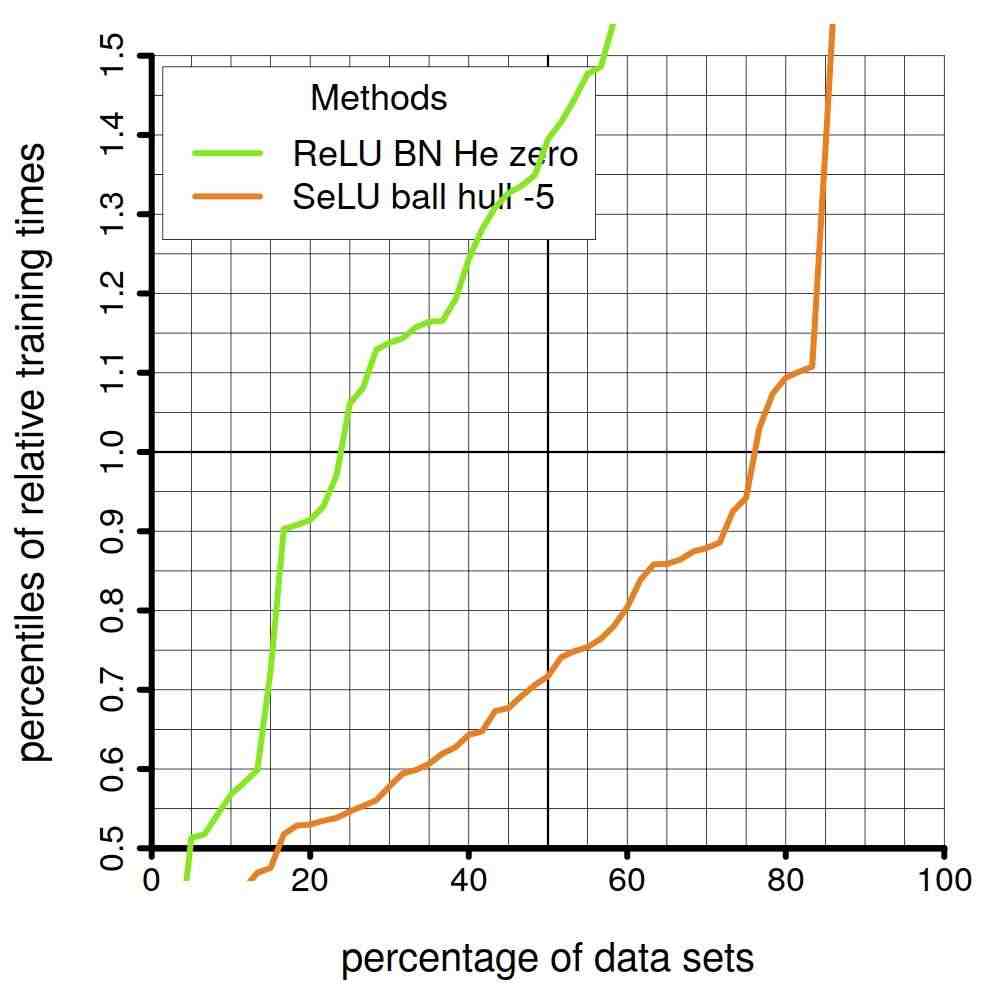}

\includegraphics[width=0.32\textwidth]{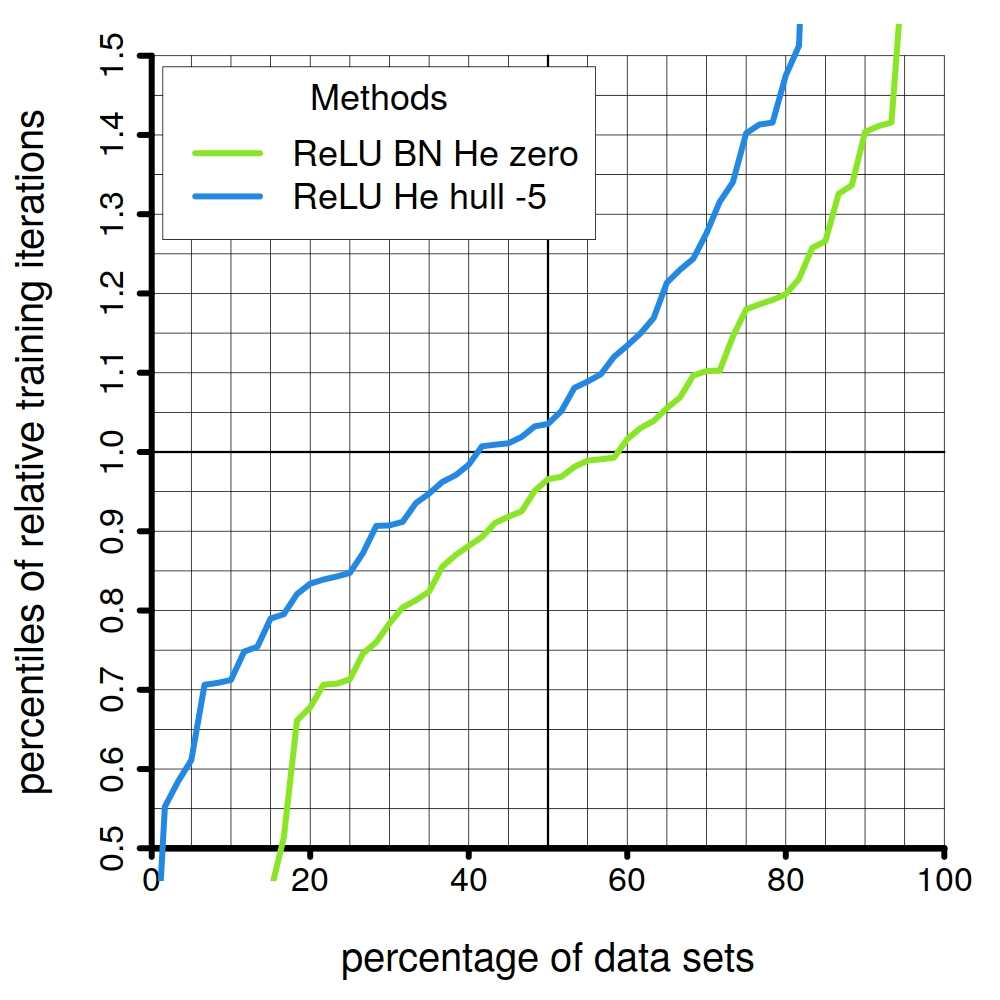}
\hspace*{-0.01\textwidth}
\includegraphics[width=0.32\textwidth]{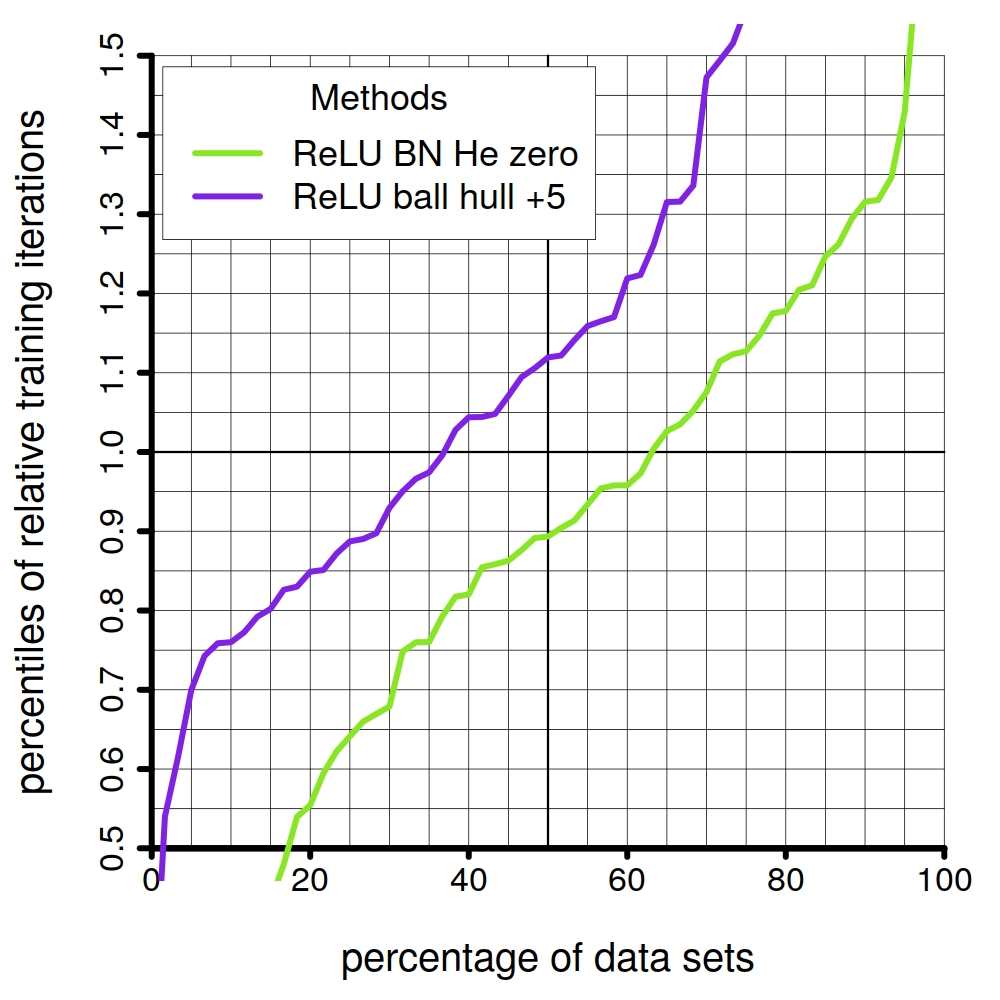}
\hspace*{-0.01\textwidth}
\includegraphics[width=0.32\textwidth]{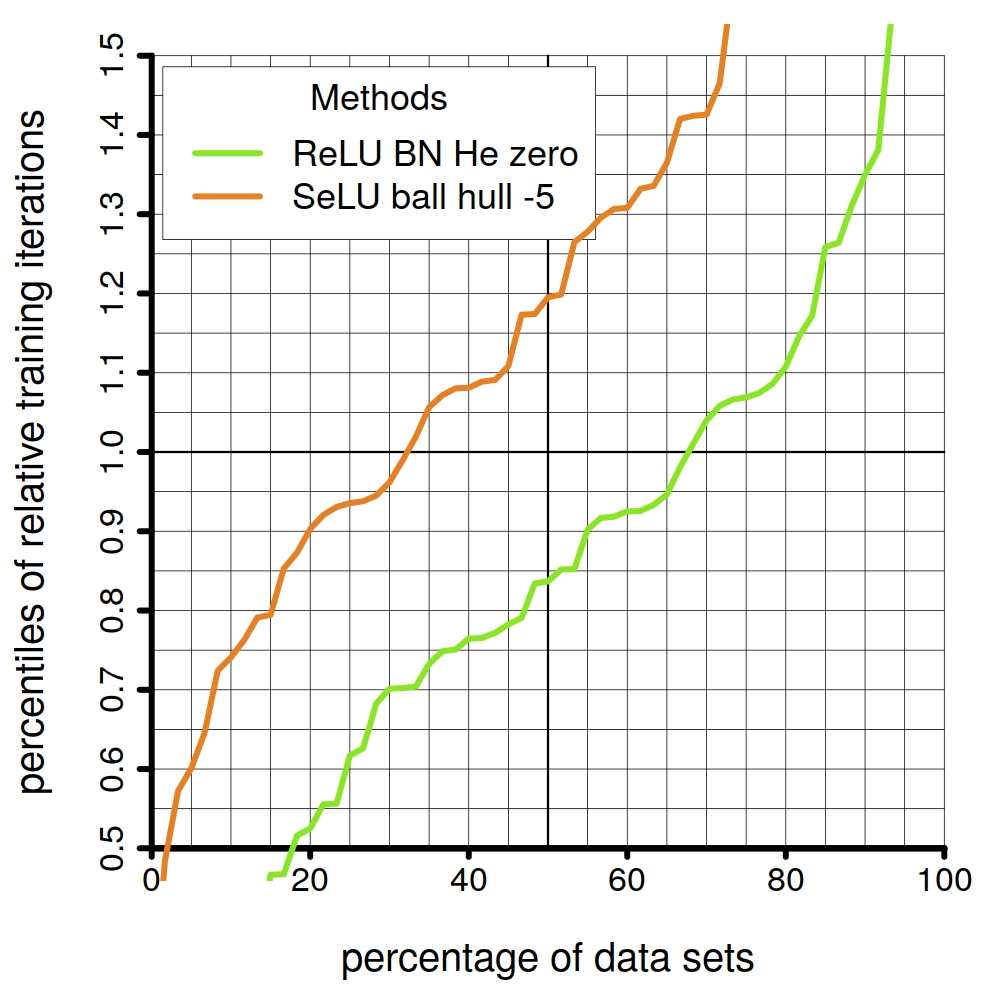}
\vspace*{-4ex}
\end{center}
\caption{Pairwise comparisons against batch normalization with standard initialization for binary classification
in terms of computational resources. The graphics, which have the same meaning as those in 
Figure \ref{figure:first-comp-relu-regress-train-resourc}, show that our new initialization strategies
require substantially less training time than \scalingname{ReLU BN He zero} does despite the fact that 
the latter needs less training iterations.
}\label{figure:first-comp-bn-train-resourc}
\end{figure}

\emph{ii).} Let us now consider the computational resources the different methods required.
Again, we begin with the regression case. Here, Figure \ref{figure:first-comp-relu-regress-train-resourc}
shows that \scalingname{ReLU sphere hull -5} is e.g.~on $90\%$ of the data sets faster than 
\scalingname{ReLU BN He zero}, and on $70\%$ of the data sets it requires less than $80\%$ of the 
training time \scalingname{ReLU BN He zero} used. Moreover, the new 
\scalingname{ReLU sphere hull -5} is also faster than \scalingname{ReLU   He zero} on around 
$75\%$ of the data sets, and on $40\%$ of the data sets  
 it requires less than $80\%$ of the 
training time \scalingname{ReLU He zero} uses. Similar, yet less pronounced, observations can be made in terms of 
training iterations the latter two methods run, which is not surprising, since the training time
per epoch should be equal for both methods. In contrast, 
\scalingname{ReLU BN He zero} requires significantly less iterations than the latter two methods,
and this indicates that the training time per iteration needs to be substantially longer 
for \scalingname{ReLU BN He zero} compared to e.g.~\scalingname{ReLU sphere hull -5}.
This is, however, not overly surprising as batch normalization adds quite a few extra computations to every step
of stochastic gradient descent.
When comparing to self-normalizing networks, Figure \ref{figure:first-comp-selu-regress-train-resourc}
shows that our new \scalingname{SeLU ball hull -5} is considerably faster  than the 
standard initialization
in terms of both time and iterations. However, only on $40\%$ of the data sets \scalingname{SeLU ball hull -5} is 
faster than \scalingname{ReLU sphere hull -5}, despite the fact that it requires less iterations on $>55\%$
of the data sets. Again, this is not overly surprising as self-normalizing networks also
add computations to each iteration of gradient descent.
\emph{In summary, \scalingname{ReLU sphere hull -5} is not only by far the best method in terms of test errors,
but it is also the most efficient method in terms of training time. In the same sense, \scalingname{SeLU ball hull -5}
outperforms the standard initialization \scalingname{SeLU SNN zero} for self-normalizing networks.}

Let us now have a look on the results for binary classification. Here, 
Figure \ref{figure:first-comp-he-train-resourc} shows that the new 
\scalingname{ReLU He hull -5} is slightly faster than both the standard
\scalingname{ReLU He zero} and our new \scalingname{ReLU   ball hull +5}, and 
not surprisingly
this 
behavior can also be found in terms of training iterations.
When combining these observations with Figure \ref{figure:first-comp-he} we thus conclude 
that \emph{\scalingname{ReLU He hull -5} outperforms both \scalingname{ReLU He zero} and 
\scalingname{ReLU   ball hull +5} in terms of test errors and required computational resources.}
Moreover, when considering self-normalizing networks, Figure \ref{figure:first-comp-selu-train_resourc}
shows that, as in the regression case, our new initialization \scalingname{SeLU ball hull -5}
leads to substantially faster training than the standard \scalingname{SeLU SNN zero}. By combining 
this with Figure \ref{figure:first-comp-selu} we thus conclude that 
\emph{our  new \scalingname{SeLU ball hull -5} outperforms the standard 
\scalingname{SeLU SNN zero} in terms of both tests errors and computational requirements.}
In contrast, when comparing \scalingname{SeLU ball hull -5} with 
\scalingname{ReLU He hull -5}  and \scalingname{ReLU   ball hull +5}, we see that 
 \scalingname{SeLU ball hull -5} requires more computational resources than the latter
 two methods, and therefore, the slight advantage of  \scalingname{SeLU ball hull -5} 
 in terms of tests errors reported in Figure \ref{figure:first-comp-selu}  comes with a price tag.
In any case, Figure \ref{figure:first-comp-bn-train-resourc} shows that all 
three new methods \scalingname{ReLU He hull -5},  \scalingname{ReLU   ball hull +5}, and
 \scalingname{SeLU ball hull -5}  are also considerably faster than the standard 
 \scalingname{ReLU BN He zero}. 
 \emph{In summary, all three of our  new initialization strategies outperform the standard methods in terms of 
 both test errors and computational requirements.}

\begin{figure}[t]
\begin{center}
\includegraphics[width=0.32\textwidth]{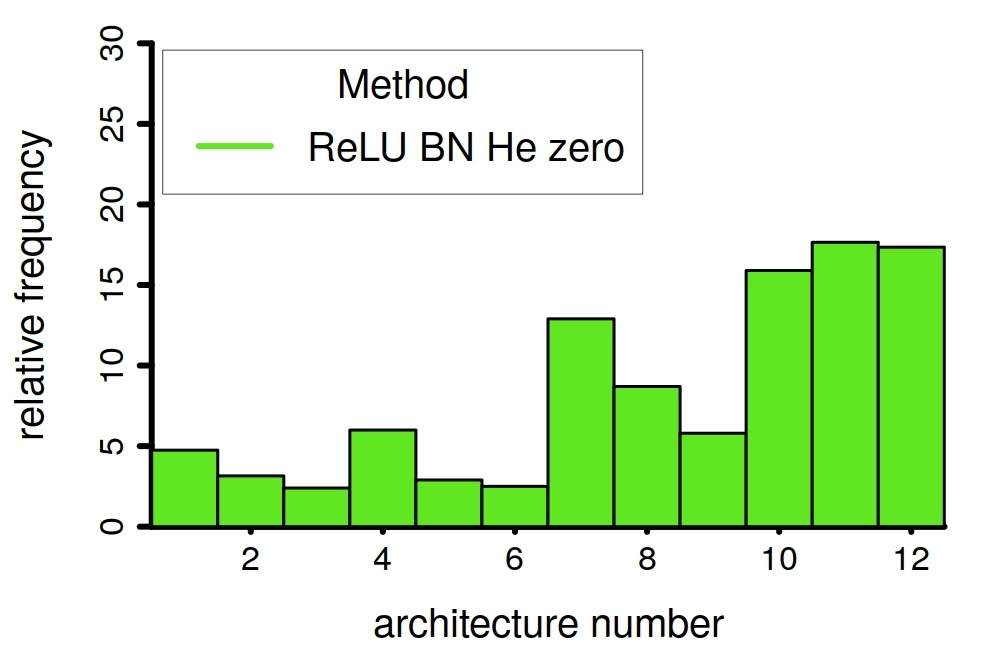}
\hspace*{-0.01\textwidth}
\includegraphics[width=0.32\textwidth]{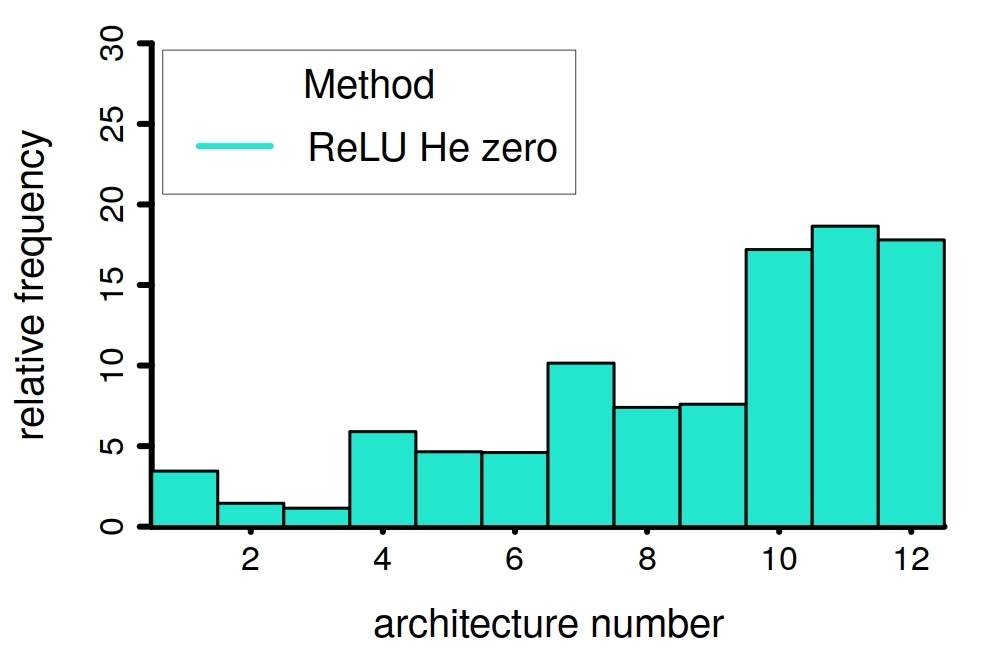}
\hspace*{-0.01\textwidth}
\includegraphics[width=0.32\textwidth]{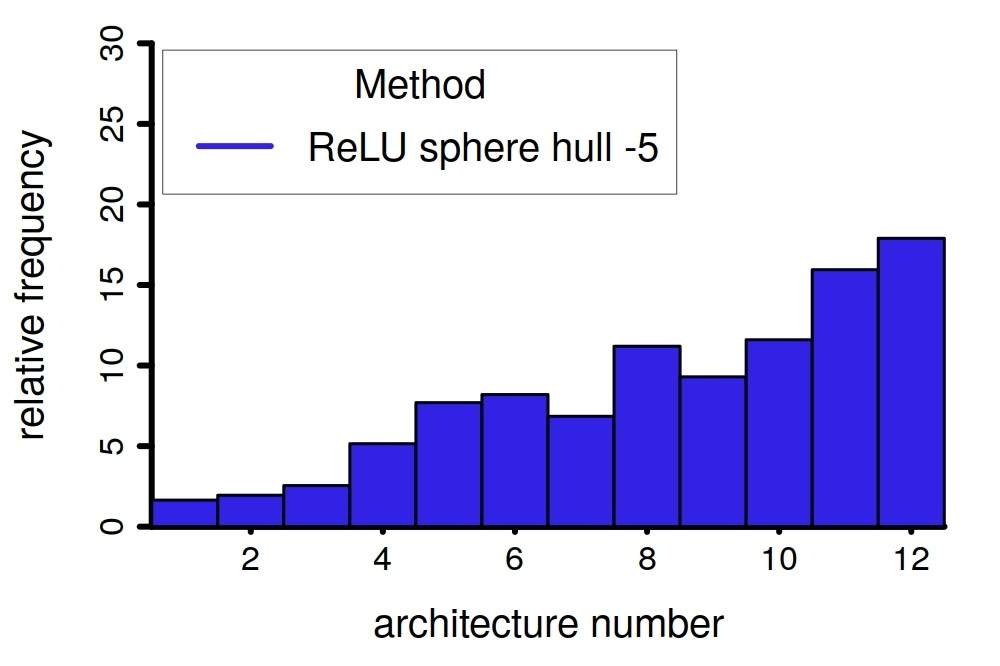}

\includegraphics[width=0.32\textwidth]{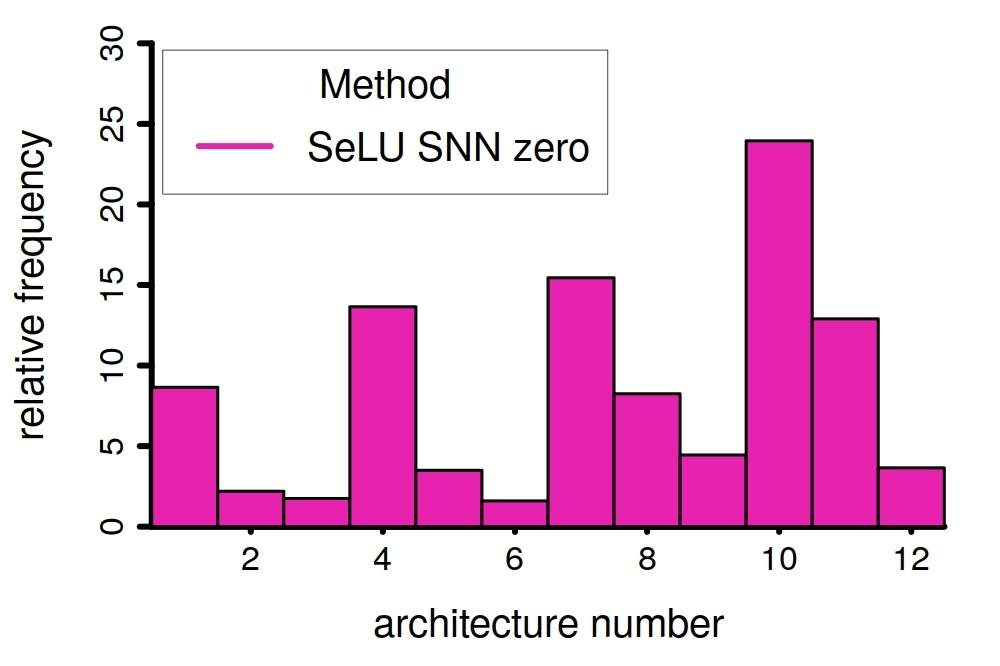}
\hspace*{-0.01\textwidth}
\includegraphics[width=0.32\textwidth]{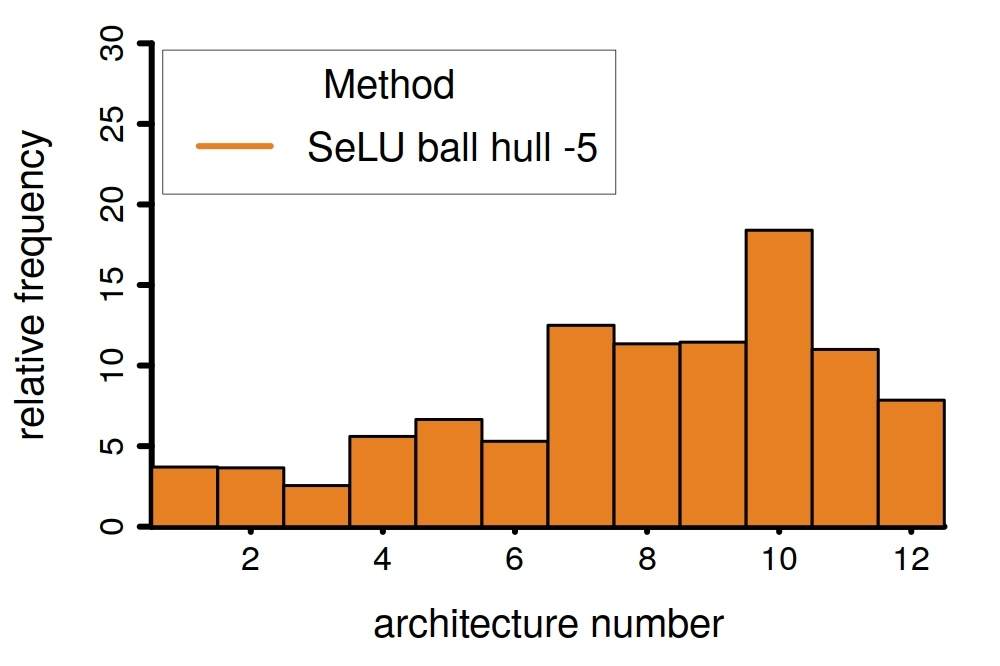}
\vspace*{-4ex}
\end{center}
\caption{Relative frequency of the architecture picked on the basis of the validation error for the  regression data sets. The histograms are based on the $40 \times 50 = 2000$ runs, and the architecture numbers in Table \ref{tab:architectures}. 
The ReLU networks 
have the tendency to pick deeper architectures, while \scalingname{SeLU SNN zero} prefers to pick narrower architectures. In addition, \scalingname{ReLU BN He zero}
slightly prefers shallower architectures.
\scalingname{SeLU ball hull -5} also has the tendency to pick deeper architectures but 
compared to the ReLU networks
this 
tendency is less pronounced.}\label{figure:arch-frequency-regress}
\end{figure}

\begin{figure}[t]
\begin{center}
\includegraphics[width=0.32\textwidth]{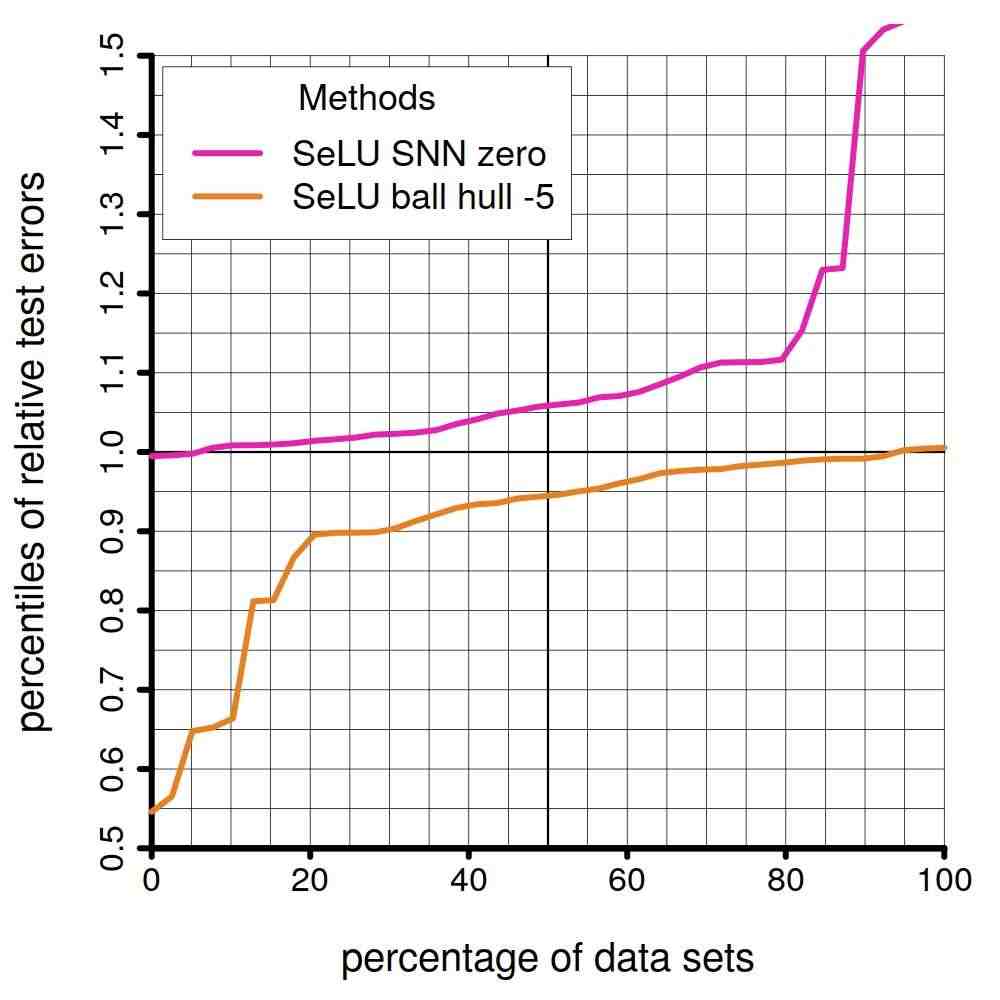}
\hspace*{-0.01\textwidth}
\includegraphics[width=0.32\textwidth]{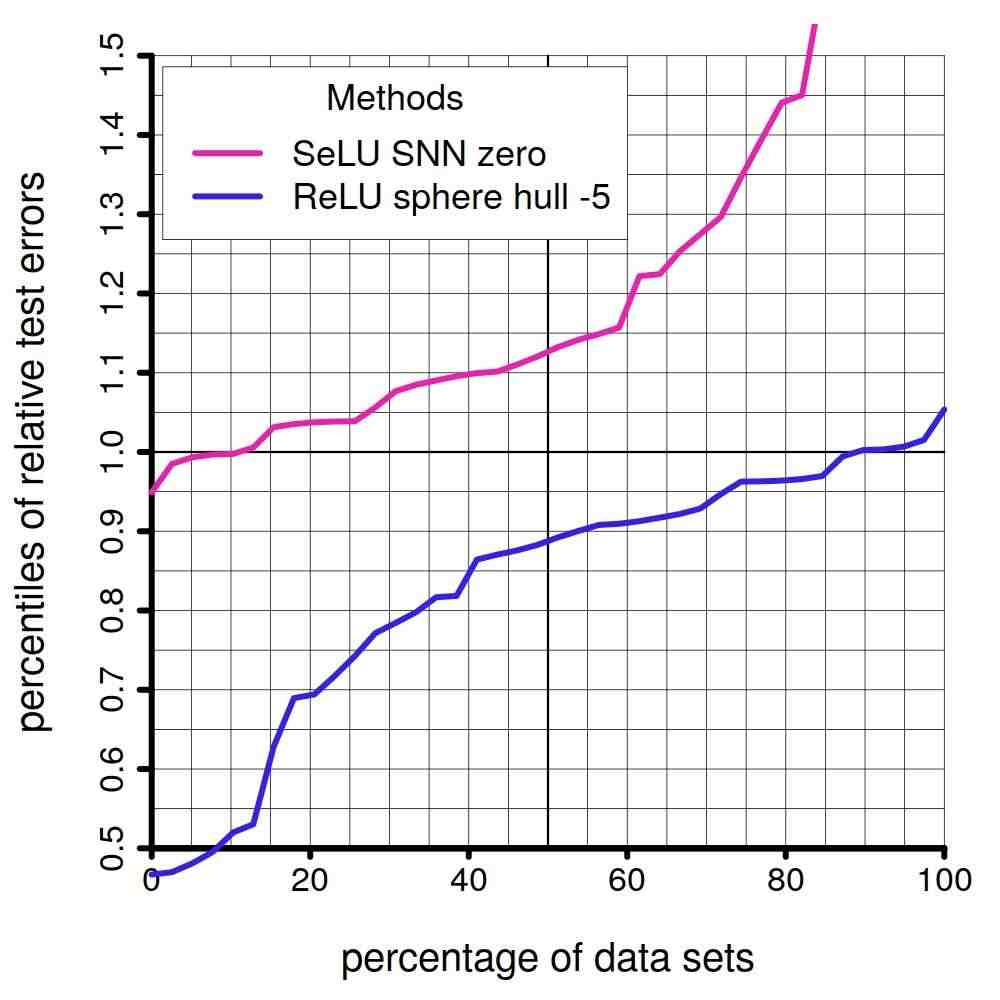}
\hspace*{-0.01\textwidth}
\includegraphics[width=0.32\textwidth]{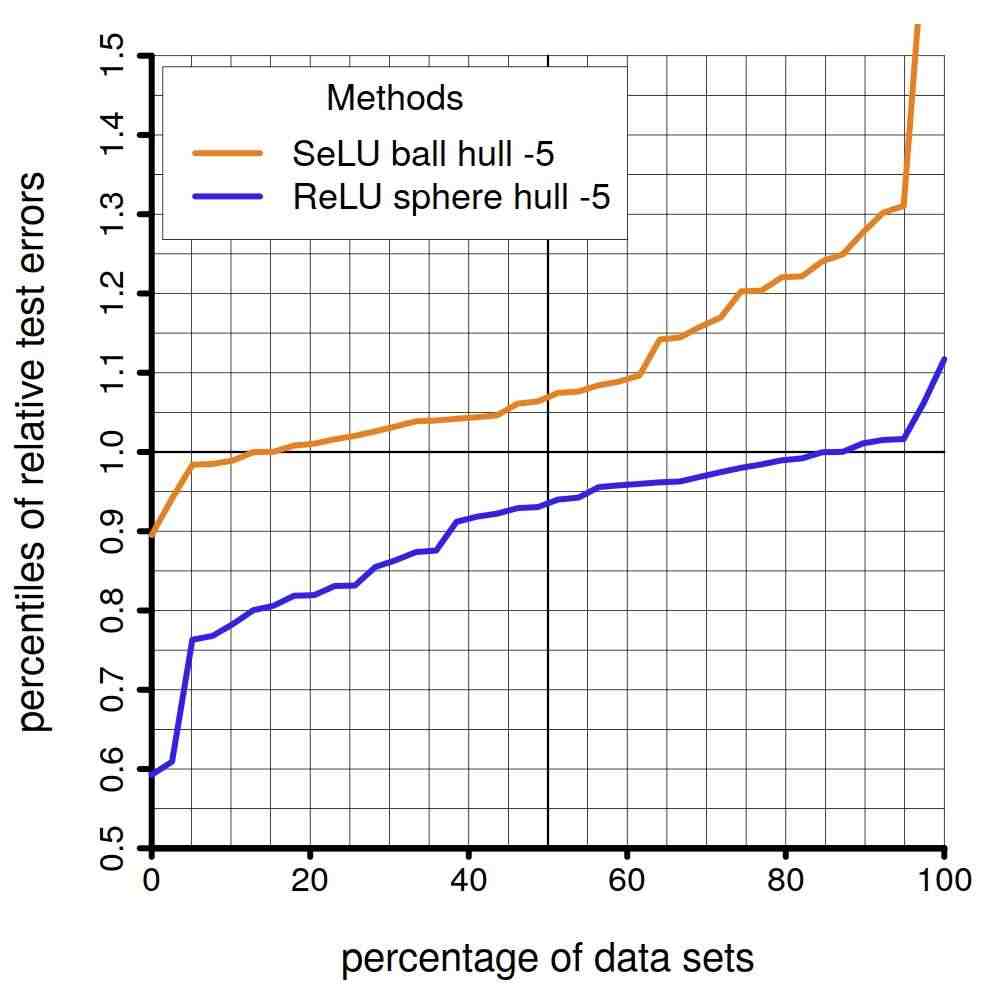}

\includegraphics[width=0.32\textwidth]{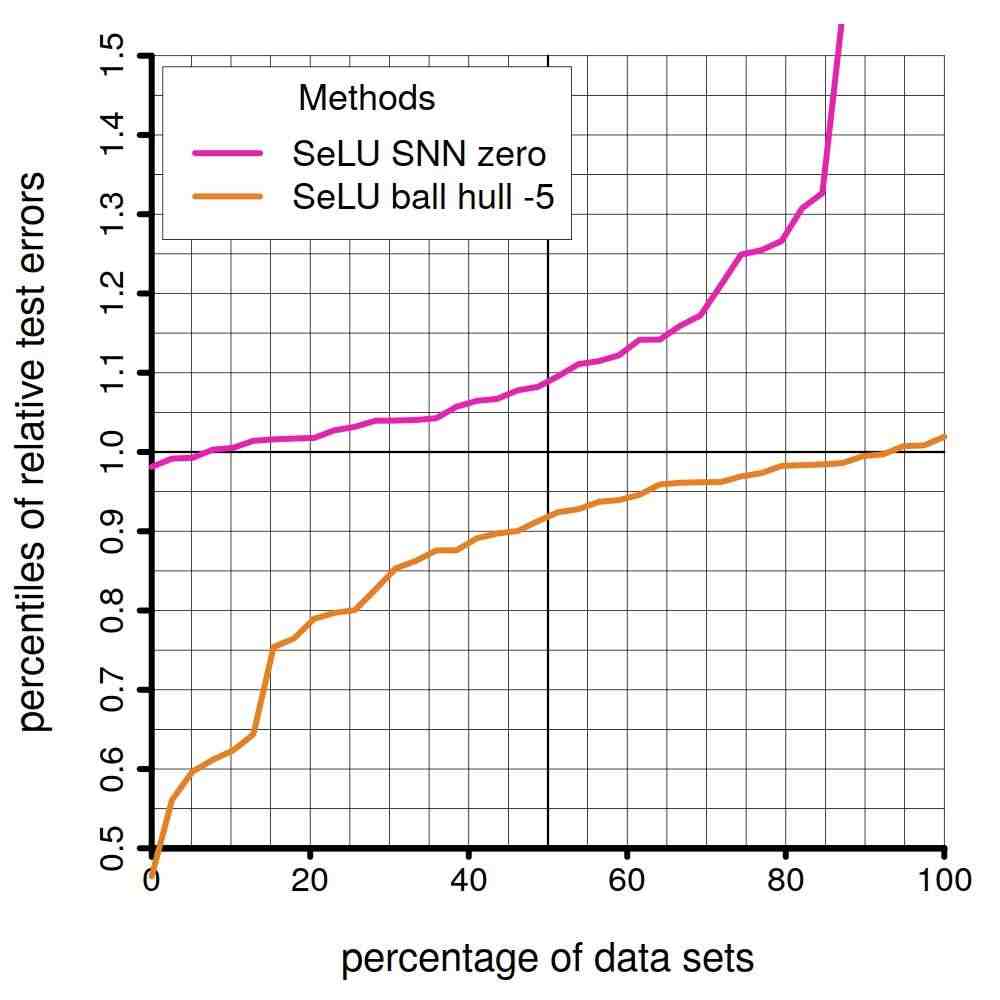}
\hspace*{-0.01\textwidth}
\includegraphics[width=0.32\textwidth]{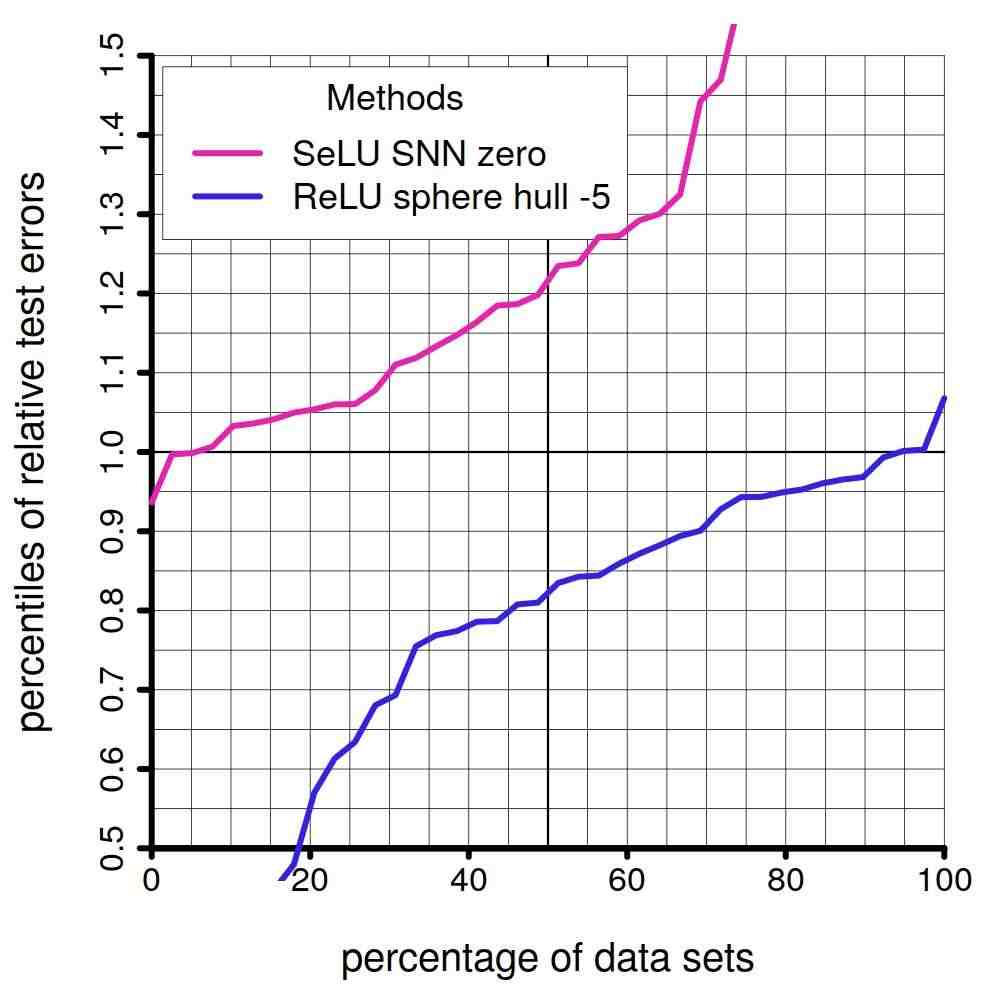}
\hspace*{-0.01\textwidth}
\includegraphics[width=0.32\textwidth]{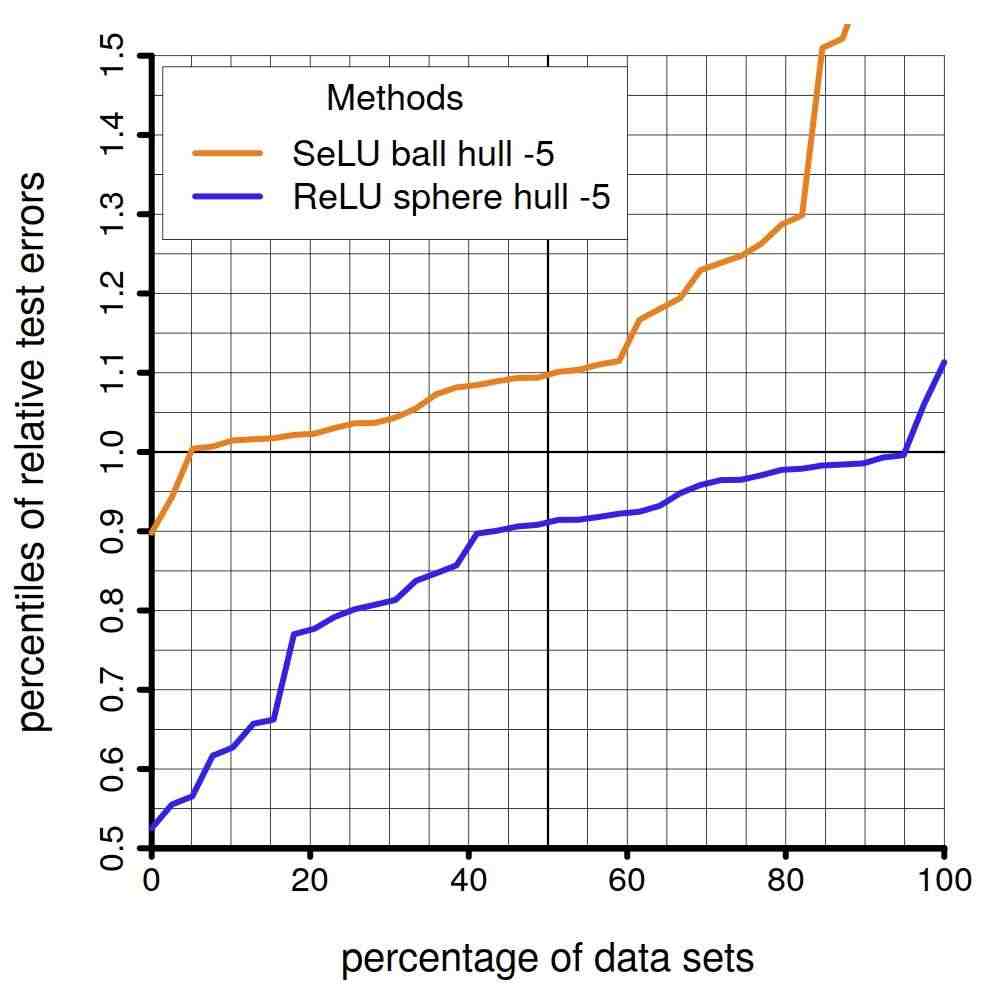}
\vspace*{-4ex}
\end{center}
\caption{Pairwise comparisons of old and new methods for regression with self-normalizing activation function
and different subsets of architectures. 
The first row displays the percentiles of $\rate_i$ if only the most narrow architectures
of each depth are considered during the selection phase, while the second row shows the corresponding results
for the widest architectures of each depth.
Although \scalingname{SeLU SNN zero} heavily prefers the narrow architectures, it is still 
almost uniformly outperformed by \scalingname{ReLU sphere hull -5} and 
\scalingname{SeLU ball hull -5} if only the narrow architectures are allowed.
Conversely, considering the widest architectures only, widens the gap between 
\scalingname{SeLU SNN zero}  and the other two methods as expected.}\label{figure:first-comp-selu-regress-archs}
\end{figure}

\emph{iii).} Let us finally investigate, how the chosen architectures influence 
our findings. In the regression case, Figure \ref{figure:arch-frequency-regress}
shows that all methods based on ReLU networks tend to pick deeper architectures and to some extend
this is also true for \scalingname{SeLU ball hull -5}. In contrast, \scalingname{SeLU SNN zero}
prefers narrower networks. One could thus ask, whether \scalingname{SeLU SNN zero} would have  
better performed in the comparisons 
if only architectures in favor of it would have been considered. 
Interestingly, a comparison between 
Figures 
\ref{figure:first-comp-selu-regress}
and
\ref{figure:first-comp-selu-regress-archs} shows that 
\scalingname{SeLU SNN zero} does benefit from such a choice of architectures, but the effect is 
rather minimal. In fact, \scalingname{SeLU SNN zero}  is still almost uniformly 
outperformed by both \scalingname{ReLU sphere hull -5} and \scalingname{SeLU ball hull -5}.

\begin{figure}[t]
\begin{center}
\includegraphics[width=0.32\textwidth]{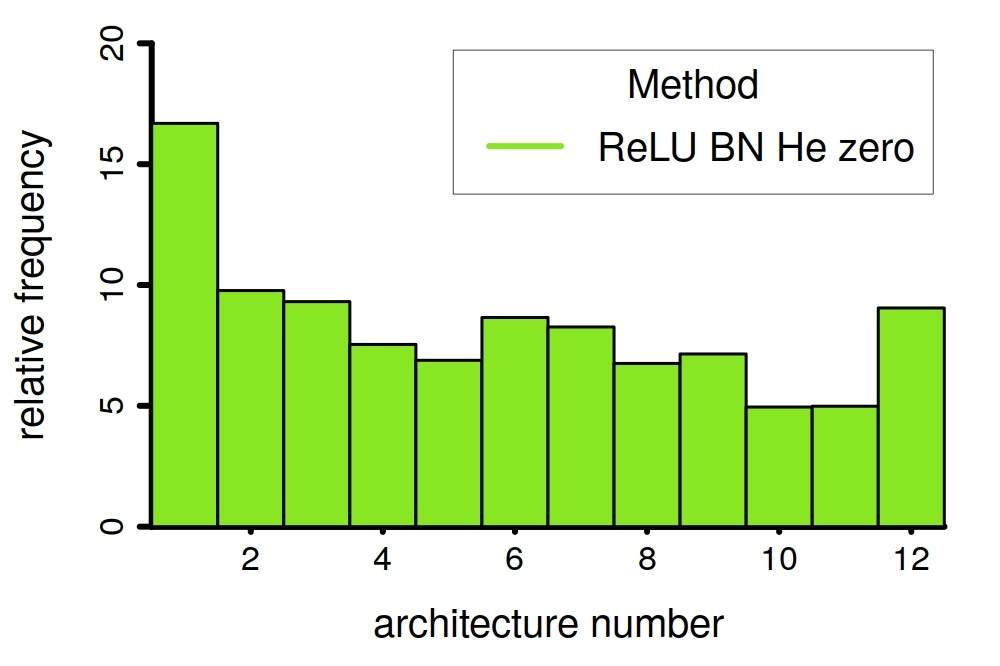}
\hspace*{-0.01\textwidth}
\includegraphics[width=0.32\textwidth]{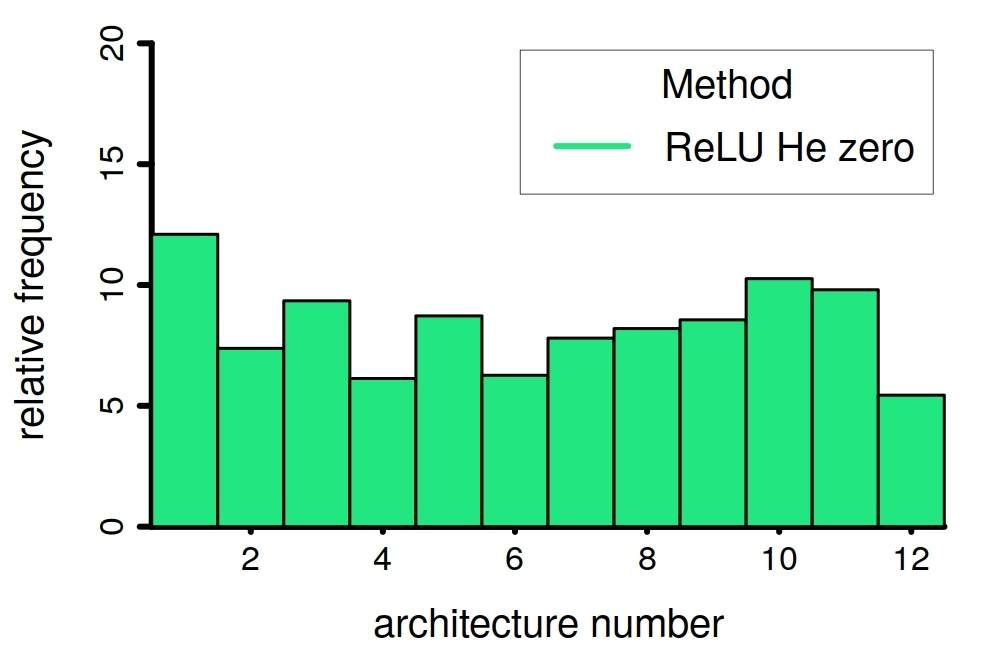}
\hspace*{-0.01\textwidth}
\includegraphics[width=0.32\textwidth]{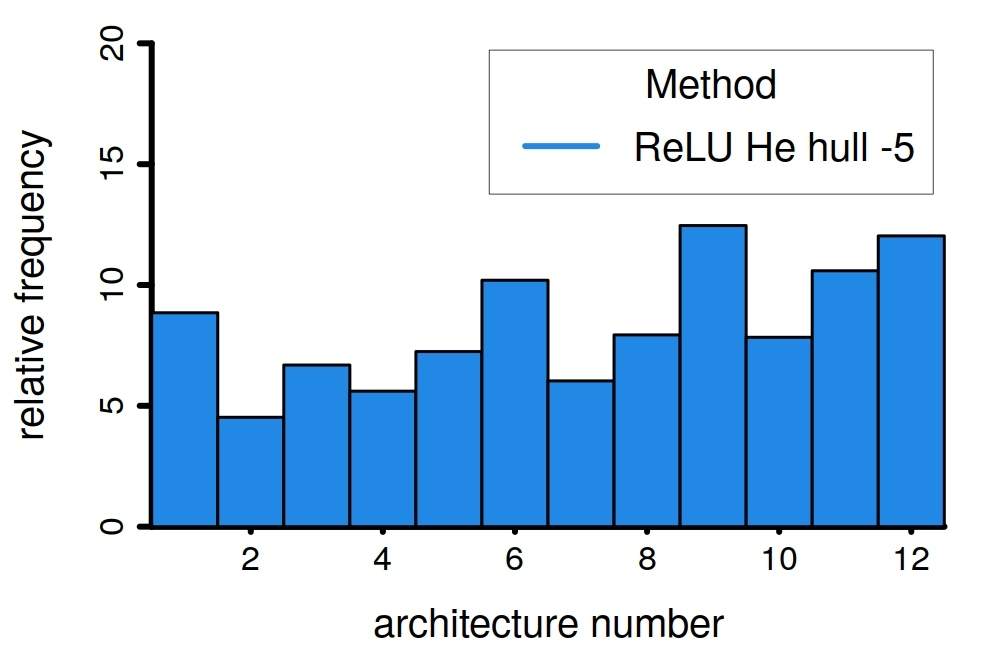}

\includegraphics[width=0.32\textwidth]{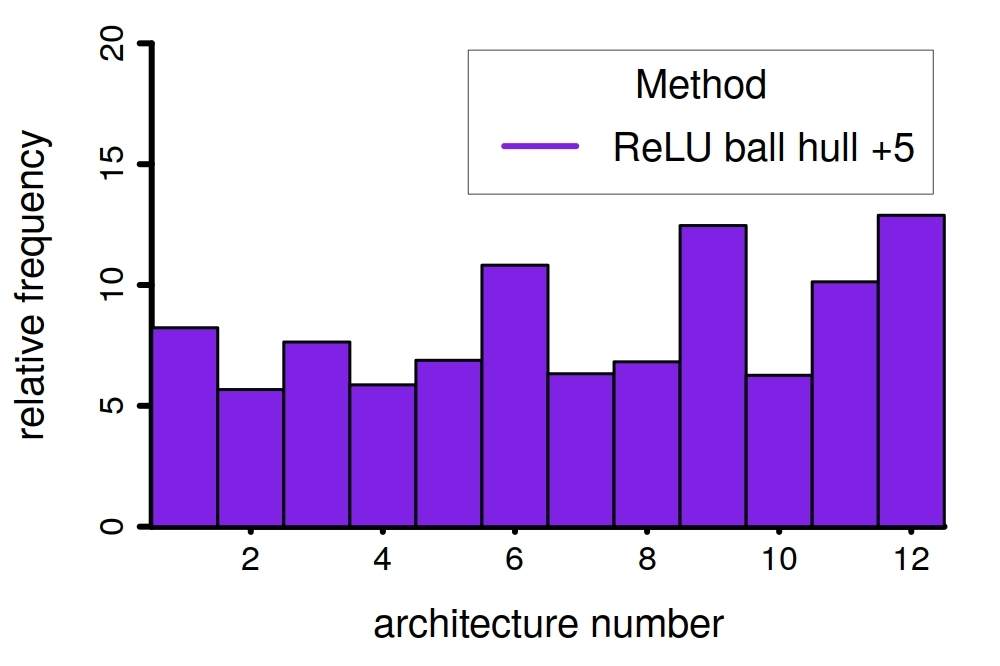}
\hspace*{-0.01\textwidth}
\includegraphics[width=0.32\textwidth]{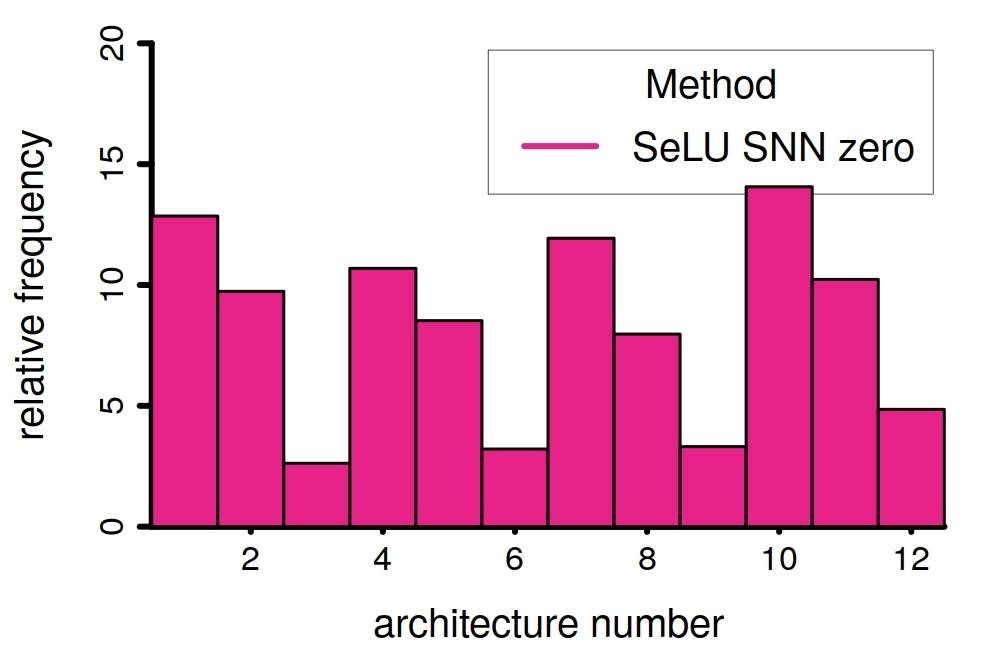}
\hspace*{-0.01\textwidth}
\includegraphics[width=0.32\textwidth]{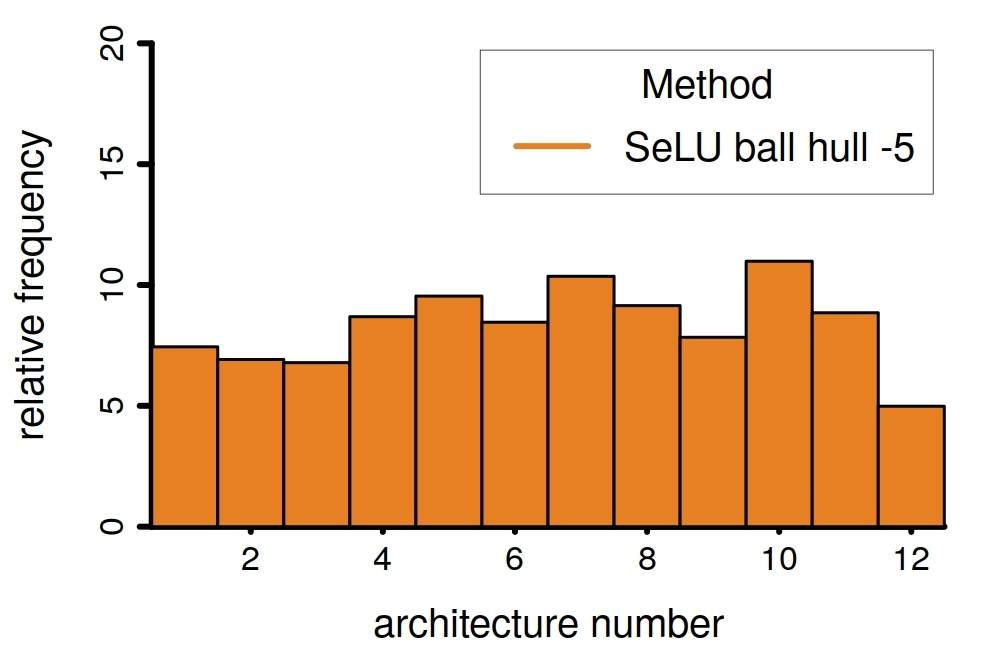}
\vspace*{-4ex}
\end{center}
\caption{Relative frequency of the architecture picked on the basis of the validation error for the binary classification data sets. The histograms are based on the $61 \times 50 = 3050$ runs, and architecture numbers of those of Table \ref{tab:architectures}. 
The strategies \scalingname{ReLU He hull -5} and \scalingname{ReLU ball hull +5} have the tendency to pick wider architectures of a given depth, while \scalingname{SeLU SNN zero} has the tendency to pick narrower architectures. In addition, \scalingname{ReLU BN He zero}
slightly prefers shallower architectures.
For the remaining initialization strategies there is no clear and simple tendency.}\label{figure:arch-frequency}
\end{figure}

\begin{figure}[tp]
\begin{center}
\includegraphics[width=0.32\textwidth]{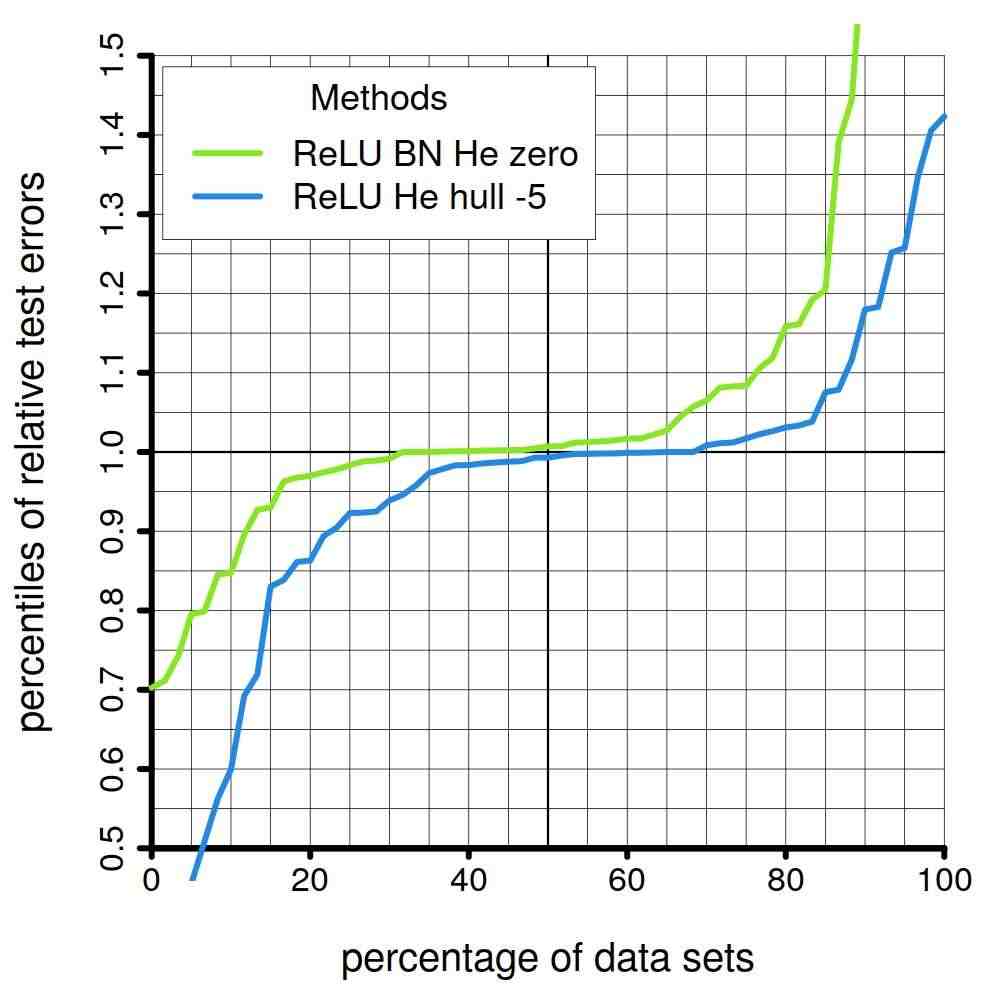}
\hspace*{-0.01\textwidth}
\includegraphics[width=0.32\textwidth]{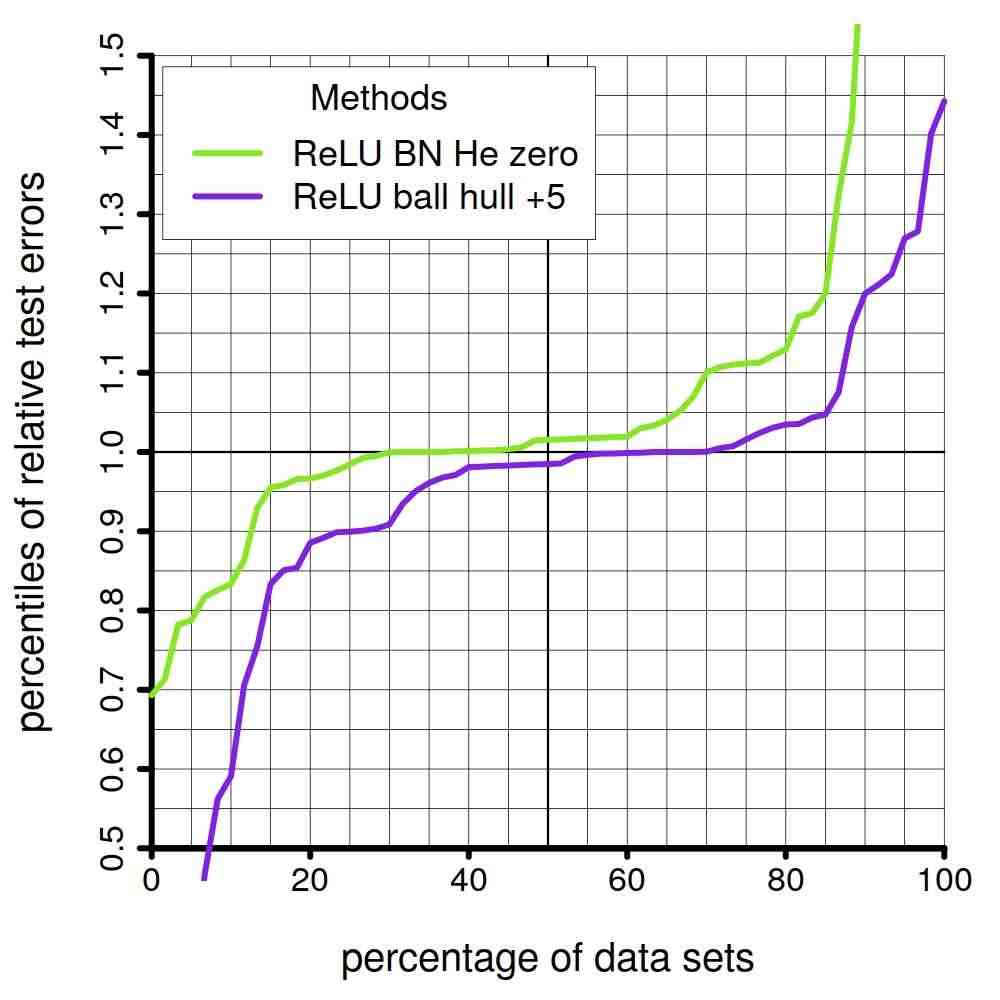}
\hspace*{-0.01\textwidth}
\includegraphics[width=0.32\textwidth]{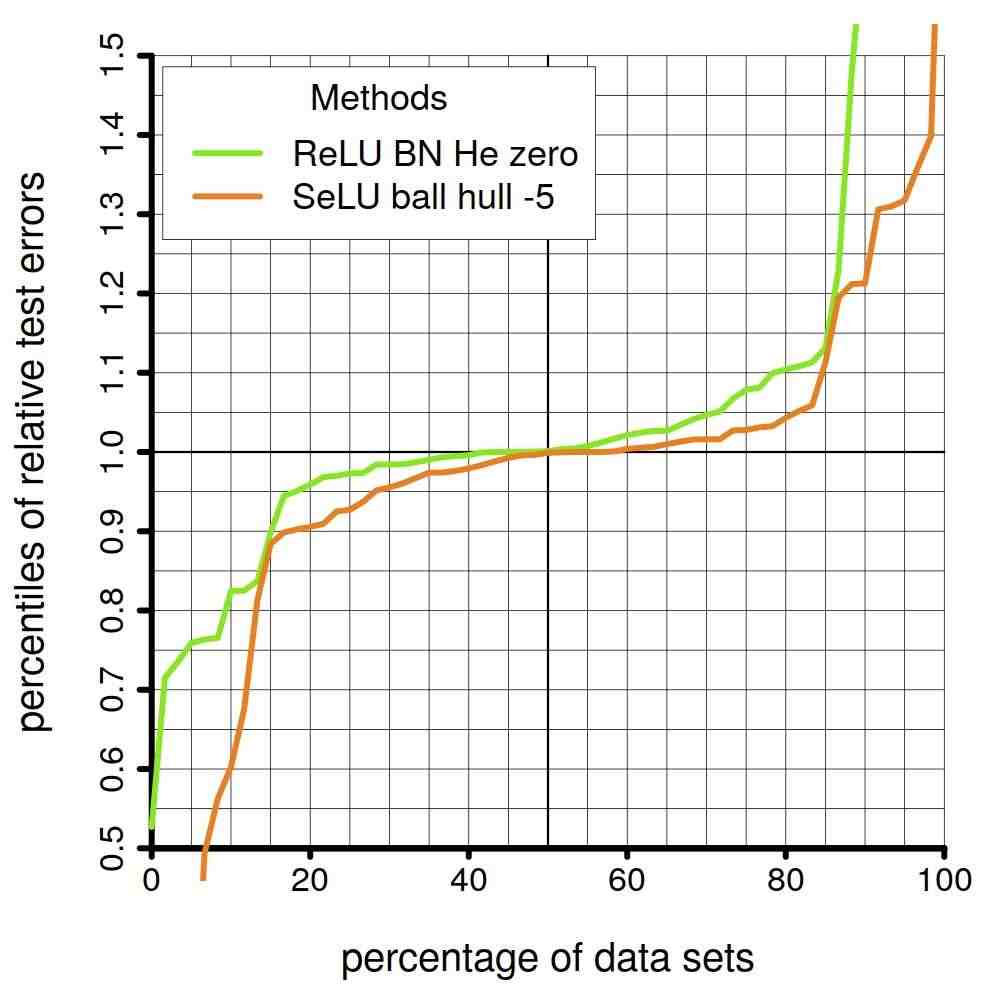}

\includegraphics[width=0.32\textwidth]{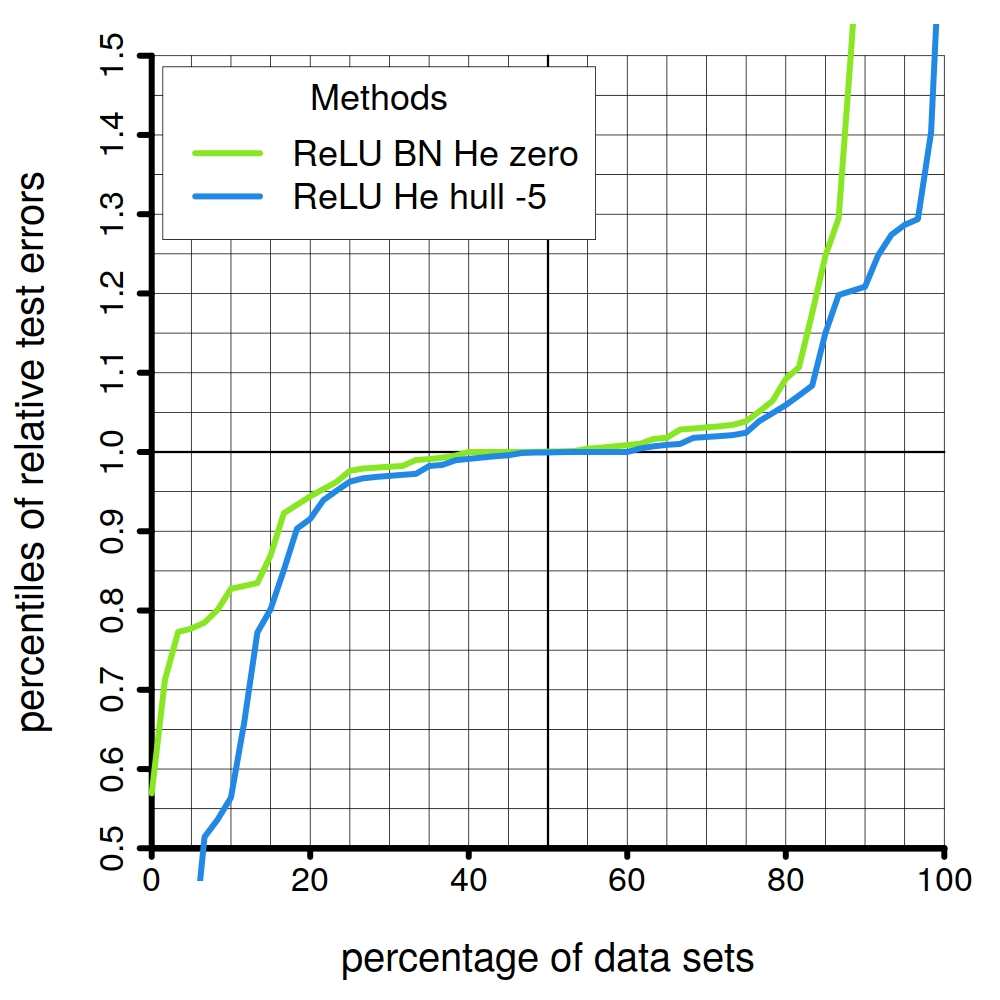}
\hspace*{-0.01\textwidth}
\includegraphics[width=0.32\textwidth]{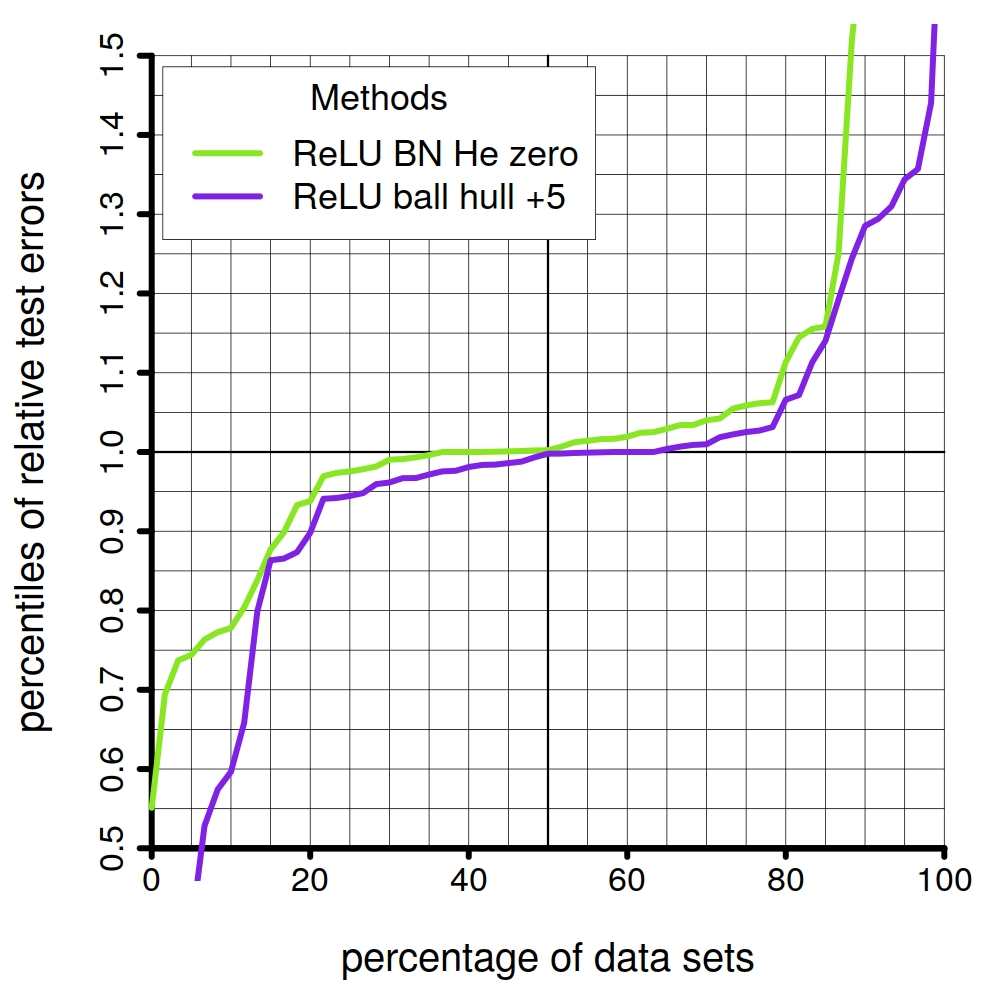}
\hspace*{-0.01\textwidth}
\includegraphics[width=0.32\textwidth]{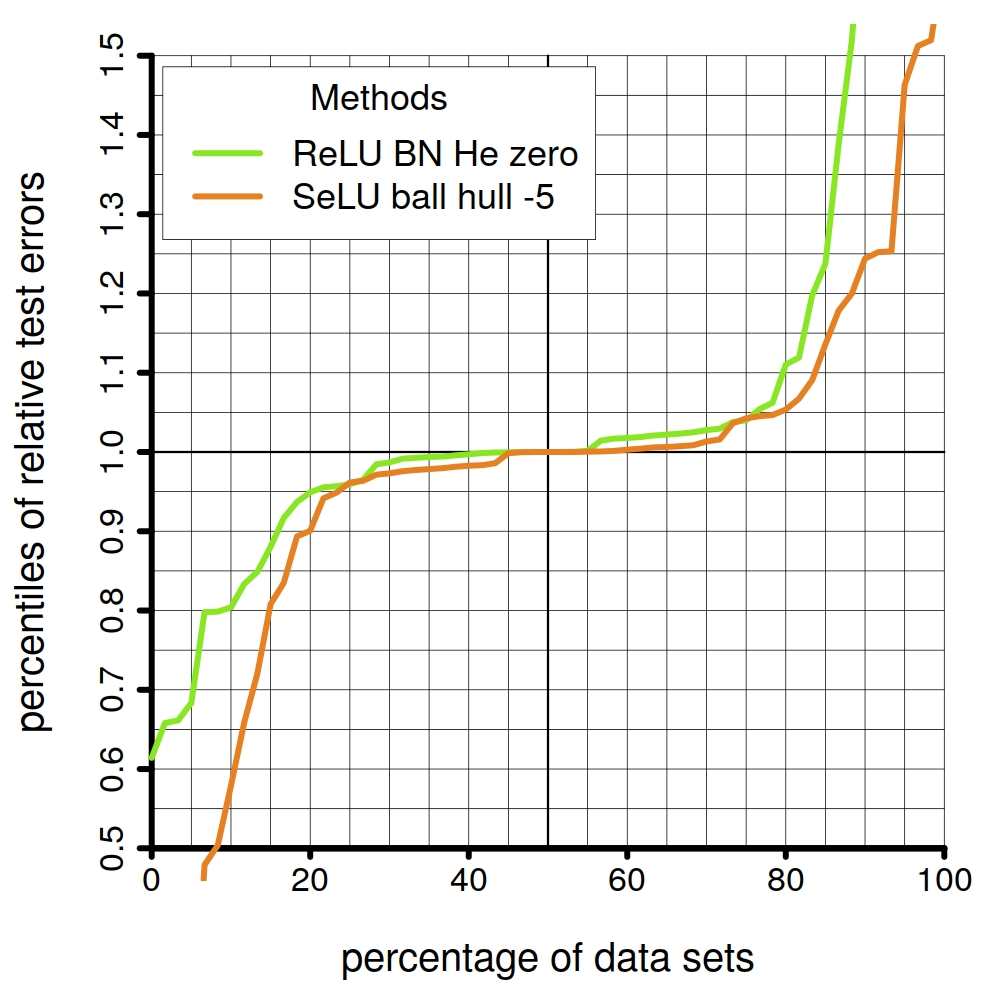}

\includegraphics[width=0.32\textwidth]{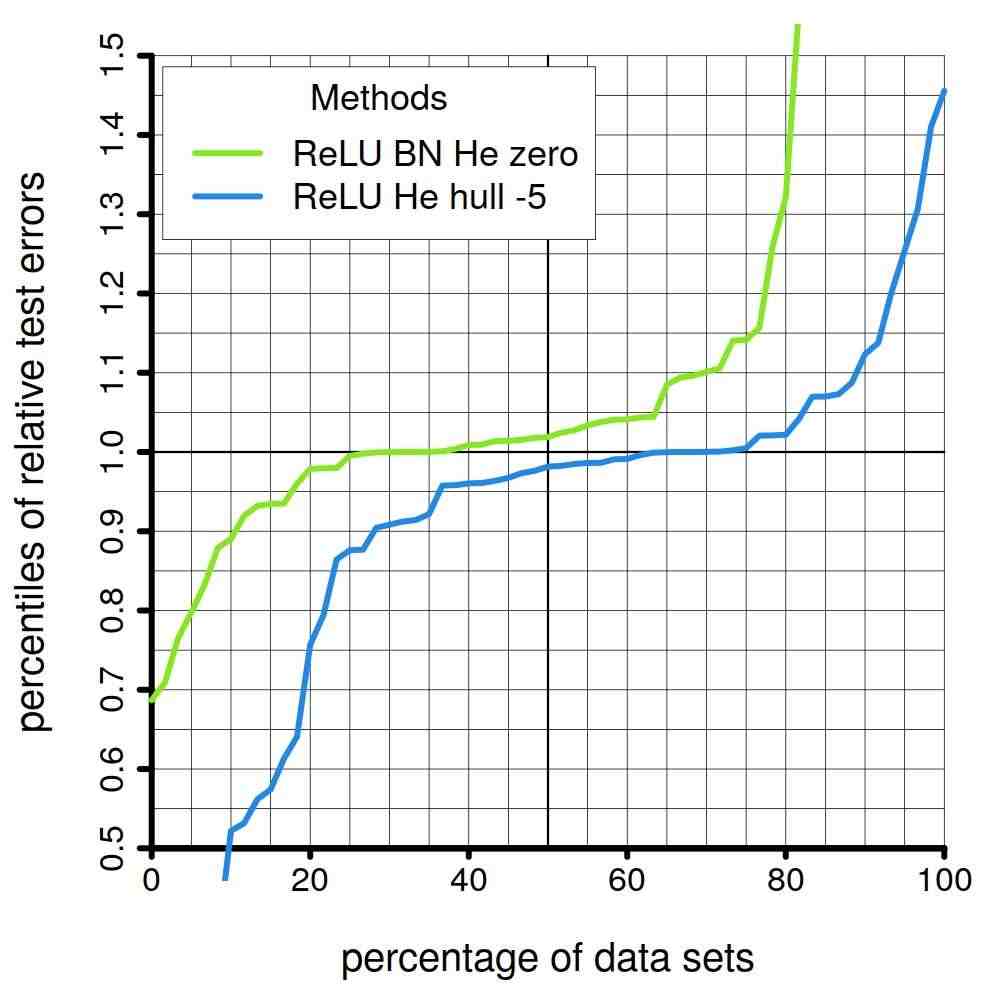}
\hspace*{-0.01\textwidth}
\includegraphics[width=0.32\textwidth]{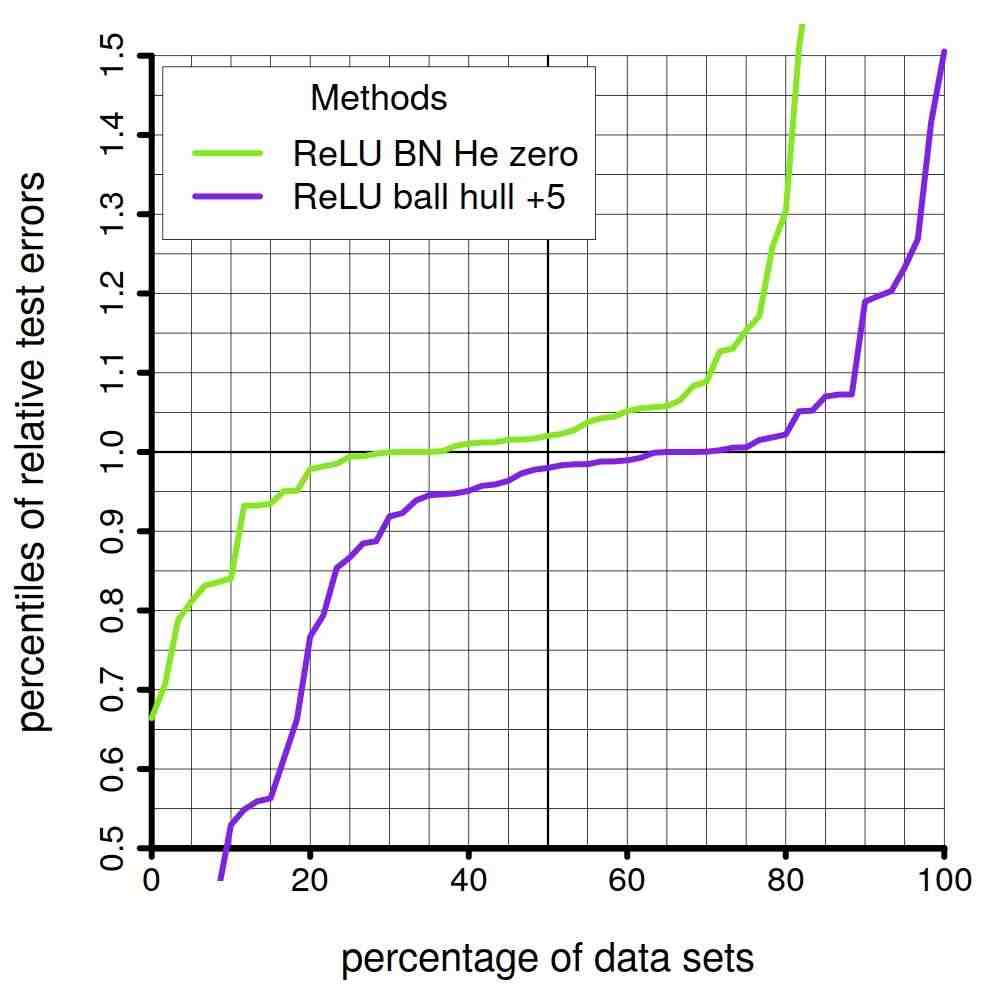}
\hspace*{-0.01\textwidth}
\includegraphics[width=0.32\textwidth]{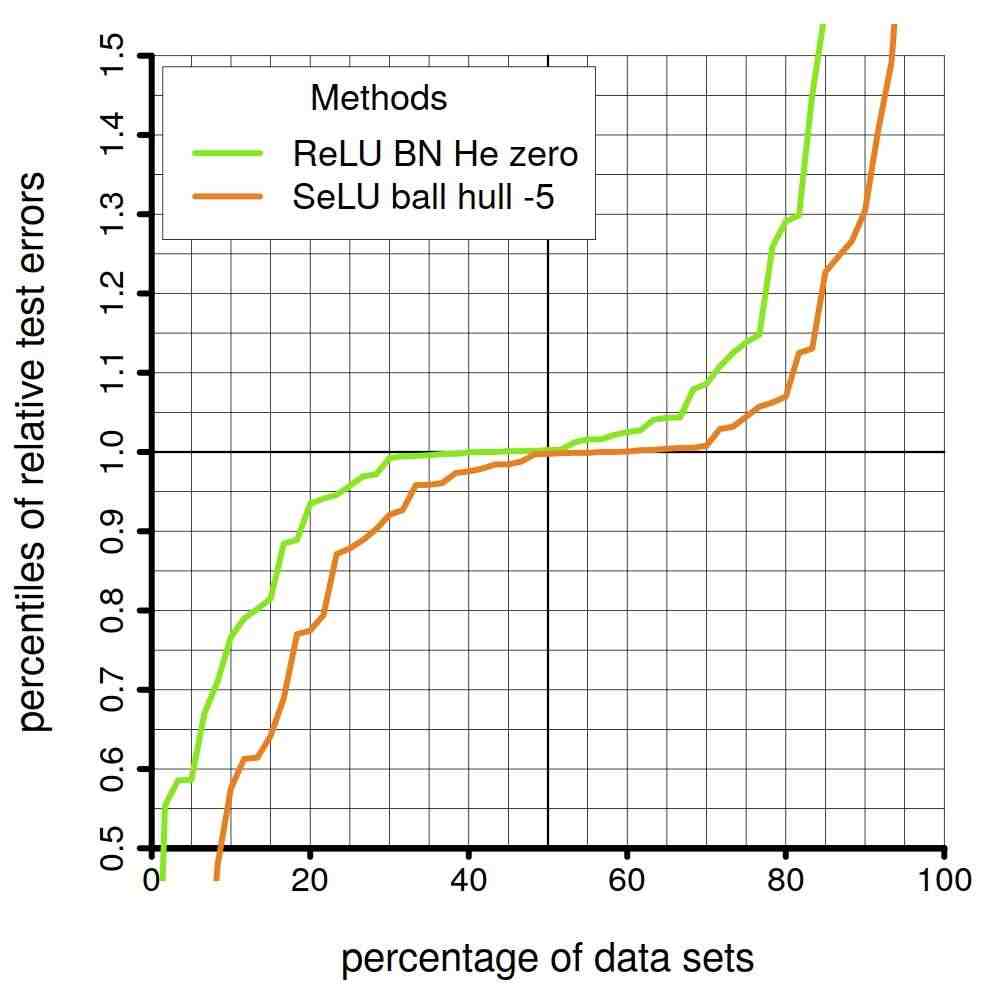}

% \includegraphics[width=0.32\textwidth]{./images/gains_and_losses_log_narrow_width_ReLU_BN1_He_zero_ReLU_He_hull_5_unlim_pat5_hist.jpg}
% \hspace*{-0.01\textwidth}
% \includegraphics[width=0.32\textwidth]{./images/gains_and_losses_log_narrow_width_ReLU_BN1_He_zero_ReLU_ball_hull_5_unlim_pat5_hist.jpg}
% \hspace*{-0.01\textwidth}
% \includegraphics[width=0.32\textwidth]{./images/gains_and_losses_log_narrow_width_ReLU_BN1_He_zero_SeLU_ball_hull_5_unlim_pat5_hist.jpg}
\vspace*{-4ex}
\end{center}
\caption{Pairwise comparisons against batch normalization with standard initialization and different subsets of used architectures in the case of binary classification.
The first row displays the results if only the architectures with two hidden layers are considered. According to Figure \ref{figure:arch-frequency}, these architectures are the ones that are most often picked with by \scalingname{ReLU BN He zero}.
The second row displays the results if only the architectures with the most narrow widths of each depth are considered. 
According to Figure \ref{figure:arch-frequency}, these architectures are the ones that are less often picked by 
\scalingname{ReLU He hull -5} and \scalingname{ReLU ball hull +5}. The third row displays the results if only the architectures with the widest widths of each depth are considered. 
According to Figure \ref{figure:arch-frequency}, these architectures are the ones that are most often picked by 
\scalingname{ReLU He hull -5} and \scalingname{ReLU ball hull +5}. Together, the graphics show that 
the new initialization strategies outperform \scalingname{ReLU BN He zero} even if the architectures are chosen in favor of \scalingname{ReLU BN He zero}. Moreover, if the architectures are chosen in favor of \scalingname{ReLU He hull -5} and \scalingname{ReLU ball hull +5}, then the difference between these methods and \scalingname{ReLU BN He zero}
becomes more pronounced.}\label{figure:first-comp-bn-arch-constraints}
\end{figure}

\begin{figure}[t]
\begin{center}
\includegraphics[width=0.32\textwidth]{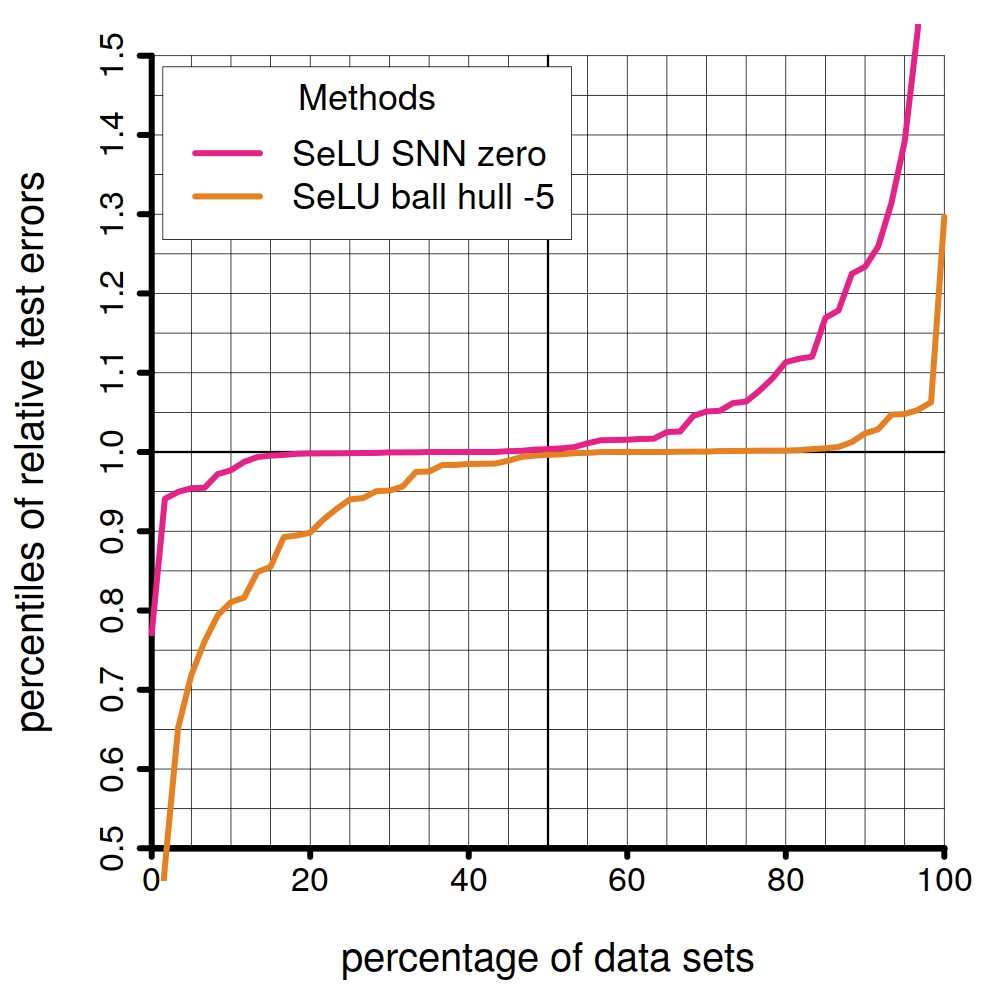}
\hspace*{-0.01\textwidth}
\includegraphics[width=0.32\textwidth]{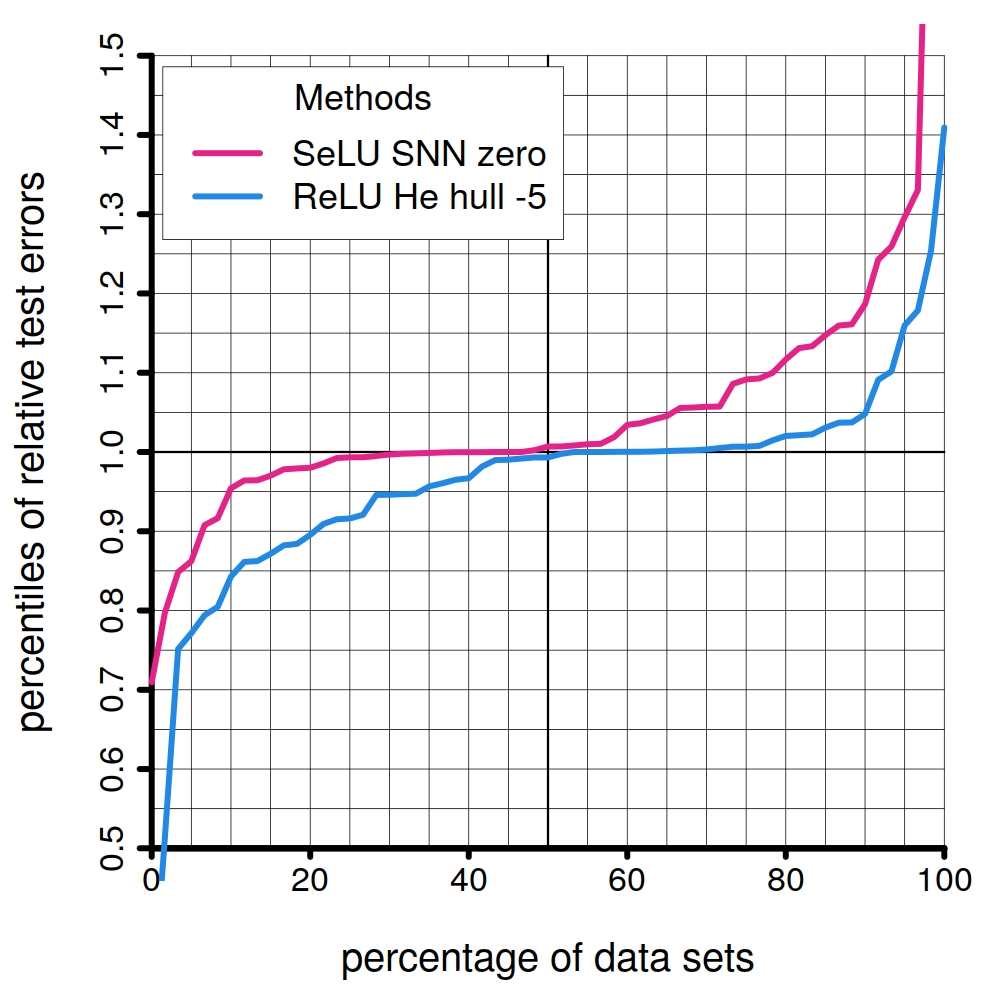}
\hspace*{-0.01\textwidth}
\includegraphics[width=0.32\textwidth]{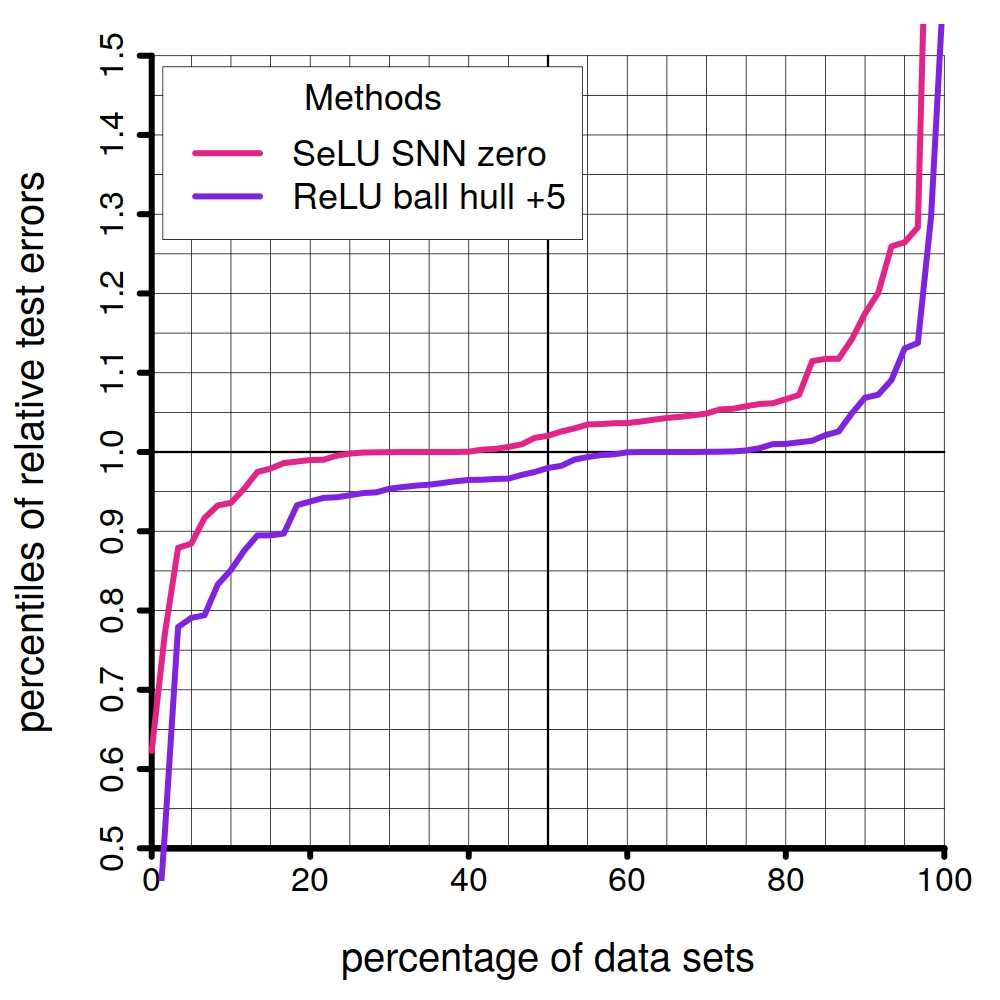}

\hspace*{-0.01\textwidth}
\includegraphics[width=0.32\textwidth]{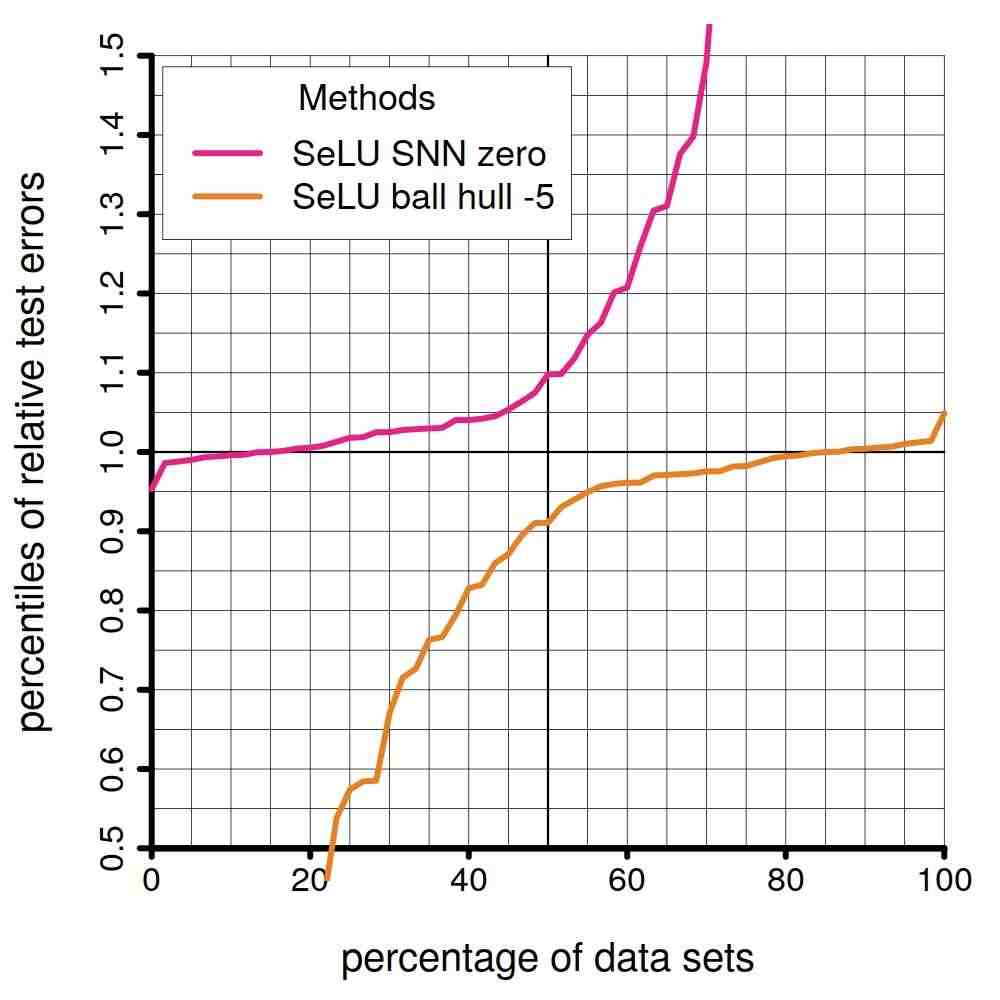}
\hspace*{-0.01\textwidth}
\includegraphics[width=0.32\textwidth]{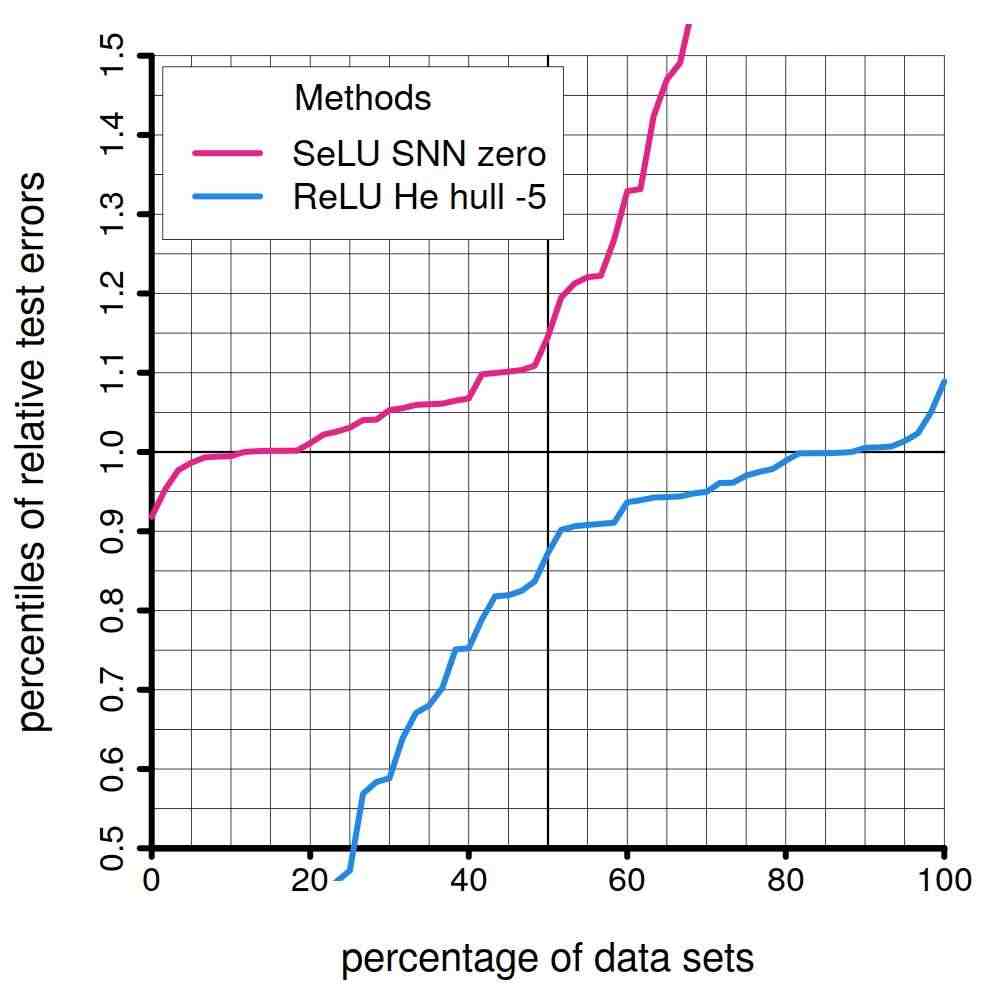}
\hspace*{-0.01\textwidth}
\includegraphics[width=0.32\textwidth]{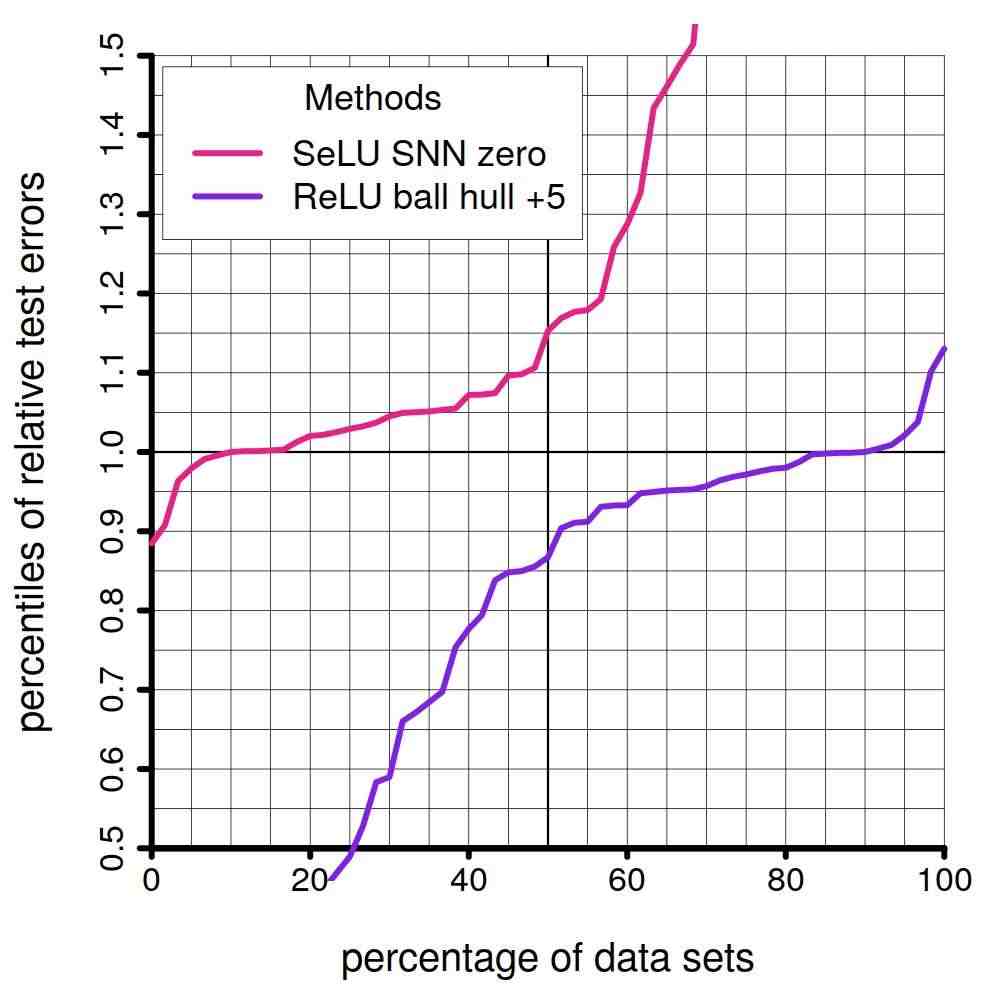}

\vspace*{-4ex}
\end{center}
\caption{Pairwise comparisons against self-normalizing networks with standard initialization and different subsets of used architectures in the case of binary classification.
The first row displays the results if only the architectures with the most narrow widths of each depth are considered. 
According to Figure \ref{figure:arch-frequency}, these architectures are the ones that are most often picked by 
\scalingname{SeLU SNN zero}. The second row displays the results if only the architectures with the widest widths of each depth are considered. 
According to Figure \ref{figure:arch-frequency}, these architectures are the ones that are most often picked by 
\scalingname{ReLU He hull -5} and \scalingname{ReLU ball hull +5}. Together, the graphics show that 
the new initialization strategies outperform \scalingname{SeLU SNN zero} even if the architectures are chosen in favor of 
\scalingname{SeLU SNN zero}. Moreover, if the architectures are chosen in favor of \scalingname{ReLU He hull -5} 
and \scalingname{ReLU ball hull +5}, then all three new initialization strategies almost uniformly outperform \scalingname{SeLU SNN zero}.}\label{figure:first-comp-selu-arch-constraints}
\end{figure}

In the classification case, the picture is again a bit more interesting. Here, Figure 
\ref{figure:arch-frequency} shows that \scalingname{ReLU He hull -5} and
\scalingname{ReLU ball hull +5} tend to prefer wider architectures, while 
\scalingname{SeLU SNN zero} again prefers narrower architectures. In addition,
\scalingname{ReLU BN He zero} slightly prefers shallower architectures, while the remaining
two methods \scalingname{ReLU He zero} and \scalingname{SeLU ball hull -5} 
do not have a clear tendency.
Interestingly, Figure \ref{figure:first-comp-bn-arch-constraints} shows that the standard
\scalingname{ReLU BN He zero} is still outperformed by all three 
new initialization strategies \scalingname{ReLU He hull -5},
\scalingname{ReLU ball hull +5}, and \scalingname{SeLU ball hull -5} 
if the architectures are restricted in favor of \scalingname{ReLU BN He zero}. 
In fact, if  only the shallowest three architectures, which are preferred by 
\scalingname{ReLU BN He zero}, are considered, then \scalingname{ReLU BN He zero}
seems to perform even slightly worse against \scalingname{ReLU He hull -5}
and \scalingname{ReLU ball hull +5}. Moreover, if these two methods are penalized 
by restricting to  narrow architectures, \scalingname{ReLU BN He zero} seem to slightly benefit 
against \scalingname{ReLU He hull -5}
and \scalingname{ReLU ball hull +5}, yet the effect is minimal  \scalingname{ReLU BN He zero} 
is still outperformed. In contrast, if \scalingname{ReLU He hull -5}
and \scalingname{ReLU ball hull +5} are favored by allowing the widest architectures only, 
then the gap between these two methods and  \scalingname{ReLU BN He zero}  clearly widens 
compared to the set-up that includes all architectures and which is shown in Figure 
\ref{figure:first-comp-bn}. Finally, Figure \ref{figure:first-comp-selu-arch-constraints}
illustrates the effects when favoring or penalizing \scalingname{SeLU SNN zero}:
If \scalingname{SeLU SNN zero} is favored by considering the narrowest architectures, only, then 
\scalingname{SeLU SNN zero} is still outperformed by all three new initialization methods, however,
the gap between e.g.~\scalingname{SeLU SNN zero} and \scalingname{SeLU ball hull -5} 
narrows a bit as a comparison between 
Figures \ref{figure:first-comp-selu} and
\ref{figure:first-comp-selu-arch-constraints} show. On the other hand, if 
\scalingname{SeLU SNN zero} is penalized by considering the widest architectures, then 
all three new initialization strategies substantially and almost uniformly outperform 
\scalingname{SeLU SNN zero}.

\emph{In summary, our overall results we obtained by considering all 12 architectures 
are rather insensitive against changes in the allowed architectures.}

% 
% \begin{figure}[t]
% \begin{center}
% \includegraphics[width=0.32\textwidth]{./images/gains_and_losses_log_ReLU_BN1_He_zero_ReLU_ball_hull_5_time_pat5.jpg}
% \hspace*{-0.01\textwidth}
% \includegraphics[width=0.32\textwidth]{./images/gains_and_losses_log_ReLU_BN1_He_zero_ReLU_He_hull_5_time_pat5.jpg}
% \hspace*{-0.01\textwidth}
% \includegraphics[width=0.32\textwidth]{./images/gains_and_losses_log_ReLU_BN1_He_zero_SeLU_ball_hull_5_time_pat5.jpg}
% 
% 
% \includegraphics[width=0.32\textwidth]{./images/gains_and_losses_log_ReLU_BN1_He_zero_ReLU_ball_hull_5_time_pat5_hist.jpg}
% \hspace*{-0.01\textwidth}
% \includegraphics[width=0.32\textwidth]{./images/gains_and_losses_log_ReLU_BN1_He_zero_ReLU_He_hull_5_time_pat5_hist.jpg}
% \hspace*{-0.01\textwidth}
% \includegraphics[width=0.32\textwidth]{./images/gains_and_losses_log_ReLU_BN1_He_zero_SeLU_ball_hull_5_time_pat5_hist.jpg}
% \vspace*{-4ex}
% \end{center}
% \caption{Pairwise comparisons against batch normalization with standard initialization under a tight time limit. The three 
% comparing columns are the same as in Figure \ref{figure:first-comp-bn}. A comparison to Figure \ref{figure:first-comp-bn} shows that batch normalization suffers substantially more under the time limit than the three new initialization strategies.}\label{figure:first-comp-bn-time}
% \end{figure}

% 
% \begin{figure}[t]
% \begin{center}
% \includegraphics[width=0.32\textwidth]{./images/gains_and_losses_log_tiny_net_ReLU_BN1_He_zero_ReLU_He_hull_5_unlim_pat5.jpg}
% \hspace*{-0.01\textwidth}
% \includegraphics[width=0.32\textwidth]{./images/gains_and_losses_log_tiny_net_ReLU_BN1_He_zero_ReLU_ball_hull_5_unlim_pat5.jpg}
% \hspace*{-0.01\textwidth}
% \includegraphics[width=0.32\textwidth]{./images/gains_and_losses_log_tiny_net_ReLU_BN1_He_zero_SeLU_ball_hull_5_unlim_pat5.jpg}
% 
% 
% \vspace*{-4ex}
% \end{center}
% \caption{Cv results for classification with the 3 tiny architectures and the 4 wide architectures.}\label{figure:first-comp-bn-time}
% \end{figure}

% \newpage
\section{Proofs}

\subsection{Proofs for Section \ref{sec:one-d-sit}}

For the proof of Proposition \ref{result:one-layer-deriv} we need the following 
trivial lemma.

\begin{lemma}\label{result:general-emp-derivative}
   Let $X\neq \emptyset$, $D:=((x_1,y_1),\dots,(x_n,y_n)) \in (X\times \R)^n$ be a data set,
$L:\R\times \R\to [0,\infty)$ be  differentiable loss function, and 
 $g:X\times \R^p\to \R$
be a function. Furthermore, let 
$v_0\in \R^p$ be a point such that $v\mapsto g(x_j,v)$ is differentiable in $v_0$
for all $j=1,\dots,n$. 
 Then $v\mapsto \RD L{g(\,\cdot\,, v)}$ is differentiable at $v_0$ and we have
\begin{displaymath}
   \frac{\partial \RD L{g(\,\cdot\,, v)}}{\partial v} (v_0) 
=  \frac 1n \sum_{j=1}^n L'\bigl(y_j, g(x_j, v_0)\bigr) \cdot \frac{\partial g}{\partial v}(x_j, v_0)\, .
\end{displaymath}
\end{lemma}

\begin{proofof}{Lemma \ref{result:general-emp-derivative}}
   Using the chain rule we obtain 
\begin{align*}
   \frac{\partial L(y_j, g(x_j, v))}{\partial v} (v_0) 
=  L'\bigl(y_j, g(x_j, v_0)\bigr)  \cdot \frac{\partial g }{\partial v}(x_j, v_0)\, ,
\end{align*}
for all $j=1,\dots,n$.
From this we easily derive the assertion.
\end{proofof}

\begin{proofof}{Proposition \ref{result:one-layer-deriv}}
Our goal is to apply Lemma \ref{result:general-emp-derivative} in a version that is extended in the sense of  
\eqref{chain-rule-extension-a} and  \eqref{chain-rule-extension-b}
to $g\in \arch 1m$.
To this end, we define $p:= 3m+1$, and for 
$(w,c,a,b) \in \R^m\times \R\times \R^m\times \R^m = \R^p$ we write 
$g(\mycdot, w,c,a,b):\R\to \R$ for the function given by \eqref{first-layer}, that is 
\begin{equation*} 
   g(x,  w,c,a,b) = \sum_{i=1}^m w_i \brelu{a_i x + b_i} +c\, , \qquad \qquad x\in \R. 
\end{equation*}
Now recall from 
 Proposition \ref{result:one-layer-repres}   that for all $x\in \R$ we have 
 \begin{align*}
   g(x,  w,c,a,b) 
   &= 
\sum_{i\in I_-: x\leq x_i^*} \,w_i(a_ix + b_i) + \sum_{i\in I_+: x\geq x_i^*} \,w_i(a_ix + b_i)
 + \sum_{i\in I\setminus I_*} w_i \relu{b_i} + c \\
 & =
 \sum_{i\in I_-: x\leq x_i^*} \,w_i\relu{a_ix + b_i} + \sum_{i\in I_+: x\geq x_i^*} \,w_i\relu{a_ix + b_i}
 + \sum_{i\in I\setminus I_*} w_i \relu{a_ix + b_i} + c\, .
\end{align*}
For $i\in I_-$,  we thus find by \eqref{relu-chain-rule-a} and  \eqref{relu-chain-rule-b}
\begin{align*}
   \frac{\partial g}{\partial a_i}(x, w,c,a,b) 
&=
w_i \cdot x \cdot \eins_{(-\infty,x_i^*)}(x)   + \partial_0 \cdot w_i \cdot x \cdot \eins_{\{x_i^*\}}(x) \, ,\\
   \frac{\partial g}{\partial b_i}(x, w,c,a,b) 
&=
w_i  \cdot \eins_{(-\infty,x_i^*)}(x)  + \partial_0 \cdot w_i  \cdot \eins_{\{x_i^*\}}(x)  \, , \\
   \frac{\partial g}{\partial w_i}(x, w,c,a,b) 
&=
(a_i \cdot x  + b_i)\cdot \eins_{(-\infty,x_i^*)}(x)\, .
\end{align*}
Analogously, for $i\in I_+$ we obtain 
\begin{align*}
   \frac{\partial g}{\partial a_i}(x, w,c,a,b) 
&=
w_i \cdot x \cdot \eins_{(x_i^*,\infty)}(x) \, + \partial_0 \cdot w_i \cdot x \cdot \eins_{\{x_i^*\}}(x) \\
   \frac{\partial g}{\partial b_i}(x, w,c,a,b) 
&=
w_i  \cdot \eins_{(x_i^*,\infty)}(x) + \partial_0 \cdot w_i  \cdot \eins_{\{x_i^*\}}(x) \\
   \frac{\partial g}{\partial w_i}(x, w,c,a,b) 
&=
(a_i \cdot x  + b_i)\cdot \eins_{(x_i^*,\infty)}(x)\, .
\end{align*}
Moreover, for $i\in I\setminus I_*$ we  find
\begin{align*}
 \frac{\partial g}{\partial a_i}(x, w,c,a,b) 
&= w_i \cdot \partial_{b_i} \cdot x    \, ,            \\
 \frac{\partial g}{\partial b_i}(x, w,c,a,b) 
&=  w_i \cdot \partial_{b_i}    \, ,            \\
 \frac{\partial g}{\partial w_i}(x, w,c,a,b) 
&= \relu{b_i}\, .
\end{align*} 
Finally, 
we have 
\begin{displaymath}
   \frac{\partial g }{\partial c}(x, w,c,a,b) = 1\, .
\end{displaymath}
Let us now fix an $i\in I_-$. Then, for all samples $(x_j,y_j)$ with 
$x_j > x_i^*$ our formulas above yield
\begin{displaymath}
   \frac{\partial g }{\partial a_i}(x_j, w,c,a,b) = \frac{\partial g }{\partial b_i}(x_j, w,c,a,b)
= \frac{\partial g }{\partial w_i}(x_j, w,c,a,b) = 0\, ,
\end{displaymath}
and consequently, Lemma \ref{result:general-emp-derivative} extended in the sense of 
\eqref{chain-rule-extension-a} and  \eqref{chain-rule-extension-b}
together with our above formulas for the partial derivatives of $g$ for $x_j\leq x_i^*$
shows the first three 
formulas.
Analogously, for 
 $i\in I_+$ and all samples $(x_j,y_j)$ with 
$x_j < x_i^*$ our formulas above yield
\begin{displaymath}
   \frac{\partial g }{\partial a_i}(x_j, w,c,a,b) = \frac{\partial g }{\partial b_i}(x_j, w,c,a,b)
= \frac{\partial g }{\partial w_i}(x_j, w,c,a,b) = 0\, ,
\end{displaymath}
and consequently, we obtain the formulas in the second case.
The remaining assertions follow even more  directly from the extended version of Lemma \ref{result:general-emp-derivative} 
and the  formulas derived for the partial derivatives of $g$.
\end{proofof}

\begin{proofof}{Corollary \ref{result:one-layer-states}}
 \ada i Without loss of generality we may assume that $x_1 = \xmin$, $x_2 = \xmax$ and $x_1 < x_3 < x_2$.
 Since $h_i$ is fully active, we then have $x_1<x_i^* < x_2$.
	By symmetry it suffices to consider the case $i\in I_+$.
% 	and 
% 	by our assumptions we also have a sample $x_3$
% 		with $x_3 \neq x_1$ and $x_3\neq x_2$. 
		Now assume that there were some 
	$\tilde a, \tilde b\in \R$ with 
	\begin{align}\label{result:one-layer-states-h1}
	   h(x_j) = \tilde a x_j + \tilde b\, 
	\end{align}
		for $j=1,2,3$. Since $x_1<x_i^*$ we also have $h_i(x_1) =0$ by 
		\eqref{right-neuron}
		and $x_2>x_i^*$ analogously gives
		\begin{align}\label{result:one-layer-states-h2}
		 h_i(x_2) = a_ix_2 + b_i \neq 0\, .
		\end{align}
Now, if $x_3 \leq x_i^*$, then
			we have 
	$h_i(x_3) = 0$ by \eqref{right-neuron}, and by \eqref{result:one-layer-states-h1} this implies 
		$\tilde a = \tilde b = 0$. Hence \eqref{result:one-layer-states-h1} would give
		$h_i(x_2) =0$, which contradicts \eqref{result:one-layer-states-h2}.
		Moreover, if $x_3 > x_i^*$, then we have $h_i(x_3) = a_ix_3 + b_i $, and by 
		\eqref{result:one-layer-states-h2} we find
		$\tilde a = a_i$ and $\tilde b = b_i$. Equation \eqref{result:one-layer-states-h1}
		then gives $h(x_1) =  a_ix_1 + b_i \neq 0$ since $x_1\neq x_i^*$, 
		which again is a contradiction. 
 
\ada{ii} If $i\in I_-$, 
		then have $x_i^* \geq \xmax$ and hence the assertion follows from 
		\eqref{left-neuron}. The case $i\in I_+$ leads 
		to $x_i^* \leq \xmin$ and \eqref{right-neuron} gives the assertion.

\ada{iii} Let us first assume that $h_i$ is inactive. 
If $i\in I_-$,  we then have $x_i^* \leq \xmin$, and therefore 
		we find $h_i(x_j) = 0$ for all $j=1,\dots,n$ by \eqref{left-neuron}. Moreover, for the subsample $D'$
		 we have 
		$x_{j_l} \geq x_i^*$ for all $l=1,\dots,k$ and therefore 
		the formulas for the gradients follow from Proposition \ref{result:one-layer-deriv}.
		The case $i\in I_+$ can be shown analogously.
		
		Let us now assume that we have $h_i(x_j) = 0$ for all $j=1,\dots,n$. Moreover, $i\in I_*$ 
		ensures $i\in I_- \cup I_+$ and $a_i\neq 0$.
		Now let $i\in I_-$ and assume that $h_i$ was not inactive. By definition, there would then exist
		a sample $x_j$ with $x_j < x_i^*$, and hence \eqref{left-neuron} together with $a_i<0$ gives 
		$h_i(x_j) = a_ix_j+b_i> a_ix_i^*+b_i = 0$. This contradicts $h_i(x_j) = 0$.
		The case $i\in I_+$ can be 
		shown analogously.
\end{proofof}

\begin{proofof}{Lemma \ref{result:one-layer-distribution-knots}}
The first two assertions are obvious. To show the first equation, we note that $h_i$ is fully active, if and only if
   $-b_i / a_i \in (\xmin, \xmax)$, and this is equivalent to   
 $b_i / a_i \in (-\xmax, -\xmin)$. This yields
 \begin{align*}
  P\bigl(\{\mbox{neuron }  h_i \mbox{ is fully active}    \} \bigr)
  &=
   P_b\otimes P_a\Bigl(\bigl\{(b_i,a_i)\in \R^2:   -b_i / a_i  \in (\xmin, \xmax)   \bigr\} \Bigr) \\
%    & =
%     P_w\otimes P_b\Bigl(\bigl\{(a_i,b_i)\in \R^2:  b_i / a_i \not\in [-\xmax, -\xmin]  \bigr\} \Bigr) \\
   & = P_b/P_a \bigl(  (-\xmax, -\xmin)  \bigr) \, .
 \end{align*}
 If $F_{P_b/P_a}$ is  continuous, 
this equation immediately implies \eqref{result:one-layer-distribution-knots-fa}. To establish \eqref{result:one-layer-distribution-knots-sa},
we first note that $h_i$ is semi-active if $a_i <0$ and $-b_i/a_i \geq \xmax$ or if $a_i>0$ and $-b_i/a_i \leq \xmin$.
Now observe that in the case $a_i<0$ the condition $-b_i/a_i \geq \xmax$ is equivalent to 
$b_i \geq - \xmax a_i$, while in the case $a_i>0$ the condition $-b_i/a_i \leq \xmin$ is equivalent to 
$b_i \geq -\xmin a_i$. Consequently, we obtain
 \begin{align*}
  P\bigl(\{\mbox{neuron }  h_i \mbox{ is semi-active}    \} \bigr)
  &=
  P_b\otimes P_a\bigl( \{(b_i,a_i)\in \R^2:  b_i \geq - \xmax a_i  \mbox{ and } a_i < 0 \} \bigr) \\
 &\qquad + P_b\otimes P_a\bigl( \{(b_i,a_i)\in \R^2:  b_i \geq -\xmin a_i  \mbox{ and } a_i > 0 \} \bigr) \\
 & = 
 F_{P_b, P_a}^-(- \xmax) +  P_b\otimes P_a\bigl( \{(b_i,a_i)\in \R^2:  a_i >0  \} \bigr) \\
 &\qquad -  P_b\otimes P_a\bigl( \{(b_i,a_i)\in \R^2:  b_i <  -\xmin a_i  \mbox{ and } a_i > 0 \} \bigr) \\
 & = F_{P_b, P_a}^-(- \xmax)  + P_a([0,\infty))  - F_{P_b, P_a}^+(- \xmin)\, ,
%  & = F_{P_b/P_a}(-\xmax) - F_{P_a, P_b}^+(- \xmax)  + 1 - F_{P_b/P_a}(-\xmin)\\
%  &\qquad- \bigl(F_{P_b/P_a}(-\xmin) - F_{P_a, P_b}^+(- \xmin)  \bigr)
  \end{align*}
  where in the last step we used $P_a(\{0\}) = 0$
  and the continuity of $F_{P_b/P_a}$, which ensures
\begin{displaymath}
 P_b\otimes P_a\bigl( \{(b_i,a_i)\in \R^2:  b_i =  -\xmin a_i  \mbox{ and } a_i > 0 \} \bigr) \leq P_b / P_a (\{-\xmin\}) = 0\, .
\end{displaymath}
 Equation \eqref{result:one-layer-distribution-knots-ia}
  immediately follows from \eqref{basic-ratio-equation}, \eqref{result:one-layer-distribution-knots-fa}, \eqref{result:one-layer-distribution-knots-sa}, and $P_a(\{0\}) = 0$,
  since each neuron $h_i$ is $P$-almost surely either fully active, or semi-active, or inactive.
  
  Finally, the implication ``$\Leftarrow$'' is part of the definition of dead neurons. Conversely, 
   since $F_{P_b/P_a}$ is  continuous, we have $P_b/P_a(\{x_j\}) = 0$ for all $j=1,\dots,n$, and hence part
  \emph{iii)} of Lemma \ref{result:one-layer-states} gives the implication ``$\Rightarrow$''.
\end{proofof}

\begin{proofof}{Theorem \ref{result:pos-bias-better}}
 \atob i{ii} Since $-\xmax\leq 0$ we find by \eqref{result:ration-distributions-h1}
 \begin{align*}
  F_{P_b,P_a}^+(-\xmax) = \int_{(0,\infty)} P_b\bigl((-\infty, -\xmax t]\bigr) \, dP_a(t) = 0
 \end{align*}
 and analogously, $-\xmin\geq 0$ implies 
 \begin{align*}
  F_{P_b,P_a}^-(-\xmin) = \int_{(-\infty,0)} P_b\bigl([ -\xmin t, \infty)\bigr) \, dP_a(t) 
  = \int_{(-\infty,0)} P_b\bigl((0,\infty)\bigr) \, dP_a(t)
  = P_a\bigl((-\infty, 0])\, ,
 \end{align*}
 where in the last step we used $P_a(\{0\}) = 0$. Now \emph{ii)} follows from \eqref{result:one-layer-distribution-knots-ia}.

 \atob {ii} i Let us assume that \emph{i)} is not satisfied. Then 
	we have $P_b((-\infty, 0]) > 0$ and consequently it holds $P_b(\{0\}) > 0$ 
	or 
 there exists a $z_0<0$ such that $F_{P_b}(z) > 0$ for all $z>z_0$. If $P_b(\{0\}) >0$, then 
	\eqref{result:ration-distributions-mass-at-zero} gives $P_b/P_a (\{0\}) > 0$,
% we obviously have $P_b/P_a = \delta_{\{0\}}$,
 and hence $F_{P_b/P_a}$ is not continuous. Since this behavior is excluded in the assumptions of our theorem,  
 it suffices to consider the second case. 
 To this end, we define 
\begin{displaymath}
   t_+ :=
\begin{cases}
   - z_0 /\xmax & \mbox{ if }  \xmax >0 \\
	\infty & \mbox{ if }  \xmax = 0\, .
\end{cases}
\end{displaymath}
Let us fix a $t\in (0,t_+)$. Then we have $-\xmax t > z_0$, and hence we find 
$F_{P_b}(-\xmax t) >0$ as well as
% 
% 
% Now observe that for $t>0$ we have  $-\xmax t > z_0$ 
%  if and only if $t < - z_0 / \xmax$. Moreover, for later we note that for
%  $t\in (0, - z_0 / \xmax)$ we have $F_{P_b}(-\xmax t) >0$, and 
%  using \eqref{result:ration-distributions-h1}
%  we further find
  \begin{align*}
  F_{P_b,P_a}^+(-\xmax) = \int_{(0,\infty)} F_{P_b}(-\xmax t)\, dP_a(t) \geq \int_{(0,t_+)} F_{P_b}(-\xmax t)\, dP_a(t) \, ,
 \end{align*}
where in the first step we used  \eqref{result:ration-distributions-h1}.
Similarly, we define 
\begin{displaymath}
   t_- :=
\begin{cases}
   - z_0 /\xmin & \mbox{ if }  \xmin >0 \\
	-\infty & \mbox{ if }  \xmin = 0\, .
\end{cases}
\end{displaymath}
For $t\in (t_-, 0)$ we then obtain $-\xmin t > z_0$. This yields  
$P_b\bigl((-\infty, -\xmin t)\bigr) > 0$ and, by 
incorporating  \eqref{result:ration-distributions-h1}, also 
% 
%  To bound $F_{P_b,P_a}^-(-\xmin)$ we define $t_0 := -\infty$ if $\xmin =0$ and $t_0 := - z_0 / \xmin$ otherwise.
%  For $t<0$, we then have 
%    $-\xmin t > z_0$ if and only if $t > t_0$. Again, for later use we note that we have 
%    for all $t\in (t_0, 0)$. Now,  \eqref{result:ration-distributions-h1} yields
  \begin{align*}
   F_{P_b,P_a}^-(-\xmin) 
   = \int_{(-\infty,0)} P_b\bigl([ -\xmin t, \infty)\bigr) \, dP_a(t)
   &=   \int_{(-\infty,0)} 1-  P_b\bigl((-\infty, -\xmin t)\bigr) \, dP_a(t)\\
% 	&= P_a((-\infty, 0]) - \int_{(-\infty,0)} P_b\bigl((-\infty, -\xmin t)\bigr) \, dP_a(t)\\
   & \leq  P_a((-\infty, 0]) - \int_{(t_-, 0)} P_b\bigl((-\infty, -\xmin t)\bigr) \, dP_a(t)\, .
  \end{align*}
  Let us fix an $\e>0$ with $\e \leq \min\{t_+, -t_- \}$.
 Plugging both estimates into \eqref{result:one-layer-distribution-knots-ia} we then obtain 
 \begin{align*}
  P (\{\mbox{neuron }  h_i \mbox{ is inactive}    \} ) 
  &= P_a((-\infty, 0]) + F_{P_b,P_a}^+(-\xmax) - F_{P_b,P_a}^-(-\xmin) \\
  &\geq  \int_{(t_-, 0)} P_b\bigl((-\infty, -\xmin t)\bigr) \, dP_a(t) + \int_{(0, t_+)} F_{P_b}(-\xmax t)\, dP_a(t) \\
  & \geq \int_{(-\e,\e)} \min\Bigl\{P_b\bigl((-\infty, -\xmin t)\bigr), F_{P_b}(-\xmax t)  \Bigr\}   \, dP_a(t)\\
  & > 0\, .
 \end{align*}
 In other words, \emph{ii)} does not hold. 
 
 \aquib {iii} {iv} This equivalence can be shown analogously. In addition, note, that for symmetric
$P_a$ it immediately follows from  
considering $(P_b)_-$ 
in the already established equivalence \emph{i)} $\Leftrightarrow$ \emph{ii)}
 in combination with  the formulas
 \eqref{result:one-layer-distribution-knots-sa}, \eqref{result:one-layer-distribution-knots-ia}, and 
 \eqref{result-ratio-distributions-inverse}.
\end{proofof}

\begin{proofof}{Theorem \ref{result:nonzero-bias}}
   We first note that $P_b(\{0\}) = 1$ implies $P_b/P_a = \d_{\{0\}}$, and hence 
   we have $P(\{x_i^* = 0  \}) = 1$. Consequently, we shown both \emph{ii)} $\Rightarrow$ \emph{i)}
	and \emph{iii)} $\Rightarrow$ \emph{i)}, and for data sets satisfying $\xmin < 0<  \xmax$ also 
	\emph{iv)} $\Rightarrow$ \emph{i)} and \emph{v)} $\Rightarrow$ \emph{i)}.

\atob i {ii} By the assumed  $P_b(\{0\}) < 1$ we conclude that there exists a $z_0>0$ with 
		\begin{align}\label{result:nonzero-bias-h1}
		   P((-\infty, -z_0]) + P([z_0,\infty)) >0\, .
		\end{align}
			We define $z := \xmax + 1$ and $\e := z_0 / z$.
			By \eqref{result:ratio-lower-bounds-negz} we then obtain
		\begin{align*}
		   P(\{x_i^* > \xmax  \}) \geq P(\{x_i^* \geq z \}) 
		&= P(\{- x_i^* \leq -z \}) \\
		& \geq P_b\bigl([ z_0, \infty)  \bigr) \cdot P_a\bigl( [-\e,0)\bigr) 
		+ P_b\bigl(- \infty,  - z_0]  \bigr) \cdot P_a\bigl( (0, \e]\bigr) \, ,
		\end{align*}
		and \eqref{result:nonzero-bias-h1} thus yields the assertion.

\atob i {ii} We first note that we again  have \eqref{result:nonzero-bias-h1}. We define 
    $z := \xmin - 1$ and $\e := -z_0 / z$. Using 
\eqref{result:ratio-lower-bounds-posz} we then obtain
		\begin{align*}
		   P(\{x_i^* < \xmin  \}) %\geq P(\{x_i^* \leq z \}) 
		\geq P(\{- x_i^* \geq -z \})  
		 \geq P_b\bigl([  z_0, \infty)  \bigr) \cdot P_a\bigl( (0, \e]\bigr) 
		+ P_b\bigl(- \infty, - z_0]  \bigr) \cdot P_a\bigl( [-\e,0)\bigr)  \, ,
		\end{align*}
  and by  \eqref{result:nonzero-bias-h1} we thus find the assertion.
  
  Finally, the implications \emph{ii)} $\Rightarrow$ \emph{iv)} and \emph{iii)} $\Rightarrow$ \emph{v)} are trivial.
\end{proofof}

\subsection{Proofs for Section \ref{sec:general}}\label{subsec:proof-general}

\begin{proofof}{Lemma \ref{result:ico-char}}
 \atob {ii} i      Let us 
     fix an $x\in x_i^* \cap \ico D$.
    Then we have $h_i(x) = 0$ and, by the definition of $\ico D$,
    there exist $\lb_1,\dots,\lb_n > 0$  with $\lb_1+\dots+\lb_n = 1$ and 
    $x=  \sum_{j=1}^n \lb_j x_j$.  
    Moreover, since $\ico D\not\subset x_i^*$ there exists a $j_1 \in \{1\dots,n\}$
    with $x_{j_1} \not \in x_i^*$, since otherwise the convexity of $x_i^*$ would imply 
    $\ico D \subset \co D \subset x_i^*$.
    Consequently, we have $x_{j_1} \in \Aip\cup \Aim$. Let us first assume that $x_{j_1} \in \Aip$. 
    Then there exists a $j_2 \in \{1\dots,n\}$ with $x_{j_2} \in \Aim$, since otherwise we would 
    find 
    \begin{displaymath}
     0 = h_i(x) = \langle a_i,x\rangle + b = \sum_{j=1}^n \lb_j \bigl( \langle a_i, x_j\rangle +b_i\bigr)
     \geq \lb_{j_1} \bigl(\langle a_i, x_{j_1}\rangle +b_i \bigr)> 0\, .
    \end{displaymath}
    Similarly, if $x_{j_1} \in \Aim$, then there also needs to exist a
    $j_2 \in \{1\dots,n\}$ with $x_{j_2} \in \Aip$, since otherwise we would 
    find 
    \begin{displaymath}
     0 = h_i(x) = \langle a_i,x\rangle + b = \sum_{j=1}^n \lb_j \bigl( \langle a_i, x_j\rangle +b_i\bigr)
     \leq \lb_{j_1} \bigl(\langle a_i, x_{j_1}\rangle +b_i \bigr)< 0\, .
    \end{displaymath}
    Consequently, we have shown the existence of the desired $j_1, j_2\in \{1\dots,n\}$.
    
    \atob i {ii} Clearly,  $\ico D \subset x_i^*$ is impossible, since this would 
    imply $D\subset \co D = \overline {\ico D} \subset \overline{x_i^*} = x_i^*$, which contradicts e.g.~$D\cap \Aip\neq\emptyset$. Therefore, it remains to show $x_i^* \cap \ico D\neq \emptyset$. 
		To this end, we define $D^+ := D\cap \Aip$, $D^- := D\cap \Aim$ and $D^0 := D\cap x_i^*$. 
		Moreover, for $t\in [0,1]$ and $j=1,\dots,n$ we define 
	\begin{displaymath}
	   \lb_j(t) :=
		\begin{dcases*}
		   \frac {1 - t}{|D^+|} & if $x_j \in D^+$\\
	\frac {t}{|D^- \cup D^0|} & if $x_j \in D^-\cup D^0.$\\
		\end{dcases*}
	\end{displaymath}
	It is easy to check that $\lb_1(t) + \dots + \lb_n(t) = 1$ and that 
	$\lb_j(t) \in (0,1)$ whenever $t\in (0,1)$. Let us now consider the function
		\begin{align*}
		   H: [0,1] & \to \R\\
				t & \mapsto  \sum_{j=1}^n \lb_j(t)\cdot \bigl(\langle a_i ,   x_j \rangle + b_i \bigr)
		\end{align*}
		Obviously, the function $H$ is continuous and since  $|D^+| \geq 1$ we further have 
		\begin{displaymath}
		   H(0) =   \frac {1}{|D^+|}  \sum_{x_j\in D^+}  \bigl(\langle a_i ,   x_j \rangle + b_i \bigr) > 0\, .
		\end{displaymath}
		Analogously, $|D^-| \geq 1$  implies
		\begin{displaymath}
		   H(1) =   \frac {1}{|D^- \cup D^0|}  \sum_{x_j\in D^- \cup D^0}  \bigl(\langle a_i ,   x_j \rangle + b_i \bigr)  
			= \frac {1}{|D^- \cup D^0|}  \sum_{x_j\in D^-}  \bigl(\langle a_i ,   x_j \rangle + b_i \bigr)  < 0 \, .
		\end{displaymath}
		The intermediate value theorem then gives a $t^\star\in (0,1)$ with $H(t^\star) = 0$ and for 
		$x^\star := \sum_{j=1}^n {\lb_j(t^\star)\cdot x_j}$ we then find both 
		$x^\star \in \ico D$ and 
		\begin{displaymath}
		   \langle a_i, x^\star\rangle + b_i 
			=    \sum_{j=1}^n \lb_j(t^\star)\cdot \bigl(\langle a_i ,   x_j \rangle + b_i \bigr)
			= H(t^\star) = 0\, .
		\end{displaymath}
		This shows $x^\star \in x_i^*$, which completes the proof.
\end{proofof}

\begin{proofof}{Lemma \ref{result:gen-layer-states}}
 \atob i {ii} 
%     Our first goal is to show that there exists $j_1, j_2\in \{1\dots,n\}$
%     with $x_{j_1} \in \Aip$ and $x_{j_2}\in \Aim$. 
%     
%     To this end, we fix an $x\in x_i^* \cap \ico D$.
%     Then we have $h_i(x) = 0$ and, by the definition of $\ico D$,
%     there exist $\lb_1,\dots,\lb_n > 0$  with $\lb_1+\dots+\lb_n = 1$ and 
%     $x=  \sum_{j=1}^n \lb_j x_j$.  
%     Moreover, since $\ico D\not\subset x_i^*$ there exists a $j_1 \in \{1\dots,n\}$
%     with $x_{j_1} \not \in x_i^*$, since otherwise the convexity of $x_i^*$ would imply 
%     $\ico D \subset \co D \subset x_i^*$.
%     Consequently, we have $x_{j_1} \in \Aip\cup \Aim$. Let us first assume that $x_{j_1} \in \Aip$. 
%     Then there exists a $j_2 \in \{1\dots,n\}$ with $x_{j_2} \in \Aim$, since otherwise we would 
%     find 
%     \begin{displaymath}
%      0 = h(x) = \langle a_i,x\rangle + b = \sum_{j=1}^n \lb_j \bigl( \langle a_i, x_j\rangle +b_i\bigr)
%      \geq \lb_{j_1} \bigl(\langle a_i, x_{j_1}\rangle +b_i \bigr)> 0\, .
%     \end{displaymath}
%     Similarly, if $x_{j_1} \in \Aim$, then there also needs to exist a
%     $j_2 \in \{1\dots,n\}$ with $x_{j_2} \in \Aip$, since otherwise we would 
%     find 
%     \begin{displaymath}
%      0 = h(x) = \langle a_i,x\rangle + b = \sum_{j=1}^n \lb_j \bigl( \langle a_i, x_j\rangle +b_i\bigr)
%      \leq \lb_{j_1} \bigl(\langle a_i, x_{j_1}\rangle +b_i \bigr)< 0\, .
%     \end{displaymath}
%     Consequently, we have shown the existence of the desired $j_1, j_2\in \{1\dots,n\}$.
    Assume that there exist $\tilde a\in \R^d$ and $\tilde b \in \R$ such that 
    for all $j=1,\dots,n$ we have 
    	\begin{align}\label{result:gen-layer-states-h1}
	     \langle \tilde a, x_j\rangle + \tilde b = h_i(x_j) =
	     \begin{cases}
	       \langle a_i, x_j\rangle +b_i & \mbox{ if } x_j \in \Aip\\
	       0 & \mbox{ else.}
	     \end{cases}
	\end{align}
	Since $h_i$ is fully active, we find  $j_1,j_2\in \{1,\dots,n\}$ with $x_{j_1} \in \Aip$
	and $x_{j_2} \in \Aim$, and the additional assumption 
	  $x_{j_0} \in \ico D$ gives us some $\lb_1,\dots,\lb_n > 0$  with $\lb_1+\dots+\lb_n = 1$ and 
    $x_{j_0}=  \sum_{j=1}^n \lb_j x_j$. Hence, 
	  a simple calculation together with \eqref{result:gen-layer-states-h1}   shows
	\begin{align}\label{result:gen-layer-states-h2}
	  \langle \tilde a, x_{j_0}\rangle + \tilde b 
	  = \sum_{j=1}^n \lb_j \bigl( \langle \tilde a, x_j\rangle +\tilde b\bigr) 
	  = \sum_{x_j \in \Aip} \lb_j \bigl( \langle \tilde a, x_j\rangle +\tilde b\bigr) 
	  = \sum_{x_j \in \Aip} \lb_j \bigl( \langle   a_i, x_j\rangle +b_i\bigr)
	  > 0\, ,
	\end{align}
	where in the last step we used that $\Aip \neq \emptyset$, $\lb_j >0$ for all $j=1,\dots,n$, and $\langle   a_i, x_j\rangle +b_i>0$
	for all $x_j\in \Aip$. By \eqref{result:gen-layer-states-h1} we conclude that 
     $x_{j_0} \in \Aip$. Moreover, a combination of 
    \eqref{result:gen-layer-states-h1} and \eqref{result:gen-layer-states-h2} yields
    \begin{displaymath}
      \sum_{j=1}^n \lb_j \bigl( \langle   a_i, x_j\rangle +b_i\bigr)
      = \langle a_i, x_{j_0}\rangle +b_i
      = \langle \tilde a, x_{j_0}\rangle + \tilde b 
      = \sum_{x_j \in \Aip} \lb_j \bigl( \langle   a_i, x_j\rangle +b_i\bigr) \, , 
    \end{displaymath}
    and this implies 
    \begin{displaymath}
     0 = \sum_{x_j \not\in \Aip} \lb_j \bigl( \langle   a_i, x_j\rangle +b_i\bigr)
     \leq  \lb_{j_2}  (\langle   a_i, x_{j_2}\rangle +b_i) < 0\, ,
    \end{displaymath}
    i.e.~we have found a contradiction. 
%     Similarly, in the case $x_{j_0} \not\in \Aip$
%     a combination of 
%     \eqref{result:gen-layer-states-h1} and \eqref{result:gen-layer-states-h2} yields
%     \begin{displaymath}
%       0
%       \geq \langle a_i, x_{j_0}\rangle +b_i
%       = \langle \tilde a, x_{j_0}\rangle + \tilde b 
%       = \sum_{x_j \in \Aip} \lb_j \bigl( \langle   a_i, x_j\rangle +b_i\bigr) 
%       \geq  \lb_{j_1}  (\langle   a_i, x_{j_1}\rangle +b_i) > 0\, , 
%     \end{displaymath}
%     which again is a contradiction. 
    Consequently,  \eqref{result:gen-layer-states-h1} cannot be true.
    
    \atob {ii} i Assume that $h_i$ was not fully active. Then it is either semi-active or inactive,
    but in both cases we have shown in front of Lemma \ref{result:gen-layer-states} that 
    $h_i$ would then behave linearly on $D$.
\end{proofof}

\begin{proofof}{Theorem \ref{result:zero-bias-general}}
 In front of Theorem \ref{result:zero-bias-general}    we have already seen that 
 $h_i$ is inactive if and only if $a_i \in -D^\star$. By the symmetry of $P_a^d$ 
%  established in 
%   Lemma \ref{result:appendix:symm-d} 
  this shows the formula for inactive neurons. Moreover, 
  the formula for fully active neurons follows as soon as we have established the formula for 
  semi-active neurons. To show the latter formula, we first observe that the 
  condition $D\subset x_i^* \cup \Aip$ is equivalent to $a_i \in D^\star$, and hence it suffices to show 
  that 
  \begin{align}\label{result:zero-bias-general-h1}
   P_a^d \bigl( \{a_i :   D\not\subset x_i^*  \}\bigr) = 1\, . 
  \end{align}
    To this end, we first observe that $x_i^*$ is a linear subspace due to our initialization
    $b_i = 0$. Consequently, 
     $D\subset x_i^*$ is equivalent to 
    $\spann D \subset x_i^*$. Moreover,  $\spann D \subset x_i^*$
    is also equivalent to $\langle a_i, x\rangle = 0$ for all $x\in \spann D$, 
    and this condition simply means $a_i \in (\spann D)^\perp$.
    Now, the sample $x_j\neq 0$ ensures $\dim (\spann D)^\perp < d$, which 
    in turn yields $\lb^d((\spann D)^\perp ) = 0$. Since $P_a^d$ is absolutely continuous 
    with respect to $\lb^d$, we conclude that $P_a^d((\spann D)^\perp ) = 0$, and the equivalences 
    discussed previously then lead to \eqref{result:zero-bias-general-h1}.
\end{proofof}

\begin{proofof}{Lemma \ref{result:nonzero-bias-lemma}}
   We first show the inequality for inactive neurons. To this end, we consider  an $a_i\in \Rd$ 
		such that the neuron described by $(a_i, b_+)$ is inactive. Then we have 
\begin{displaymath}
   \langle a_i, x_j\rangle + b_+ \leq 0\, , \qquad \qquad j=1,\dots, n\, .
\end{displaymath}
Since $b_-< b_+$ we then see that $\langle a_i, x_j\rangle + b_- \leq 0$
for all samples $x_j$, and consequently, the neuron described by 
$(a_i, b_-)$ is inactive, too.
This shows the first inequality.

The proof of the second inequality is similar: Indeed, assume that we have an $a_i\in \Rd$ such that the
neuron described by $(a_i, b_-)$ is semi-active. Then, for all samples $x_j$ we have 
\begin{displaymath}
   \langle a_i, x_j\rangle + b_- \geq 0\, ,
\end{displaymath}
 and there is one sample $x_{j_0}$ such that we even have  $\langle a_i, x_j\rangle + b_- > 0$.
Using $b_+>b_-$ we conclude that  $ \langle a_i, x_j\rangle + b_+ > 0$ for all samples $x_j$, and this shows that 
neuron described by $(a_i, b_+)$ is also semi-active.
\end{proofof}

\renewcommand{\thesection}{\Alph{section}}
\renewcommand{\theequation}{\Alph{section}.\arabic{equation}}

\setcounter{section}{0}
\section{Appendix: Ratio Distributions}

The goal of this appendix is to collect some, probably known, results about ratio 
distributions as well as some new results about $F_{P,Q}^-$ and $F_{P,Q}^+$.

We begin with a proposition that collects various structural properties. 
For its formulation we define, for a given probability measure $\nu$ on $\R$,
the probability measure $\nu_-$ on $\R$ by 
\begin{displaymath}
 \nu_-(A) := \mu(-A)\, , \qquad\qquad \mbox{for all measurable $A\subset \R$.}
\end{displaymath}
Obviously,  $\nu$ is symmetric if $\nu_- = \nu$.

% -----------------------------------------------------------------------------------------------------------------------------------------

\begin{proposition}\label{result:ration-distributions}
   Let $P$ and $Q$ be two probability measures on $\R$ with $Q(\{0\}) = 0$.
	Then  it holds
	\begin{align}\label{result:ration-distributions-mass-at-zero}
	   P/Q \bigl( \{0\}\bigr) = P \bigl( \{0\}\bigr) 
	\end{align}
	and, for all $z\in \R$,  we have 
\begin{align}\label{result:ration-distributions-h1}
	F_{P,Q}^-(z)  = \int_{(-\infty,0)} P\bigl([ zt, \infty)\bigr) \, dQ(t)
\qquad \mbox{ and } \qquad
       F_{P,Q}^+(z)  =  \int_{(0,\infty)} P\bigl((-\infty, zt]\bigr) \, dQ(t) 
       \, .
\end{align}
 Moreover, 
the following statements
	are true:
\begin{enumerate}
   \item If $Q$ is symmetric, then $P/Q$ is symmetric and we have 
\begin{align}\label{result:ration-distributions-h2}
   F_{P/Q}(z) 
&=
\begin{dcases*}
   \frac 12 - \int_{(0,\infty)} P\bigl((zt,-zt)\bigr) \, dQ(t) & if $z<0$ \\
	\frac 12 + \int_{(0,\infty)} P\bigl([-zt,zt]\bigr) \, dQ(t)  & if $z\geq 0$\, .
\end{dcases*}
\end{align}
Moreover, for all $z\in \R$ we have 
\begin{align}\label{result-ratio-distributions-inverse}
 F_{P_-/Q}^+(z) & = F_{P/Q}^-(z)\,,   &
  F_{P_-/Q}^-(z) & = F_{P/Q}^+(z)\, , &
   F_{P_-/Q}(z) & = F_{P/Q}(z)\, .
\end{align}
In particular, if, in addition, $P$ is also symmetric, 
then for all $z\in \R$ we have 
\begin{equation}\label{pqp-sym}
F_{P,Q}^-(z)  = F_{P,Q}^+(z)  = \frac 12 F_{P/Q}(z)\, . 
\end{equation}

\item If $P$ and $Q$ are Lebesgue absolutely continuous with densities $f_P$ and $f_Q$, then 
$P/Q$ is Lebesgue absolutely continuous with density
\begin{align}\label{result:ration-distributions-h3}
   f_{P/Q}(z) = \int_\R |t| \,f_P(tz) \, f_Q(t)\, dt\qquad\qquad \mbox{for $\lb$-almost all $z\in \R$.}
\end{align}
\item If $P$ is a Dirac distribution at $b>0$, that is $P=\d_{\{b\}}$, then 
we have both
\begin{align}\label{cdf-pq}
   F_{P/Q}(z) 
&=
\begin{dcases*}
     F_Q(0) -  \lim_{\e\searrow 0} F_Q(bz^{-1} - \e) & if $z<0$  \\
      F_Q(0) & if $z=0$\\
	F_Q(0) + 1 - \lim_{\e\searrow 0} F_Q(bz^{-1} - \e) & if $z> 0$
\end{dcases*}
\end{align}
and
\begin{align}\label{b-pqp}
  F_{P,Q}^+(z)
  & = 
  \begin{cases}
   0 & \mbox{ if } z\leq 0 \\
   1 - \lim_{\e\searrow 0} F_Q(bz^{-1} - \e)& \mbox{ if } z> 0\, .
  \end{cases}
\end{align}
In particular, if $F_Q$ is continuous, then taking $\lim_{\e\searrow 0}$ is superfluous in all three cases, and $F_{P/Q}$
is continuous.
Finally, if 
$Q$ is even Lebesgue absolutely continuous and has a  
density $f_Q$ that is piecewise continuous, then $P/Q$ is Lebesgue absolutely 
continuous and its density is $\lb$-almost surely given by 
\begin{align}\label{pdf-pq}
   f_{P/Q}(z) 
   =
   bz^{-2}  f_Q(bz^{-1}) \, .
\end{align}
\end{enumerate}
\end{proposition}

\begin{proofof}{Proposition \ref{result:ration-distributions}}
We begin by showing \eqref{result:ration-distributions-mass-at-zero}.
To this end, we consider, like in the derivation of \eqref{basic-ratio-equation},
 the projections $\pi_X, \pi_Y:\R^2\to \R$
defined by $\pi_X(x,y) := x$ and $\pi_Y(x,y) := y$. 
With respect to the product measure $P\otimes Q$ on $\R^2$ these projections
 $\pi_X$ and $\pi_Y$ are
independent random variables  and their distributions are $P$ and $Q$.
Using $Q(\{0\}) = 0$ this leads to 
\begin{align*} 
   P/Q \bigl( \{0\}\bigr) 
= P\otimes Q \Bigl( \frac{\pi_X}{\pi_Y} = 0\Bigr)  
= P\otimes Q \bigl( \{ (x,y) \in \R^2 : x/y = 0  \} \bigr)
&= P\otimes Q \bigl( \{ (x,y) \in \R^2 : x  = 0  \} \bigr) \\
&= P \bigl( \{0\}\bigr) \, .
\end{align*}
Moreover, 
for the proof of the first equality in \eqref{result:ration-distributions-h1}, we simply note that 
\begin{align*}
  F_{P,Q}^-(z) =   P\otimes Q \bigl( \{ (x,y) \in \R^2 : x\geq zy \mbox{ and } y<0 \} \bigr) 
& = \int_\R \int_\R \eins_{[zy, \infty)}(x) \eins_{(-\infty, 0)}(y) \, dP(x) dQ(y)\\ 
& =  \int_{(-\infty,0)} P\bigl([ zy, \infty)\bigr) \, dQ(y)\, .
\end{align*}
The second equality can be shown analogously.

\ada i Our first goal is to show \eqref{result:ration-distributions-h2}. 
To this end, we first observe that the symmetry of $Q$ gives
% Since $Q$ is symmetric we first observe that 
\begin{displaymath}
    \int_{(-\infty,0)} P\bigl([ zt, \infty)\bigr) \, dQ(t)  =  \int_{(0,\infty)} P\bigl([ -zt, \infty)\bigr) \, dQ(t) \, .
\end{displaymath}
Let us first consider the case 
 $z<0$. Using \eqref{basic-ratio-equation} and \eqref{result:ration-distributions-h1} we then obtain
\begin{align*}
    F_{P/Q}(z) 
=  F_{P,Q}^+(z)  +  F_{P,Q}^-(z) 
 = \int_{(0,\infty)} \hspace*{-2ex}P\bigl((-\infty, zt]\bigr) +  P\bigl([ -zt, \infty)\bigr) \, dQ(t)
 = \int_{(0,\infty)} \hspace*{-2ex}1 -  P\bigl((zt,-zt)\bigr) \, dQ(t)\, ,
\end{align*}
and since we have $Q([0,\infty)) = 1/2$ we get \eqref{result:ration-distributions-h2} 
in the case $z<0$. The case $z\geq 0$ can be shown analogously, namely
\begin{align*}
    F_{P/Q}(z) 
 = \int_{(0,\infty)} P\bigl((-\infty, zt]\bigr) +  P\bigl([ -zt, \infty)\bigr) \, dQ(t)
 = \int_{(0,\infty)} 1 +  P\bigl([-zt,zt]\bigr) \, dQ(t)\, .
\end{align*} 

For the proof of the symmetry of $P/Q$ we recall that a distribution $\nu$ on $\R$
is  symmetric, if and only if  $\nu((-\infty, -z]) = \nu([z,\infty))$ for all $z>0$,
and the latter is equivalent to
\begin{align}\label{result:ration-distributions-ph1}
F_\nu(-z) + \lim_{\e\searrow 0} F_\nu(z-\e) = 1\, , \qquad \qquad z>0\, .
\end{align}
Now observe that, for $z>0$, the already established \eqref{result:ration-distributions-h2}
and
Beppo Levi's theorem yield
\begin{align*}
  \lim_{\e\searrow 0} F_{P/Q}(z-\e)
 = \frac 12 +
 \lim_{\e\searrow 0} \int_{(0,\infty)} P\bigl([-(z-\e)t,(z-\e)t]\bigr) \, dQ(t) 
=\frac 12 +\int_{(0,\infty)} P\bigl((-zt,zt)\bigr) \, dQ(t) \, ,
\end{align*}
and by combining this with \eqref{result:ration-distributions-h2} applied to $-z$
we find \eqref{result:ration-distributions-ph1} for $\nu= P/Q$. Consequently $P/Q$ is symmetric.

Let us now establish \eqref{result-ratio-distributions-inverse}. We begin with 
 the second equality in \eqref{result-ratio-distributions-inverse}.
To this end, we simply use the already established \eqref{result:ration-distributions-h1} and the symmetry of $Q$ to obtain
\begin{align}\nonumber
F_{P,Q}^+(z) = \int_{(0,\infty)} P\bigl((-\infty, zt]\bigr) \, dQ(t)
 &= 
\int_\R   \eins_{(0,\infty)}(t)\,  P\bigl((-\infty, zt]\bigr)  \,  dQ(t)\\ \nonumber
 &= 
\int_\R   \eins_{(0,\infty)}(-t)\,  P\bigl((-\infty, -zt]\bigr)  \,  dQ(t)\\ \nonumber
 &= 
\int_{(-\infty,0)}     P_-\bigl([zt, \infty)\bigr)  \,  dQ(t)\\ \nonumber
& = 
F_{P_-,Q}^-(z)\, .
\end{align}
Now using $(P_-)_- = P$, we also find the first equality in \eqref{result-ratio-distributions-inverse}, namely 
$F_{P_-/Q}^+  = F_{(P_-)_-/Q}^-  = F_{P/Q}^-$, and using these just established identities in combination
with \eqref{basic-ratio-equation} gives $ F_{P_-/Q} = F_{P/Q}$. Finally, if $P$ is symmetric,
we find $F_{P/Q}^-  = F_{P_-/Q}^+  = F_{P/Q}^+$ and using \eqref{basic-ratio-equation} yet another time
gives  $F_{P,Q}^+  = \frac 12 F_{P/Q}$.

\ada {ii} For a proof we refer to  \cite[Theorem 3.1]{Curtiss41a}, but the assertion 
can also be quickly derived from \eqref{basic-ratio-equation} and  \eqref{result:ration-distributions-h1}.
% Although a proof of \eqref{result:ration-distributions-h3} has already been shown
% in \cite[Theorem 3.1]{Curtiss41a} we briefly present a self-contained one for the sake 
% of completeness. To this end, observe that 
% for $z\in \R$ an application of \eqref{result:ration-distributions-h1} gives
% \begin{align*}
%    P/Q\bigl((-\infty, z]\bigr) 
% &=  \int_0^\infty P\bigl((-\infty, zt]\bigr) \, dQ(t) + \int_{-\infty}^0 P\bigl([ zt, \infty)\bigr) \, dQ(t) \\
% & =  \int_0^\infty \int_{-\infty}^{tz} f_P(s) f_Q(t) \, ds\, dt  + \int_{-\infty}^0 \int_{tz}^{\infty} f_P(s) f_Q(t) \, ds\, dt \\
% & =  \int_0^\infty \int_{-\infty}^{z} t f_P(st) f_Q(t) \, ds\, dt + \int_{-\infty}^0 \int_{z}^{-\infty} t f_P(ts) f_Q(t) \, ds\, dt  \\
% & =  \int_0^\infty \int_{-\infty}^{z} t f_P(st) f_Q(t) \, ds\, dt - \int_{-\infty}^0 \int_{-\infty}^{z} t f_P(ts) f_Q(t) \, ds\, dt \, .
% \end{align*}
% Now the assertion follows by an application of Fubini's theorem and some simple transformations.

\ada {iii} We begin by  showing \eqref{cdf-pq}. In the case $z=0$ we first observe that 
  $P((-\infty, 0]) = 0$ and $P([0,\infty)) = 1$, 
  and hence  \eqref{basic-ratio-equation} together with \eqref{result:ration-distributions-h1}
  gives 
  \begin{align*}
   F_{P/Q}(0) =  \int_{(0,\infty)} P\bigl((-\infty, 0]\bigr) \, dQ(t)  + 
    \int_{(-\infty,0)} P\bigl([ 0, \infty)\bigr) \, dQ(t) 
    = Q\bigl((-\infty, 0)\bigr) = F_Q(0)\, ,
  \end{align*}
    where in the last step we used $Q(\{0\})= 0$.
%   
%   
%   
%   the case $z=0$ immediately follows 
% from \eqref{result:ration-distributions-h1}.
Let us now consider the case $z<0$. For $t>0$ we then find 
\begin{displaymath}
 P\bigl((-\infty,zt]\bigr) = \eins_{[b, \infty)}(zt) = 0\, ,
\end{displaymath}
while for $t<0$ we obtain
$P([zt,\infty)) = \eins_{(-\infty,b]}(zt) = \eins_{[bz^{-1},\infty)}(t)$. By \eqref{basic-ratio-equation} and
\eqref{result:ration-distributions-h1} we thus get 
\begin{displaymath}
 F_{P/Q}(z)  =  \int_{(-\infty,0)} \eins_{[bz^{-1},\infty)}(t) \, dQ(t) 
 = Q\bigl( [bz^{-1}, 0] \bigr) = F_Q(0) -  \lim_{\e\searrow 0} F_Q(bz^{-1} - \e)\, .
\end{displaymath}
Similarly, for $z>0$ and $t>0$ we find 
\begin{displaymath}
 P\bigl((-\infty,zt]\bigr) = \eins_{[bz^{-1}, \infty)}(t)   
\end{displaymath}
while for $z>0$ and $t<0$ we get $P([zt,\infty)) =1$. Together with \eqref{basic-ratio-equation} and \eqref{result:ration-distributions-h1}
this yields
\begin{align*}
    F_{P/Q}(z) =  \int_{(0,\infty)}  \eins_{[bz^{-1}, \infty)}(t)   \, dQ(t) + Q\bigl( ( -\infty, 0]\bigr)
&= Q\bigl( [bz^{-1}, \infty) \bigr) + F_Q(0)\\
&= F_Q(0) + 1 - \lim_{\e\searrow 0} F_Q(bz^{-1} - \e)) \, ,
\end{align*}
and hence we have shown \eqref{cdf-pq}. 

To establish \eqref{b-pqp}, it suffices to observe that 
% Moreover, an easy calculation shows
\begin{align*}
  F_{P,Q}^+(z)
%   & = 
%   Q \bigl(\{ y\in \R: b/y \in (-\infty, z] \mbox{ and } y>0   \} \bigr)\\
   = 
  Q \bigl(\{ y\in \R: b/y \leq  z \mbox{ and } y>0   \} \bigr)
  & = 
  \begin{cases}
   0 & \mbox{ if } z\leq 0 \\
   Q \bigl([ bz^{-1},  \infty)\bigr) & \mbox{ if } z> 0\, .
  \end{cases}
\end{align*}
% and hence we obtain \eqref{b-pqp}.

Let us finally assume that the density $f_Q$ is piecewise
continuous, that is 
\begin{displaymath}
N:= \{t\in \R: f_Q \mbox{ is not continuous at } t\}
\end{displaymath} 
is finite.
Then the fundamental
theorem of calculus shows that $F_Q$ is differentiable on $\R\setminus N$
and $F'_Q(z) = f_Q(z)$ for all $z\not\in N$. Using \eqref{cdf-pq} we consequently find for 
% Consequently, for all 
$z_0\neq 0$ with $bz_0^{-1}\not \in N$ that
\begin{align*}
   F_{P/Q}'(z_0) 
   = - \frac {\partial F_Q(bz^{-1})}  {\partial z} (z_0)
   =
   bz_0^{-2}  f_Q(bz_0^{-1}) \, .
\end{align*}
In particular, $F'_{P/Q}$ is piecewise continuous and  the fundamental
theorem of calculus  thus shows that $F_{P/Q}'$ is a Lebesgue density of $P/Q$.
\end{proofof}

% -----------------------------------------------------------------------------------------------------------------------------------------

The next lemma establishes some simple lower bounds on certain 
ratio probabilities.

\begin{lemma}\label{result:ratio-lower-bounds}
   Let $P$ and $Q$ be two probability measures on $\R$ with $Q(\{0\}) = 0$.
	 Then for all $\e>0$ the following two estimates hold:
\begin{align}\label{result:ratio-lower-bounds-negz}
   P/Q \bigl((-\infty, z]\bigr) 
& \geq P\bigl([-\e z, \infty)  \bigr) \cdot Q\bigl( [-\e,0)\bigr) 
		+ P\bigl(- \infty, \e z]  \bigr) \cdot Q\bigl( (0, \e]\bigr) 
& \mbox{ if } z < 0 \\   \label{result:ratio-lower-bounds-posz}
   P/Q \bigl( [z, \infty)\bigr) 
& \geq P\bigl([\e z, \infty)  \bigr) \cdot Q\bigl( (0, \e]\bigr) 
		+ P\bigl(- \infty, -\e z]  \bigr) \cdot Q\bigl( [-\e,0)\bigr) 
& \mbox{ if } z > 0\, .
\end{align}
\end{lemma}

\begin{proofof}{Lemma \ref{result:ratio-lower-bounds}}
   We begin by showing \eqref{result:ratio-lower-bounds-negz}. To this end, we observe that for 
$z<0$ and $y\in [-\e,0)$ we have $zy \leq -\e z$, and hence we find 
\begin{align*}
    P\otimes Q \bigl(\{(x,y)\in \R^2: x\geq zy \mbox{ and } y < 0   \}    \bigr) 
& \geq  
	P\otimes Q \bigl(\{(x,y)\in \R^2: x\geq zy \mbox{ and } y\in [-\e,0)   \}    \bigr) \\
& \geq  
P\otimes Q \bigl(\{(x,y)\in \R^2: x\geq - \e z \mbox{ and } y\in [-\e,0)   \}    \bigr) \\
& = 
 P\bigl([-\e z, \infty)  \bigr) \cdot Q\bigl( [-\e,0)\bigr) \, .
\end{align*}
Analogously, we obtain for $z<0$ that
\begin{align*}
    P\otimes Q \bigl(\{(x,y)\in \R^2: x\leq zy \mbox{ and } y > 0   \}    \bigr) 
% & \geq  
% 	P\otimes Q \bigl(\{(x,y)\in \R^2: x\leq zy \mbox{ and } y\in (0, \e]   \}    \bigr) \\
& \geq  
P\otimes Q \bigl(\{(x,y)\in \R^2: x\leq  \e z \mbox{ and } y\in (0, \e]   \}    \bigr) \\
& = 
P\bigl(- \infty, \e z]  \bigr) \cdot Q\bigl( (0, \e]\bigr)  \, .
\end{align*}
Combining both estimates with \eqref{basic-ratio-equation} then yields 
\eqref{result:ratio-lower-bounds-negz}. The proof of \eqref{result:ratio-lower-bounds-posz}
is similar, namely, for $z>0$ we have 
\begin{align*}
    P\otimes Q \bigl(\{(x,y)\in \R^2: x\geq zy \mbox{ and } y > 0   \}    \bigr) 
% & \geq  
% 	P\otimes Q \bigl(\{(x,y)\in \R^2: x\leq zy \mbox{ and } y\in (0, \e]   \}    \bigr) \\
& \geq  
P\otimes Q \bigl(\{(x,y)\in \R^2: x\geq  \e z \mbox{ and } y\in (0, \e]   \}    \bigr) \\
& = 
P\bigl([\e z, \infty)  \bigr)  \cdot Q\bigl( (0, \e]\bigr)  
\end{align*}
as well as
\begin{align*}
    P\otimes Q \bigl(\{(x,y)\in \R^2: x\leq zy \mbox{ and } y < 0   \}    \bigr) 
& \geq  
P\otimes Q \bigl(\{(x,y)\in \R^2: x\leq - \e z \mbox{ and } y\in [-\e,0)   \}    \bigr) \\
& = 
P\bigl(- \infty, -\e z]  \bigr) \cdot Q\bigl( [-\e,0)\bigr) \, .
\end{align*}
Combining these two estimates with an equation analogous to \eqref{basic-ratio-equation},
namely 
\begin{align*}
  P/Q \bigl( [z, \infty)\bigr) 
&= P\otimes Q \bigl( \{ (x,y) \in \R^2 : x\geq zy \mbox{ and } y>0 \} \bigr)\\
 &\qquad + P\otimes Q \bigl( \{ (x,y) \in \R^2 : x\leq zy \mbox{ and } y<0 \} \bigr) \, ,
\end{align*}
then yields \eqref{result:ratio-lower-bounds-posz}.
\end{proofof}

% -----------------------------------------------------------------------------------------------------------------------------------------

\begin{example}[Normal Ratios]\label{example:normal-ratio}
  For $\s_P, \s_Q>0$ consider the distributions  $P:=\ca N(0,\s_P^2)$ and $Q: = \ca N(0,\s_Q^2)$.
Then we have 
\begin{displaymath}
   f_{P/Q}(z) = \frac{\s_P\cdot \s_Q}{\pi} \cdot \frac{1}{\s_Q^2 z^2+ \s_P^2}\, , \qquad \qquad z\in \R,
\end{displaymath}
see e.g.~\cite[Equation (3.3)]{Curtiss41a}. In other words, $P/Q$ is the Cauchy distribution with 
location parameter $x_0 = 0$ and scale parameter $\g = \s_P /\s_Q$.
This leads to 
\begin{displaymath}
   F_{P/Q}(z) = \frac 1 \pi \arctan\Bigl(\frac{\s_Q\cdot z}{\s_P} \Bigr) + \frac 12 \, , \qquad \qquad z\in \R.
\end{displaymath}
Finally, by the symmetry of $\ca N(0,\s_P^2)$ and  $\ca N(0,\s_Q^2)$ and  Equation
\eqref{pqp-sym} we find 
$F_{P,Q}^-(z)  = F_{P,Q}^+(z)  = \frac 12 F_{P/Q}(z)$ for all $z\in \R$.
\end{example}

% -----------------------------------------------------------------------------------------------------------------------------------------

\begin{example}[Normal Denominators]\label{example:normal-denominator}
 For $b>0$ and $\s>0$ consider the distributions   $P:=\d_{\{b\}}$ and 
 $Q: = \ca N(0,\s^2)$. Since $Q$ has the Lebesgue density $f_Q$ given by
 $f_Q(t) = \frac{1}{\sqrt{2\pi\s^2}} \exp(-\frac{t^2}{2\s^2})$ for $t\in \R$,
 we  then find using   \eqref{pdf-pq} that
\begin{align*}
   f_{P/Q}(z) 
   =
   \frac{b}{\sqrt{2\pi\s^2}\, z^2} \exp\Bigl(-\frac{b^2}{2\s^2 z^2}\Bigr) 
%    = 
%     \frac{\a}{\sqrt{\pi}\, z^2} \exp\Bigl(-\frac{\a^2}{ z^2}\Bigr) 
   \, , \qquad \qquad z\neq 0.
\end{align*}
% where $\a:= \frac b{\sqrt{2}\, \s}$. 
Note that this density has two modes at $\pm \frac b{\sqrt{2}\, \s}$ 
and it vanishes at $0$. Furthermore, part \emph{iii)} of Proposition \ref{result:one-layer-distribution-knots}
shows that 
\begin{align*}
   F_{P/Q}(z) 
&=
\begin{dcases*}
     \frac 12 - \Phi\Bigl(\frac b{\s z}\Bigr)  & if $z<0$  \\
      \frac 12 & if $z=0$\\
    \frac 12  + \Phi\Bigl(-\frac b{\s z}\Bigr)  & if $z> 0$\, , \\
\end{dcases*}\\
  F_{P,Q}^+(z)
  & = 
  \begin{dcases*}
   0 & if  $z\leq 0$ \\
   1 - \Phi\Bigl(\frac b{\s z}\Bigr) & if  $z> 0$\, ,
  \end{dcases*}
\end{align*}
where $\Phi$ denotes the cumulative distribution function of the standard normal distribution $\ca N(0,1)$.
\end{example}

% -----------------------------------------------------------------------------------------------------------------------------------------

\begin{example}[Asymmetric Uniform Ratios]\label{example:asym-uniform-ratio}
For $\b>0$ and $\a>0$ consider the distributions   $P:=\ca U[0,\b]$ and $Q:=\ca U[-\a,\a]$.
Plugging their densities $f_P = \frac 1 \b\eins_{[0,\b]}$ and  $f_Q= \frac 1 {2a} \eins_{[-a,a]}$ into 
\eqref{result:ration-distributions-h3} gives
\begin{align}\label{uniform-ratio-dens}
 f_{P/Q}(z) =  \frac 1 {4\a \b}  \min\bigl\{\a^2, \b^2z^{-2}   \bigr\} 
 =
 \begin{dcases*}
     \frac \a{4\b} & \mbox{if} $z\in \bigl[ -\frac \b\a, \frac \b\a  \bigr]$\\
	\frac \b{4\a}\cdot z^{-2}  & \mbox{if} $z<-\frac \b\a$ \mbox{ or } $z> \frac \b\a$\, ,
\end{dcases*}
\end{align}
and by integrating this density we obtain
\begin{align}\label{uniform-ratio-cdf}
 F_{P/Q}(z)
 =
 \begin{dcases*}
 - \frac \b{4\a z} & \mbox{if}  $z\leq-\frac \b\a$\\
    \frac {2\b+\a z}{4\b} & \mbox{if} $z\in \bigl[ -\frac \b\a, \frac \b\a  \bigr]$\\
	1- \frac \b{4\a z}  & \mbox{if}   $z\geq \frac \b\a$\, .
\end{dcases*}
\end{align}
Finally, for $z\leq 0$, Equation \eqref{result:ration-distributions-h1} immediately shows $F_{P_b,P_a}^+(z) = 0$, and for 
$z>0$ we obtain
\begin{align*}
 F_{P_b,P_a}^+(z) 
 = \frac 1{2\a\b}\int_0^\a  \int_0^{zt \wedge \b} 1\, ds\, dt
 & =  \frac 1{2\a\b}\int_0^\a  \min\{zt, \b\}\,  dt \, .
\end{align*}
Now, in the case $z\leq \frac \b\a$ we have $zt \leq \b$ for all $t\in [0,\a]$, and hence we find $F_{P_b,P_a}^+(z) = \frac\a{4\b}z$, while
in the case $z\geq \frac \b\a$, we obtain
\begin{align*}
 F_{P_b,P_a}^+(z) 
  =  \frac 1{2\a\b}\int_0^{\b z^{-1}}  zt\,  dt + \frac 1{2\a\b} \int_{\b z^{-1}}^\a \b \, dt
  = \frac 1{2\a\b} \cdot \frac {\b^2}{2 z} + \frac \b{2\a\b}\Bigl(\a-\frac\b z   \Bigr)
  = \frac 12 -   \frac {\b}{4 \a z} \, .
\end{align*}
Summarizing these calculation, we have found
\begin{displaymath}
 F_{P/Q}^+(z)
 =
 \begin{dcases*}
 0& \mbox{if}  $z\leq 0$\\
    \frac\a{4\b}z & \mbox{if} $z\in \bigl[0, \frac \b\a  \bigr]$\\
	\frac 12 -   \frac {\b}{4 \a z}  & \mbox{if}   $z\geq \frac \b\a$\, .
\end{dcases*}
\end{displaymath}
\end{example}

% -----------------------------------------------------------------------------------------------------------------------------------------

\begin{example}[Symmetric Uniform Ratios]\label{example:sym-uniform-ratio}
For $\b>0$ and $\a>0$ consider the distributions   $P:=\ca U[-\b,\b]$ and $Q:=\ca U[-\a,\a]$.
Then an easy calculation shows that the density $f_{P/Q}$ coincides with 
\eqref{uniform-ratio-dens}, and therefore $F_{P/Q}$ coincides with \eqref{uniform-ratio-cdf}.
Unlike in Example \ref{example:asym-uniform-ratio}, however, the distributions $P$ and $Q$ are now symmetric,
and therefore part \emph{i)} of Proposition \ref{result:ration-distributions} shows that $F_{P,Q}^+= \frac 1 2F_{P/Q}$.
\end{example}

% -----------------------------------------------------------------------------------------------------------------------------------------

\begin{example}[Uniform Denominators]\label{example:uniform-denominator}
 For $b>0$ and $\a>0$ consider the distributions   $P:=\d_{\{b\}}$ and $Q:=\ca U[-\a,\a]$. Then
 the Lebesgue density $f_Q$ of $Q$ is given by $f_Q= \frac 1 {2a} \eins_{[-a,a]}$ and
by  \emph{iii)} of Proposition \ref{result:one-layer-distribution-knots} 
 therefore we obtain  that
 \begin{align*}
   f_{P/Q}(z) 
&=
\begin{dcases*}
     0 & \mbox{if} $z\in \bigl[ -\frac b\a, \frac b\a  \bigr]$\\
	\frac b{2\a}\cdot z^{-2}  & \mbox{if} $z<-\frac b\a$ \mbox{ or } $z> \frac b\a$\, .
\end{dcases*}
\end{align*}
Moreover, using $F_Q(z) = \max\{0, \min\{1, \frac{\a+z}{2\a}\}\}$,
part \emph{iii)} of Proposition \ref{result:one-layer-distribution-knots}
further yields
\begin{align*}%\label{uniform-denom-cdf}
 F_{P/Q}(z)
 &=
 \begin{dcases*}
 - \frac b{2\a z} & \mbox{if}  $z\leq-\frac b\a$\\
    \frac 12 & \mbox{if} $z\in \bigl[ -\frac b\a, \frac b\a  \bigr]$\\
	1- \frac b{2\a z}  & \mbox{if}   $z\geq \frac b\a$\\
\end{dcases*}\\ \nonumber
  F_{P,Q}^+(z)
  & = 
  \begin{dcases*}
   0 & if  $z\leq \frac b\a$ \\
   1 - \frac b{2\a z} & if  $z\geq \frac b\a$\, .
  \end{dcases*}
\end{align*}
\end{example}

\section{Appendix: Geometric set operations}\label{appendix:geometry}

Given a set $A\subset \Rd$, the convex hull $\ca A$ of $A$ is the smallest convex set
containing $A$. If $A$ is finite, say 
 $A=\{y_1,\dots,y_k\}$, then $\ca A$ can be computed by 
\begin{displaymath}
 \co A = \biggl\{ y\in \R^d: \exists  \lb_1,\dots,\lb_k \geq 0 \mbox{ with } \lb_1+\dots+\lb_k = 1 \mbox{ and } y=  \sum_{j=1}^k \lb_j y_j   \biggr\}\, .
\end{displaymath}
Moreover, a set $K\subset \Rd$ is called a cone, if for all $x\in K$ and $\a >0$ we 
have $\a x\in K$. It is well known, see e.g.~\cite[page 30]{MaTi03} that a cone $K$ is convex, if and only if $K+K\subset K$. Moreover, the conic hull $\coni A$ of a set $A$ is the smallest convex cone containing $A$. If $A$ is of the form 
 $A=\{y_1,\dots,y_k\}$, then $\coni A$ can be computed by 
\begin{displaymath}
 \coni A = \biggl\{ y\in \R^d: \exists  \lb_1,\dots,\lb_k \geq 0 \mbox{ with }  y=  \sum_{j=1}^k \lb_j y_j   \biggr\}\, ,
\end{displaymath}
and for such sets $A$, both $\co A$ and $\coni A$ are closed.
Moreover, for all $A\subset \Rd$ we have $\co A \subset \coni A$.
In addition, given an $A\subset \Rd$, 
the dual cone of $A$  is defined 
by 
\begin{displaymath}
 A^\star := \bigl\{ y\in \Rd: \langle y, x\rangle \geq 0 \mbox{ for all } x\in A\bigr\}\, .
\end{displaymath}
Obviously, $A^\star$
 is always a convex, closed cone. Moreover, we obviously have $(\coni A)^\star \subset A^\star$,
and since the converse implication is also straightforward, we actually have $(\coni A)^\star = A^\star$.
Furthermore, $A^{\star \star} = A$ if and only if $A$ is a closed convex cone,
see e.g.~\cite[page 43]{MaTi03}, and by combining these 
results we find $A^{\star \star} = (\coni A)^{\star \star}= \coni A$ for arbitrary sets $A\subset \Rd$.
Finally, it is straightforward to check that $([0,\infty)^d)^\star = [0,\infty)^d$, cf.~also \cite[page 91]{MaTi03}.
The next lemma, which is probably well-known but could not be found in the literature,
 characterizes the finite sets $A$ satisfying $\coni A = [0,\infty)^d$.

\begin{lemma}\label{result:full-cone-char}
Let $A = \{x_1,\dots,x_n\}$ be a finite set with $A\subset [0,\infty)^d$. Then 
the following statements are equivalent 
\begin{enumerate}
 \item We have $\coni A = [0,\infty)^d$.
 \item For all $k=1,\dots,d$ there exist $j_k\in \{1,\dots,n\}$ and $\a_k\geq 0$
such that $\a_k x_{j_k} = e_k$.
\end{enumerate}
\end{lemma}

\begin{proofof}{Lemma \ref{result:full-cone-char}}
 We first observe that \emph{i)} holds if and only if for all 
 $y\in [0,\infty)^d$ there exist $\lb_1,\dots,\lb_n\geq 0$ such that 
 $y= \lb_1x_1+ \dots + \lb_nx_n$.  Since for $y\in [0,\infty)^d$ there also exist $\mu_1,\dots,\mu_d\geq 0$ with 
 $y=\mu_1 e_1 + \dots+\mu_d e_d$, we conclude that \emph{i)}  holds if and only if
 the following condition is satisfied:
 
 \vspace*{2ex}
 \textbf{EC)\/} For all $k=1,\dots,d$ there exist $\a_{k,1},\dots,\a_{k,n}\geq 0$ with $e_k = \a_{k,1} x_1 + \dots+ \a_{k,n} x_n$.
 
 \vspace*{2ex}
 \noindent
 \atob {ii} i Clearly, \emph{ii)} implies Condition \textbf{EC)\/}, and this implies \emph{i)} as previously discussed.
 
 \noindent
 \atob i {ii} It suffices to show that \textbf{EC)\/} implies \emph{ii)}. To this end, we denote
   the $l$-th coordinate of $x_j$ by $x_{j,l}$, that is $x_j = (x_{j,1},\dots,x_{j,d})$. 
   Moreover, we fix a $k\in \{1,\dots,d\}$ and by  \textbf{EC)\/} we 
   choose  $\a_{k,1},\dots,\a_{k,n}\geq 0$ such that 
   \begin{align}\label{result:full-cone-char-h2}
    e_k = \a_{k,1} x_1 + \dots+ \a_{k,n} x_n\, .
   \end{align}
   Considering \eqref{result:full-cone-char-h2}
    for the $l$-th coordinate with $l\neq k$ then gives
 \begin{align}\label{result:full-cone-char-h1}
  0 = \sum_{j=1}^n \a_{k,j} x_{j,l}\, .
 \end{align}
Now, our assumptions guarantee  $\a_{k,j} x_{j,l} \geq 0$, and hence \eqref{result:full-cone-char-h1}
implies $a_{k,j} = 0$ for all $j\in \{1,\dots, n\}$ with $x_{j, l} > 0$.
Consequently, in the $k$-th coordinate, \eqref{result:full-cone-char-h2} reduces to 
\begin{displaymath}
 1 = \sum_{j\in J} \a_{k,j} x_{j,l}\, ,
\end{displaymath}
where $J := \{j: x_{j,l} = 0 \mbox{ for all } l\neq k\}$. However, this equation implies that there 
is a $j\in J$ with $\a_{k,j} x_{j, k} > 0$. The choice $\a_k := x_{j,k}^{-1}$ then yields 
$\a_k x_{j_k} = e_k$.
\end{proofof}

The conic hull of an  $A\subset \Rd$ is the smallest convex cone containing $A$. If one is only interested 
in the smallest cone containing $A$, one needs to consider the ray of $A$, which is 
\begin{displaymath}
   \ray A:= \bigl\{tx: x\in A,\, t> 0\bigr\} \, .
\end{displaymath}
The next simple lemma, which is needed for Lemma \ref{result:projection-measure}, shows that 
$\ray A$ is measurable if $A\subset \Sd$ is measurable.

\begin{lemma}\label{result:measurable-ray}
 Let $A\subset \Sd$ be measurable. Then $\ray A$ is also measurable.
\end{lemma}

\begin{proofof}{Lemma \ref{result:measurable-ray}}
%  Obviously, it suffices to show that $\ray A \setminus \{0\}$ is measurable. 
%  To this end, 
 We
 define $\phi: (0,\infty) \times \Rd\setminus \{0\}\to \Rd$ by $\phi(t,x) := tx$.
 Then the pre-image of $A$ under $\phi$ is given by 
 \begin{align}\label{result:measurable-ray-h1}
  \phi^{-1}(A) := \bigl\{  \bigl(\snorm x_2^{-1}, x\bigr) : x\in \ray A     \bigr\}\, .
 \end{align}
Indeed, if we have an $(t,x) \in  \phi^{-1}(A)$, then $tx = \phi(t,x) \in A$ implies both $x\in t^{-1} A \subset \ray A$
and $\snorm{tx} = 1$. This shows the inclusion ``$\subset$''. 
Conversely, if $x\in \ray A$, then there exist $t>0$ and $y\in A$ with $x = ty$.
This implies $t^{-1} \snorm x_2 = \snorm y_2 = 1$, and hence $t = \snorm x_2$.
The latter gives  $\snorm x_2^{-1}\cdot  x = y \in A$.

Let us now consider the map $\psi:\Rd\setminus \{0\} \to (0,\infty) \times \Rd\setminus \{0\}$
defined by $\psi(x) := (\snorm x_2^{-1}, x)$. Clearly, $\psi$  injective, and therefore 
\eqref{result:measurable-ray-h1} yields $\psi^{-1}(\phi^{-1}(A)) = \ray A$.
Now the measurability of $A$, $\phi$, and $\psi$ shows that $ \ray A$ is measurable.
\end{proofof}

\begin{lemma}\label{result:ray-algebra}
 For all $A, B\subset \Sd$ the following statements are true:
 \begin{enumerate}
  \item If $A\cap B = \emptyset$, then $\ray A\cap \ray B = \emptyset$.
  \item We have $\ray (A\cup B) = \ray A \cup \ray B$.
 \end{enumerate}
\end{lemma}

\begin{proofof}{Lemma \ref{result:ray-algebra}}
 \ada i Let us assume that there is a $z \in \ray A\cap \ray B$. Then we find some
 $x\in A$, $y\in B$, $s,t>0$ such that $z = sx$ and $z = ty$. This implies 
 $\snorm z_2 = |s| \cdot \snorm x_2 = s$ and $\snorm z_2 = |t| \cdot \snorm y_2 = t$, and consequently,
 we find $x = s^{-1} z = s^{-1} t y = y$. In other words, we have $x\in A\cap B$.
 
 \ada {ii} The inclusion ``$\subset$'' is essentially trivial, and the converse inclusion follows from 
 $A\subset A\cup B$ and  the monotonicity of $\ray (\cdot)$.
\end{proofof}

\section{Appendix: Some function classes}\label{appendix:function-classes}

\begin{definition}
   A function $f:\Rd \to \R^m$ is called positively homogeneous, if for all $\a>0$ and all $x\in \Rd$ we have 
\begin{displaymath}
   f(\a x) = \a f(x)\, .
\end{displaymath}
\end{definition}

Obviously, every linear function $f:\Rd \to \R^m$ is   positively homogeneous. Moreover, every norm
$\snorm \cdot :\Rd \to \R$ is  positively homogeneous, and $\relu\cdot:\R\to \R$ is also  positively homogeneous.
Combining these examples with the help of the following trivial lemma gives a wealth of 
positively homogeneous functions.

\begin{lemma}\label{result:pos-homo-permanent}
   Let $f, g:\Rd \to \R^m$,  $h:\R^m \to \R^k$, and $f_1,\dots,f_m:\Rd \to \R$ be    positively homogeneous functions and $\lb\in \R$. 
Then the following functions
are also  positively homogeneous:
\begin{align*}
   \lb f + g & : \Rd \to \R^m \\
		h\circ f & : \R^m \to \R^k\\
	(f_1,\dots,f_m) &:\Rd \to \R^m\, .
\end{align*}
% Finally, if $g_i:\Rd\to \R^m$ are positively homogeneous and there exists a function $g^*:\Rd\to \R^m$ 
\end{lemma}

The next simple lemma shows that positively homogeneous   functions vanish at the origin.

\begin{lemma}\label{result:pos-homo-zero}
   Let $f :\Rd \to \R^m$  be  a  positively homogeneous   function. Then we have 
		$f(0) = 0$.
\end{lemma}

\begin{proofof}{Lemma \ref{result:pos-homo-zero}}
   For $x = 0$ and $\a= 2$ we have $f(0) = f(2 x) = 2f(x) = 2 f(0)$. This implies  $f(0) = 0$.
\end{proofof}

In the following, we denote the set of positively homogeneous, $\R$-valued  functions $f:\Rd\to \R$
by ${\cal PH} (\Rd)$
Moreover, for a subset $X\subset \Rd$ we define 
\begin{displaymath}
 {\cal PH} (X)  := \bigl\{  f:X\to \R\, \bigl|\,   \exists\,  g\in {\cal PH}(\Rd) \mbox{ such that } g_{|X} = f\bigr\}\, .
\end{displaymath}
The next result will be used to show that ${\cal PH} (X)$ is a closed subset of some commonly considered spaces.

% \begin{lemma}\label{result:pos-homo-closed-lemma}
%  Let $X\subset \Rd$ be  subset and   $(f_n)\subset {\cal PH} (X)$
%  be a sequence   for which there exists an $N\subset X$ 
%  such that for all $x\in X\setminus N$ the following limit exists
%  \begin{displaymath}
%  f(x) :=  \lim_{n\to \infty}  f_n(x) \, .
%  \end{displaymath}
% Then there exists a function $f^*\in {\cal PH} (X)$
%  with $f^*(x) = f(x)$ for all  $x\in X\setminus N$. 
% \end{lemma}
% 
% 
% 
% \begin{proofof}{Lemma \ref{result:pos-homo-closed-lemma}}
%  Without loss of generality we may assume that $(f_n(x))$ does not converge    for all $x\in   N$.
%  Let us pick $g_n\in {\cal PH}(\Rd) \mbox{ such that } (g_n)_{|X} = f_n$ for all $n\geq 1$.
% %  Moreover, by Lemma \ref{result:pos-homo-zero} we now that $f_n(0 ) = 0$ for all $n\geq 0$ provided that $0\in X$.
% %  Consequently, we have $0\not\in N$.
%  Our first goal is to show 
%  \begin{displaymath}
%   X\cap \ray N = N\, .
%  \end{displaymath}
%  Obviously, it suffices to check $X \cap \ray N \subset N$. To this end, we fix an $x\in X\cap  \ray N$. Then there exist
%  $t>0$ and $y\in N$ with $x = ty$. Now the homogeneity ensures $f_n(x) = g_n(x) = tg_n(y) =   t f_n(y)$, and since 
%  $(f_n(y))$ does not converge, we conclude that $(f_n(x))$  does not converge, that is $x\in N$. 
%  
%  For $x\in  \ray X \setminus \ray N =  \ray (X\setminus N)$ 
% \end{proofof}

\begin{lemma}\label{result:pos-homo-closed-lemma}
 Let $X\subset \Rd$ be  subset and   $(f_n)\subset {\cal PH} (X)$
 be a sequence  such that there exists a function $f:X\to \R$ with 
 \begin{displaymath}
 f(x) =  \lim_{n\to \infty}  f_n(x)  
 \end{displaymath}
 for all $x\in X$.
Then we have $f\in {\cal PH} (X)$.
\end{lemma}

\begin{proofof}{Lemma \ref{result:pos-homo-closed-lemma}}
 Let us pick $g_n\in {\cal PH}(\Rd) \mbox{ such that } (g_n)_{|X} = f_n$ for all $n\geq 1$.
 For $y\in \ray X$ we then find some $t>0$ and $x\in X$ with $y = tx$. This gives 
 \begin{displaymath}
  g_n(y) = t g_n(x) = t f_n(x) \to t f(x)\, .
 \end{displaymath}
 Since this holds for all possible such representations $y = tx$, 
 while the limit $\lim_{n\to \infty} g_n(y)$ is unique,
 we conclude that setting $g(y) := t f(x)$
  is actually
    independent of the chosen representation $y = tx$ of $y$. 
%     To this end, let us assume that we have 
%     another $\tilde t>0$ and $\tilde x\in X$ with $y = \tilde t \tilde x$.
%     Then we have 
%     \begin{displaymath}
%      \tilde t f(\tilde x) = \lim_{n\to \infty} \tilde t f_n(\tilde x) = \lim_{n\to \infty} \tilde t g_n(\tilde x) = \lim_{n\to \infty}  g_n(y) =  t f(x)\, .
%     \end{displaymath}
%     Consequently, the definition of $g(y)$ is indeed independent of the chosen representation, 
    Now, this definition
     ensures $g_n(y) \to g(y)$ for all $y\in \ray X$. For $y\in \ray X$ and $\a>0$ we thus  have 
    \begin{displaymath}
     g(\a y) = \lim_{n\to \infty} g_n(\a y) = \a \lim_{n\to \infty} g_n(  y) = \a g(y)\, .
    \end{displaymath}
    Let us define $g(y) := 0$ for all $y\not\in \ray X$. Since for such $y$ we have 
    $g(\a y) = 0 = \a g(y)$ for all $\a > 0$, we then see that $g$ is positively homogeneous.
    Moreover, for $x\in X$ we have 
    \begin{displaymath}
     g(x) = \lim_{n\to \infty} g_n(x) = \lim_{n\to \infty} f_n(x) = f(x)\, ,
    \end{displaymath}
 and thus we find
    $g_{|X} = f$. This gives $f\in {\cal PH} (X)$.
\end{proofof}

\begin{corollary}\label{result:pos-homo-closed-cx}
 Let $X\subset \Rd$ be a compact subset. Then ${\cal PH}(X) \cap C(X)$ is a closed
 subspace of $C(X)$.
\end{corollary}

\begin{proofof}{Corollary \ref{result:pos-homo-closed-cx}}
 Let us fix a sequence $(f_n) \subset {\cal PH}(X) \cap C(X)$ and an $f\in C(X)$ with $\inorm{f_n-f} \to 0$.
 Then Lemma \ref{result:pos-homo-closed-lemma} shows $f\in {\cal PH}(X)$ and hence we have $f\in {\cal PH}(X) \cap C(X)$.
\end{proofof}

\begin{corollary}\label{result:pos-homo-closed-lp}
 Let $P$ be a probability measure on $\Rd$.
%  that is absolutely continuous with respect to the Lebesgue measure $\lbd$.
 Then, for all $p\in [1,\infty)$, the space 
   ${\cal PH}(\Rd) \cap \Lx pP$ is a closed
 subspace of $\Lx pP$.
\end{corollary}

\begin{proofof}{Corollary \ref{result:pos-homo-closed-lp}}
 Let us fix a sequence $(f_n) \subset {\cal PH}(\Rd) \cap \Lx pP$ and an $f\in \Lx pP$ with $\snorm{f_n-f}_{\Lx pP} \to 0$.
 By considering a subsequence we may assume without loss of generality that there exists a measurable
 $N\subset \Rd$ with $P(N) = 0$ and $f_n(x) \to f(x)$ for all $x\in X:= \Rd\setminus N$. 
 Lemma \ref{result:pos-homo-closed-lemma} then shows $f_{|X}\in {\cal PH}(X)$, and  thus we can pick a  $g\in {\cal PH}(\Rd)$
 with $g_{|X} = f_{|X}$.
 Since $P(X) = 1$ we conclude both $\snorm{f_n-g}_{\Lx pP} \to 0$ and $g\in \Lx pP$.
\end{proofof}

\section{Appendix: Some specific distributions}

Recall, that for  $a\in \R$ 
 the incomplete gamma function is defined by 
$$
\Gamma (a,x) := \int_x^\infty \eul^{-t} \, t^{a-1} dt\, , \qquad\qquad x\geq 0.
$$
Note that $\Gamma (a,x)<\infty$ for all $x>0$, and for $a>0$ 
 the gamma function 
$\Gamma (a) := \Gamma(a,0)$ is also finite. Our first result 
provides bounds on certain ratios of the gamma function.

\begin{lemma}\label{result:gautschi}
 For all $d\geq 1$ we have 
\begin{displaymath}
   \sqrt {d-1/2}  \leq   \frac {\sqrt 2 \cdot \Gamma(\frac {d+1} 2) }{\Gamma(\frac d2)} \leq \sqrt {d- 1/4}\, .
\end{displaymath}
\end{lemma}

\begin{proofof}{Lemma \ref{result:gautschi}}
   We first note that in \cite{Kershaw83a} the following refinement of 
Gautschi's inequality, see \cite{Gautschi59a}, has been established:
\begin{displaymath}
	\sqrt{x + 1/4} \,  \leq\,   
\frac {\Gamma(x+1)}{\Gamma(x + 1/2)} 
\, \leq \, 
\sqrt{x + \frac{\sqrt 3 -1}{2}  }\, , \qquad \qquad x>0.
\end{displaymath}
Moreover, using $\G(1) = 1$ and $\G(1/2) = \sqrt\pi$ one easily checks that this inequality also holds for 
$x=0$.
Considering $x :=   d/2 - 1/2$ and using $\sqrt 3 - 2 \leq - 1/4$ for the upper bound
then gives the assertion.
\end{proofof}

The gamma function can also be approximated by a generalization of Stirling's formula. Namely we have 
\begin{align}\label{gamma-stirling}
   \Gamma(a) =  \sqrt{\frac{2\pi}{a}} \cdot \Bigl(\frac{a}{\eul}  \Bigr)^{a} \cdot \eul^{\mu(a)}\, ,
\end{align}
where $0<\mu(a) < \frac 1 {12a}$.
It is also well-known that for fixed $a>0$ and $x\geq a$ the incomplete gamma function satisfies
\begin{displaymath}
\min\{1,a\} x^{a-1} \eul^{-x} \leq \G(a,x) \leq \max\{1,a\} x^{a-1} \eul^{-x}\, ,
\end{displaymath}
see
e.g.~\cite[Lemma A.1.1]{StCh08}. Note that for $a>1$, $\d>0$, and   $x = (1+\d)a$, the upper bounds 
reads as 
\begin{align}\label{result:incomplete-gamma-bound-upper}
 \G(a, (1+\d)a) \leq a^a (1+\d)^{a-1}  \eul^{- (1+\d)a}\, .
\end{align}
The following lemma presents an alternative upper bound on $\G(a,x)$, which improves the last inequality.
In \cite{BoCh09a}, it is attributed to 
\cite{NaPa00a}, but since the latter article is difficult to obtain, we present 
the short proof for the sake of completeness.

\begin{lemma}\label{result:incomplete-gamma-bound-refined}
 For all $a>1$ and $x>a-1$ we have 
 \begin{align*}
  \G(a,x) \leq  \frac x {x-a+1} \cdot x^{a-1}\,    \eul^{-x}\, .
 \end{align*}
\end{lemma}

\begin{proofof}{Lemma \ref{result:incomplete-gamma-bound-refined}}
 We define $B:= \frac x {x-a+1}$. A simple calculation then shows 
 $a-1 = \frac {B-1}B x$, and by 
  integration by parts we find 
\begin{align*}
\Gamma (a,x)
=
\int_x^\infty \eul^{-t} t^{a-1}\,  dt
= 
-\eul^{-t} t^{a-1}\Bigr|_x^\infty + (a-1)\int_x^\infty \eul^{-t} t^{a-2}\,  dt
& =
x^{a-1} \eul^{-x}  +   \frac {B-1}B \, x \int_x^\infty \eul^{-t} t^{a-2}\,  dt \\
&\leq
x^{a-1} \eul^{-x}  +   \frac {B-1}B \cdot \Gamma (a,x) \, ,
\end{align*}
Now the assertion follows by a simple transformation.
\end{proofof}

To appreciate Lemma \ref{result:incomplete-gamma-bound-refined} we apply it to $a>1$, $\d>0$, and 
$x = (1+\d)a$. This gives 
\begin{align}\label{result:incomplete-gamma-bound-refined-upper}
\G(a,(1+\d) a) 
\leq \frac{ 1+\d}{1+a\d} \cdot   a^a (1+\d)^{a-1} \eul^{-(1+\d) a}\, .
\end{align}
In other words, compared to \eqref{result:incomplete-gamma-bound-upper} the new bound is better by the factor of $\frac{ 1+\d}{1+a\d}$, and for cases with $\d\to 0$ and
$a\d\to \infty$
this extra factor changes the asymptotics.

% 
% \begin{lemma}\label{result:incomplete-gamma-bound}
%    For all $a>1$ and $\d>0$  we have 
% \begin{displaymath}
%    \bigl( (1+\d) a \bigr)^{a-1} \eul^{-(1+\d) a} 
% \leq \G(a,(1+\d) a) 
% \leq \frac{1}{1+a\d} \cdot   \bigl( (1+\d) a \bigr)^{a} \eul^{-(1+\d) a}\, .
% \end{displaymath}
% \end{lemma}
% 
% 
% 
% 
% 
% \begin{proofof}{Lemma \ref{result:incomplete-gamma-bound}}
%    In \cite{NaPa00a}, see also \cite{BoCh09a}, the following bound, which holds for all
%  for $a>1$, $B>1$, and $x> \frac B{B-1}(a-1)$ is shown
% \begin{align}\label{result:incomplete-gamma-bound-h1}
%    x^{a-1} \eul^{-x} \leq \G(a,x) \leq B x^{a-1} \eul^{-x}\, .
% \end{align}
% Now consider $x = a(1+\d)$. 
% For $\e>0$ we then define $B:= (1+\e) \cdot \frac{a+a\d}{1+a\d}$. 
% Now, $a>1$ implies $B>1$, and since we also have 
% $B> \frac{a+a\d}{1+a\d}$ we can   quickly check that $x> \frac B{B-1}(a-1)$.
% Consequently, 
% \eqref{result:incomplete-gamma-bound-h1} holds for  $x = a(1+\d)$, i.e.~we have
% \begin{displaymath}
%    \bigl( (1+\d) a \bigr)^{a-1} \eul^{-(1+\d) a} \leq \G(a,(1+\d) a) 
% \leq B   \bigl( (1+\d) a \bigr)^{a-1} \eul^{-(1+\d) a}
% =  \frac{1+\e}{1+a\d} \cdot   \bigl( (1+\d) a \bigr)^{a} \eul^{-(1+\d) a}
% \end{displaymath}
% Letting $\e\to 0$ then yields the assertion.
% \end{proofof}

\begin{lemma}\label{result:expected-euclid-norm}
 Let $X_1,\dots,X_d$ be i.i.d.~random variables with $X_i \sim \ca N(0, \s^2)$ for some $\s>0$.
 For $X := (X_1,\dots,X_d)\in \Rd$ we then have 
 \begin{align*}
  \E \snorm X_2 & = \s \cdot \frac {\sqrt 2 \cdot \Gamma(\frac {d+1} 2) }{\Gamma(\frac d2)} \, , &
   \var \snorm X_2 & = d \s^2 -  \s^2 \cdot \biggl( \frac {\sqrt 2\cdot \Gamma(\frac {d+1} 2) }{\Gamma(\frac d2)} \biggr)^2  \, .
 \end{align*}
Moreover, the mode is given by  $\mode \snorm X_2 = \s \sqrt{d-1}$  and the density of the distribution of $\snorm X_2$ is 
\begin{align}\label{dens-euclid-norm}
 f(x) = \eins_{[0,\infty)}(x) \cdot   \frac {2^{1-d/2}}{\Gamma(\frac d2)} \cdot  \s^{-d}\cdot  x^{d-1} \eul^{- \s^{-2}x^2/2}\, , \qquad \qquad z\in \R\, .
\end{align}
Finally, for all $s>0$ we have 
\begin{align}\label{dev-euclid-norm}
   P\bigl( \snorm X_2 \geq s\bigr) = \frac {\Gamma(\frac d2, \frac {s^2}{2\s^2})}{\Gamma(\frac d2)}\, .
\end{align}
\end{lemma}

\begin{proofof}{Lemma \ref{result:expected-euclid-norm}}
 Let us consider $Y_i := \s^{-1} X_i$ and $Y := (Y_1,\dots,Y_d)$. Then $Z:= \snorm Y_2$
 is $\chi_d$-distributed, and it is well-known that 
 \begin{align*}
  \E Z &= \frac {\sqrt 2 \cdot \Gamma(\frac {d+1} 2) }{\Gamma(\frac d2)} \, , &
  \mode Z &= \sqrt{d-1}\, , &
  \var Z & = d - 2 \cdot \biggl( \frac {\Gamma(\frac {d+1} 2) }{\Gamma(\frac d2)} \biggr)^2  \, ,
 \end{align*}
    where the latter can also 
    be easily derived using the formula for $\E Z$ and the fact $Z^2 \sim \chi_d^2$, and thus  $\E Z^2 = d$.
    Moreover, the distribution of $Z$ has the Lebesgue density
    \begin{displaymath}
     f_Z(z) = \eins_{[0,\infty)}(z) \cdot  \frac {2^{1-d/2}}{\Gamma(\frac d2)} \cdot    z^{d-1} \eul^{-z^2/2}\, , \qquad \qquad z\in \R\, .
    \end{displaymath}
    Now, \eqref{dens-euclid-norm} easily follows from $\snorm X_2 = \s Z$ and $f(x) = |\s|^{-1} f_Z(x/\s)$ for $x\in \R$.
		For the proof of \eqref{dev-euclid-norm} we first note that \eqref{dens-euclid-norm} yields
	\begin{align*}
	    P\bigl( \snorm X_2 \geq s\bigr)  
		= \int_s^\infty f(x)\, dx
		= \frac {2^{1-d/2}}{\Gamma(\frac d2)} \cdot \s^{-d}\cdot  \int_s^\infty  x^{d-1} \eul^{- \s^{-2}x^2/2} \, dx \, .
	\end{align*}
	Moreover, by the substitution $t:= \p(x) := \frac {x^2}{2\s^2}$ we obtain 
	\begin{align*}
	    \int_s^\infty  x^{d-1} \eul^{- \s^{-2}x^2/2} \,dx
	=  2^{d/2-1} \cdot\s^{d}\cdot \int_s^\infty \bigl( \p(x)\bigr)^{d/2 -1} \eul^{- \p(x)}  \p'(x)\, dx
	 = 2^{d/2-1} \cdot\s^{d}\cdot \int_{\p(s)}^\infty t^{d/2 -1} \eul^{- t}  \, dt\, ,
	\end{align*}
	and combining both equations gives \eqref{dev-euclid-norm}.
\end{proofof}

The next lemma, which will be needed 
to compute the directions of randomly initialized weight vectors,
  is a classical result from multi-dimensional calculus, see e.g.~\cite[Satz §14.8]{Forster17}.

\begin{lemma}\label{result:sphere-integration}
	Let $\Sd$ be the Euclidean sphere in $\Rd$ and $\sd$ be the surface measure on $\Sd$.
  For all Lebesgue integrable functions $f:\Rd\to [0,\infty)$ we then have
	\begin{displaymath}
	   \int_\Rd f \, d\lbd = \int_\Sd \int_0^\infty f(r \xi)\,  r^{d-1} \, dr\, d\sd(\xi)\, .
	\end{displaymath}
\end{lemma}

% For the following theorem we define $\ray A:= \{tx: x\in A,\, t\geq 0\}$ for any $A\subset \Rd$.
% Obviously, if $A$ is measurable, then $\ray A$ is also measurable.

For the following theorem recall that we have seen in Lemma \ref{result:measurable-ray} that 
$\ray A$ is measurable for all measurable $A\subset \Sd$.

\begin{theorem}\label{result:projection-measure}
   Let $f:\Rd\to [0,\infty)$ be a Lebesgue probability density. On $\Sd$ we define 
\begin{displaymath}
   \mu_f (A) := \int_{\ray A} f \, d\lbd\, ,
\end{displaymath}
for all measurable $A\subset \Sd$. Then the following statements hold true:
\begin{enumerate}
   \item $\mu_f$ is a probability measure.
		\item The measure $\mu_f$ is absolutely continuous with respect to the surface measure $\sd$ on $\Sd$.
			Moreover, the $\sd$-density $h$ of $\mu_f$ is $\sd$-almost surely given by 
		\begin{displaymath}
		   h(\xi) = \int_0^\infty f(r \xi) \, r^{d-1}\, dr \, , \qquad \qquad \xi \in \Sd.
		\end{displaymath}
\end{enumerate}
\end{theorem}

\begin{proofof}{Theorem \ref{result:projection-measure}}
\ada i This immediately follows from Lemma \ref{result:ray-algebra} and $\ray \Sd = \Rd\setminus\{0\}$.

 \ada {ii} Using Lemma \ref{result:sphere-integration} we see that the measure $\mu_f$ can be evaluated 
 by 
\begin{align*}
   \mu_f (A) 
& = \int_{\Rd} \eins_{\ray A} \cdot f \, d\lbd \\
& = \int_\Sd \int_0^\infty  \eins_{\ray A}(r\xi ) \cdot   f(r \xi)\,  r^{d-1} \, dr\, d\sd(\xi)\\
& = \int_\Sd \eins_{A}(\xi ) \cdot  \int_0^\infty     f(r \xi)\,  r^{d-1} \, dr\, d\sd(\xi) \\
& = \int_ A  h(\xi) \, d\sd(\xi)
\end{align*}
where in the second to  last step we used $\eins_{\ray A}(r\xi )  = \eins_{A}(\xi )$ for all $\xi \in \Sd$ and all $r>0$. 
\end{proofof}

Let us now recall some facts about sub-Gaussian random variables. To this end, we denote, for a 
given a random variable $X$,  its sub-Gaussian norm by $\sgnorm X$, that is 
\begin{displaymath}
   \sgnorm X := \inf\Bigl\{t> 0: \E \exp\bigl(X^2/t^2   \bigr)   \leq 2  \Bigr\}\, .
\end{displaymath}
We say that $X$ is sub-Gaussian if $\sgnorm X< \infty$.
Some simple properties of $\sgnorm\cdot$ and characterizations of sub-Gaussian random variables
can be found in \cite[Chapter 2.5]{Vershynin18}. In particular recall from there that 
$\sgnorm X \leq \frac 1 {\sqrt{\ln 2}} \inorm X$ for all bounded $X$. In addition, we have 
$\sgnorm{\a X} = |\a| \cdot \sgnorm X$
whenever $\sgnorm X< \infty$. Finally, since the moment generating function of a $Z\sim \chi_1^2$ 
is 
\begin{displaymath}
   \E \eul^{tZ}              = \frac 1 {\sqrt{1-2t}}\, , \qquad \qquad t\in [0,1/2),
\end{displaymath}
it is easy to check that $\sgnorm X = \sqrt{8/3}$ for $X\sim \ca N(0,1)$.

Now, 
 \cite[Theorem 3.1.1]{Vershynin18}, or more precisely, Inequality (3.3) in its proof, provides
the following concentration of the Euclidean norm. Note that some tedious calculations along the lines of its proof
will actually provide a value for the universal constant in this estimate.

\begin{theorem}\label{result:euclid-norm-conc}
	There exists a universal constant $K>0$ such that for all $d\geq 1$, all
   i.i.d.~random variables $X_1,\dots,X_d$   with $c_X := \max\{1, \sgnorm {X_1}\}<\infty$ and $\E X_1^2 = 1$,
  and all $t>0$ the random vector $X=(X_1,\dots,X_d)$ satisfies
	\begin{displaymath}
	   P\Bigl( \bigl| \,\snorm X_2 - \sqrt d \,\, \bigr| \geq t  \Bigr) \leq 2 \exp \biggl( -\frac {Kt^2}{c_X^4}  \biggr)  \, .
	\end{displaymath}
\end{theorem}

\begin{lemma}\label{result:half-sym-distribution}
 Let $Y$ be a symmetric random variable. Then we have $2 \E \relu Y^2 = \E Y^2$.
\end{lemma}

\begin{proofof}{Lemma \ref{result:half-sym-distribution}}
 Let $P_Y$ be the distribution of $Y$. 
 Then a simple calculation shows
 \begin{displaymath}
  \E \relu Y^2 = \int_\R \relu y^2 \, dP_Y(y) = \int_{(0,\infty)} y^2 \, dP_Y(y) \, .
 \end{displaymath}
 Moreover, an analogous calculation shows 
 \begin{displaymath}
  \E   Y^2 = \int_\R   y^2 \, dP_Y(y) = \int_{(-\infty, 0)} y^2 \, dP_Y(y) + \int_{(0,\infty)} y^2 \, dP_Y(y) 
  = \int_{(0, \infty)} y^2 \, dP_Y(y) + \int_{(0,\infty)} y^2 \, dP_Y(y)  \, ,
 \end{displaymath} 
 where in the last step we used the symmetry of $P_Y$. By combining both equations we obtain  the assertion. 
\end{proofof}

% -------------------------------------------------------------------------------------------------------------------------------

\newcommand{\regressiondatacharacteristics}{Characteristics of the regression data sets. The naive error is the standard deviation of the labels. Note that this equals the root mean square error one obtains when predicting all labels by the label mean.}
\begin{table}[h!]
{\tablefontsize
\centering
\begin{tabular}{|r|r|r|r|}
\hline
&Size&Dimension&Naive Error \\
\hline
air-quality-bc&8991&10&.23426 \\
\hline
air-quality-co2&7674&10&.24630 \\
\hline
air-quality-no2&7715&10&.28620 \\
\hline
air-quality-nox&7718&10&.28838 \\
\hline
appliances-energy&19735&29&.19163 \\
\hline
bejing-pm25&41757&12&.18521 \\
\hline
bike-sharing-casual&17379&12&.26868 \\
\hline
bike-sharing-total&17379&12&.37169 \\
\hline
carbon-nanotubes-u&10721&5&.63039 \\
\hline
carbon-nanotubes-v&10721&5&.63114 \\
\hline
carbon-nanotubes-w&10721&5&.57816 \\
\hline
chess-krvk&28056&22&4.63674 \\
\hline
cycle-power-plant&9568&4&.45208 \\
\hline
facebook-comment-volume&40949&52&.05440 \\
\hline
five-cities-beijing-pm25&19062&14&.24504 \\
\hline
five-cities-chengdu-pm25&21074&14&.19888 \\
\hline
five-cities-guangzhou-pm25&20074&14&.16495 \\
\hline
five-cities-shanghai-pm25&21436&14&.16315 \\
\hline
five-cities-shenyang-pm25&19038&14&.13064 \\
\hline
gas-sensor-drift-class&13910&128&1.72854 \\
\hline
gas-sensor-drift-conc&13910&128&.34322 \\
\hline
indoor-loc-alt&21048&520&.60629 \\
\hline
indoor-loc-lat&21048&520&.49678 \\
\hline
indoor-loc-long&21048&520&.62644 \\
\hline
insurance-benchmark&9822&85&.23686 \\
\hline
naval-propulsion-comp&11934&14&.58878 \\
\hline
naval-propulsion-turb&11934&14&.60000 \\
\hline
nursery&12960&8&1.23560 \\
\hline
online-news-popularity&39644&58&.02757 \\
\hline
parkinson-motor&5875&19&.47159 \\
\hline
parkinson-total&5875&19&.44588 \\
\hline
protein-tertiary-structure&45730&9&.58272 \\
\hline
skill-craft&3338&18&1.44795 \\
\hline
sml2010-dining&4137&17&.37693 \\
\hline
sml2010-room&4137&17&.37896 \\
\hline
wall-follow-robot-2&5456&2&1.00466 \\
\hline
wall-follow-robot-24&5456&24&1.00466 \\
\hline
wall-follow-robot-4&5456&4&1.00466 \\
\hline
wine-quality-all&6497&12&.87319 \\
\hline
wine-quality-white&4898&11&.88555 \\
\hline
\end{tabular}
\caption{\regressiondatacharacteristics}
\label{regressiondatacharacteristics}
}
\end{table}

\newcommand{\logdatacharacteristics}{Characteristics of the classification data sets. The naive error is the classification error one obtains when predicting all labels by the most frequent label in the data set.}
\begin{table}[h!]
{\tablefontsize
\centering
\begin{tabular}{|r|r|r|r|}
\hline
&Size&Dimension&Naive Error \\
\hline
abalone&2870&8&.46760 \\
\hline
adult&45222&88&.24784 \\
\hline
anuran-calls-families&6585&22&.32878 \\
\hline
anuran-calls-genus&5743&22&.27738 \\
\hline
anuran-calls-species&4599&22&.24375 \\
\hline
avila&12495&10&.31397 \\
\hline
bank-marketing&41579&29&.11407 \\
\hline
\end{tabular}
\caption{\logdatacharacteristics}
\label{logdatacharacteristics}
}
\end{table}

\newcommand{\regressionAllunlimpatfivecverrorAvA}{Comparison of all considered methods on the regression data sets.
The numbers in the upper part of the table are the average root mean squared test errors
the corresponding standard deviations. Red and orange entries indicate the best-performing method for each data set,
where red entries indicate those methods that are significantly better than the second best method according 
to a paired two-sample $t$-test with a  significance level of $95\%$. Similarly, blue and green entries indicate the
worst-performing method, and blue entries indicate those methods that are significantly worse than the
 best method according 
to a paired two-sample $t$-test with with a significance level of $95\%$.}
\begin{table}[h!]
{\tablefontsize
\centering
\begin{tabular}{|r|r|r|r|r|r|}
\hline
&ReLU BN He zero&ReLU He zero&ReLU sphere hull -5&SeLU SNN zero&SeLU ball hull -5 \\
\hline
air-quality-bc&\!\!\significantworseentry{.02530} {\sdsize\mypm\hspace*{0.2ex}.00292}\!\!&\!\!.01196 {\sdsize\mypm\hspace*{0.2ex}.00180}\!\!&\!\!\significantbetterentry{.00440} {\sdsize\mypm\hspace*{0.2ex}.00145}\!\!&\!\!.01194 {\sdsize\mypm\hspace*{0.2ex}.00147}\!\!&\!\!.00500 {\sdsize\mypm\hspace*{0.2ex}.00237}\!\! \\
\hline
air-quality-co2&\!\!\significantworseentry{.08348} {\sdsize\mypm\hspace*{0.2ex}.00397}\!\!&\!\!.08002 {\sdsize\mypm\hspace*{0.2ex}.00367}\!\!&\!\!\betterentry{.07965} {\sdsize\mypm\hspace*{0.2ex}.00344}\!\!&\!\!.08288 {\sdsize\mypm\hspace*{0.2ex}.00379}\!\!&\!\!.08301 {\sdsize\mypm\hspace*{0.2ex}.00382}\!\! \\
\hline
air-quality-no2&\!\!.11367 {\sdsize\mypm\hspace*{0.2ex}.00402}\!\!&\!\!.10787 {\sdsize\mypm\hspace*{0.2ex}.00335}\!\!&\!\!\significantbetterentry{.10280} {\sdsize\mypm\hspace*{0.2ex}.00428}\!\!&\!\!\significantworseentry{.11682} {\sdsize\mypm\hspace*{0.2ex}.00413}\!\!&\!\!.11251 {\sdsize\mypm\hspace*{0.2ex}.00483}\!\! \\
\hline
air-quality-nox&\!\!.09470 {\sdsize\mypm\hspace*{0.2ex}.00494}\!\!&\!\!.09003 {\sdsize\mypm\hspace*{0.2ex}.00510}\!\!&\!\!\significantbetterentry{.08819} {\sdsize\mypm\hspace*{0.2ex}.00520}\!\!&\!\!\significantworseentry{.09709} {\sdsize\mypm\hspace*{0.2ex}.00517}\!\!&\!\!.09496 {\sdsize\mypm\hspace*{0.2ex}.00656}\!\! \\
\hline
appliances-energy&\!\!\significantworseentry{.16003} {\sdsize\mypm\hspace*{0.2ex}.00480}\!\!&\!\!.15697 {\sdsize\mypm\hspace*{0.2ex}.00521}\!\!&\!\!\significantbetterentry{.15340} {\sdsize\mypm\hspace*{0.2ex}.00503}\!\!&\!\!.15975 {\sdsize\mypm\hspace*{0.2ex}.00544}\!\!&\!\!.15571 {\sdsize\mypm\hspace*{0.2ex}.00500}\!\! \\
\hline
\end{tabular}
\caption{\regressionAllunlimpatfivecverrorAvA}
\label{regressionAllunlimpatfivecverrorAvA}
}
\end{table}

\newcommand{\logAllunlimpatfivecverrorAvA}{Comparison between \heetal~with and without batch normalization, the original SeLU, and our new initialization strategies.
Our new strategies outperform the three standard approaches both in terms of ranking and reliability.}
\begin{table}[h!]
{\tablefontsize
\centering
\begin{tabular}{|r|r|r|r|r|r|r|}
\hline
&ReLU BN He zero&ReLU He zero&ReLU He hull -5&ReLU ball hull +5&SeLU SNN zero&SeLU ball hull -5 \\
\hline
abalone&\!\!.19530 {\sdsize\mypm\hspace*{0.2ex}.02380}\!\!&\!\!.19628 {\sdsize\mypm\hspace*{0.2ex}.01635}\!\!&\!\!\betterentry{.18828} {\sdsize\mypm\hspace*{0.2ex}.01255}\!\!&\!\!.18963 {\sdsize\mypm\hspace*{0.2ex}.01213}\!\!&\!\!\significantworseentry{.19743} {\sdsize\mypm\hspace*{0.2ex}.01377}\!\!&\!\!.19012 {\sdsize\mypm\hspace*{0.2ex}.01211}\!\! \\
\hline
adult&\!\!.15258 {\sdsize\mypm\hspace*{0.2ex}.00406}\!\!&\!\!.15405 {\sdsize\mypm\hspace*{0.2ex}.00345}\!\!&\!\!\significantworseentry{.15461} {\sdsize\mypm\hspace*{0.2ex}.00345}\!\!&\!\!.15337 {\sdsize\mypm\hspace*{0.2ex}.00399}\!\!&\!\!\significantbetterentry{.14984} {\sdsize\mypm\hspace*{0.2ex}.00383}\!\!&\!\!.15054 {\sdsize\mypm\hspace*{0.2ex}.00338}\!\! \\
\hline
anuran-calls-families&\!\!\significantbetterentry{.00714} {\sdsize\mypm\hspace*{0.2ex}.00239}\!\!&\!\!.00949 {\sdsize\mypm\hspace*{0.2ex}.00254}\!\!&\!\!.00876 {\sdsize\mypm\hspace*{0.2ex}.00254}\!\!&\!\!.00960 {\sdsize\mypm\hspace*{0.2ex}.00234}\!\!&\!\!\significantworseentry{.01034} {\sdsize\mypm\hspace*{0.2ex}.00294}\!\!&\!\!.01016 {\sdsize\mypm\hspace*{0.2ex}.00359}\!\! \\
\hline
anuran-calls-genus&\!\!\betterentry{.00230} {\sdsize\mypm\hspace*{0.2ex}.00137}\!\!&\!\!.00251 {\sdsize\mypm\hspace*{0.2ex}.00135}\!\!&\!\!.00249 {\sdsize\mypm\hspace*{0.2ex}.00128}\!\!&\!\!.00270 {\sdsize\mypm\hspace*{0.2ex}.00155}\!\!&\!\!\significantworseentry{.00320} {\sdsize\mypm\hspace*{0.2ex}.00159}\!\!&\!\!.00306 {\sdsize\mypm\hspace*{0.2ex}.00147}\!\! \\
\hline
anuran-calls-species&\!\!.00143 {\sdsize\mypm\hspace*{0.2ex}.00117}\!\!&\!\!.00126 {\sdsize\mypm\hspace*{0.2ex}.00138}\!\!&\!\!.00128 {\sdsize\mypm\hspace*{0.2ex}.00100}\!\!&\!\!\significantworseentry{.00148} {\sdsize\mypm\hspace*{0.2ex}.00127}\!\!&\!\!.00141 {\sdsize\mypm\hspace*{0.2ex}.00119}\!\!&\!\!\betterentry{.00117} {\sdsize\mypm\hspace*{0.2ex}.00107}\!\! \\
\hline
\end{tabular}
\caption{\logAllunlimpatfivecverrorAvA}
\label{logAllunlimpatfivecverrorAvA}
}
\end{table}

% \newcommand{\logAlltimepatfivecverrorAvA}{Comparison between \heetal~with batch normalization, the original SeLU, and our new method
% with a tight time budget, in which the   allowed training time for a data set and a  method equals the $10\%$-percentile 
% of all run times for that data set.
% Our new method significantly outperforms the two standard approaches both in terms of ranking and reliability.
% In particular, the difference to standard initialization plus batch normalization is pronounced since 
% the normalization requires significant computational resources.
% }
% \input{./tables_log/All_time_pat5_cv_error_AvA}

% \input{./tables_log/All_unlim_pat5_cv_error}

% \input{./tables_log/All_unlim_pat5_cv_time_AvA}

% \input{./tables_log/All_time_pat5_cv_error}
% \input{./tables_log/All_time_pat5_cv_error_AvA}
% \input{./tables_log/All_time_pat5_cv_time_AvA}
% 
% % \input{./tables_log_more/All_unlim_pat5_cv_error}
% % \input{./tables_log_more/All_unlim_pat5_cv_error_AvA}
% % \input{./tables_log_more/All_unlim_pat5_cv_time_AvA}

% % \input{./tables_hinge/All_unlim_pat5_cv_error_AvA}

% \newcommand{\regressionAllunlimpatfivecverror}{bla}
% \input{./tables_regression/All_unlim_pat5_cv_error}

% \input{./tables_log/All_unlim_pat5_cv_error_AvA}

% \newcommand{\logAllunlimpatfivecverrorAvA}{Comparison between \heetal~with and without batch normalization, the original SeLU, and our new initialization strategies.
% Our new strategies outperform the three standard approaches both in terms of ranking and reliability.}
% \input{./tables_ls_svm/All_cv_error_AvA}

% \include{todos}

\end{document}